\documentclass[preprint]{elsarticle}
\usepackage{graphicx}
\usepackage{tabularx}
\usepackage{algorithm}
\usepackage{algorithmic}
\usepackage{amssymb}
\usepackage{subfigure}
\usepackage{array}
\usepackage{lscape}
\usepackage{dcolumn,longtable,hhline,colortbl}
\usepackage[rotateright]{rotating}
\usepackage{booktabs,makecell,multirow}
\usepackage[section]{placeins}
\usepackage{enumerate}
\usepackage{color,colortbl}
\usepackage{hyperref}
\usepackage{lineno}
\usepackage[margin=2.5cm]{geometry}
\usepackage{color}
\usepackage{soul}
\usepackage{url}
\usepackage[table]{xcolor}
\usepackage{makecell}
\usepackage{amsthm}
\usepackage{multirow}
\usepackage{bm}
\usepackage{autoaligne}
\usepackage{mathtools}
\usepackage[flushleft]{threeparttable}
\usepackage{setspace}
\usepackage{appendix}
\usepackage{nomencl}
\usepackage{amsmath}
\usepackage{amssymb}
\usepackage{subfigure}
\usepackage{enumitem}
\usepackage{longtable}
\usepackage{multirow}
\usepackage{enumitem}
\usepackage[english]{babel}
\usepackage[autostyle]{csquotes}
\usepackage{upgreek}
\usepackage{subfigure}
\usepackage{lscape}
\usepackage{pdflscape}
\usepackage{tabularray}
\usepackage{bbm}
\usepackage{nomencl}
\usepackage{multicol}
\usepackage{tikz}
\usetikzlibrary{arrows}
\biboptions{sort&compress}
\tikzset{>=latex} 
\colorlet{myred}{red!80!black}
\colorlet{myblue}{blue!80!black}
\colorlet{mygreen}{green!60!black}
\colorlet{mydarkred}{myred!40!black}
\colorlet{mydarkblue}{myblue!40!black}
\colorlet{mydarkgreen}{mygreen!40!black}
\tikzstyle{node}=[very thick,circle,draw=myblue,minimum size=22,inner sep=0.5,outer sep=0.6]
\tikzstyle{connect}=[->,thick,mydarkblue,shorten >=1]
\tikzset{ 
  node 1/.style={node,mydarkgreen,draw=mygreen,fill=mygreen!25},
  node 2/.style={node,mydarkblue,draw=myblue,fill=myblue!20},
  node 3/.style={node,mydarkred,draw=myred,fill=myred!20},
}

\usetikzlibrary{arrows,tikzmark,shadows}

\tikzset{
  comment/.style={
    draw,
    fill={rgb:red,255;green,173;blue,0},
    text=white,
    rounded corners,
    drop shadow,
    align=left,
  },
}
\usepackage{listings}

\usepackage{xcolor}

\definecolor{codegreen}{rgb}{0,0.6,0}
\definecolor{codegray}{rgb}{0.5,0.5,0.5}
\definecolor{codepurple}{rgb}{0.58,0,0.82}
\definecolor{backcolour}{rgb}{0.95,0.95,0.92}

\lstdefinestyle{mystyle}{
  backgroundcolor=\color{backcolour}, commentstyle=\color{codegreen},
  keywordstyle=\color{magenta},
  numberstyle=\tiny\color{codegray},
  stringstyle=\color{codepurple},
  basicstyle=\ttfamily\footnotesize,
  breakatwhitespace=false,         
  breaklines=true,                 
  captionpos=b,                    
  keepspaces=true,                                  
  numbersep=5pt,   
  numbers=none, 
  showspaces=false,                
  showstringspaces=false,
  showtabs=false,                  
  tabsize=2
}

\newcommand{\afigwidth}{1.28in}

\usepackage{xfrac}
\usepackage{etoolbox}

\makenomenclature

\renewcommand\nomgroup[1]{%
  \item[\bfseries
  \ifstrequal{#1}{A}{List of acronyms}{%
  \ifstrequal{#1}{C}{List of mathematical notations}}%
]}

\renewcommand*{\nompreamble}{\begin{multicols}{2}}
\renewcommand*{\nompostamble}{\end{multicols}}

\usepackage[fontsize=11pt]{scrextend}

\renewcommand\theadalign{bc}
\renewcommand\theadfont{\bfseries}
\renewcommand\theadgape{\Gape[4pt]}
\renewcommand\cellgape{\Gape[4pt]}
\DeclareMathOperator*{\argmin}{argmin}

\makeatletter
\hypersetup{
    colorlinks=true,
    linkcolor=blue,
    filecolor=magenta,
    urlcolor=cyan,
}

\makeatletter
\def\@xfootnote[#1]{%
  \protected@xdef\@thefnmark{#1}%
  \@footnotemark\@footnotetext}
\makeatother

\newsavebox\CBox

\newcolumntype{L}[1]{>{\raggedright\let\newline\\\arraybackslash\hspace{0pt}}m{#1}}
\newcolumntype{C}[1]{>{\centering\let\newline\\\arraybackslash\hspace{0pt}}m{#1}}
\newcolumntype{R}[1]{>{\raggedleft\let\newline\\\arraybackslash\hspace{0pt}}m{#1}}

\journal{Mechanical Systems and Signal Processing}

\begin{document}

\begin{frontmatter}

\title{Uncertainty Quantification in Machine Learning for Engineering Design and Health Prognostics:
A Tutorial}

\author[address1]{Venkat Nemani}
\author[address2]{Luca Biggio}
\author[address3]{Xun Huan}
\author[address7]{Zhen Hu}
\author[address8]{Olga Fink}
\author[address4]{Anh Tran}
\author[address5]{Yan Wang}
\author[address9,address10]{Xiaoge Zhang\corref{label1}}
\ead{xiaoge.zhang@polyu.edu.hk}
\author[address11]{Chao Hu\corref{label1}}
\ead{chao.hu@uconn.edu}

\cortext[label1]{Correspondence authors.}

\address[address1]{Department of Mechanical Engineering, Iowa State University, Ames, IA 50011, USA}
\address[address2]{Data Analytics Lab, ETH, Zürich, Switzerland}
\address[address3]{Department of Mechanical Engineering, University of Michigan, Ann Arbor, MI 48109, USA}
\address[address7]{Department of Industrial and Manufacturing Systems Engineering, University of Michigan-Dearborn, Dearborn, MI 48128, USA}
\address[address8]{Intelligent Maintenance and Operations Systems, EPFL, Lausanne, 12309, Switzerland}
\address[address4]{Scientific Machine Learning, Sandia National Laboratories, Albuquerque, NM 87123, USA}
\address[address5]{George W. Woodruff School of Mechanical Engineering, Georgia Institute of Technology, Atlanta, GA 30332, USA}
\address[address9]{Department of Industrial and Systems Engineering, The Hong Kong Polytechnic University, Kowloon, Hong Kong}
\address[address10]{Center for Advances in Reliability and Safety (CAiRS), New Territories, Hong Kong}
\address[address11]{Department of Mechanical Engineering, University of Connecticut, Storrs, CT 06269, USA}

\begin{abstract}
On top of machine learning (ML) models, uncertainty quantification (UQ) functions as an essential layer of safety assurance that could lead to more principled decision making by enabling sound risk assessment and management. The safety and reliability improvement of ML models empowered by UQ has the potential to significantly facilitate the broad adoption of ML solutions in high-stakes decision settings, such as healthcare, manufacturing, and aviation, to name a few. In this tutorial, we aim to provide a holistic lens on emerging UQ methods for ML models with a particular focus on neural networks and the applications of these UQ methods in tackling engineering design as well as prognostics and health management problems. Toward this goal, we start with a comprehensive classification of uncertainty types, sources, and causes pertaining to UQ of ML models. Next, we provide a tutorial-style description of several state-of-the-art UQ methods: Gaussian process regression, Bayesian neural network, neural network ensemble, and deterministic UQ methods focusing on spectral-normalized neural Gaussian process. Established upon the mathematical formulations, we subsequently examine the soundness of these UQ methods quantitatively and qualitatively (by a toy regression example) to examine their strengths and shortcomings from different dimensions. Then, we review quantitative metrics commonly used to assess the quality of predictive uncertainty in classification and regression problems. Afterward, we discuss the increasingly important role of UQ of ML models in solving challenging problems in engineering design and health prognostics. Two case studies with source codes available on GitHub are used to demonstrate these UQ methods and compare their performance in the life prediction of lithium-ion batteries at the early stage (case study 1) and the remaining useful life prediction of turbofan engines (case study 2).
\end{abstract}

\begin{keyword}
Machine learning\sep Uncertainty quantification \sep Engineering design \sep Prognostics and health management
\end{keyword}
\end{frontmatter}

\nomenclature[A, 01]{ARD}{Automatic relevance determination}
\nomenclature[A, 02]{BNN}{Bayesian neural network}
\nomenclature[A, 03]{DL}{Deep learning}
\nomenclature[A, 04]{DNN}{Deep neural network}
\nomenclature[A, 04]{ECE}{Expected calibration error}
\nomenclature[A, 05]{EI}{Expected improvement}
\nomenclature[A, 06]{ELBO}{Evidence lower bound}
\nomenclature[A, 07]{GAN}{Generative adversarial network}
\nomenclature[A, 08]{GPR}{Gaussian process regression}
\nomenclature[A, 09]{HMC}{Hamiltonian Monte Carlo}
\nomenclature[A, 10]{KL}{Kullback–Leibler}
\nomenclature[A, 11]{MC}{Monte Carlo}
\nomenclature[A, 12]{MCMC}{Markov chain Monte Carlo}
\nomenclature[A, 13]{MFVI}{Mean-field variational inference}
\nomenclature[A, 14]{ML}{Machine learning}
\nomenclature[A, 15]{MSE}{Mean squared error}
\nomenclature[A, 16]{NLL}{Negative log-likelihood}
\nomenclature[A, 17]{OOD}{Out of distribution}
\nomenclature[A, 18]{PDF}{Probability density function}
\nomenclature[A, 19]{PHM}{Prognostics and health management}
\nomenclature[A, 20]{RUL}{Remaining useful life}
\nomenclature[A, 21]{SNGP}{Spectral-normalized neural Gaussian process}
\nomenclature[A, 22]{SVGD}{Stein variational gradient descent}
\nomenclature[A, 23]{UQ}{Uncertainty quantification}
\nomenclature[A, 24]{VAE}{Variational autoencoder}
\nomenclature[A, 25]{VI}{Variational inference}

\nomenclature[C, 01]{$\mathcal{D} = \left\{ {\left( {{\mathbf{x}_1},{y_1}} \right),\left( {{\mathbf{x}_2},{y_2}} \right), \cdots ,\left( {{\mathbf{x}_N},{y_N}} \right)} \right\}$}{Training data}
\nomenclature[C, 02]{$D$}{Number of features (dimensions) in a single input $\mathbf{x}$}
\nomenclature[C, 03]{$\varepsilon$}{A random noise variable following a zero-mean Gaussian distribution}
\nomenclature[C, 04]{$\mathbb{E}\left[ \bullet  \right]$}{Expectation of $\bullet$}

\nomenclature[C, 05]{$k(\mathbf{x}, \mathbf{x}')$}{Covariance function or kernel in GPR depicting the covariance between function outputs at $\mathbf{x}$ and $\mathbf{x}'$}

\nomenclature[C, 06]{$\lambda$}{Parameter to be optimized in the variational distribution $q$}
\nomenclature[C, 07]{$l$}{Length scale parameter of a kernel}
\nomenclature[C, 08]{$N$}{Number of training samples}
\nomenclature[C, 09]{$p$}{Probability density}
\nomenclature[C, 10]{$p \left(\bm{\uptheta} \right)$}{Prior distribution of $\bm{\uptheta}$}
\nomenclature[C, 11]{$p(\bm{\uptheta}\rvert \mathcal{D})$}{Posterior distribution of $\bm{\uptheta}$ given the training data $\mathcal{D}$}
\nomenclature[C, 12]{$p(\mathbf{y}\rvert \bm{\uptheta}, \mathbf{X})$}{Likelihood function indicating the probability of observing $\mathbf{y}$ given the parameters $\bm{\uptheta}$ and inputs $\mathbf{X}$}

\nomenclature[C, 13]{$q \left(\bm{\uptheta}; \lambda \right)$}{A variational distribution parameterized by $\lambda$ to approximate the posterior distribution $p(\bm{\uptheta}\rvert \mathcal{D})$}
\nomenclature[C, 14]{$\sigma_\mathrm{f}$}{Signal amplitude parameter of a kernel}
\nomenclature[C, 15]{$\sigma_\varepsilon$}{Standard
deviation of a random noise variable $\varepsilon$}
\nomenclature[C, 16]{$\bm{\uptheta}$}{Set of tunable parameters in an ML model}
\nomenclature[C, 17]{$\bm{\uptheta}^*$}{Set of optimal parameters in an ML model after tuning}

\nomenclature[C, 18]{$\bm{\mathrm{X}} = \left\{ {{\mathbf{x}_1},{\mathbf{x}_2}, \cdots ,{\mathbf{x}_N}} \right\}$}{Inputs (or input points) in a training dataset for BNN}
\nomenclature[C, 19]{$\mathbf{X}_\mathrm{t}$}{Matrix representation of inputs in training data, i.e., $\mathbf{X}_\mathrm{t} = [\mathbf{x}_1, \dots, \mathbf{x}_{N}]^\mathrm{T} \in \mathbb{R}^{N \times D}$}
\nomenclature[C, 20]{$\bm{\mathrm{x}}$}{A single input, $\bm{\mathrm{x}} \in \mathbb{R}^{N}$}
\nomenclature[C, 21]{$\mathbf{x}_*$}{A test point}

\nomenclature[C, 22]{$y$}{A single observation/target, $y \in \mathbb{R}^{1}$}

\nomenclature[C, 23]{$\bm{\mathrm{y}} = \left\{ {{y_1},{y_2}, \cdots ,{y_N}} \right\}$}{Observations/targets in a training dataset to be predicted by an ML model}
\nomenclature[C, 24]{$\mathbf{y}_\mathrm{t}$}{Matrix representation of target output in training data, that is $\mathbf{y}_\mathrm{t} = [y_1, \dots, y_{N}]^\mathrm{T} \in \mathbb{R}^{N}$}

\printnomenclature


\section{Introduction}
\label{sec:introduction}
In recent years, data-driven machine learning (ML) models have become increasingly prevalent across a wide range of engineering fields. Two application domains of interest to this tutorial are engineering design and post-design health prognostics. The ML community has devoted significant efforts toward creating deep learning (DL) models that yield improved prediction accuracy over earlier DL models on publicly available, large, standardized datasets, such as MNIST \citep{lecun1998gradient}, ImageNet \citep{deng2009imagenet}, Places \citep{zhou2014learning}, and Microsoft COCO \citep{lin2014microsoft}. Among these DL models are deep neural networks (DNNs), known for their ability to extract high-level abstracted features from large volumes of data automatically achieved through multiple layers of neurons and activation functions in an end-to-end fashion. 

Despite record-breaking prediction accuracy on some fixed sets of test samples (i.e., images in the case of computer vision), these neural networks typically have difficulties in generalizing to data not observed during model training. Suppose test samples come from a distribution substantially different from the training distribution, where most of the training samples are located. These test samples can be called out-of-distribution (OOD) samples. Trained neural network models tend to produce large prediction errors on these OOD samples. Despite considerable efforts, such as domain adaption~\citep{blitzer2007biographies,glorot2011domain,li2021knowledge}, aimed at improving the generalization performance of neural network models, the issue of poor generalizability still persists. Another limitation that adds to the challenge is that complex ML models, such as DNNs, are mostly black-box in nature. It is generally preferred to use simpler models (e.g., linear regression and decision tree) that are easier to interpret unless more complex models can be justified with non-incremental benefits (e.g., substantially improved accuracy). In recent years, the growing availability of large volumes of data has made complex models, which are often significantly more accurate than simple models, the obvious better choice in many ML applications where prediction accuracy is the priority. Consequently, black-box ML models that are hard to understand are increasingly deployed, particularly in big data applications. Some efforts have been made to address the lack of interpretability, with notable explanation algorithms such as SHAP \citep{lundberg2017unified} and Grad-CAM \citep{selvaraju2017grad} and a good review of interpretable ML \cite{molnar2020interpretable}. Despite these recent efforts, many complex ML models are still implemented as black-box models and cannot explain their predictions to the end user for various reasons. This limitation makes it extremely intricate for the end user to understand the decision mechanism behind a neural network’s prediction. Given these two limitations (difficulties in extrapolating to OOD samples and lack of interpretability), it is vital to quantify the predictive uncertainty of a trained ML model and communicate this uncertainty to end users in an easy-to-understand way. To enhance algorithmic transparency and trustworthiness, uncertainty quantification (UQ) and interpretation should ideally be performed together, with UQ providing information on the confidence of complex machine learning models in making predictions. This integration allows for a better understanding of often difficult-to-interpret models and their predictions.

Let us first look at typical ways to express and communicate predictive uncertainty. A simple case is with classification problems, where the probability of the model-predicted class can depict model confidence at a prediction. For example, a fault classification model may predict a bearing to have an inner race fault with a 90\% probability/confidence. In regression problems, predictive uncertainty is often communicated as confidence intervals, shown as error bars on graphs visualizing predictions. For instance, we could train a probabilistic ML model to predict the number of weeks a rolling element bearing can be used before failure, i.e., the remaining useful life (RUL). An example prediction may be 120 $\pm$ 15, in weeks, which represents a two-sided 95\% confidence interval (i.e., $\sim$1.96 standard deviations subtracted from or added to the mean estimate assuming the model-predicted RUL follows a Gaussian distribution). A narrower confidence interval comes from lower predictive uncertainty, which suggests higher model confidence. 

One clear advantage of UQ is that it helps end users determine when they can trust predictions made by the model and when they need extra caution while making decisions based on these predictions. This is especially important when incorrect decisions can lead to severe financial losses or even life-threatening outcomes. Towards this end, the integration of UQ in ML models, as well as the sound quantification and calibration of uncertainty in ML model prediction, has a viable potential to tackle a central research question the ML community confronts -- safety assurance of ML models~\citep{jimenez2020drug,guo2022hierarchical,khan2018review,thelen2022comprehensivepart1}. In fact, the absence of essential performance characteristics (e.g., model robustness and safety assurance) has emerged as the fundamental roadblock to limiting ML's application scope in risk-insensitive areas, while its adoptions in high-stakes, high-reward decision environments (e.g., healthcare, aviation, and power grid) are still in the infancy stage primarily because of the reluctance of end users to delegate critical decision making to machine intelligence in cases where the safety of patients or critical engineering systems might be put at stake~\cite{begoli2019need,rudin2019stop,sensoy2018evidential,zhang2022airport}. Towards the translation of ML solutions in high-risk domains, UQ offers an additional dimension by extending the traditional discipline of statistical error analysis to capture various uncertainties arising from limited or noisy data, missing variables, incomplete knowledge, etc. This development has wide-ranging implications for supporting quantitative and precise risk management in high-stakes decision-making settings, particularly concerning potential model failures and decision limitations of ML algorithms. However, the evaluation of ML model performance on most benchmarking datasets focuses exclusively on some form of prediction accuracy on a fixed test dataset; it rarely considers the quality of predictive uncertainty. As a result, UQ of ML models is typically pushed to the sidelines, yielding the centerlines to prediction accuracy. In reality, underestimating uncertainty (overconfidence) can create trust issues, while overestimating uncertainty (underconfidence) may result in overly conservative predictions, ultimately diminishing the value of ML.

More recently (approximately since 2015), there has been growing interest in approaches to estimating the predictive uncertainty of deep learning models, for example, in the form of class probability for classification and predicted variance for regression, as discussed earlier. The growing interest can be attributed to failure cases where trained ML models produced unexpectedly incorrect predictions on test samples while communicating high confidence in the predictions \citep{hullermeier2021aleatoric} and those where models changed their predictions substantially in response to minor, unimportant changes to samples (or so-called \emph{adversarial samples}) \citep{szegedy2013intriguing}. Two pioneering studies that stimulated many subsequent efforts created two widely used approaches to UQ of neural networks: (1) Monte Carlo (MC) dropout as a computationally efficient alternative to traditional Bayesian neural network \citep{gal2016dropout} and (2) neural network ensemble consisting of multiple independently trained neural networks, each predicting a mean and standard deviation of a Gaussian target  \citep{lakshminarayanan2017simple}. Another notable early study highlighted differences between aleatory and epistemic uncertainty and discussed situations where quantifying aleatory uncertainty is important and where quantifying epistemic uncertainty is important \citep{kendall2017uncertainties}. A common understanding in the ML community towards these two types of uncertainty has been the following: aleatory uncertainty can be considered \emph{data uncertainty} and represents inherent randomness (e.g., measurement noise) in observations of the target that an ML model is tasked with predicting; epistemic uncertainty can be treated as \emph{model uncertainty} and results from having access to only limited training data, which makes it not possible to learn a precise model. As discussed in Sec. \ref{sec:aleatory_epistemic}, aleatory and epistemic uncertainty could encompass more sources and causes than the well-known data and model uncertainty. 

The engineering design community has a long history of applying Gaussian process regression (GPR) or kriging, an ML method with UQ capability, to build cheap-to-evaluate surrogates of expensive simulation models for simulation-based design, dating back to the early 2000s \citep{ jin2001comparative, queipo2005surrogate, wang2007review}. GPR has an elegant way of quantifying aleatory and epistemic uncertainty and can produce high uncertainty on OOD samples. However, the UQ capability of GPR is typically not used to detect OOD samples or quantify the epistemic uncertainty of a final built surrogate. Rather, it is leveraged in an adaptive sampling scheme to encourage sampling in highly uncertain and critical regions of the input space (exploration) to minimize the number of training samples for either (1) building an accurate surrogate within some lower and upper bounds of input variables (local or global surrogate modeling) \citep{jin2002sequential,bichon2008efficient,echard2011ak} or (2) finding a globally optimally design for some expensive-to-evaluate black-box objective function \citep{jones1998efficient,shahriari2016taking}. Additionally, little effort is made to evaluate the quality of UQ for a trained GPR model, likely because the model makes predictions on samples within predefined design bounds and does not need to extrapolate much (low epistemic uncertainty). Other classical surrogate modeling methods, such as standard artificial neural networks and support vector machines, are generally less capable of quantifying predictive uncertainty, especially epistemic uncertainty. These methods and GPR are typically used to build surrogates that act as  ``deterministic” transfer functions and allow propagating aleatory uncertainty in input variables to derive the uncertainty in the model output, known as uncertainty propagation \citep{lee2009comparative}. The recent two years have seen efforts applying DNNs to surrogate modeling for reliability analysis \citep{chakraborty2020simulation, li2020deep, zhang2022simulation}. Similarly, these DNNs do not have built-in UQ capability and are typically used as deterministic functions primarily for uncertainty propagation. 

For over two decades, the prognostics and health management (PHM) community has used ML methods with built-in UQ capability as part of the health forecasting/RUL prediction process. Early applications include the Bayesian linear regression for aircraft turbofan engine prognostics \citep{coble2008prognostic}, the relevance vector machine, a probabilistic kernel regression model of an identical function form to the support vector machine \citep{tipping2001sparse}, for battery prognostics \citep{saha2008prognostics,wang2013prognostics,CHANG2022109166} and general purpose prognostics \citep{wang2012generic, hu2012ensemble}, and GPR for battery prognostics  \citep{liu2013prognostics, richardson2017gaussian, thelen2022augmented}. UQ of ML models for PHM is perceived to have more significance than that for engineering design, mainly due to (1) the more likely lack of sufficient training data, given an expensive and time-consuming process to collect run-to-failure data for training ML models for health prognostics, (2) the higher need to extrapolate to unseen operating conditions in PHM applications, and (3) the higher criticality of consequences from incorrectly made maintenance decisions. Two representative reviews of UQ work in the field of PHM can be found in \citep{ sankararaman2015significance, sankararaman2015uncertainty}. Both reviews seem to focus on identifying uncertainty sources in health prognostics and discussing ways to propagate these sources of uncertainty to derive the probability distribution of RUL.

\begin{figure}[!ht]
    \centering
    \includegraphics[scale=0.9]{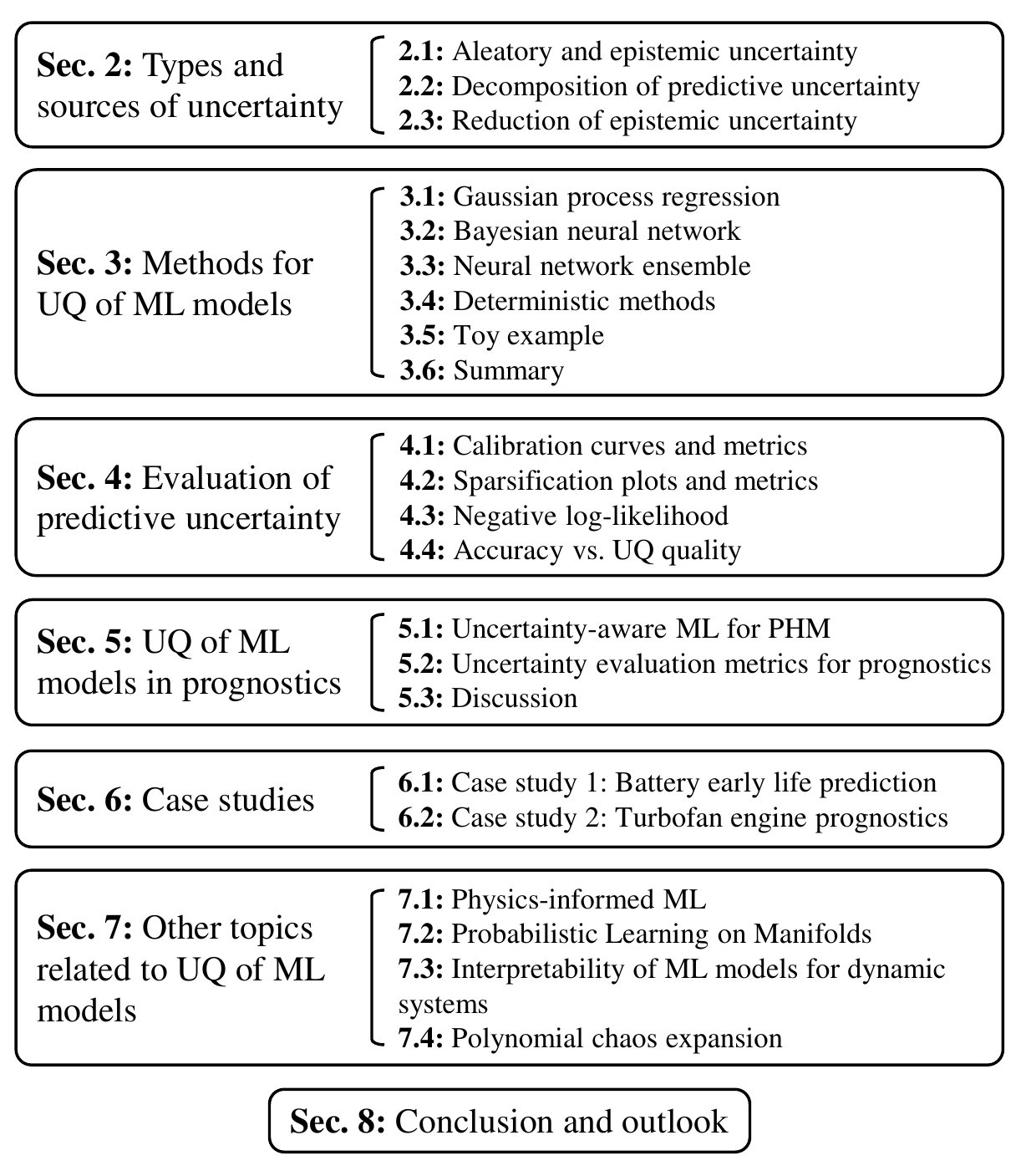}
    \caption{Overview of the organization of the tutorial paper.}
    \label{fig:outline}
\end{figure}

Within this paper, we seek to provide a comprehensive overview of emerging approaches for UQ of ML models and a brief review of applications of these approaches to solve engineering design and health prognostics problems. As for the ML models, our tutorial focuses on neural networks due to their increasing popularity amongst academic researchers and industrial practitioners. In essence, we look at methods to quantify the predictive uncertainty of neural networks, i.e., methods for UQ of neural networks. This focus differs from the notion of ``ML for UQ" where UQ of engineered systems or processes becomes the primary task, and ML models are built only to serve the primary purpose of UQ. Figure \ref{fig:outline} shows an outline of this tutorial paper. Our tutorial possesses four unique properties that distinguish it from recent reviews on UQ of ML models in the ML community \citep{abdar2021review, bhatt2021uncertainty,gawlikowski2021survey}, computational physics community~\citep{psaros2023uncertainty}, and PHM community~\citep{sankararaman2015significance,sankararaman2015uncertainty}.
\begin{itemize}
\item First, we give a detailed classification of uncertainty types, sources, and causes (Sec. \ref{sec:aleatory_epistemic}) and discuss ways to reduce epistemic uncertainty (Sec. \ref{sec:reduction}). Our classification and discussion complement the theoretical and data science-oriented discussions in the ML community and provide more context for researchers and practitioners in the engineering design and PHM communities. Additionally, we provide an easy-to-understand explanation of the process of decomposing the total predictive uncertainty of an ML model into aleatory and epistemic uncertainty, leveraging simple mathematical examples (Sec. \ref{sec:decomposition}).
\item Second, we provide a tutorial-style description and a qualitative and quantitative comparison of emerging UQ approaches developed in the ML community over the past eight years. This tutorial-style description covers both methodologies (Sec. \ref {sec:UQ_methods}) and their implementations on real-world case studies (Sec. \ref{sec:case_studies}). The tutorial style also applies to our discussion on methods and metrics for assessing the quality of predictive uncertainty (Sec. \ref{sec:uncertainty_evaluation}), an increasingly important exercise in UQ of ML models.
\item Third, although our tutorial focuses primarily on UQ methods for ML models, it additionally briefly covers a collection of recent studies that apply some of the emerging UQ approaches to solve challenging problems in engineering design (\ref {sec:UQ_design}) and health prognostics (Sec. \ref {sec:UQ_PHM}). This review is meaningful because as the adoption of ML techniques in design and prognostics rapidly increases, we also expect to see an increasing need for UQ of ML models. Note that deep neural network architectures, originally created for computer vision tasks based on large image datasets, can be readily adopted in engineering design tasks, such as surrogate modeling for reliability analysis \cite{bichon2008efficient,echard2011ak} and generative designs \cite{zhang20193d, chen2018b, chen2019synthesizing}, and PHM tasks, such as fault diagnostics \citep{he2017deep,hoang2019survey,lu2023physics,hou2022interpretable,sinitsin2022intelligent} and RUL prediction \citep{deutsch2017using,yu2019remaining,li2019deep}. We hope to provide observations and insights that can help guide researchers in the engineering design and PHM communities in choosing and implementing the UQ methods suitable for specific applications. This unique and distinct application area distinguishes our tutorial paper from a recent review paper on UQ of ML models~\citep{psaros2023uncertainty}, which explored the use of ML with UQ for solving partial differential equations and learning neural operators.
\item Fourth, we share, on GitHub, our code for implementing several UQ methods on one toy regression example (Sec. \ref{sec:toy_example}) and two real-world case studies on health prognostics (Sec. \ref{sec:case_studies}). Our implementations have been thoroughly verified to have quality on par with high quality implementations by the ML community. Some of our implementations are directly built on top of code shared by the ML community. We anticipate our code will allow researchers and practitioners in the engineering design and PHM communities to replicate results, customize existing UQ methods to specific applications, and test new methods. Moving forward, we plan to make continuous improvements to the codebase, e.g., by polishing lines of code and adding new methods as they become available. 
\end{itemize}

Our tutorial paper is concluded in Sec. \ref{sec:conclusion_outlook}, where we also discuss directions for future research.

\section{Types and sources of uncertainty}
\label{sec:types_sources_uncertainty}

This section first provides the definitions of different types of uncertainty and a summary of their sources and causes, and then discusses the methods to decompose and reduce the predictive uncertainty of ML models.

\subsection{Aleatory and epistemic uncertainty}
\label{sec:aleatory_epistemic}

Uncertainty, in general, can be classified into two types: aleatory uncertainty and epistemic uncertainty \cite{der2009aleatory}. This classification of uncertainty originated in the engineering domain for risk and reliability analysis \cite{der2009aleatory} and is also applicable to the ML domain \cite{kendall2017uncertainties,hullermeier2021aleatoric}. The definitions and sources of these two types of uncertainty are summarized as follows.
\begin{enumerate}[label=\roman*.]
        \item {\bf{Aleatory uncertainty}}: It stems from natural variability and is irreducible by nature \cite{der2009aleatory}. This type of uncertainty captures the noise inherent in physical systems \citep{gal2017concrete}. A typical example of aleatory uncertainty is the noise in sensor measurements, which would persist even if more data were collected. In ML, aleatory uncertainty represents the inherently stochastic nature of an input, an output, or the dependency between these two \cite{hullermeier2021aleatoric}. Example causes of aleatory uncertainty include variability of material properties from one specimen to another, variability of response from different runs of the same experiment, variability in classes for classification problems, and variability of the output for regression problems. This type of uncertainty is usually modeled as a part of the likelihood function in a probabilistic ML model. The predictions of the ML model is also probabilistically distributed \citep{gal2017concrete}. This way of capturing the observation uncertainty (sometimes termed \emph{data uncertainty}) is leveraged by several UQ methods, such as homoscedastic (Eq. (\ref{eq:GaussianObs})) and heteroscedastic (Eq. (\ref{eq:GaussianObsHetero})) GPRs discussed in Sec. \ref{sec:gpr} and neural network ensemble (Eq. (\ref{eq:GaussianObsHetero})) discussed in Sec. \ref{sec:neuralnetworkensemble}.

        \item {\bf{Epistemic uncertainty}}: This type of uncertainty is attributed to things one could know in principle but remain unknown in practice due to a lack of knowledge. It is reducible by nature \cite{der2009aleatory}. Common causes of epistemic uncertainty in the engineering domain include model simplification, model-form selection, computational assumptions, lack of information about certain model parameters, and numerical discretization. ML models generally have similar epistemic uncertainty sources as engineering models. In particular, the epistemic uncertainty in ML models can be further classified into the following two categories:
        \begin{enumerate}
            \item {\it {Model-form uncertainty}} is due to the simplification and approximation procedures involved in ML model construction. It is usually associated with the choices of model types, such as the architectures and activation functions of neural networks and the model forms of kernel functions in GPR models.  
            \item {\it {Parameter uncertainty}} is associated with model parameters and arises from the model calibration and training processes. Major causes of parameter uncertainty include a lack of enough training data, inherent bias in the training data due to low data fidelity, and difficulties in converging to optimal solutions faced by training algorithms. 
        \end{enumerate}
\end{enumerate}

Table \ref{tab:epistemic uncertainty} summarizes the common sources and associated causes of the above two types of uncertainty in ML. When the test dataset falls outside the training data distribution, the ML model predictions likely have high epistemic uncertainty since the performance of ML models is typically poorer in extrapolation than in interpolation. When the test data in some regions of the input space are associated with higher measurement noise, they can lead to higher aleatory uncertainty. Additionally, data of output used to train an ML model could deviate from the true values of the output. When the error is caused by random noise of measurement, it will lead to aleatory uncertainty in the output. However, when there is also bias in the data, the error causes additional epistemic uncertainty. For instance, when the bias is caused by low data fidelity representing the data's low accuracy, this bias will result in epistemic uncertainty, which is reducible by adding high-fidelity data for training. 

\begin{table*}[!ht]
\centering
\caption{Types, sources, and causes of uncertainty in ML}
\begin{tabular}{p{4cm}|p{4.5cm}|p{6cm}}
\hline \hline
\multicolumn{1}{l|}{\bf{Type}}               & \bf{Source} & \bf{Cause(s)} \\ \hline 
\multirow{3}{*}{Aleatory uncertainty}  & Observational uncertainty (model input and output)   & Measurement noise (e.g., sensor noise in measuring inputs/outputs of ML models)            \\ \cline{2-3} 
                                        & Natural variability (model input)   & Variability in material properties, manufacturing tolerance, variability in loading and environmental conditions, etc. \\ \cline{2-3} 
                                        & Lack of predictive power (model input)    &  Dimension reduction, non-separable classes in input space (classification), etc.     \\ \hline
\multirow{2}{*}{Epistemic uncertainty} & Parameter uncertainty   &  Limited training data, local optima of ML model parameters, low-fidelity training data*, etc.                    \\ \cline{2-3} 
                                        & Model-form uncertainty    & Choices of neural network architectures and activation and other functions, missing input features, etc. 
                                                             \\ \hline \hline
\end{tabular}

 \begin{tablenotes}
          \footnotesize   
          \item[*] * Data fidelity is the accuracy with which data quantifies and embodies the characteristics of the source \cite{fidelity2022}. \\
    \end{tablenotes}

\label{tab:epistemic uncertainty}
\end{table*}

Note that aleatory uncertainty could exist in the input, output, or both of an ML model. A common practice of dealing with aleatory uncertainty in the inputs is propagating the uncertainty to the output after constructing the ML model. The aleatory uncertainty in the output, however, is more challenging to tackle, since it needs to be accounted for during the training of an ML model (see more detailed discussion in Secs. \ref{sec:gpr} and \ref{sec:neuralnetworkensemble}). Uncertainty propagation of input aleatory uncertainty to the output is not the focus of this paper. We mainly focus on accounting for aleatory uncertainty in the output during the training of an ML model. Moreover, it is worth mentioning that aleatory uncertainty and epistemic uncertainty often coexist, making it difficult to separate them. Even though some efforts have been made in recent years to separate these two types of uncertainty, for example, by using the variance decomposition method (see Sec. \ref{sec:decomposition}) that has been extensively studied in the global sensitivity analysis field \cite{saltelli2004sensitivity,sobol1990sensitivity,sobol2001global}, a clean and complete separation of these two types of uncertainty may only be possible for some cases when there are no complicated interactions between aleatory and epistemic uncertainty sources. We are interested in separating these two types of uncertainty often because we are usually concerned about when the \enquote{prediction accuracy} of ML models becomes so low that model prediction cannot be trusted. These “break-down” cases are typically associated with high epistemic uncertainty, the quantification of which would help identify low-confidence predictions by the ML models and avoid making sub-optimal or even incorrect decisions whose consequences could be very costly and even life-threatening in safety-critical applications. 

Suppose we cannot separate these two types of uncertainty and only look at their combination. In that case, we only have access to the total predictive uncertainty of an ML model, which can be used to measure the model’s confidence in predicting at a test point, given both noise sources in the environment and the reducible uncertainty arising from a lack of training data. The total predictive uncertainty is often what commercially available ML solutions produce as ML outputs (e.g., the probability mass function of the predicted health class for health diagnostics and the variance of the remaining useful life estimate for health prognostics). 

\subsection{Decomposition of predictive uncertainty}
\label{sec:decomposition}
From the above discussion, we can intuitively and qualitatively tell the difference between aleatory (irreducible) and epistemic (reducible) uncertainty. Some recent studies also attempted to estimate these two types of uncertainty quantitatively. To this end, it is essential to decompose the total predictive uncertainty into aleatory and epistemic components \cite{gal2016uncertainty,depeweg2018decomposition,smith2018understanding}. Let us consider the simplest form of a probabilistic ML model, a linear regression model. This model is parameterized by weights and biases, concatenated into a vector $\bm{\uptheta}$. Then, we can express this linear regression model in the following form: 
\begin{equation}
    \hat y({\bf{x}}) = f({\bm{\mathrm{x}}};{\bm{\uptheta }}) = {{\bm{\uptheta }}^\text{T}}{\bf{x}} + \varepsilon ,
    \label{eq:linear}
\end{equation}
where $\varepsilon \sim \mathcal{N}\left( {{\bf{0}},{\sigma ^2}{\bf{I}}} \right)$ is the Gaussian noise variable with ${\bf{I}}$ denoting an ${D} \times {D}$ identity matrix. Note that applying an activation function to the linear term ${{\bm{\uptheta}}^\text{T}}{\bf{x}}$ introduces nonlinearity to the regression model, making it a building block in a neural network. 

If we make a Bayesian treatment of Eq. (\ref{eq:linear}), we will start with a prior distribution $p({\bm{\uptheta}})$ over model parameters ${\bm{\uptheta}}$ and then infer a posterior from a training dataset $\mathcal{D}$, $p({\bm{\uptheta}}|\mathcal{D})$. Essentially, we build a Bayesian linear regression model, from which we can derive the predictive distribution of $y$ at a given training/validation/test point $\bf{x}$ via marginalization:
\begin{equation}
    p(y|{\bf{x}},\mathcal{D}) = \int {p(y|{\bf{x}},{\bm{\uptheta}})p({\bm{\uptheta}}|\mathcal{D})d{\bm{\uptheta}}} .
    \label{eq:marginal}
\end{equation}

To make the discussion more concrete and easier to understand, we further assume that Eq. (\ref{eq:linear}) is a two-dimensional model (i.e., $D=2$) and the posterior of ${\bm{\uptheta}}$ is jointly Gaussian: $p({\bm{\uptheta}}|\mathcal{D}) = \mathcal{N}({{\bf{\mu }}_{\bm{\uptheta}}},\;{{\bf{\Sigma }}_{\bm{\uptheta}}})$ with ${{\bf{\mu }}_{\bm{\uptheta}}} = {[{\mu _{{\theta_1}}},{\mu _{\theta_2}}]^\text{T}}$ and a covariance matrix ${{\bf{\Sigma }}_{\bm{\uptheta}}} = \left[ \begin{matrix} {\sigma _{{\theta _1}}^2} & {\rho \sigma _{\theta _1}\sigma _{\theta _2}}  \cr 
   {\rho \sigma _{\theta _1}\sigma _{\theta _2}} & {\sigma _{{\theta _2}}^2}  \cr 
\end{matrix}
\right]$. The predicted $y$ then follows a Gaussian distribution given by:
\begin{equation}
   p(y|{\bf{x}},\mathcal{D}) = \mathcal{N}({\mu _{{\theta _1}}}{x_1} + {\mu _{{\theta _2}}}{x_2},\;\sigma _{{\theta _1}}^2x_1^2 + \sigma _{{\theta _2}}^2x_2^2 + 2\rho {\sigma _{{\theta _1}}}{\sigma _{{\theta _2}}}{x_1}{x_2} + {\sigma ^2}) .
    \label{eq:y_dis}
\end{equation}

For classification problems, we typically use {\it{differential entropy}} as a measure of uncertainty \cite{malinin2018predictive}; for regression problems, a typical choice is {\it{variance}} of a Gaussian output \citep{murphy2022probabilistic}. Since we deal with a regression problem, we use variance to measure uncertainty in this example. The total predictive uncertainty is measured as the predicted variance
\begin{equation}
  {\mathcal{U}_\text{total}} = Var(y|{\bf{x}},\mathcal{D})=\sigma _{{\theta _1}}^2x_1^2 + \sigma _{{\theta _2}}^2x_2^2 + 2\rho {\sigma _{{\theta _1}}}{\sigma _{{\theta _2}}}{x_1}{x_2} + {\sigma ^2} .
    \label{eq:total_U}
\end{equation}

The aleatory uncertainty can be measured as the variance of the Gaussian noise (intrinsic in the data)
\begin{equation}
{\mathcal{U}_\text{aleatory}} = {\sigma ^2}.
    \label{eq:aleatory}
\end{equation}

Then, the epistemic uncertainty can be estimated by subtracting the aleatory uncertainty from the total predictive uncertainty
\begin{equation}
{\mathcal{U}_\text{epistemic}} = {\mathcal{U}_\text{total}} - {\mathcal{U}_\text{aleatory}} = \sigma _{{\theta _1}}^2x_1^2 + \sigma _{{\theta _2}}^2x_2^2 + 2\rho {\sigma _{{\theta _1}}}{\sigma _{{\theta _2}}}{x_1}{x_2}.
    \label{eq:epistemic}
\end{equation}

It can be seen from the above equation that the epistemic  uncertainty is dependent on  (1) the posterior variances ($\sigma _{{\theta _1}}^2$ and $\sigma _{{\theta _2}}^2$) and covariance ($\rho {\sigma _{{\theta _1}}}{\sigma _{{\theta _2}}}$) of the model parameters ${\bm{\uptheta}}$ and (2) values of the input variables ($x_1$ and $x_2$). The noise variance, which measures the intrinsic uncertainty in the data, does not affect and has nothing to do with the epistemic uncertainty. 

Using the law of total variance or variance-based sensitivity analysis \cite{saltelli2010variance}, we can generalize Eqs. (\ref{eq:total_U}) through (\ref{eq:epistemic}) for uncertainty decomposition:
\begin{equation}
\underbrace {Var(y|{\bf{x}},\mathcal{D})}_{{\mathcal{U}_\text{total}}} = \underbrace {{\mathbb{E}_{{\bm{\uptheta}}\sim p({\bm{\uptheta}}|\mathcal{D})}}[Var(y|{\bf{x}},{\bm{\uptheta}})]}_{{\mathcal{U}_\text{aleatory}}} + \underbrace {Va{r_{{\bm{\uptheta}}\sim p({\bm{\uptheta}}|\mathcal{D})}}[\mathbb{E}(y|{\bf{x}},{\bm{\uptheta}})]}_{{\mathcal{U}_\text{epistemic}}},
    \label{eq:decomp}
\end{equation}
where ${\mathbb{E}(y|{\bf{x}},{\bm{\uptheta}})}$ and ${Var(y|{\bf{x}},{\bm{\uptheta}})}$ are the mean and variance of $y$ at $\bf{x}$ for a given realization of ${\bm{\uptheta}}$. The first term on the right-hand side of Eq. (\ref{eq:decomp}), ${{\mathbb{E}_{{\bm{\uptheta}}\sim p({\bm{\uptheta}}|\mathcal{D})}}[Var(y|{\bf{x}},{\bm{\uptheta}})]}$, computes the average of the variance of $y$, ${Var(y|{\bf{x}},{\bm{\uptheta}})}$, over ${p({\bm{\uptheta}}|\mathcal{D})}$. This term does not consider any contribution of parameter (${\bm{\uptheta}}$) uncertainty to the variance of $y$, as the expectation operation, ${{\mathbb{E}_{{\bm{\uptheta}}\sim p({\bm{\uptheta}}|\mathcal{D})}}[\cdot]}$, take out the contribution of the variation in $\bm{\uptheta}$. It only captures the intrinsic data noise ($\varepsilon $) and therefore represents the aleatory uncertainty. The second term, ${Va{r_{{\bm{\uptheta}}\sim p({\bm{\uptheta}}|\mathcal{D})}}[\mathbb{E}(y|{\bf{x}},{\bm{\uptheta}})]}$, computes the variance of ${\mathbb{E}(y|{\bf{x}},{\bm{\uptheta}})}$ for ${{\bm{\uptheta}}\sim p({\bm{\uptheta}}|\mathcal{D})}$. The expectation operation, ${\mathbb{E}(y|{\bf{x}},{\bm{\uptheta}})}$, essentially takes out the contribution by the data noise ($\varepsilon $). Therefore, this second term measures epistemic uncertainty. For classification problems, similar expressions can be derived for the uncertainty metric of differential entropy, as demonstrated in some earlier work (see, for example, \cite{gal2016uncertainty, depeweg2018decomposition, smith2018understanding}). 

\begin{figure}[!ht]
    \centering
    \includegraphics[scale=0.8]{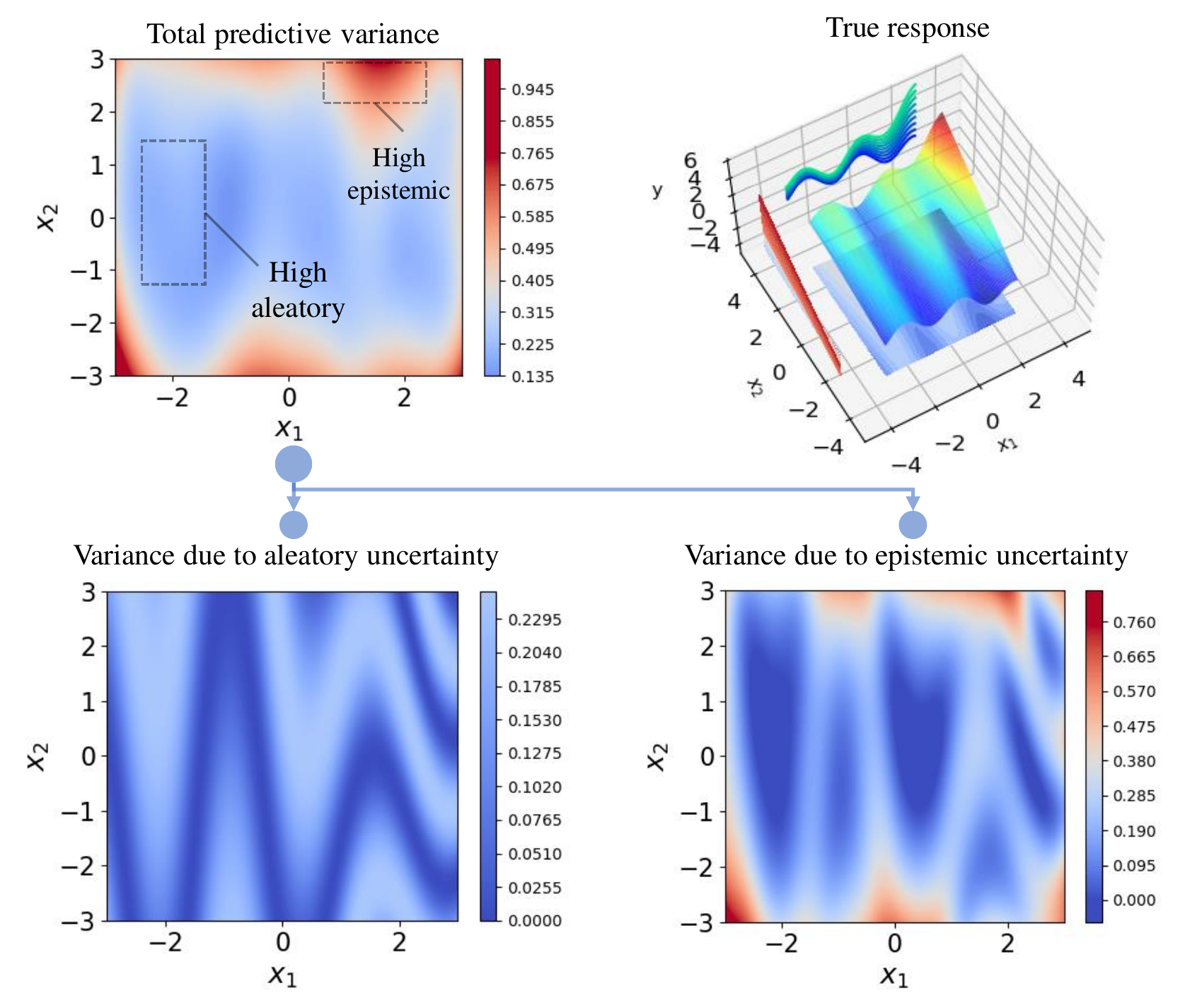}
    \caption{An example of uncertainty decomposition using variance-decomposition based method.}
    \label{fig:uncertainty_decomp}
\end{figure}

Figure \ref{fig:uncertainty_decomp} shows an example of uncertainty decomposition using the above variance decomposition method for a mathematical problem. The true model is a two-dimensional function as depicted in the top-right graph of Fig. \ref{fig:uncertainty_decomp} and this function has the following closed form: $y({\bf{x}}) = {\frac{1}{20}}({(1.5 + {x_1})^2} + 4) \times (1.5 + {x_2}) - \frac{\sin {{5 \times (1.5 + {x_1})} }}{2}$. In this example,  the true model is assumed to be unknown and needs to be learned from training data using an ML model. Due to inherent sensor noise, observational uncertainty is present in the output of the training data. It is modeled as a random variable following a Gaussian distribution as $\varepsilon({\bf{x}}) \sim \mathcal{N}(0,\; 0.5|\text{sin}(y({\bf{x}}))|^2)$. Based on 50 training samples, a GPR model is constructed. The total predicted variance of the resulting ML model is shown in the upper left graph of Fig. \ref{fig:uncertainty_decomp}. This graph shows that the predicted variance is high for some regions and low for others. Since both aleatory and epistemic uncertainty exists and only the total predictive uncertainty is visualized, it is difficult to tell if the uncertainty (the total predicted variance) in a certain region could be further reduced. 

Decomposing the total predicted variance into variances due to aleatory uncertainty and epistemic uncertainty, respectively, as shown in the lower half of this figure, allows us to identify regions with high aleatory uncertainty and those with high epistemic uncertainty. If a region with high epistemic uncertainty is the prediction region of interest, we can reduce the uncertainty to improve the prediction confidence of the ML model (see the uncertainty reduction methods in Sec. \ref{sec:reduction}). However, if a region with high aleatory uncertainty and low epistemic uncertainty is the prediction region of interest, it would be difficult to further reduce the total predictive uncertainty. In that case, risk-based decision making needs to be employed to account for the irreducible aleatory uncertainty when deriving optimal decisions (see, for example, decision-making scenarios in engineering design, as discussed in \ref{sec:UQ_design}, and in PHM, as discussed in Sec. \ref{sec:UQ_PHM}).

\subsection{Reduction of epistemic uncertainty}
\label{sec:reduction}
As mentioned in Sec. \ref{sec:aleatory_epistemic}, epistemic uncertainty is reducible. Suppose an ML model has low prediction accuracy and confidence due to high epistemic uncertainty, resulting in sub-optimal or even incorrect decisions. In that case, it is necessary to reduce the epistemic uncertainty. Commonly used strategies for the reduction of epistemic uncertainty can be roughly divided into the following two groups according to the source of epistemic uncertainty of interest.
\begin{enumerate}[label=(\alph*)]
    \item{{\bf{Reducing parameter uncertainty}}}
    \begin{enumerate}[label=\roman*]
    \item {\bf{Adding more training data}}: Having access to limited training data usually leads to uncertainty in ML model parameters. The model-parameter uncertainty is part of epistemic uncertainty. It can be reduced by increasing the training data size, e.g., via data augmentation using physics-based models \cite{shorten2019survey} or simply by collecting and adding more experimental data to the training set. Let us assume the added training data is as clean as the existing data. In that case, the epistemic uncertainty component of the predictive uncertainty becomes smaller, while the aleatory uncertainty is expected to remain at a similar level. Suppose that, in a different case, the added training data contains more noise than the existing data. In that case, we still expect lower epistemic uncertainty in regions of the input space where the added data lie but higher aleatory uncertainty in these regions.
    
    \item {\bf{Adding physics-informed loss or physical constraints for ML model training}}: Incorporating physical laws as new loss terms or imposing physical constraints, such as boundedness, monotonicity, and convexity for interpretable latent variables for ML model training, may allow us to obtain a more accurate estimate of ML model parameters. Although this physics-informed/constrained ML approach may not directly reduce epistemic uncertainty in ML predictions, it helps to reduce the training data size required to build a robust ML model that produces accurate predictions across a wide range of input settings. Specifically, enforcing principled physical laws into an ML model considerably prunes the search space of model parameters as parameters violating these constraints are discarded immediately. As a result, physical constraints contribute to reducing parameter uncertainty to some extent by complementing the insufficient training data and narrowing down the feasible region of these parameters. This benefit becomes especially relevant when training data is lacking and has been reported in recent review papers in various engineering fields, such as computational physics  \cite{karniadakis2021physics}, digital twin \cite{thelen2022comprehensive}, and reliability engineering \cite{xu2022physics}, and in research papers published in recent special issue collections on health diagnostics/prognostics \cite{hu2023special} and the broader topic of reliability and safety \cite{pinn_si}. For over-parameterized ML models such as neural networks, it is possible to simultaneously reduce bias and variance in the model parameters \cite{malashkhia2022physics}. For simpler models such as GPR, utilizing additional information such as gradient information \cite{deng2020multifidelity}, orthogonality \cite{plumlee2018orthogonal}, and monotonicity \cite{tran2023monotonic} as constraints in kernel construction can also improve the prediction accuracy. 

    \item {\bf{Adopting better strategies for ML model training}}: If a better starting point can be used when training an ML model, the optimization process may yield a more accurate estimate of the model parameters. Similar to adding physics-informed loss terms, this strategy can also indirectly reduce epistemic uncertainty. A popular example of this strategy is transfer learning, where the model trained in one domain is used as a starting point for training a model in another domain (e.g., transfer of weights and biases in selected neural network layers) \cite{raghu2019transfusion}. Another strategy is to use better optimization algorithms when the number of parameters to be optimized is large. Global optimization in high-dimensional search spaces is always challenging. Algorithms such as stochastic gradient descent can have better convergence than traditional quasi-Newton methods in training deep neural networks \cite{bottou2012stochastic}. Reformulating model training with multiple loss terms as minimax problems to adjust the focus of different loss terms can also improve convergence \cite{liu2021dual}.
    \end{enumerate}
    \item{{\bf{Reducing model-form uncertainty}}}
    \begin{enumerate}[label=\roman*]
        \item {\bf{Identifying better input features}}: In practical applications, an important step in training ML models is the selection of input features with strong predictive power according to domain knowledge, expert opinions, or exploratory analysis \cite{cai2018feature,chandrashekar2014survey}. Identifying input features with higher predictive power and using them as input features allows us to reduce the model-form uncertainty of ML models.
        \item {\bf{Choosing better model architecture/kernel functions}}: All models are wrong, but some are useful \cite{box1979all}. An appropriately chosen model architecture can better approximate the true underlying function than many other model architectures. A commonly used method is, therefore, to choose better model architecture or kernel function through tuning or model validation. It can reduce model-form uncertainty to some degree. 
        \item {\bf{Adding high-fidelity data}}: An obvious way to reduce model-form uncertainty caused by bias in the training data is by adding high-fidelity data, thereby reducing the overall epistemic uncertainty. Such strategies have been widely adopted in the ML field in the context of multi-fidelity surrogate modeling/ML \cite{tran2020multi,pilania2017multi,liu2019multi,liu2023multi} and transfer learning \cite{huang2022transfer}.
    \end{enumerate}
\end{enumerate}

\begin{figure}[!ht]
    \centering
    \includegraphics[scale=0.62]{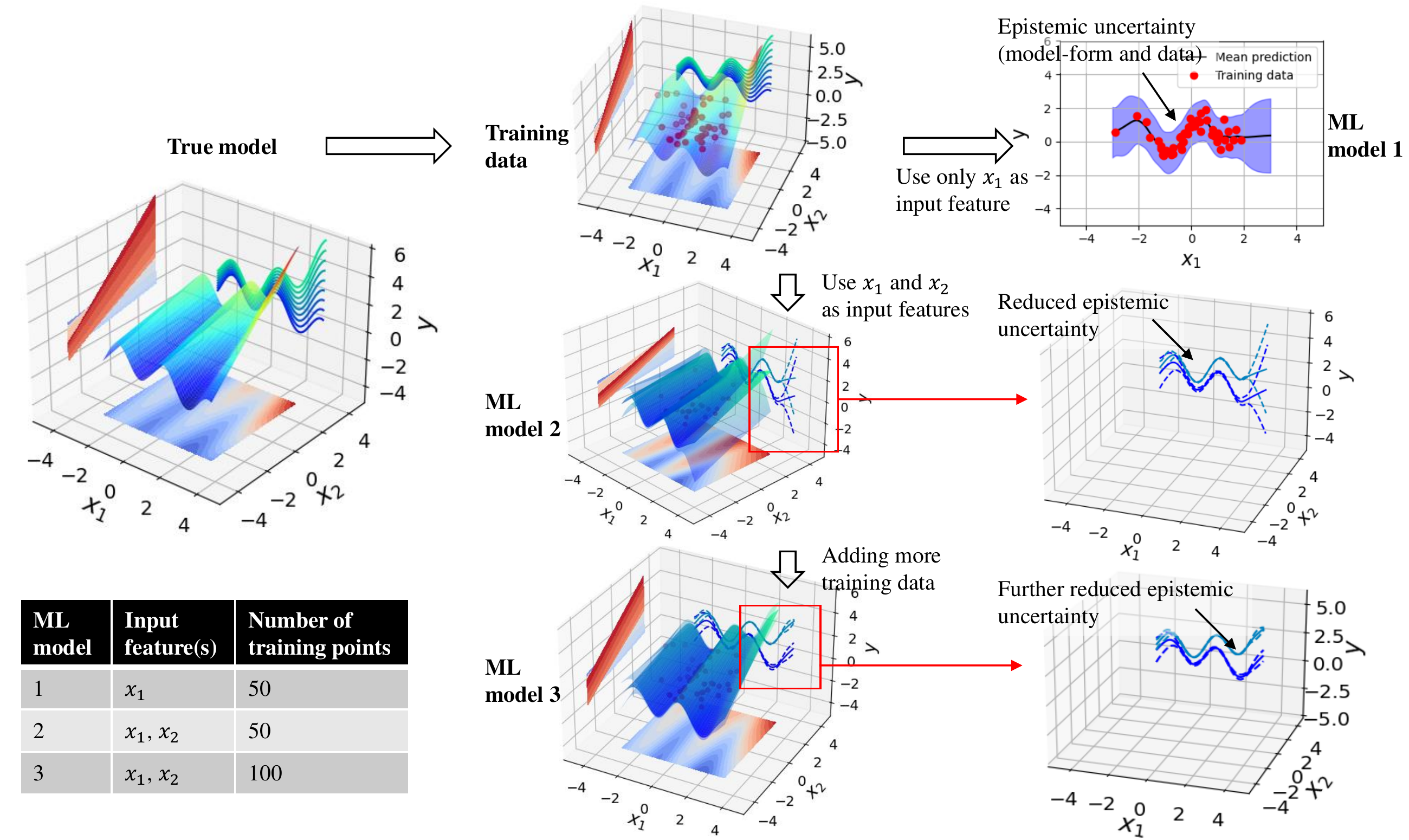}
    \caption{Types of uncertainty sources in ML models and the process of reducing epistemic uncertainty (i.e., methods (b).i and (a).i described in Sec. \ref{sec:reduction}).}
    \label{fig:uncertainty}
\end{figure}

Next, we use the two-dimensional example given in Fig. \ref{fig:uncertainty_decomp} to illustrate the process of reducing epistemic uncertainty. As shown in Fig. \ref{fig:uncertainty}, a group of training points is first generated from a known mathematical function. Then, an ML model with only $x_1$ as the input feature is constructed based on this group of training data. As shown in this figure, the resulting ML model (i.e., {\it ML model 1}) has considerable epistemic uncertainty due to the combined effect of model-form uncertainty and model-parameter uncertainty. In particular, the model-form uncertainty is caused by the fact that the underlying model used to generate this dataset has two input variables ($x_1$ and $x_2$) while ML model 1 only uses $x_1$ as its input feature. Model-parameter uncertainty stems from the limited number of training samples (i.e., 50 in this example). In order to reduce the epistemic uncertainty (model-form uncertainty), we then include both $x_1$ and $x_2$ as the input features, and another ML model labeled {\it ML model 2} is constructed using the same group of training data. As illustrated in Fig. \ref{fig:uncertainty}, adding input feature $x_2$ (i.e., strategy (b).i as described above) substantially reduces the epistemic uncertainty in regions within the training sample distribution. If we increase the size of the training data to $100$ (i.e., strategy (a).i), a third ML model ({\it ML model 3}) can be built based on this larger training dataset. As expected, the epistemic component of the predictive uncertainty is shown to decrease further due to the reduction of model-parameter uncertainty.

\section{Methods for UQ of ML models}
\label{sec:UQ_methods}

Data-driven ML models, most notably neural networks, have demonstrated unprecedented performance in establishing associations and correlations from large volumes of data in high-dimensional space via multiple layers of neurons and activation functions stacked together~\cite{lecun2015deep}. While ML has progressed on a fast track, it is still far away from fulfilling the stringent conditions of mission-critical applications~\cite{begoli2019need,zhang2022towards}, such as medical diagnostics, self-driving, and health prognostics of critical infrastructures, where safety and correctness concerns are salient. In addition to safety and reliability concerns, we are only able to collect a limited amount of data to train an ML model in a broad range of applications due to practical constraints on physical experiments and computational resources. To address some of these challenges, it is of paramount importance to establish principled and formal UQ approaches so that we can quantitatively analyze the uncertainty in ML model predictions arising from scarce and noisy training data as well as model parameters and structures in a sound manner. Accurate quantification of uncertainty in ML model predictions substantially facilitates the risk management of ML models in high-stakes decision-making environments~\cite{he2016deep,zhang2020bayesian_DSS,zhang2022explainable,cheng2020quantifying}.

\begin{figure}[!ht]
    \centering
    \includegraphics[scale=0.55]{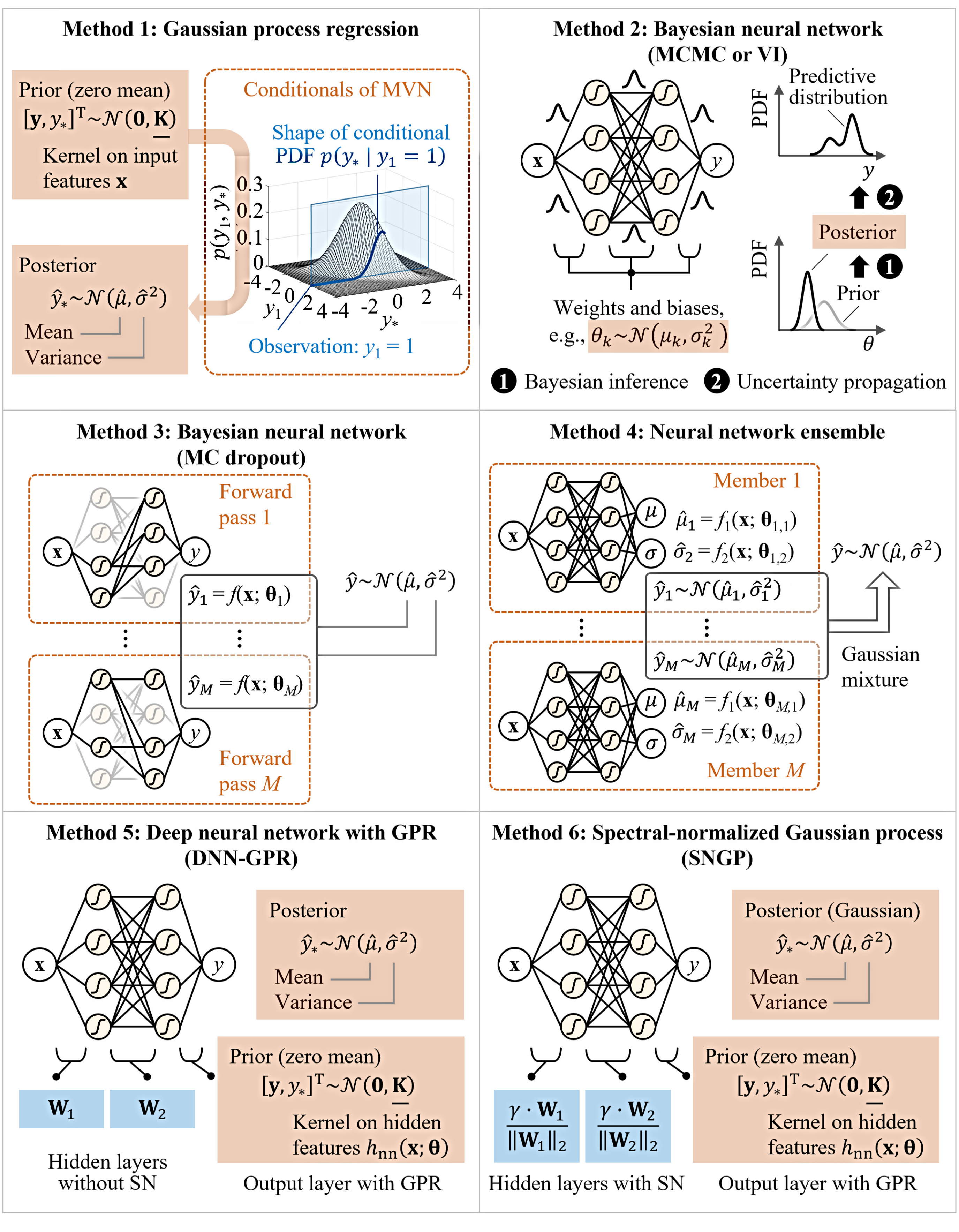}
    \caption{Graphical comparison of six state-of-the-art UQ methods introduced in Sec. \ref{sec:UQ_methods}. These methods are GPR (method 1), BNN via MCMC or VI (method 2), BNN via MC dropout (method 3), neural network ensemble (method 4), DNN with GPR -- DNN-GPR (method 5), and SNGP (method 6). In method 1, MVN standards for the multivariate normal distribution, or equivalently, the multivariate Gaussian distribution used in the main text. In methods (5) and (6), SN stands for spectral normalization.} 
    \label{fig:UQ_methods_schematic}
\end{figure}

In particular, when dealing with input samples in the region of input space with low signal-to-noise ratios or when handling the so-called OOD samples (input points sampled from a distribution very different from the training distribution), most ML models are prone to produce erroneous predictions~\cite{maartensson2020reliability}. If the uncertainty of an ML model can be quantified appropriately, it could lead to more principled decision making by enabling ML models to automatically detect samples for which there is high uncertainty. In fact, principled ML models are expected to yield high uncertainty (low confidence) in their predictions when the ML model predictions are likely to be wrong~\cite{tagasovska2019single,osawa2019practical}. Having uncertainty estimates that appropriately reflect the correctness of predictions is essential to identifying these “difficult-to-predict” samples that need to be examined cautiously, possibly with the eyes of a domain expert. This section provides a detailed, tutorial-style introduction of state-of-the-art methods for estimating the predictive uncertainty of data-driven ML models. As graphically summarized in Fig. \ref{fig:UQ_methods_schematic}, these UQ methods are GPR (Sec. \ref{sec:gpr}), Bayesian neural network (BNN) (Sec. \ref{sec:bnn}), neural network ensemble (Sec. \ref{sec:neuralnetworkensemble}), and deterministic methods focusing on SNGP (Sec. \ref{sec:deterministic_uq}).

\subsection{Gaussian process regression}
\label{sec:gpr}

GPR can be viewed as a generalized Bayesian inference, extending from an inference about a finite set of random variables to an inference about functions (each being an infinite-dimensional vector of random variables) ~\citep{rasmussen2006gaussian}. This generalized Bayesian inference works with a joint probability distribution of a random function (i.e., an infinite-dimensional random vector) rather than a joint distribution of a finite-dimensional random vector. Comprehensive and critical reviews are provided by Rasmussen~\cite{rasmussen2006gaussian}, Brochu et al.~\cite{brochu2010tutorial}, and Shahriari et al.~\cite{shahriari2016taking}. For complete details about GPR, readers are referred to the seminal textbook by Rasmussen~\cite{rasmussen2006gaussian}.

\subsubsection{Basics of Gaussian process regression}
\label{sec:basics_gpr}

\paragraph{a. Introduction to Gaussian process and Gaussian process prior} \mbox{}

A Gaussian process is a collection of random variables over some domain, where any finite subset of these variables follows a joint (multivariate) Gaussian distribution. Intuitively, the Gaussian process also defines a probability distribution for an unknown function, and this function comprises a collection of (infinitely many) random variables. Let $f(\mathbf{x})$ be the unknown function, 
where $\mathbf{x} \in \mathbb{R}^D$ is a $D$-dimensional input, 
then for any finite set of input ($\mathbf{x}$) points of this function, for example, 
$\mathbf{X}_\mathrm{t} = [\mathbf{x}_1, \dots, \mathbf{x}_{N}]^\mathrm{T}  \in \mathbb{R}^{N \times D}$, their corresponding outputs 
$f(\mathbf{X}_\mathrm{t}) = [f(\mathbf{x}_1), \dots, f(\mathbf{x}_N)]^\mathrm{T} \in \mathbb{R}^{N}$ follow a joint Gaussian distribution. 

GPR starts from a Gaussian process prior for the unknown function: $f(\mathbf{x}) \sim \mathcal{GP}(m(\mathbf{x}),k(\mathbf{x}, \mathbf{x}'))$ \citep{rasmussen2006gaussian}. This Gaussian process prior is fully characterized by a (prior) mean function $m(\mathbf{x}): \mathbb{R}^D \mapsto \mathbb{R}$ and a (prior) covariance function $k(\mathbf{x}, \mathbf{x}'):\mathbb{R}^D \times \mathbb{R}^D \mapsto \mathbb{R}$. 
The mean function $m(\mathbf{x})$ defines the prior mean of $f$ at any given input point $\mathbf{x}$, i.e.,
\begin{equation}
m(\mathbf{x}) = \mathbb{E}[f(\mathbf{x})].
\label{eq:priormean}
\end{equation}
The prior mean of the Gaussian process is often set as zero everywhere, $m(\mathbf{x}) = 0$, for the ease of computing the posterior. If the prior mean is a non-zero function, a trick is subtracting the prior means from the observations and function means (which we want to predict), thereby maintaining the ``zero-mean" condition. The covariance function $k(\mathbf{x}, \mathbf{x}')$, also called the \emph{kernel} in GPR, captures how the function values at two input points, $\mathbf{x}$ and $\mathbf{x}'$, linearly depend on each other. It takes the following form 
\begin{equation}
k(\mathbf{x},\mathbf{x}') = \mathbb{E}\left[ \left(f(\mathbf{x}) - m(\mathbf{x})\right) \left(f(\mathbf{x}') - m(\mathbf{x}')\right) \right].
\label{eq:priorcov}
\end{equation}

When the prior mean is zero, the kernel fully defines the shape (e.g., smoothness and patterns) of functions sampled from the prior and posterior. 

\paragraph{b. Kernel (covariance function)} \mbox{}

Probably the most commonly used kernel is the squared exponential kernel (a.k.a. the radial basis function kernel and the Gaussian kernel), defined as
\begin{equation}
k(\mathbf{x}, \mathbf{x}') = \sigma_f^2 \exp{\left(-\frac{\|\mathbf{x}-
\mathbf{x}'\|^2}{2l^2} \right)}.
\label{eq:sekernel}
\end{equation}
where the two kernel parameters, or two hyperparameters of the GPR model, are the signal amplitude $\sigma_f$ ($\sigma_f^2$ is called signal variance) and length scale $l$. $\sigma_f^2$ sets the upper limit of the prior variance and covariance and should take a large value if $f(\mathbf{x})$ spans a large range vertically (along the y-axis). It can be observed that the covariance between $f(\mathbf{x})$ and $f(\mathbf{x}')$ decreases as $\mathbf{x}$ and $\mathbf{x}'$ get farther apart. When $\mathbf{x}$ is extremely far from $\mathbf{x}'$, they have a very large Euclidian distance, and thus, $k(\mathbf{x}, \mathbf{x}') \approx 0$, i.e., the covariance between their function values approaches 0. Therefore, when predicting $f$ at a new input point, observations far away in the input space will have a minimum influence. When a new input is OOD, it has a very low covariance with any training point, meaning that the training observations contribute minimally to reducing the prior variance of the function value at the OOD point, leading to high epistemic uncertainty. This kernel-enabled characteristic has important implications for the \emph{distance awareness} property of GPR. On the other extreme, if two input points are extremely close, i.e., $\mathbf{x} \approx \mathbf{x}'$, then $k(\mathbf{x}, \mathbf{x}')$ becomes very close to its maximum, meaning $f(\mathbf{x})$ and $f(\mathbf{x}')$ have an almost perfect correlation. Function values of neighbors being highly correlated ensures smoothness in the GPR model, which is desirable because we often want to fit smooth functions to data.

The squared exponential kernel in Eq. (\ref{eq:sekernel}) uses the same length scale $l$ across all $D$ dimensions. An alternative approach is to assign a different length scale $l_d$ for each input dimension $x_d$, known as \emph{automatic relevance determination (ARD)} \citep{neal2012bayesian}. The resulting \emph{ARD squared exponential kernel} takes the following form
\begin{equation}
k(\mathbf{x}, \mathbf{x}') = \sigma_f^2 \exp{\left(-\frac{1}{2} \sum_{d=1}^{D}\frac{\left(x_d-
x_d'\right)^2}{l_d^2}\right)},
\label{eq:ardsekernel}
\end{equation}
where the $(D + 1)$ kernal parameters are the $D$ length scales, $l_1, \dots, l_D$, and the signal amplitude, $\sigma_f$. The ARD squared exponential kernel is also known as the \emph{anisotropic variant} of the (isotropic) squared exponential kernel. Each length scale determines how relevant an input variable is to the GPR model. If $l_d$ is learned to take a very large value, the corresponding input dimension $x_d$ is deemed irrelevant and contributes minimally to the regression. It is worth noting that  the squared exponential kernel is a special case of a more general class  of kernels called Mat\'ern kernels. See \ref{sec:gpr_kernels_extended} for an extended discussion of kernels.

\paragraph{c. Drawing random sample functions} \mbox{}

After defining a mean and a covariance function (kernel), we can draw sample functions from the Gaussian process prior without any observations of the function output. We can also sample function values from the Gaussian process posterior (i.e., the conditional Gaussian process conditioned on observed data), an essential task in GPR. Let us look at sampling functions from a Gaussian prior; a similar process can be followed to draw samples from a Gaussian process posterior. It is practically impossible to generate a perfectly continuous function from the prior, simply because this continuous function theoretically consists of an infinitely sized vector, which is not possible to sample. Alternatively, we can sample function values at a finite, densely populated set of input points and use these function values to reasonably approximate the continuous function. This approximation is acceptable in practice, given that we only need to predict $f$ at a finite set of input points. Since a Gaussian process entails this finite collection of random variables (i.e., the $f$ values at the finite set of input points) follow a multivariate Gaussian distribution, we can conveniently sample the function values from multivariate Gaussian.

Suppose we wish to sample function values at $N_\mathrm{*}$ input points, $\mathbf{x}_1^{*}, \dots, \mathbf{x}_{N_\mathrm{*}}^{*}$, from the prior. These input points could become new, unseen test points in a regression setting, and we use a subscript/superscript asterisk to distinguish them from training points. We start by defining an $N_\mathrm{*}$-by-$D$ matrix $\mathbf{X}_*$ where each row contains an input point, i.e., $\mathbf{X}_* = [\mathbf{x}_1^{*}, \dots, \mathbf{x}_{N_*}^{*}]^\mathrm{T}$. For simplicity, we assume the multivariate Gaussian prior has zero means ($m(\mathbf{x}) = 0$), so we only need to obtain the covariances between the function values at these $N_\mathrm{*}$ input points. Using the squared exponential kernel, we can derive the following covariance matrix
\begin{equation}\label{eq:cov_matrix}
\mathbf{K}_{\mathbf{X_{*}},\mathbf{X_{*}}} = \left[{\begin{array}{*{20}{c}}
{k(\mathbf{x}_1^{*}, \mathbf{x}_1^{*})} &{\cdots} &{k(\mathbf{x}_1^{*}, \mathbf{x}_{N_\mathrm{*}}^{*})}\\
{ \vdots} & {\ddots} &{\vdots}\\
{k(\mathbf{x}_{N_\mathrm{*}}^{*}, \mathbf{x}_1^{*})} &{\cdots} &{k(\mathbf{x}_{N_\mathrm{*}}^{*}, \mathbf{x}_{N_\mathrm{*}}^{*})}\\
\end{array}}\right].
\end{equation}

\begin{figure}[!ht]
    \centering
    \includegraphics[scale=0.86]{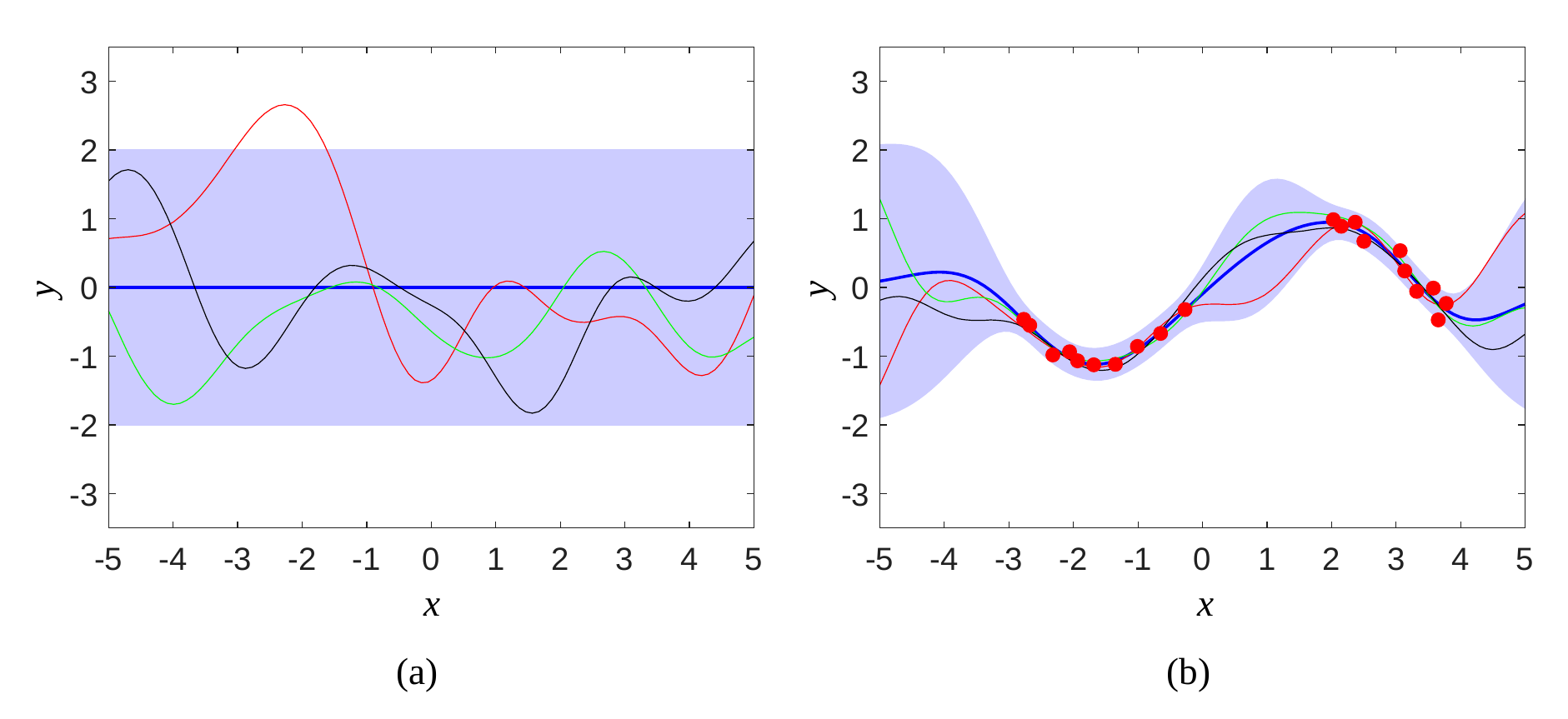}
    \caption{Sample functions drawn a Gaussian process prior (a) and posterior (b). The GPR model uses the squared exponential kernel with a length scale ($l$) of 1 and a signal amplitude ($\sigma_\mathrm{f}$) of 1, and a Gaussian observation model with a noise standard deviation ($\sigma_{\varepsilon}$) of 0.1. The means are shown collectively as a solid blue line/curve, and $\sim$95\% confidence intervals (means plus and minus two standard deviations) are shown collectively as a light blue shaded area. 20 training observations are generated by corrupting a sine function with a white Gaussian noise term, $y = \mathrm{sin}(0.9x) + \varepsilon$ with $\varepsilon \sim \mathcal{N} \left(0, 0.1^2 \right)$; these observations are shown as red dots.}
    \label{fig:samples_prior_posterior}
\end{figure}

Now we can draw random samples of the function values at the $N_\mathrm{*}$ input points $\mathbf{X}$ from $\mathcal{GP}(\mathbf{0},k(\mathbf{x}, \mathbf{x}'))$ by sampling from the following multivariate Gaussian distribution: $\mathbf{f}_* \sim \mathcal{N}(\mathbf{0},\mathbf{K}_{\mathbf{X}_*,\mathbf{X}_*})$. Each sample ($\mathbf{f}_*$) consists of ${N_*}$ function values, i.e., $\mathbf{f}_* = f(\mathbf{X}_*) = [f(\mathbf{x}_1^*), \dots, f(\mathbf{x}_{N_*}^*)]^\mathrm{T}$. The most commonly used numerical procedure to sample from a multivariate Gaussian distribution consists of two steps: (1) generate random samples (vectors) from the multivariate ($D$-dimensional) standard normal distribution, $\mathcal{N}(\mathbf{0},\mathbf{I})$, and (2) transform these random samples linearly based on the mean vector of the target multivariate Gaussian and the Cholesky decomposition of its covariance matrix (see further details in Sec. A.2 (Gaussian Identities) of Ref. \citep{rasmussen2006gaussian}). Figure \ref{fig:samples_prior_posterior}(a) shows three sample functions randomly drawn from a Gaussian process prior.

\paragraph{d. Making predictions at new points} \mbox{}
 
In practice, often, we only have noisy observations of $f(\mathbf{x})$, for example, through the following Gaussian observation model:
\begin{equation}
y = f(\mathbf{x}) + \varepsilon,
\label{eq:GaussianObs}
\end{equation}
where $\varepsilon$ is a zero-mean Gaussian noise, i.e., $\varepsilon \sim \mathcal{N}(0,\sigma_{\varepsilon}^2)$. The above additive Gaussian form will also be commonly used for other UQ methods in the upcoming sections. The $N$ noisy observations can be conveniently written in a vector form: $\mathbf{y}_\mathrm{t} = [y_1, \dots, y_N]^\mathrm{T} \in \mathbb{R}^{N}$. Note that these observations are sometimes called \emph{targets} in a regression setting. In GPR, we want to infer the input ($\mathbf{x}$) - target ($y$) relationship from the noisy observations; we may also be interested in learning the input ($\mathbf{x}$) - output ($f$) relationship in some cases. 

The Gaussian observation model in Eq. (\ref{eq:GaussianObs}) portrays an observation as two components: a \emph{signal term} and a \emph{noise term}. The signal term $f(\mathbf{x})$ carries the \emph{epistemic uncertainty} 
(see Sec. \ref{sec:aleatory_epistemic}) about $f(\mathbf{x})$, which can be reduced with additional observations of $f$ at a finite set of training points (e.g., $\mathbf{x}_1, \dots, \mathbf{x}_{N}$). The noise term $\varepsilon$ represents the inherent mismatch between signal and observation (e.g., due to measurement noise; see Table \ref{tab:epistemic uncertainty}), which is a type of \emph{aleatory uncertainty} (see Sec. \ref{sec:aleatory_epistemic}) and cannot be reduced from additional observations. In some cases, observations may be noise-free, corresponding to a special case where $\sigma_{\varepsilon} = 0$. In other words, we have access to the true function ($f$) output in these cases.

Now it is time to look at how to make predictions of function values $\mathbf{f}_*$ for $N_*$ new, unseen input points $\mathbf{X}_*$, given a collection of training observations, $\mathcal{D} = \left\{ {\left( {{\mathbf{x}_1},{y_1}} \right),\left( {{\mathbf{x}_2},{y_2}} \right), \dots, \left( {{\mathbf{x}_N},{y_N}} \right)} \right\}$, equivalently expressed as $\mathcal{D} = \left\{\mathbf{X}_\mathrm{t}, \mathbf{y}_\mathrm{t}\right\}$. These predictions can be made by drawing samples from the Gaussian process posterior, $p(f|\mathcal{D})$. We denote the function values at the training inputs as $\mathbf{f_{\mathrm{t}}} = f(\mathbf{X_{\mathrm{t}}}) = [f(\mathbf{x}_1), \dots, f(\mathbf{x}_N)]^\mathrm{T}$. Again, according to the definition of a Gaussian process, the function values at the training inputs and those at the new inputs are jointly Gaussian (prior without using observations), written as
\begin{equation}\label{eq:joint_Gaussian_prior}
\left[ \begin{array}{l}
\mathbf{f}_{\mathrm{t}}\\
\mathbf{f}_*
\end{array} \right] \sim
\mathcal{N}\left(\mathbf{0},\left[ {\begin{array}{*{20}{c}}
{\mathbf{K}_{\mathbf{X}_{\mathrm{t}},\mathbf{X}_{\mathrm{t}}}}&{\mathbf{K}_{\mathbf{X}_{\mathrm{t}},\mathbf{X}_*}}\\
{\mathbf{K}_{\mathbf{X}_*,\mathbf{X}_{\mathrm{t}}}}&{\mathbf{K}_{\mathbf{X}_*,\mathbf{X}_*}}
\end{array}} \right] \right), 
\end{equation}
where $\mathbf{K}_{\mathbf{X}_{\mathrm{ t }},\mathbf{X}_{\mathrm{t}}}$ is the covariance matrix between the $f$ values at the training points, expressed by simply replacing $\mathbf{X}$ in Eq. (\ref{eq:cov_matrix}) with $\mathbf{X}_{\mathrm{t}}$, $\mathbf{K}_{\mathbf{X}_{\mathrm{t}},\mathbf{X}_*}$ is the covariance matrix between the training points and new points (also called the cross-covariance matrix), $\mathbf{K}_{\mathbf{X}_*,\mathbf{X}_{\mathrm{t}}} = \mathbf{K}_{\mathbf{X}_{\mathrm{t}},\mathbf{X}_*}^\mathrm{T}$, and $\mathbf{K}_{\mathbf{X}_*,\mathbf{X}_*}$ is the covariance matrix between the new points. 

As shown in the Gaussian observation model in Eq. (\ref{eq:GaussianObs}), we assume all observations contain an additive independent and identically distributed (i.i.d.) Gaussian noise with zero mean and variance $\sigma_{\varepsilon}^2$. Under this assumption, the covariance matrix for the training \emph{observations} needs the addition of the noise variance to each diagonal element, i.e., $\mathbf{y}_{\mathrm{t}} \sim \mathcal{N}(\mathbf{0},\mathbf{K}_{\mathbf{X}_{\mathrm{t}},\mathbf{X}_{\mathrm{t}}}+\sigma_{\varepsilon}^2\mathbf{I})$, where $\mathbf{I}$ denotes the identity matrix of size $N$ whose diagonal elements are ones and off-diagonal elements are zeros. It then follows that the training observations (known) and the function values at the new input points (unknown) follow a slightly revised version of the multivariate Gaussian prior shown in Eq. (\ref{eq:joint_Gaussian_prior}), expressed as 
\begin{equation}\label{eq:joint_Gaussian_prior_noisy}
\left[ \begin{array}{l}
\mathbf{y}_{\mathrm{t}}\\
\mathbf{f}_*
\end{array} \right] \sim
\mathcal{N}\left(0,\left[ {\begin{array}{*{20}{c}}
{\mathbf{K}_{\mathbf{X}_{\mathrm{t}},\mathbf{X}_{\mathrm{t}}}+\sigma_{\varepsilon}^2\mathbf{I}}&{\mathbf{K}_{\mathbf{X}_{\mathrm{t}},\mathbf{X}_*}}\\
{\mathbf{K}_{\mathbf{X}_*,\mathbf{X}_{\mathrm{t}}}}&{\mathbf{K}_{\mathbf{X}_*,\mathbf{X}_*}}
\end{array}} \right] \right). 
\end{equation}

Now we want to ask the following question: ``given the training dataset $\mathcal{D}$ and new test points $\mathbf{X}_*$, what is the posterior distribution of the new, unobserved function values  $\mathbf{f}_*$?”. It has been shown that conditionals of a multivariate Gaussian are also multivariate Gaussian (see, for example, Sec. 3.2.3 of the probabilistic ML book \citep{murphy2022probabilistic}). Therefore, the posterior distribution $p(\mathbf{f}_*|\mathcal{D}, \mathbf{X}_*)$ is multivariate Gaussian. The posterior mean $\mathbf{\overline f}_*$ and covariance $cov(\mathbf{f}_*)$ can be derived based on the well-known formulae for conditional distributions of multivariate Gaussian, leading to the following:
\begin{equation}
\label{eq:posteriorMeanWithZeroMean}
\mathbf{\overline f}_* = \mathbf{K}_{\mathbf{X}_{\mathrm{t}},\mathbf{X}_*}^\mathrm{T} (\mathbf{K}_{\mathbf{X}_{\mathrm{ t }},\mathbf{X}_{\mathrm{t}}} + \sigma_{\varepsilon}^2 \mathbf{I})^{-1} \mathbf{y}_\mathrm{t},
\end{equation}
and
\begin{equation}
\label{eq:posteriorVarianceWithZeroMean}
cov(\mathbf{f}_*) = \mathbf{K }_{\mathbf{X}_*,\mathbf{X}_*} - \mathbf{K}_{\mathbf{X}_{\mathrm{t}},\mathbf{X}_*}^\mathrm{T} (\mathbf{K}_{\mathbf{X}_{\mathrm{t}},\mathbf{X}_{\mathrm{t}}}  + \sigma_{\varepsilon}^2 \mathbf{I})^{-1} \mathbf{K}_{\mathbf{X}_{\mathrm{t}},\mathbf{X}_*}.
\end{equation}
It is worth noting that this posterior distribution is also a Gaussian process, called a Gaussian process posterior. So we have $f(\mathbf{x})|\mathcal{D} \sim \mathcal{GP}(m_\mathrm{post}(\mathbf{x}),k_\mathrm{post}(\mathbf{x}, \mathbf{x}'))$, where the mean and kernel functions of this Gaussian process posterior take the following forms:
\begin{equation}
\label{eq:posteriorGPMeanWithZeroMean}
m_\mathrm{post}(\mathbf{x}) = \mathbf{K}_{\mathbf{X}_\mathrm{t},\mathbf{x}}^\mathrm{T} (\mathbf{K}_{\mathbf{X}_{\mathrm{ t }},\mathbf{X}_{\mathrm{t}}} + \sigma_{\varepsilon}^2 \mathbf{I})^{-1} \mathbf{y}_\mathrm{t},
\end{equation}
and
\begin{equation}
\label{eq:posteriorGPKernelWithZeroMean}
k_\mathrm{post}(\mathbf{x}, \mathbf{x}') = k(\mathbf{x},\mathbf{x}') - \mathbf{K}_{\mathbf{X}_{\mathrm{t}},\mathbf{x}}^\mathrm{T} (\mathbf{K}_{\mathbf{X}_{\mathrm{t}},\mathbf{X}_{\mathrm{t}}}  + \sigma_{\varepsilon}^2 \mathbf{I})^{-1} \mathbf{K}_{\mathbf{X}_{\mathrm{t}},\mathbf{x}'}.
\end{equation}

It can be observed from Eqs. (\ref{eq:posteriorMeanWithZeroMean}) and (\ref{eq:posteriorVarianceWithZeroMean}) that the key to making predictions with a Gaussian process posterior is calculating the three covariance matrices, $\mathbf{K}_{\mathbf{X}_{\mathrm{t}},\mathbf{X}_{\mathrm{t}}}$, 
$\mathbf{K}_{\mathbf{X}_{\mathrm{t}},\mathbf{X}_*}$, and $\mathbf{K }_{\mathbf{X}_*,\mathbf{X}_*}$. Difficulties in computation usually arise when performing a matrix inversion on a large covariance matrix $\mathbf{K}_{\mathbf{X}_{\mathrm{ t }},\mathbf{X}_{\mathrm{t}}}$ with many training observations. Much effort has been devoted to solving this matrix inversion problem, resulting in many approximation methods, such as covariance tapering \citep{furrer2006covariance, kaufman2008covariance} and low-rank approximations \citep{cressie2008fixed, banerjee2008gaussian}, mostly applied to handle large spatial datasets. Another important issue associated with the matrix inversion is that the covariance matrix could become ill-conditioned, most likely due to some training points being too close and providing redundant information. Two common strategies to invert an ill-conditioned covariance matrix are (1) performing the Moore–Penrose inverse or pseudoinverse using the singular value decomposition \citep{jones1998efficient} and (2) applying ``nugget" regularization, i.e., adding a small positive constant (e.g., $10^{-6}$) to each diagonal element of the covariance matrix to make it better conditioned while having a negligible effect on the calculation \citep{neal1997monte, andrianakis2012effect}. Oftentimes, adding the variance of the Gaussian noise $\sigma_{\varepsilon}^2$, as shown in Eqs. (\ref{eq:posteriorMeanWithZeroMean}) and (\ref{eq:posteriorVarianceWithZeroMean}), serves the purpose of ``nugget" regularization.

Following the numerical procedure described in Sec. \ref{sec:basics_gpr}.c, we can generate random samples of $\mathbf{f}$ from the Gaussian process posterior. For example, we can sample function values at the $N_\mathrm{*}$ input points, $\mathbf{x}_1^{*}, \dots, \mathbf{x}_{N_\mathrm{*}}^{*}$, by sampling from a multivariate Gaussian with mean $\mathbf{\overline f}_*$ and covariance $cov(\mathbf{f}_*)$. It is possible that the Cholesky decomposition needs to be performed on an ill-conditioned posterior covariance matrix $cov(\mathbf{f}_*)$. This issue can be tackled by applying ``nugget" regularization or adopting an alternative sampling procedure that centers around defining and sampling from a zero-mean, unconditional Gaussian process, as described in Refs. \citep{le2014bayesian,
menz2021variance,
wei2023expected}. Figure \ref{fig:samples_prior_posterior}(b) shows three sample functions drawn from a Gaussian process posterior after collecting 20 noisy observations of a 1D sine function.

We have been looking at the posterior of noise-free function values. To derive the posterior over the noisy observations, $p(\mathbf{y}_*|\mathcal{D}, \mathbf{X}_*)$, we add a vector of i.i.d. zero-mean Gaussian noise variables to the $\mathbf{f}_*$ posterior, producing a multivariate Gaussian with the same means (Eq. (\ref{eq:posteriorMeanWithZeroMean})) and a different covariance matrix whose diagonal elements increase by $\sigma_{\varepsilon}^2$ compared to the covariance matrix in Eq. (\ref{eq:posteriorVarianceWithZeroMean}). It is also straightforward to make predictions on a noise-free Gaussian process using Eqs. (\ref{eq:posteriorMeanWithZeroMean}) and (\ref{eq:posteriorVarianceWithZeroMean}). We can simply take out the noise variance term $\sigma_{\varepsilon}^2\mathbf{I}$ and use $\mathbf{y}_* = \mathbf{f}_*$. As is discussed in \ref{sec:UQ_design}, GPR with noise-free observations is widely used to build cheap-to-evaluate surrogates of computationally expensive computer simulation models in engineering design applications such as model calibration, reliability analysis, sensitivity analysis, and optimization. The observations in these applications are free of noise because we have direct access to the true underlying function (i.e., the computer simulation model) that we want to approximate. In contrast, as will be discussed in Sec. \ref{sec:UQ_PHM}, many applications of GPR in health prognostics require the consideration of noisy observations, as we often do not have access to the true targets (e.g., health indicator) but can only obtain noisy measurements or estimates of these targets. 

Now let us look back at the distance awareness property of GPR. Suppose a new input point $\mathbf{x}_*$ keeps moving away from the training distribution $\mathcal{D}$. In that case, the Euclidean distance between $\mathbf{x}_*$ and any input point  ${\mathbf{x}}_i$ in $\mathcal{D}$, i.e., $dist(\mathbf{x}_*, \mathbf{x}_i), \forall i=1, \dots, N$, constantly increases. All elements in the cross-covariance matrix and, more strictly, the cross-covariance vector $\mathbf{k}_{\mathbf{X}_{\mathrm{t}},\mathbf{x}_*} = [k(\mathbf{x}_1, \mathbf{x}_*), \dots, k (\mathbf{ x }_N, \mathbf{x}_*)]^\mathrm{T}$ quickly approach zero. Given that neither the training-data covariance matrix $\mathbf{K}_{\mathbf{X}_{\mathrm{t}},\mathbf{X}_{\mathrm{t}}}$ nor the new-data covariance (variance in this case) $k(\mathbf{x}_*,\mathbf{x}_*)$ experiences any changes, the posterior mean $\overline f_*$ will approach zero (i.e., the prior mean), and more importantly, the posterior variance $var(f_*)$ will approach its maximum allowed value $\sigma_{\mathrm{f}}^2$. This observation of the GPR model behavior is significant for UQ because it means that a GPR model naturally yields high-uncertainty predictions for OOD samples falling outside of the training distribution.

\paragraph{e. Optimizing hyperparameters} \mbox{}

Suppose we choose the squared exponential kernel as the covariance function. In that case, we will have three unknown hyperparameters that need to be estimated based on training data. These parameters are the characteristic length scale ($l$), signal amplitude ($\sigma_\mathrm{f}$), and noise standard deviation ($\sigma_\varepsilon$), i.e., $\bm{\uptheta} = [l, \sigma_\mathrm{f}, \sigma_\varepsilon]^\mathrm{T}$. Estimating these hyperparameters can be regarded as \emph{training} a GPR model. As it is often difficult yet not much value-added to obtain the full Bayesian posterior of $\bm{\uptheta}$, we typically choose to obtain a \emph{maximum a posteriori probability (MAP)} estimate of $\bm{\uptheta}$, a point estimate at which the log marginal likelihood $\log{p(\mathbf{y}_{\mathrm{t}}|\mathbf{X}_{\mathrm{t}},\bm{\uptheta})}$ reaches the largest value. Assuming the prior is uniform, the log marginal likelihood function of the posterior takes the following form \citep{rasmussen2006gaussian}: 
\begin{equation}
\log{p(\mathbf{y}_{\mathrm{t}}|\mathbf{X}_{\mathrm{t}},\bm{\uptheta})} = \underbrace{- \frac{1}{2} (\mathbf{y}_{\mathrm{t}}^\mathrm{T} (\mathbf{K}_{\mathbf{X}_{\mathrm{ t }},\mathbf{X}_{\mathrm{t}}} + \sigma_{\varepsilon}^2 \mathbf{I})^{-1} \mathbf{y}_\mathrm{t}}_{\text{Model-data fit}}
\underbrace{- \frac{1}{2} \log{| \mathbf{K}_{\mathbf{X}_{\mathrm{ t }},\mathbf{X}_{\mathrm{t}}} + \sigma_{\varepsilon}^2 \mathbf{I} |}}_{\text{Complexity penalty}}
\underbrace{- \frac{N}{2} \log{(2\pi)}}_{\text{Constant}},
\label{eq:LogMarginalLikelihood}
\end{equation}
The first term on the right-hand side, the so-called ``model-data fit" term, quantifies how well the model fits the training observations. The second term, called the ``complexity penalty'' term, quantifies the model complexity where a smoother covariance matrix with a smaller determinant is preferred~\cite{rasmussen2006gaussian}. The third and last term is a normalization constant and indicates that the likelihood of data tends to decrease as the training data size increases~\cite{shahriari2016taking}. 
It should be noted that the cost complexity of Eq. (\ref{eq:LogMarginalLikelihood}) is $\mathcal{O}(N^3)$ to compute the inverse of the covariance matrix $\mathbf{K}_{\mathbf{X}_{\mathrm{ t }},\mathbf{X}_{\mathrm{t}}}$ and the space complexity is $\mathcal{O}(N^2)$ to store this matrix. Hyperparameter optimization significantly influences the accuracy of GPR. See \ref{sec:gpr_hyperparameters_effect} for an illustrated example on the effect of hyperparameter optimization. 

\subsubsection{UQ capability and some limitations of Gaussian process regression}
\label{sec:uq_gpr}

GPR is capable of capturing both aleatory and epistemic uncertainty. For regression problems, the posterior variance for a query or test point, shown in Eq. (\ref{eq:posteriorVarianceWithZeroMean}), is an elegant expression of the total predictive uncertainty. The variance of the additive white noise, $\sigma_{\varepsilon}^2$, is a measure of the aleatory uncertainty. If this noise variance is assumed to be a constant (learned from the observations), the GPR model is called a ``homoscedastic'' model. In contrast, a heteroscedastic GPR model represents the noise variance as a function of the input variables $\mathbf{x}$ \cite{le2005heteroscedastic}. Assuming a squared exponential kernel is used, the epistemic uncertainty is determined mainly by $\mathbf{k}_{\mathbf{X}_{\mathrm{t}},\mathbf{x}_*}$, the covariance vector between the training points and query point, as discussed at the end of Sec. \ref{sec:basics_gpr}.d. The farther away the query point is from the training points, the smaller the elements of $\mathbf{k}_{\mathbf{X}_{\mathrm{t}},\mathbf{x}_*}$ and the larger the epistemic uncertainty. Therefore, using a distance-based covariance (or kernel) function and according to conditionals of a multivariate Gaussian, a GPR model produces low epistemic uncertainty at query or test points close to observations used for training and high epistemic uncertainty at query points far away from any training observation. This distance awareness property makes GPR an ideal choice for highly reliable OOD detection for problems of low dimensions and small training sizes. The aleatory and epistemic uncertainty components of the posterior variance determine how wide the confidence interval of a model prediction at the query point should be, reflecting the total predictive uncertainty.

Despite the highly desirable distance awareness property and OOD detection capability, GPR does not always produce posterior variances that reliably measure the predictive uncertainty. The reliability of UQ by GPR depends on many factors, such as the test point where a prediction is made, the  behavior of the underlying function to be fitted, and the choices of the kernel and hyperparameters. For example, a necessary condition for reliable UQ by a GPR model is properly choosing its kernel and optimizing the resulting hyperparameters (e.g., the variance of the additive white noise, $\sigma_{\varepsilon}^2$, measures the aleatory uncertainty and should be optimized for accurate UQ). As discussed earlier, GPR can detect OOD test points, especially those far from the training distribution. However, the high posterior variances at these ``extreme" test points may still not accurately measure the prediction accuracy. Specifically, as a test point moves away from the training distribution, the posterior variance will start to ``saturate" at its peak value, as discussed in detail in Sec. \ref{sec:basics_gpr}.d; in contrast, the prediction error at this test point may continue to rise due to an increasing degree of extrapolation, and so should an ``ideal" estimate of the predictive uncertainty. Although GPR may not produce reliable UQ in such an extreme extrapolation scenario, it is important to take a step back and keep in mind that extrapolating to an extensive degree goes against the purpose for which GPR was originally introduced, i.e., interpolation \citep{stein1999interpolation, rasmussen2006gaussian}.

Standard GPR generally does not scale well to large training datasets (large $N$) because its training complexity is $\mathcal{O}(N^3)$. This scalability issue originates from the computation of the inverse and determinant of the $N \times N$ covariance matrix $\mathbf{K}_{\mathbf{X}_{\mathrm{t}},\mathbf{X}_{\mathrm{t}}}$ during model training (i.e., hyperparameter optimization), as shown in Eq. (\ref{eq:LogMarginalLikelihood}). This scalability issue motivated considerable effort in examining local and global approximation methods to scale GPR to large training datasets while maintaining prediction accuracy and UQ quality. Interested readers may refer to a recent review on scalable GPR in \citep{liu2020gaussian}. Another limitation of GPR is its lack of scalability to high input dimensions (high $D$). This limitation stems from two issues. First, training a GPR model in a high-dimensional input space typically requires optimizing a large number of hyperparameters. This is because an ARD kernel form often needs to be chosen to deal with high-dimensional problems. As a result, the number of hyperparameters increases linearly with the number of input variables (e.g., a GPR model with the ARD squared exponential kernel shown in Eq. (\ref{eq:ardsekernel}) has $(D + 2)$ hyperparameters). A direct consequence is that a large quantity of training samples (high $N$) is needed to optimize the many hyperparameters, leading to a large covariance matrix. As discussed earlier, inverting this large covariance matrix and calculating its determinant have high computational complexity. Second, maximizing the log marginal likelihood (see Eq. (\ref{eq:LogMarginalLikelihood})) with a large number of hyperparameters becomes a high-dimensional optimization problem. Solving this high-dimensional problem requires many function evaluations, each involving one-time covariance matrix inversion and determinant calculation. Attempts to improve GPR's scalability to high-dimensional problems include (1) projecting the original, high-dimensional input onto a much lower-dimensional subspace and building a GPR model in the subspace \citep{wang2016bayesian,tripathy2016gaussian}, (2) defining a new kernel with a substantially smaller number of parameters identified with partial least squares \citep{bouhlel2016improving}, and (3) adopting an additive kernel in place of a tensor product kernel in Eq. (\ref{eq:ardsekernel}) \citep{durrande2012additive}. More detailed discussions on scaling GPR to high-dimensional problems can be found in a recent review \citep{binois2022survey}.

As a final note, since this tutorial focuses on UQ of neural networks, it is relevant and interesting to discuss connections between GPR and neural networks. Considerable efforts have been made to establish such connections. Some of these efforts are briefly discussed in \ref{sec:gpr_nn}.

\subsection{Bayesian neural network}
\label{sec:bnn}
We will first introduce the non-Bayesian (frequentist) training of a DNN, and contrast it against the Bayesian training used to form the BNNs. Consider a DNN $f:\mathbb{R}^D \mapsto \mathbb{R}$ with tunable parameters $\bm{\uptheta}$, and its prediction at an $\mathbf{x}$ is written as $\hat{y} = f(\mathbf{x};\bm{\uptheta})$. In non-Bayesian (frequentist) training, $\bm{\uptheta}$ are treated as deterministic, but unknown, parameters (i.e. \textit{not} random variables). 
An estimator for $\bm{\uptheta}$ can then be created from a training dataset $\mathcal{D} = \left\{ {\left( {{\mathbf{x}_1},{y_1}} \right),\left( {{\mathbf{x}_2},{y_2}} \right), \cdots ,\left( {{\mathbf{x}_N},{y_N}} \right)} \right\}$
by minimizing a loss function shown below:
\begin{align}
\bm{\uptheta}^{\star} =  \argmin_{\bm{\uptheta}} \,\mathcal{L}(\bm{\uptheta}; \mathcal{D}). \label{e:DNN_loss}
\end{align}

For example, a commonly used loss function for regression problems is the mean squared error (MSE) defined below:
\begin{align}
\bm{\uptheta}^{\star}_{\text{MSE}} = \argmin_{\bm{\uptheta}} \,\frac{1}{N} \sum_{i=1}^{N} || y_{i}-f(\mathbf{x}_{i};\bm{\uptheta}) ||_2^2. \label{e:DNN_MSE}
\end{align}

With the gradient of $f$ accessible through back-propagation~\cite{Rumelhart1986}, 
the loss minimization is typically solved numerically using stochastic gradient descent~\cite{Robbins1951,LeCun2012}. Once $\bm{\uptheta}^{\star}$ is found, prediction at a new point $\mathbf{x}_{*}$ can be made via $\hat{y}_{*}=f(\mathbf{x}_{*};\bm{\uptheta}^{\star})$. These predictions, however, are single-valued and do not have quantified uncertainty.

A Bayesian training~\cite{Berger1985,Bernardo2000,Sivia2006} of DNNs, also known as Bayesian deep learning~\cite{MacKay1992,neal2012bayesian,Graves2011,Blundell2015,gal2016uncertainty}, produces a \emph{Bayesian neural network} or \emph{BNN}. The Bayesian approach views $\bm{\uptheta}$ as a random variable with the goal to find the entire distribution of plausible $\bm{\uptheta}$ values that could have generated the observed data $\mathcal{D}$. Following Bayes' rule, the prior probability density function (PDF) $p(\bm{\uptheta})$ (``before''-uncertainty in $\bm{\uptheta}$) is updated to the posterior PDF $p(\bm{\uptheta}|\mathcal{D})$ (``after''-uncertainty in $\bm{\uptheta}$) conditioned on the training data $\mathcal{D}$. Mathematically, we have:
\begin{align}
    p(\bm{\uptheta}|\mathcal{D}) = 
    \frac{p(\mathcal{D}|\bm{\uptheta})p(\bm{\uptheta})}{p(\mathcal{D})}
    =
    \frac{p(\bm{\mathrm{y}}|\bm{\uptheta},\bm{\mathrm{X}})p(\bm{\uptheta})}{p(\bm{\mathrm{y}}|\bm{\mathrm{X}})}, \label{e:Bayes}
\end{align}
where we separate the training dataset $\mathcal{D}=\{\bm{\mathrm{X}},\bm{\mathrm{y}}\}$ into their inputs $\bm{\mathrm{X}} = \left\{ {{\mathbf{x}_1},{\mathbf{x}_2}, \cdots ,{\mathbf{x}_N}} \right\}$ and corresponding outputs $\bm{\mathrm{y}} = \left\{ {{y_1},{y_2}, \cdots ,{y_N}} \right\}$. Note that in the GPR section (Sec. \ref{sec:gpr}), $\mathbf{X}_{\mathrm{t}}$ and 
$\mathbf{X}_*$ denote matrices comprising input points and $\mathbf{y}_{\mathrm{t}}$ and $\mathbf{y}_*$ denote \emph{vectors} consisting of observations. In this BNN section, $\mathbf{X}$ and $\mathbf{y}$ denote \emph{sets} of input points and observations, respectively, to be consistent with the literature on Bayesian inference and BNN. In the above, $p(\bm{\mathrm{y}}|\bm{\uptheta},\bm{\mathrm{X}})$ is the likelihood and $p(\bm{\mathrm{y}}|\bm{\mathrm{X}})$ is the marginal likelihood (model evidence). The Bayesian problem and the BNN entail solving for
the posterior $p(\bm{\uptheta}|\mathcal{D})$.
We further discuss each term in the Bayes' rule in Eq.~\eqref{e:Bayes} below.

The prior $p(\bm{\uptheta})$ can be formed in an informative or non-informative manner. The former allows one to inject domain knowledge and expert opinions on the probable values of $\bm{\uptheta}$, formally through the methods of prior elicitation~\cite{OHagan2006a}. However, these methods are difficult to use on DNN parameters $\bm{\uptheta}$ due to their abstract and high-dimensional nature. The latter generates a prior following guiding principles for desirable properties (e.g., 
Jeffreys' prior~\cite{Jeffreys1946}, 
maximum entropy prior~\cite{Jaynes1968}). In practice, isotropic Gaussian is often adopted for their convenience, but caution must be taken to consider their pitfalls and appropriateness as BNN priors~\cite{Fortuin2022}.

The likelihood $p(\bm{\mathrm{y}}|\bm{\uptheta},\bm{\mathrm{X}})$ commonly follows a data (observation) model with an additive independent Gaussian noise (similar to Eq.~\eqref{eq:GaussianObs} in the GPR case):
$y_i = f(\bm{\mathrm{x}}_i;\bm{\uptheta})+\varepsilon$,
where $\varepsilon\sim \mathcal{N}(0,\sigma_{\varepsilon}^2)$. In the implementation, we often work with the log-likelihood, which is computed as:
\begin{align}
\log p(\bm{\mathrm{y}}|\bm{\uptheta},\bm{\mathrm{X}})=\sum_{i=1}^{N} \log p(y_i|\bm{\uptheta},\bm{\mathrm{x}}_i)
=-N\log(\sqrt{2\pi}\sigma_{\varepsilon})-\frac{1}{2\sigma_{\varepsilon}^2}\sum_{i=1}^{N}||y_i-f(\bm{\mathrm{x}}_i;\bm{\uptheta})||_2^2. 
\label{e:BNN_likelihood}
\end{align}
We can see that finding the mode of the Gaussian (log)-likelihood above (i.e. the $\bm{\uptheta}$ that maximizes Eq.~\eqref{e:BNN_likelihood})
is equivalent to the MSE minimization in Eq.~\eqref{e:DNN_MSE}; hence,  $\bm{\uptheta}^{\star}_{\text{MSE}}$ is also known as the maximum likelihood estimator. 
Furthermore, adding a regularization term to Eq.~\eqref{e:DNN_MSE} serves the role of a prior, and in a similar fashion, a regularized loss minimization is also known as a maximum a-posteriori (MAP) estimator (e.g., L2-regularization is the MAP with a Gaussian prior, L1-regularization is the MAP with a Laplace prior).

The marginal likelihood $p(\bm{\mathrm{y}}|\bm{\mathrm{X}})$ in the denominator of Eq.~\eqref{e:Bayes} is a (normalization) constant for the posterior that integrates the numerator: $p(\bm{\mathrm{y}}|\bm{\mathrm{X}})=\int p(\bm{\mathrm{y}}|\bm{\uptheta},\bm{\mathrm{X}})p(\bm{\uptheta})\,d\bm{\uptheta}$. As it requires a non-trivial integration, this term is highly difficult to estimate. Fortunately, Bayesian computation algorithms are often designed to avoid the marginal likelihood altogether; we will describe examples of these algorithms in the upcoming sections. 

Lastly, once the Bayesian posterior $p(\bm{\uptheta}|\mathcal{D})$ is obtained, the posterior uncertainty can be propagated through the BNN at a new point $\bm{\mathrm{x}}_{*}$ via, for example, MC sampling. Importantly, we draw the distinction between the \emph{posterior-pushforward} and \emph{posterior-predictive} distributions. 
The posterior-pushforward is $p(\hat{y}_{*}|\bm{\mathrm{x}}_{*}, \mathcal{D}) = p(f(\bm{\mathrm{x}}_{*}; \bm{\uptheta})|\bm{\mathrm{x}}_{*}, \mathcal{D})$. It describes the uncertainty on 
$\hat{y}_{*}$ (i.e. the ``clean'' prediction from the DNN) as a result of the uncertainty in $\bm{\uptheta}$. In contrast, the posterior-predictive is
$p({y}_{*}|\bm{\mathrm{x}}_{*}, \mathcal{D}) = p(\,[f(\bm{\mathrm{x}}_{*}; \bm{\uptheta})+\varepsilon]\, |\bm{\mathrm{x}}_{*}, \mathcal{D})$, it describes the uncertainty on ${y}_{*}$ (i.e. the noisy observed quantity). Hence, the former incorporates epistemic parametric uncertainty, while the latter further augments aleatory data uncertainty to the new prediction. The two distributions can be easily confused with each other, with the danger of improper UQ assessments where one might incorrectly expect the posterior-pushforward uncertainty to ``capture'' the noisy observation data. 

In the following sections, we introduce several major types of Bayesian computational methods for solving the Bayesian posterior: Markov chain Monte Carlo or MCMC (posterior sampling), variational inference (posterior approximating), and MC dropout.

\subsubsection{Markov chain Monte Carlo}
\label{sec:mcmc}

The classical method for solving the Bayesian problem is to sample the posterior distribution using Markov chain Monte Carlo (MCMC)~\cite{Andrieu2003,Various2011}. 
MCMC establishes a Markov chain $\{\bm{\uptheta}^{(n)}\},n\in\mathbb{N}$ from a transition kernel (i.e. proposal distribution) such that the chain converges to the posterior $p(\bm{\uptheta}|\mathcal{D})$ regardless of its initial position $\bm{\uptheta}^{(0)}$. Most importantly, ergodicity theorems ensure that the empirical average of MCMC samples, $\frac{1}{N_s}\sum_{n=1}^{N_s} h(\bm{\uptheta}^{(n)})$, converges to the posterior expectation $\mathbb{E}_{\bm{\uptheta}|\mathcal{D}}[h(\bm{\uptheta})]$ almost surely.
The most fundamental MCMC is the Metropolis-Hastings (MH) algorithm~\cite{Metropolis1953a,Hastings1970a}, which forms the basis of many advanced MCMC variants. 
Hamiltonian Monte Carlo (HMC) \cite{Neal2011a,Betancourt2017}, an advanced type of MCMC with improved mixing properties, is more commonly used for BNNs. Drawing intuition from physics, HMC introduces an auxiliary momentum variable to form a system of Hamiltonian dynamics that can generate trajectories following the high-probability regions of the posterior (the so-called \emph{typical set}). However, the effect of concentration of measure brings the typical set to become more singular with increasing dimension, and even HMC has only been exercised for $\bm{\uptheta}$ that is hundreds-dimensional~\cite{neal2012bayesian, Chen2014, Zhang2018}. This is still orders of magnitude shorter than modern DNNs that can easily have millions, even billions, of tunable parameters. While MCMC methods are theoretically appealing due to their asymptotic convergence to the true posterior, the Markov chains can be very difficult to mix for high dimensions in practice. As a result, they see limited usage in BNNs.

\subsubsection{Variational Inference}
\label{sec:vi}

A more scalable approach to the Bayesian inference problem can be found through variational inference (VI)~\cite{Blei2017,Zhang2019}. In contrast to MCMC sampling, the idea of VI is to approximate the posterior within a parametric family of distributions (e.g., a family of Gaussian distributions). In this section, we will start by defining the optimization problem that describes the best posterior approximation, then introduce some examples of numerical algorithms to solve the VI problem.

Denoting a variational distribution (for approximating the posterior) using $q(\bm{\uptheta};\lambda)$ parameterized by $\lambda$, VI seeks the best posterior-approximation $q(\bm{\uptheta};\lambda^{\star})$ that minimizes the Kullback-Leibler (KL) divergence between $q(\bm{\uptheta};\lambda)$ and $p(\bm{\uptheta}|\mathcal{D})$, that is:
\begin{align}
    \lambda^{\star} = \argmin_{\lambda} \, D_{\mathrm{KL}}\left[\, q(\bm{\uptheta};\lambda) \,||\, p(\bm{\uptheta}|\mathcal{D})\, \right] 
    . \label{e:VI_KL}
\end{align} 

A popular choice for the variational distribution is the independent (mean-field) Gaussian:
$q(\bm{\uptheta}; \lambda) = \prod_{k=1}^K q(\bm{\uptheta}_k; \lambda_k)= \prod_{k=1}^K \mathcal{N}(\bm{\uptheta}_k; \mu_k,\sigma_k^2)$,
where $K$ is the total number of parameters in the DNN. The independence structure allows the joint PDF to be factored into a product of univariate Gaussian marginals, and so the variational parameters are $\lambda = {\{\mu_k, \sigma_k\}},k=1,\ldots,K$ that encompasses the mean and standard deviation of each component of $\bm{\uptheta}$, for a total of $2K$ variational parameters. As a result,  mean-field simplifies to a diagonal global covariance matrix (instead of dense covariance) in the approximate posterior, and it is unable to capture any correlation among the $\bm{\uptheta}_k$'s. 
More expressive representations of $q(\bm{\uptheta};\lambda)$ are also possible, for example via normalizing flows~\cite{Rezende2015} and transport maps~\cite{Marzouk2016} that parameterize the mapping from the posterior random variable $\bm{\uptheta}$ to a standard normal reference random variable. 

Given the variational distribution $q(\bm{\uptheta}; \lambda)$, Eq.~\eqref{e:VI_KL} can be further simplified as follows:
\begin{align} 
    \lambda^{\ast}  &= \argmin_{\lambda}\, D_{\mathrm{KL}}\left[\,q(\bm{\uptheta};\lambda) \,||\, p(\bm{\uptheta}|\mathcal{D})\,\right]
    \nonumber\\
    &= \argmin_{\lambda}\, \mathbb{E}_{q(\bm{\uptheta};\lambda)} \left[ \ln q(\bm{\uptheta};\lambda) - \ln \left( \frac{p(\bm{\mathrm{y}}|\bm{\uptheta},\bm{\mathrm{X}})p(\bm{\uptheta})}{p(\bm{\mathrm{y}}|\bm{\mathrm{X}})} \right) \right]
    \nonumber\\
    &=\argmin_{\lambda}\, \underbrace{D_{\mathrm{KL}}\left[\,q(\bm{\uptheta};\lambda) \,||\, p(\bm{\uptheta})\,\right] -\mathbb{E}_{q(\bm{\uptheta};\lambda)} \left[ \ln p(\bm{\mathrm{y}}|\bm{\uptheta},\bm{\mathrm{X}}) \right]  }_{-\, {\text{Evidence Lower Bound (ELBO)}}}
    ,\label{e:ELBO}
\end{align}
where going from the second to the third equation, the log-denominator's contribution $\mathbb{E}_{q(\bm{\uptheta};\lambda)} \left[\ln p(\bm{\mathrm{y}}|\bm{\mathrm{X}})\right]=\ln p(\bm{\mathrm{y}}|\bm{\mathrm{X}})$ is omitted since it is constant with respect to both $\lambda$ and $\bm{\uptheta}$ and its exclusion does not change 
the minimizer. The resulting expression in Eq.~\eqref{e:ELBO} is the negative of the well-known \emph{Evidence Lower Bound (ELBO)}. 
The first term of ELBO acts as a regularization to keep $q(\bm{\uptheta};\lambda)$ close to the prior.
The second term of ELBO involves the log-likelihood of generating the observed data under DNN parameters $\bm{\theta}\sim q(\bm{\uptheta};\lambda)$; hence it measures the expected model-data fit.

In general, it is impossible to evaluate the ELBO analytically, and Eq.~\eqref{e:ELBO} must be solved numerically. The simplest approach is to use MC sampling to estimate the ELBO, which only entails sampling $\bm{\theta}\sim q(\bm{\uptheta};\lambda)$. Often, further simplifications can be made by analyically computing the first term, which involves only the prior and variational distribution. Furthermore, the gradient of ELBO with respect to $\lambda$ may be derived (e.g., see~\cite{Blundell2015} for Gaussian $q$) or obtained through automatic differentiation, allowing one to take advantage of gradient-based optimization algorithms (e.g., stochastic gradient descent) to solve Eq.~\eqref{e:ELBO}.


The Stein variational gradient descent (SVGD)~\cite{Liu2016} is another VI variant offering a flexible particle approximation to the posterior distribution. 
SVGD leverages the relationship between the gradient of the KL divergence in Eq.~\eqref{e:VI_KL} to the Stein discrepancy, the latter which can be approximated using a set of particles. An update procedure can then formed to iteratively ascent along a perturbation direction
$
    \bm{\uptheta}_i^{\ell+1} \leftarrow \bm{\uptheta}_i^{\ell} + \epsilon_{\ell} \hat{\varphi}^*({\bm{\uptheta}_i^{\ell}})
$, 
where $\bm{\uptheta}_i^{\ell}, \; i=1,\ldots,N_\mathrm{p}$, denotes the $i$-th particle at the $\ell$-th iteration, $\epsilon_{\ell}$ is the learning rate, and the perturbation direction is defined as:
\begin{align}
    \hat{\varphi}^*(\bm{\uptheta}) = \frac{1}{N_\mathrm{p}}\sum_{j=1}^{N_\mathrm{p}} \left[k(\bm{\uptheta}_j^{\ell}, \bm{\uptheta}) \nabla_{\bm{\uptheta}_j^{\ell}}\ln p(\bm{\uptheta}_j^{\ell} \mid \mathcal{D}) + \nabla_{\bm{\uptheta}_j^{\ell}}k(\bm{\uptheta}_j^{\ell}, \bm{\uptheta})\right],
\end{align}
with $k(\cdot,\cdot)$ being a positive definite kernel (e.g., radial basis function kernel in Eq.~\eqref{eq:sekernel})
Notably, the gradient of the log-posterior in the above equation can be evaluated via the sum of gradients of log-likelihood and log-prior, since the gradient of the log-marginal-likelihood with respect to $\bm{\uptheta}$ is zero. The overall effect is an iterative transport of a set of particles to best match the target posterior distribution $ p(\bm{\uptheta}|\mathcal{D})$. 
Building upon the SVGD, advanced methods of Stein variational Newton~\cite{Detommaso2018,Leviyev2022} that makes use of second-order (Hessian) information, and projected SVGD~\cite{Chen2020} that finds low dimensional data-informed subspaces, have also been proposed.

\begin{figure}[!ht]
\includegraphics[width=\textwidth]{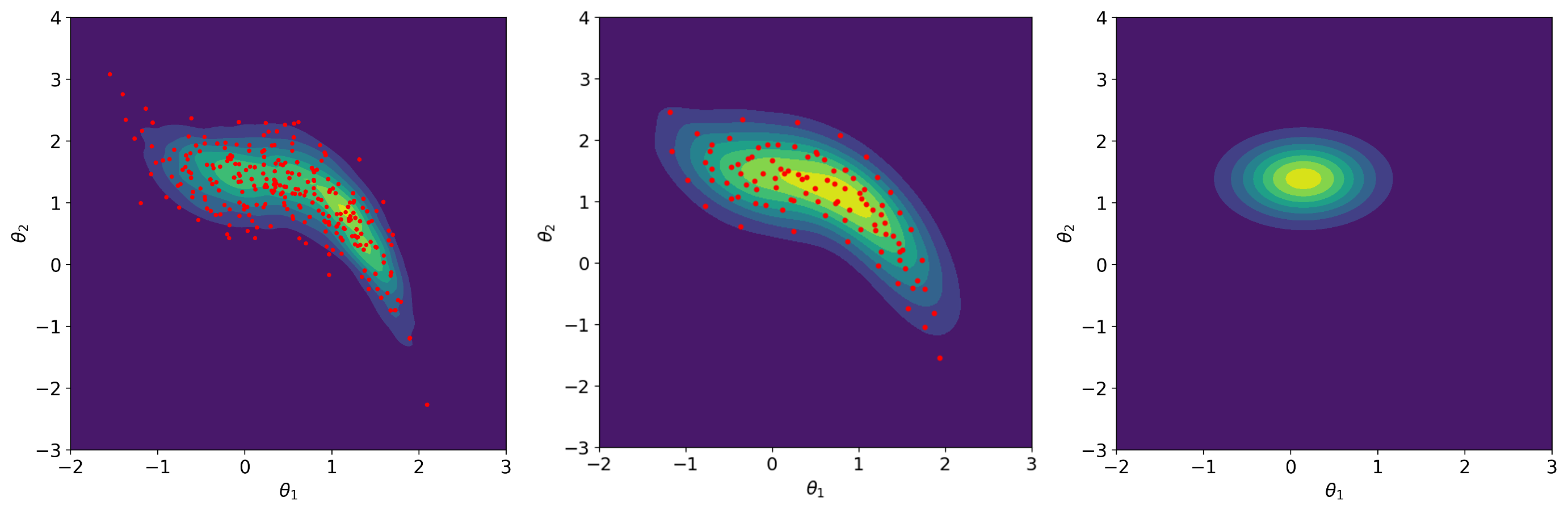}
\caption{Illustration of Bayesian posterior obtained from (left) MCMC, (middle) SVGD, and (right) mean-field Gaussian VI for a simple low-dimensional Bayesian inference test problem. 
}
\label{fig:Bayesian_posteriors}
\end{figure}

Figure \ref{fig:Bayesian_posteriors} compares the different Bayesian posteriors obtained from a simple low-dimensional Bayesian inference test problem using MCMC, SVGD, and mean-field Gaussian VI. MCMC and SVGD provide sample/particle representations of the posterior distribution, while VI produces an analytical Gaussian approximation of the PDF. Both MCMC and SVGD are able to capture non-Gaussian and correlated structure, although SVGD is more restrictive in the number of particles it can use due to higher memory requirement. However, SVGD and VI are more scalable to higher $\bm{\uptheta}$ dimensions than MCMC.

We note that another variant of VI can arise from the reverse KL divergence $D_{\mathrm{KL}}\left[\, p(\bm{\uptheta}|\mathcal{D}) \,||\, q(\bm{\uptheta}; \lambda) \, \right]$ (in contrast to the $D_{\mathrm{KL}}\left[\, q(\bm{\uptheta};\lambda) \,||\, p(\bm{\uptheta}|\mathcal{D})\, \right]$ from Eq.~(\ref{e:VI_KL})). Notable algorithms from this formulation include expectation propagation~\cite{minka2001expectation}, assumed density filtering~\cite{lauritzen1992propagation}, and moment matching~\cite{opper1999bayesian}; in particular, expectation propagation has been shown to be quite effective in logistic-type models in general.

\subsubsection{MC dropout}\label{sec:mcdropout}
Although the Bayesian approach offers an elegant and principled way to model and quantify the uncertainty in neural networks, it typically comes with a prohibitive computational cost. As introduced earlier, MCMC and VI are two commonly used methods to perform Bayesian inference over the parameters of neural network. However, Bayesian inference with MCMC and variational inference in DNNs suffers from extremely time-consuming computational burden and poor scalability. Specifically, in the case of MCMC, estimating the uncertainty of neural network prediction with respect to a given input requires to draw a large number of samples from the posterior distributions of thousands or even millions of neural network parameters and propagate these samples through the neural network~\cite{shen2021variational}. Compared with MCMC, VI is much faster and has better scalability as it recasts the inference of posterior distributions of neural network parameters as an optimization problem. However, VI unfortunately doubles the parameters to be estimated for the same neural network. In addition, it is intricate to derive and formulate the optimization problem, much less optimization regarding the high-dimensional problem consumes a large amount of time before convergence~\cite{gal2016dropout}.


Beyond MCMC and VI, further scalability can be achieved through the MC dropout method. Initially proposed as a regularization technique to prevent the overfitting of DNNs~\citep{srivastava2014dropout}, MC dropout has been shown to approximate the posterior predictive distribution under a particular Bayesian setup~\citep{gal2016dropout}. Procedurally, MC dropout follows the same deterministic DNN training in Eq.~\eqref{e:DNN_loss}, except that it forms new sparsely connected DNNs from the original DNN (see method 3 in Fig. \ref{fig:UQ_methods_schematic}) by multiplying every weight with an independent Bernoulli random variable. Hence, each weight has some probability of becoming zero (i.e., the weight being dropped). 
These Bernoulli random variables are re-sampled (i.e., a new, randomized sparse DNN is formed) for every training sample and for every forward pass of the model.
At test time, the prediction at a new point $\bm{\mathrm{x}}_{*}$ can also be repeated with multiple forward passes each with a new, randomized sparse DNN resulting from the dropout operation. An ensemble of predictions can thus be obtained to estimate the uncertainty. 
Practical implementation of MC dropout in probabilistic programming languages is often realized by adding a dropout layer after each fully-connected layer. 

The connection from MC dropout to a Bayesian setup is detailed in~\cite{gal2016dropout,gal2016theoretically}. Those works show that the loss function following the dropout procedure corresponds to a single-sample MC approximation to the VI objective (i.e., the ELBO in Eq.~\eqref{e:ELBO}),
where the variational posterior of the DNN weights is a Bernoulli mixture of two independent Gaussians of fixed covariance.
Furthermore, the prior of each DNN weight is assumed to follow a standard normal distribution, and the likelihood is based on the additive Gaussian noise model in Eq.~\eqref{e:BNN_likelihood}. Established upon such a setup, in MC dropout, the variational distribution $q \left(\bm{\uptheta}; \lambda \right)$ for approximating the posterior distribution $p(\bm{\uptheta}|\mathcal{D})$ becomes a factorization over the weight matrices $\bm{\mathrm{W}}_{i}$ of the layers $1$ to $L$. Mathematically, the variational distribution $q \left(\bm{\uptheta}; \lambda \right)$ takes the following multiplicative form:
\begin{equation}
q \left(\bm{\uptheta}; \lambda \right) = \prod\limits_{i=1}^{L} {{q_{{\bm{\mathrm{M}}_{i}}}}\left( {{\bm{\mathrm{W}}_{i}}} \right)},
\end{equation}
where ${q_{{\bm{\mathrm{M}}_{i}}}} \left(\bm{\mathrm{W}}_i \right)$ denotes the density associated with the weight matrices $\bm{\mathrm{W}}_i$ of layer $i$, 
and under MC dropout, it emerges as a Gaussian mixture model consisting of two independent 
Gaussian components with a fixed and identical variance, as shown below
\cite{gal2016dropout,gal2016theoretically}:
\begin{equation}
q_{\bm{\mathrm{M}}_i}\left( {\bm{\mathrm{W}}_i} \right) = {p_i}\underbrace{\mathcal{N}\left( {\bm{\mathrm{M}}_{i},{\sigma ^2}\bm{\mathrm{I}}_i} \right)}_\text{First Gaussian} + \left( {1 - {p_i}} \right)\underbrace{\mathcal{N}\left( {\bm{0},{\sigma ^2}\bm{\mathrm{I}}_i} \right)}_\text{Second Gaussian}. \label{eq:GMMs_MCdropout}
\end{equation}

In the above, $\bm{\mathrm{M}}_{i}$ is the mean of the first Gaussian, which is a vectorization of $n_{i-1} \times n_{i}$ values pertaining to the weight matrix $\bm{\mathrm{W}}_i$ of size $n_{i-1} \times n_{i}$ ($n_i$ denotes the number of units in the $i$-th layer; when $i=0$, it denotes the number of inputs), $\sigma$ is the standard deviation parameter specified by the end user, $\bm{\mathrm{I}}_i$ is the identity matrix,
$\mathcal{N}$ denotes the normal distribution, and $p_i$ $(p_i \in [0, 1])$ is the dropout rate associated with the set of links connecting two consecutive layers of the neural network. Under this VI perspective, MC dropout corresponds to optimizing $\lambda=\bm{\mathrm{M}}_{i}$, while both $\sigma$ and $p_i$ have fixed user-chosen values and are not part of the variational parameter set.

In the MC dropout implementation, for each element of $\bm{\mathrm{W}}_i$,
we sample a $\upsilon$ according to a Bernoulli distribution with a prescribed dropout rate $p_i$, that is $\upsilon \sim \text{Bernoulli}(p_i)$.
If the binary variable $\upsilon = 0$, it indicates that link connecting the $i$-th and $(i+1)$-th layers is dropped out. 
This operation corresponds to choosing one of the two Gaussians from the mixture model in Eq. \eqref{eq:GMMs_MCdropout}, and hence
MC dropout can serve as an approximation to the Bayesian posterior in BNNs. 

A major advantage of MC dropout is that it is very straightforward to implement, requiring only a few lines of modification to insert the $\bm{z}$'s to an existing DNN setup and often conveniently available as a dropout layer in many  programming environments. Furthermore, its ease of implementation is agnostic of the neural network architecture, and can be readily adopted for many polular types of neural networks such as convolutional neural network (CNN) and recurrent neural network (RNN)~\cite{gal2016theoretically,gal2015bayesian}. 
Another major advantage of MC dropout is its low computational cost and high scalability since its training procedure is effectively identical to an ordinary, non-Bayesian training of DNNs but with randomized sparse networks. These appealing properties collectively contribute to the growing popularity of MC dropout in practice.

MC dropout also has some limitations. One disadvantage is that the quality of the uncertainty generated by MC dropout is highly dependent on the choice of several hyperparameters~\cite{osband2016risk,alarab2021illustrative,caldeira2020deeply}, such as the dropout rate and number of dropout layers. Thus, these hyperparameters need to be fine tuned.
Along this front, we also have similar findings in Section~\ref{sec:toy_example} that MC dropout exhibits poor stability to the dropout rate, training epochs, and the number of trainable network parameters (see~\ref{appendix_MCDropout} for more details). Regardless of the instability, the uncertainty produced by MC dropout exhibits a consistent difficulty in detecting OOD instances. Note that other approximation inference methods, such as MFVI, have a pathology that is slightly different from MC dropout with respect to the soundness of the quantified uncertainty, see Section~\ref{sec:toy_example} for more details. As highlighted by~\citet{foong2020expressiveness}, the pathology of UQ in approximation methods is solely attributed to the restrictiveness of approximating family, while exact inference methods, such as MCMC, do not have such a problem. Another disadvantage of MC dropout is that users do not have the option to inject their prior knowledge by specifying the prior or likelihood function because there is no mechanism for MC dropout to integrate such information---as a result, MC dropout can only represent a narrow spectrum of Bayesian problems. A further side effect of this limitation is that users may be hindered from critically thinking about the prior and likelihood altogether, which may lead to claims of a Bayesian solution without actually having a Bayesian problem setup. Finally, some researchers~\cite{verdoja2020notes} have argued that MC dropout is not Bayesian because the variational distribution fails to converge to ground-truth posterior distribution on closed-form benchmarks. 

\subsection{Neural network ensemble}
\label{sec:neuralnetworkensemble}

Ensemble learning is a well-established technique to prevent overfitting and mitigate the poor generalizability of ML models~\cite{opitz1999popular}. An essential step in constructing ensemble models is to train multiple individual models independently and aggregate predictions from these individual models to derive the final prediction. When building an ML model ensemble, it is of paramount importance to retain a high degree of diversity among the individual models to achieve desirable performance improvement~\cite{dietterich2000ensemble}. Such diversity can be achieved through a broad spectrum of means that can be grouped into two principal categories: (1) randomization approaches, such as bagging (a.k.a. bootstrapping), where ensemble members are trained on different bootstrap samples of the original training set or a random subset of original features~\cite{breiman1996bagging}; and (2) boosting approaches: boosting learns from the errors of previous iterations by increasing the importance of those wrongly predicted training instances, thus sequentially and incrementally constructing an ensemble~\cite{schapire2013boosting}.

In the context of deep learning, building an ensemble of neural networks entails independently training multiple neural networks with an identical architecture. Due to the easiness of implementation, neural network ensemble has been pervasively used to characterize the uncertainty of neural network predictions~\cite{thelen2022comprehensive,zhang2019ensemble}. In particular, well-calibrated uncertainty estimates tend to yield higher uncertainty on OOD data than on samples sufficiently similar to the distribution of training data. On this front, the uncertainty of a neural network ensemble is principled to some extent in the sense that this ensemble is inclined to produce higher uncertainty estimates (e.g., entropy in the case of classification problems) for OOD instances~\cite{lakshminarayanan2017simple}. The appealing feature of neural network ensembles in producing higher uncertainty for OOD instances has been actively exploited s a prevailing 
means to detect dataset shifts in the ML community because the data collected under a shifted environment typically displays salient patterns that are substantially different from the data that the ensemble neural networks are trained with~\cite{zhang2022explainable,lakshminarayanan2017simple,ovadia2019can}.

\subsubsection{Aleatory uncertainty: training each network individually}
\label{sec:neuralnetwork_ensemble_aleatory}

We consider a popular configuration of neural network ensemble where each individual neural network in an ensemble outputs two quantities denoting the predicted mean $\widehat \mu \left( \mathbf{x}_i \right)$ and variance $\widehat \sigma^2 \left(\mathbf{x}_i \right)$ with respect to an input $\mathbf{x}_i$ in its final output layer (see Fig.~\ref{fig:UQ_methods_schematic} for an overview on the architecture of the individual neural network). In this configuration, the predictive distribution of each network is often assumed to be Gaussian; therefore, the final output layer is sometimes called \textit{Gaussian layer}. Such a configuration enables characterizing observational noise of aleatory nature associated with target values. 

Let us take a closer look at the aleatory uncertainty, more specifically, the observational noise pertaining to each target observation. The simplest case is that we assume the same amount of noise or aleatory uncertainty for every input $\mathbf{x}_i$, also known as \emph{homoscedasticity} or \emph{homogeneity of variance} in statistics (similar to the homoscedastic case for GPR discussed in Sec. \ref{sec:gpr}). To represent the relationship between input $\mathbf{x}_i$ and observation $y_i$, we can use the Gaussian observation model given in Eq. (\ref{eq:GaussianObs}), substituting $\mathbf{x}$ with $\mathbf{x}_i$ and $y$ with $y_i$. In this model, a random noise term $\varepsilon$, often modeled as a zero-mean Gaussian noise, shifts the target away from the true value $f\left( \mathbf{x}_i \right)$ to the observed value ${y_i}$. In this simplest case, the variance of random noise $\varepsilon$ takes the same value $\sigma_{\varepsilon}^2$ for every input and is thus a constant. Although we could learn $\sigma_{\varepsilon}$ together with the neural network parameters $\bm{\uptheta}$, this simplest case may not be realistic as some regions of the input space may have larger measurement noise than other regions. 

A more realistic case is one where the noise variance depends on $\mathbf{x}_i$. The basic idea is to tailor aleatory uncertainty to each input, making the uncertainty input-dependent. This heteroscedastic case is also briefly discussed in Sec. \ref{sec:gpr} where heteroscedastic GPR is the focus of the discussion. The observation model now becomes the following:
\begin{equation}
{y_i} = f\left( \mathbf{x}_i \right) + \varepsilon \left( \mathbf{x}_i \right),
\label{eq:GaussianObsHetero}
\end{equation}
where the variance of the noise term $\varepsilon \left( \mathbf{x}_i \right)$, $\sigma_{\varepsilon}^2\left( \mathbf{x}_i \right)$, is now a function of $\mathbf{x}_i$. It turns out that a neural network can be trained to learn the mapping from $\mathbf{x}$ to $\sigma_{\varepsilon}^2$ \citep{kendall2017uncertainties,lakshminarayanan2017simple}. It then follows that we can train a neural network with parameters $\bm{\uptheta}$ that learns to predict both the mean $\mu \left( \mathbf{x}_i \right)$ and variance $\sigma^2 \left(\mathbf{x}_i \right)$ of the target for each input $\mathbf{x}_i$. This neural network has two outputs, predicted mean $\widehat \mu\left( \mathbf{x}_i;\bm{\uptheta} \right)$ and variance $\widehat \sigma^2 \left(\mathbf{x}_i;\bm{\uptheta} \right)$, which fully characterise a Gaussian predictive distribution, i.e., $\widehat y_i \sim \mathcal{N}\left( {\widehat \mu \left( \mathbf{x}_i;\bm{\uptheta} \right)},{\widehat \sigma^2 \left(\mathbf{x}_i;\bm{\uptheta} \right)} \right)$

Before optimizing the network parameters $\bm{\uptheta}$, we need to define a proper scoring rule that measures the quality of predictive (aleatory) uncertainty. For regression problems, a typical choice of a proper scoring rule is the likelihood function ${p}\left( {\left. {y_i} \right|{\mathbf{x}_i;\bm{\uptheta}}}\right)$ whose logarithmic transformation takes the following form~\cite{lakshminarayanan2017simple,nix1994estimating}:
\begin{equation}\label{eq:log_likelihood}
\log p\left( {\left. {{y_i}} \right|{\mathbf{x}_i};\bm{\uptheta}} \right) = -\frac{{\log \widehat\sigma ^2\left( \mathbf{x}_i;\bm{\uptheta} \right)}}{2} - \frac{{{{\left( {y_i - {\widehat\mu}\left( \mathbf{x}_i;\bm{\uptheta} \right)} \right)}^2}}}{{2\widehat\sigma^2\left( \mathbf{x}_i;\bm{\uptheta} \right)}} - \text{constant}.
\end{equation}

Given a training dataset consisting of $N$ input-output pairs, $\mathcal{D} = \left\{ {\left( {{\mathbf{x}_1},{y_1}} \right),\left( {{\mathbf{x}_2},{y_2}} \right), \cdots ,\left( {{\mathbf{x}_N},{y_N}} \right)} \right\}$, 
$\bm{\uptheta}$ can be optimized by minimizing the following negative log-likelihood (NLL) loss on the entire training data, which is equivalent to maximizing the negative counterpart of the likelihood function in Eq. (\ref{eq:log_likelihood}), after being summed up over all $N$ training samples.
\begin{equation}\label{eq:ensemble_nn_loss}
    \mathcal{L} \left(\bm{\uptheta} \right) = \sum\limits_{i = 1}^N {\left[ {\frac{{\log \widehat\sigma^2\left( \mathbf{x}_i;\bm{\uptheta} \right)}}{2} + \frac{{{{\left( {y_i - {\widehat\mu}\left( \mathbf{x}_i;\bm{\uptheta} \right)} \right)}^2}}}{{2\widehat\sigma^2\left( \mathbf{x}_i;\bm{\uptheta} \right)}}} \right]}, 
\end{equation}
where the constant term in Eq. (\ref{eq:log_likelihood}) is omitted for brevity because it has nothing to do with the optimization of $\bm{\uptheta}$. 

\subsubsection{Epistemic uncertainty: using an ensemble of independently trained networks}
\label{sec:neuralnetwork_ensemble_epistemic}
As discussed in Sec. \ref{sec:neuralnetwork_ensemble_aleatory}, the neural network ensemble approach captures aleatory uncertainty by training a neural network that produces a Gaussian output (or another type of probability distribution) for each input. This modeling process improves over traditional deterministic approaches that only produce a point estimate. Plus, the network-predicted variance varies according to the input, making it possible to capture input-dependent observational noise. One limitation is that minimizing the loss function in Eq. (\ref{eq:ensemble_nn_loss}) yields a single vector of network parameters. Therefore, the resulting neural network cannot capture the uncertainty related to the network parameters because all parameters are deterministic. This treatment becomes an issue when only limited training data are available. These cases are more realistic than having abundant training data, and when training data are of limited quantities, epistemic uncertainty is high and cannot be ignored. One widely used way to capture epistemic uncertainty is to assume and estimate uncertainty in the parameters of a neural network model, also known as model parameter uncertainty or network parameter uncertainty. 

After tuning the neural network parameters $\bm{\uptheta}$, at the time of prediction, each individual neural network generates a pair of outputs $\left(\widehat\mu \left( \mathbf{x}_{*} \right), \widehat\sigma \left(\mathbf{x}_{*} \right) \right)$ with respect to an unseen instance $\mathbf{x}_{*}$, where $\widehat\sigma\left(\mathbf{x}_{*} \right)$ explicitly quantifies the aleatory uncertainty in model prediction arising from the random noise $\varepsilon \left( \cdot  \right)$ associated with the target value. Next, to quantify the epistemic uncertainty associated with the neural network parameters $\bm{\uptheta}$, we can build an ensemble of neural networks, for example, by adopting the randomization strategy (random parameter initialization and mini-batch sampling) that attains a diverse set of neural networks. Suppose the neural network ensemble is composed of $M$ individual neural networks, then the ensemble model produces $M$ pairs of $\left(\widehat\mu_m \left( \mathbf{x}_{*} \right), \widehat\sigma_m \left(\mathbf{x}_{*} \right) \right)$ $\left(m = 1, 2, \cdots, M \right)$ for the given input $\mathbf{x}_{*}$. The $M$ pairs of predictions $\left(\widehat\mu_m \left( \mathbf{x}_{*} \right), \widehat\sigma_m \left(\mathbf{x}_{*} \right) \right)$ can be viewed as a mixture of Gaussian distributions. Thus, we can use a single Gaussian distribution to approximate the mixture of Gaussian distributions as long as the mean and variance of the single Gaussian distribution are the same as the mean and variance of the mixture. Assuming that each individual neural network in the ensemble carries an equal weight, we have the mean and variance of the ensemble-predicted single Gaussian distribution as:
\begin{equation}
\label{eq:ensembleMean_and_Variance}
\begin{array}{l}
\overline \mu  \left( \mathbf{x}_{*} \right) = \frac{1}{M}\sum\limits_{m = 1}^M {{\widehat\mu _m}} \left( \mathbf{x}_{*} \right),\\
{\overline \sigma  ^2}\left( \mathbf{x}_{*} \right) = \frac{1}{M}\sum\limits_{m = 1}^M {\left( { \widehat\sigma_m^2 \left(\mathbf{x}_{*} \right) + \widehat\mu_m^2 \left(\mathbf{x}_{*} \right) - \overline \mu^2  \left( \mathbf{x}_{*} \right)} \right)}. 
\end{array}
\end{equation}

In the ensemble of neural networks, both the aleatory and epistemic uncertainty can be measured in a straightforward way. Specifically, the aleatory uncertainty arising from the noise associated with the observation $y$ is reflected in the variance $\widehat\sigma_m \left( \mathbf{x}_{*} \right)$ predicted by each individual neural network. In contrast, the epistemic uncertainty associated with the network structure and parameters is manifested mainly as the difference with respect to $\widehat\mu_m \left( \mathbf{x}_{*} \right)$ of the $M$ neural networks because each individual neural network is initialized with a random set of weights and biases and trained with a random mini-batch data for the gradient descent algorithm. Such randomness introduces a sufficient amount of diversity among the individual models. Thus, the difference between the individual mean predictions ${\widehat\mu _m} \left(\mathbf{x}_{*} \right)$ that dominates the epistemic uncertainty characterizes the structural and parametric uncertainty pertaining to the neural network. 

An interesting question about neural network ensembles is why training multiple neural networks of an identical architecture independently with just random initializations can capture epistemic uncertainty. The answer lies in that training a neural network with a large number of parameters (e.g., weights and biases) is an extremely intricate large-scale optimization problem in a high-dimensional space, and stochastic gradient descent-based algorithms oftentimes converge to different sets of parameter values $\bm{\uptheta}$ that are locally optimal~\citep{fort2019deep}. As mentioned earlier, network training involves two sources of randomness: (1) random parameters initialization at the beginning of model training and (2) random perturbations of the training data to produce mini-batches of data in stochastic gradient descent As a result, the locally optimal parameters $\bm{\uptheta}$ vary from one trained neural network to another. Suppose $M$ independent training runs give rise to $M$ different local minima for the network parameters, which then lead to the creation of $M$ individual members of an ensemble, as shown in Eq. (\ref{eq:ensembleMean_and_Variance}). From the optimization perspective, the randomness in the initialization of neural network parameters and the sampling of mini-batch data encourages the optimization algorithm to explore different modes of the function space of a neural network. As a result, the predicted means of these $M$ networks may differ substantially in some regions of the input space, while the predicted variances may still be similar, resulting in high epistemic uncertainty. These regions are typically located outside the training data distribution. Test samples falling into these regions are called OOD samples (as previously defined in Sec. \ref{sec:introduction}), where ensemble predictions must be taken cautiously and are often untrustworthy.

\subsection{Deterministic methods}
\label{sec:deterministic_uq}

A recent line of effort attempted to estimate the predictive uncertainty of neural networks using deterministic UQ methods. These methods require only a \emph{single forward pass} on a neural network with deterministic parameters (weights and biases) to produce probabilistic outputs (e.g., predicted mean and variance for regression). A resulting benefit that makes these methods uniquely attractive is high computational efficiency (test time), particularly suitable for safety-critical applications with stringent real-time inference requirements (e.g., high-rate structural health monitoring and prognostics \citep{dodson2022high} and autonomous driving \citep{kiran2021deep}). Examples of these deterministic methods include deterministic uncertainty quantification (DUQ) \citep{van2020uncertainty}, deep deterministic uncertainty (DDU) \citep{mukhoti2021deterministic}, deterministic uncertainty estimation (DUE) \citep{van2021feature}, and spectral-normalized neural Gaussian process (SNGP) \citep{liu2020simple, fortuin2021deep}. This section first discusses distance awareness in the hidden space, which is a fundamental property of many deterministic methods, then provides a brief overview of how distance-aware feature representation (hidden layers) and uncertainty prediction (output layer) are achieved in SNGP.

\subsubsection{Feature collapse and hidden-space regularization}
\label{sec:feature_collapse}
The idea fundamental to many recently developed deterministic methods is (input) distance-aware representations in the latent (or hidden) space, achieved by regularizing the learned latent representations of a neural network such that distances between points in the input space are preserved in the hidden space. The need for distance-aware latent representations comes from a recently reported phenomenon called \emph{feature collapse} \citep{van2020uncertainty}, where some OOD points in the input space are mapped through feature extraction to in-distribution points in the hidden space, leading to overconfident predictions at these OOD points. Feature collapse must be combatted for feature representations in the hidden space to be useful for epistemic uncertainty estimation and OOD detection. One option is imposing a bi-Lipschitz constraint on the feature extractor (i.e., a neural network excluding its output layer). The term ``bi-Lipschitz” means a two-sided constraint on the Lipschitz constant of a feature extractor that determines how much distances in the input space contract (small Lipschitz, feature collapse) and expand (large Lipschitz, small changes in input resulting in drastic changes in latent features). 

We now briefly describe the math pertaining to a bi-Lipschitz constraint. Suppose we take any two input points  $\mathbf{x}$ and $\mathbf{x}'$ from a training dataset and let $h_\mathrm{nn}(\cdot)$ denote a function mapping an input into latent features (i.e., right after the activation function in the last hidden layer of a neural network). A bi-Lipschitz constraint on the mapping function $h$ for any training input pairs looks like: 
\begin{equation}
Lip^\mathrm{lb} || \mathbf{x} - \mathbf{x}' ||_\mathrm{input} \leq || h_\mathrm{nn}(\mathbf{x}) - h_\mathrm{nn}(\mathbf{x}') ||_\mathrm{hidden} \leq Lip^\mathrm{ub} || \mathbf{x} - \mathbf{x}' ||_\mathrm{input}. 
\label{eq:bi_Lipschitz_constraint}
\end{equation}
where $Lip^\mathrm{lb}$ and $Lip^\mathrm{ub}$ are, respectively, the lower and upper bounds imposed on the Lipschitz constant of the feature extractor $h_\mathrm{nn}(\cdot)$, and $||\cdot ||_\mathrm{input}$ and $||\cdot ||_\mathrm{hidden}$ are, respectively, the distance metrics chosen for the input and hidden spaces. Setting the lower bound $Lip^\mathrm{lb}$ ensures that  latent representations are distance sensitive, i.e., if $\mathbf{x}$ and $\mathbf{x}'$ are relatively far apart in the input space, they also have a  relatively large distance in the hidden space. This sensitivity regularization allows the feature extractors to preserve input distances and directly counteracts the feature collapse issue by preventing OOD points from overlapping with in-distribution feature representations. Setting the upper bound $Lip^\mathrm{ub}$ ensures that hidden representations are smooth, i.e., small distance changes in the input space do not result in drastically large distance changes in the hidden space. This smoothness enforcement leads to feature extractors that generalize well and are robust to adversarial attacks. As for the distance metric, the Euclidean distance $dist(\cdot, \cdot)$ is often a good choice for measuring distances between input points and even those between hidden representations, except for image-like data. The Euclidean distance has recently been adopted as the distance metric in several deterministic UQ methods \citep{liu2020simple, van2020uncertainty,fortuin2021deep}. 

The feature-space regularization via a bi-Lipschitz constraint shown in Eq. (\ref{eq:bi_Lipschitz_constraint}) can be implemented during model training by applying either of the following two methods: (1) gradient penalty, originally introduced for training generative adversarial networks (GANs) \citep{gulrajani2017improved} and then adopted for deterministic uncertainty estimation \citep{van2020uncertainty}, and (2) spectral normalization, originally proposed again for training GANs \citep{miyato2018spectral} and then adopted for deterministic uncertainty estimation \citep{liu2020simple, fortuin2021deep, mukhoti2021deterministic, van2021feature}. In the rest of this subsection, we will briefly go over the application of spectral normalization in SNGP. We will also discuss the use of GPR as the output layer by SNGP to produce an uncertainty estimate based on distances in the ``regularized” hidden space.

\subsubsection{Spectral normalization for distance preservation in hidden space}
\label{sec:spectral_normalization}

The algorithm of SNGP enforces the lower bound of the Lipschitz constant in Eq. (\ref{eq:bi_Lipschitz_constraint}) simply by using network architectures with residual connections (e.g., residual networks) while imposing the upper bound using spectral normalization. Briefly, for each hidden layer, spectral normalization first calculates the spectral norm of the weight matrix $\mathbf{W}$ (i.e., the largest singular value of $\mathbf{W}$), denoted as $||\mathbf{W}||_2$, and then normalizes $\mathbf{W}$ using its spectral norm as:
\begin{equation}
\widehat{\mathbf{W}}_\mathrm{sn} = \frac{\gamma \cdot \mathbf{W}}{\lVert \mathbf{W} \rVert_2}, 
\label{eq:spectral_weight_normalization}
\end{equation}
where $\gamma$ is the upper bound of the spectral norm (i.e., $\lVert\mathbf{W}\rVert_2 \leq \gamma$), also called the \emph{spectral norm upper bound}, which effectively enforces an upper bound on the Lipschitz constant of the mapping function in the hidden layer. The weight matrix needs to be spectral-normalized only when its spectral norm exceeds the upper bound, i.e., when $||\mathbf{W}||_2 > \gamma$ \citep{liu2020simple}. Introducing the spectral norm upper bound gives rise to the flexibility to balance the expressiveness and distance awareness of the resulting spectral-normalized feature extractor. Specifically, when $\gamma$ takes a small value ($\gamma < 1$), the feature extractor tends to contract toward identity mapping, thereby limiting the ability of the feature extractor to learn complex nonlinear mapping, critically important for achieving high prediction accuracy on the training distribution; when $\gamma$ is large ($\gamma \gg 1$), the feature extractor is allowed to expand and be more expressive but may not preserve input distances. However, in reality, this flexibility may become a limitation against adoption, as $\gamma$ needs to be carefully tuned to balance accuracy/generalizability and distance awareness. 

\subsubsection{Gaussian process regression output layer for distance-aware prediction}
\label{sec:GPR_output_layer}

As discussed in Sec. \ref{sec:spectral_normalization}, the feature extraction layers of a neural network can be encouraged to preserve distances in the input space through a combination of residual connections and spectral normalization. Now we can make the predictive uncertainty of this neural network (input) distance-aware by replacing the last (output) layer with a GPR model that takes the learned hidden features as the input. Let us start by using the squared exponential kernel in Eq. (\ref{eq:sekernel}) as the base kernel. We replace the input points $\mathbf{x}$ and $\mathbf{x}'$ with their ``distance-aware" feature representations in the hidden space, $h_\mathrm{nn}(\mathbf{x};\bm{\uptheta})$ and $h_\mathrm{nn} (\mathbf{x’};\bm{\uptheta})$, where $h_\mathrm{nn}(~\cdot~;\bm{\uptheta})$ denotes the feature extraction part of a neural network parameterized by $\bm{\uptheta}$, i.e., the neural network up to the last hidden layer. The resulting kernel takes the following form:
\begin{equation}
k_\mathrm{nn}(\mathbf{x}, \mathbf{x}') = k(h_\mathrm{nn}(\mathbf{x};\bm{\uptheta}), 
h_\mathrm{nn} (\mathbf{x}';\bm{\uptheta})) = \sigma_\mathrm{f}^2 \exp{\left(-\frac{\| h_\mathrm{nn}(\mathbf{x};\bm{\uptheta})-
h_\mathrm{nn} (\mathbf{x}';\bm{\uptheta})\|^2}{2l^2} \right)}.
\label{eq:sekernel_nn}
\end{equation} \\
When the neural network is a DNN (e.g., with $>$ 5 hidden layers), the above kernel can sometimes be called a \emph{deep kernel}. The prior and posterior derivations follow the standard procedures described in Secs. \ref{sec:basics_gpr}.c and \ref{sec:basics_gpr}.d. Essentially, we perform a GPR in the learned, distance-preserving feature space instead of the input space. The resulting GPR model yields the posterior variance of a test input $\mathbf{x}_*$ based on its Euclidean distances from all training points in the hidden space, leveraging the distance awareness property of GPR, extensively discussed in Secs. \ref{sec:basics_gpr}.b and \ref{sec:basics_gpr}.d, to make the output layer distance aware. Intuitively speaking, let us suppose $\mathbf{x}_*$ keeps moving away from the training distribution. The value of the hidden-space kernel between any training input $\mathbf{x}_i$ and $\mathbf{x}_*$, $k_\mathrm{nn}(\mathbf{x}_i, \mathbf{x}_*)$, will become smaller and smaller given the distance preservation property of $h_\mathrm{nn}(\cdot)$. At some point, this kernel value will quickly approach zero. As a result, the posterior variance at $\mathbf{x}_*$ will keep increasing and eventually approach its maximum value $\sigma_\mathrm{f}^2$. This scenario suggests the distance awareness property of SNGP makes it an ideal tool for OOD detection.

To make inference computationally tractable, SNGP applies two approximations to the GPR output layer: (1) expanding the GPR model into simpler Bayesian linear models in the space of random Fourier features and (2) approximating the resulting posterior via Laplace approximation \citep{liu2020simple}. It is noted that another deterministic UQ method named DUE also uses spectral normalization plus residual connections to encourage a bi-Lipschitz mapping to the hidden space and GPR in the output layer. The only major difference is that DUE uses variational inducing point approximation for GPR in place of the random Fourier feature expansion \citep{van2021feature}. 

\subsubsection{Discussion on deterministic UQ methods}
\label{sec:discussion_deterministic_methods}

Deterministic methods run only a single forward pass for UQ and are computationally more attractive than BNN and neural network ensemble. These deterministic approaches also thrive at OOD detection thanks to their distance awareness property. However, they typically cannot separate aleatory and epistemic uncertainty. Additionally, they may require modifications to the network architecture (e.g., adding residual connections to enforce the Lipschitz lower bound in SNGP \citep{mukhoti2021deterministic,
liu2020simple}) and training procedure (e.g., to accommodate spectral normalization) with additional hyperparameters (e.g., the spectral norm upper bound $\gamma$, length scale $l$, and signal amplitude $\sigma_\mathrm{f}$). Finally, it was reported that deterministic methods such as SNGP may produce substantially lower-accuracy UQ (e.g., higher values of the ECE defined in Sec. \ref{sec:calibration_metrics}) than more mature methods such as MC dropout and neural network ensemble \citep{postels2021practicality,van2022benchmarking}. Findings from these recent benchmarking studies call for more effort to investigate the calibration performance of deterministic approaches and, in particular, to evaluate how accurately the predictive uncertainty can be used as a proxy for model accuracy for in-distribution, around-distribution, and OOD data.

\subsection{Toy example}
\label{sec:toy_example}
Following the above discussions on several popular methods for UQ of ML models, we now consider a toy 2D regression problem to compare the performance of these UQ methods quantitatively. The functional relationship  between $y$ and $\bf{x}$ underlying this toy example takes the following form: $y({\bf{x}}) = {1 \over {20}}({(1.5 + {x_1})^2} + 4) \times (1.5 + {x_2}) - \sin {{5 \times (1.5 + {x_1})} \over 2}$. To train an ML model, we randomly generate 800 samples from the following two bivariate Gaussian distributions, with 400 samples randomly drawn from either distribution, and use these 800 samples as the training data. 
\begin{equation}\label{eq:bivariate_distributions}
\mathcal{N}\left( {\left[ \begin{array}{l}
8\\
3.5
\end{array} \right],\left[ {\begin{array}{*{20}{c}}
{0.4}&{ - 0.32}\\
{ - 0.32}&{0.4}
\end{array}} \right]} \right), \; \mathcal{N}\left( {\left[ \begin{array}{l}
-2.5\\
-2.5
\end{array} \right],\left[ {\begin{array}{*{20}{c}}
{0.4}&{ - 0.32}\\
{ - 0.32}&{0.4}
\end{array}} \right]} \right).
\end{equation}

\begin{figure}[!ht]
    \subfigure[GPR]{\includegraphics[width=0.3\textwidth]{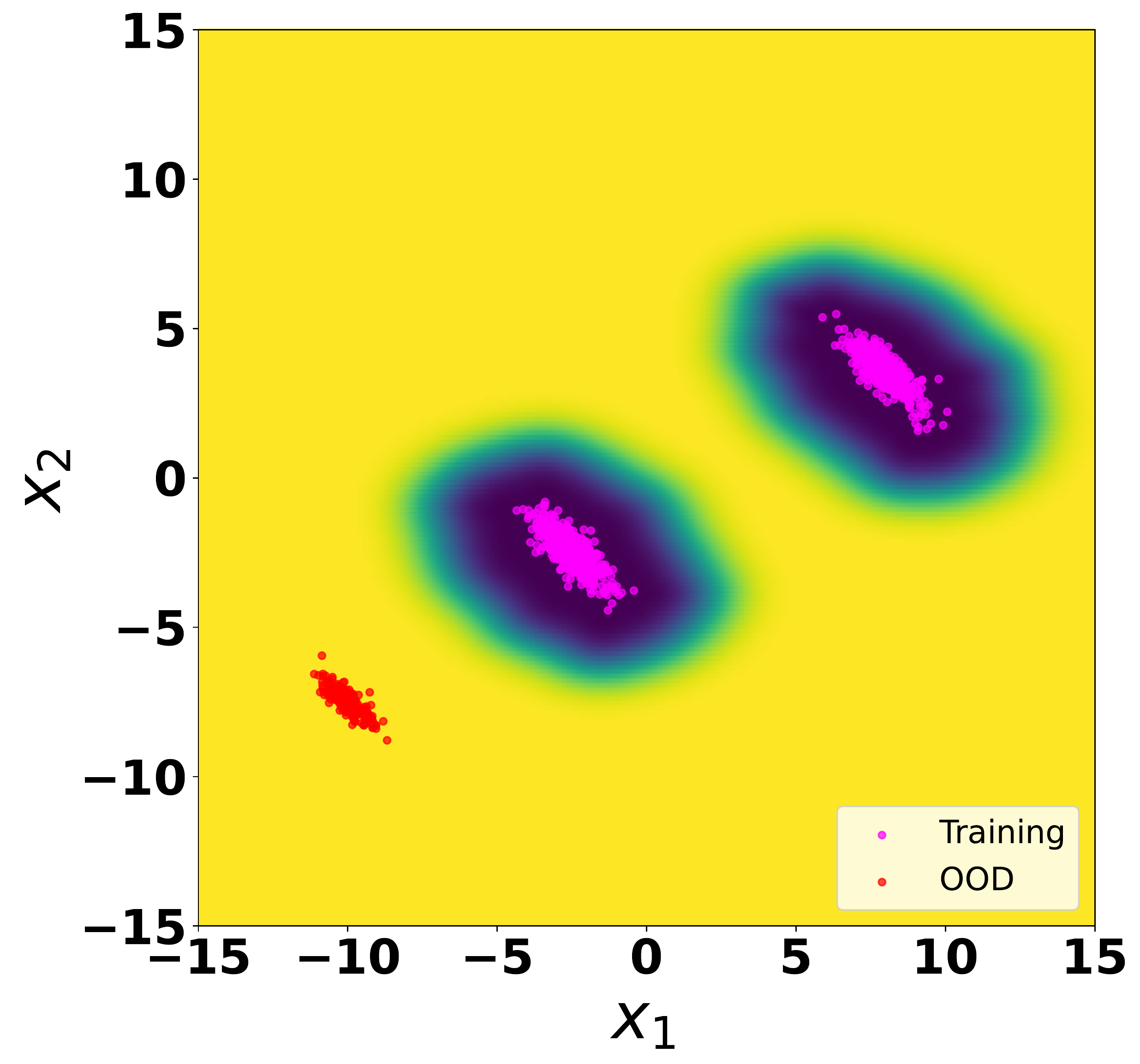}}
    \subfigure[MFVI]{\includegraphics[width=0.3\textwidth]{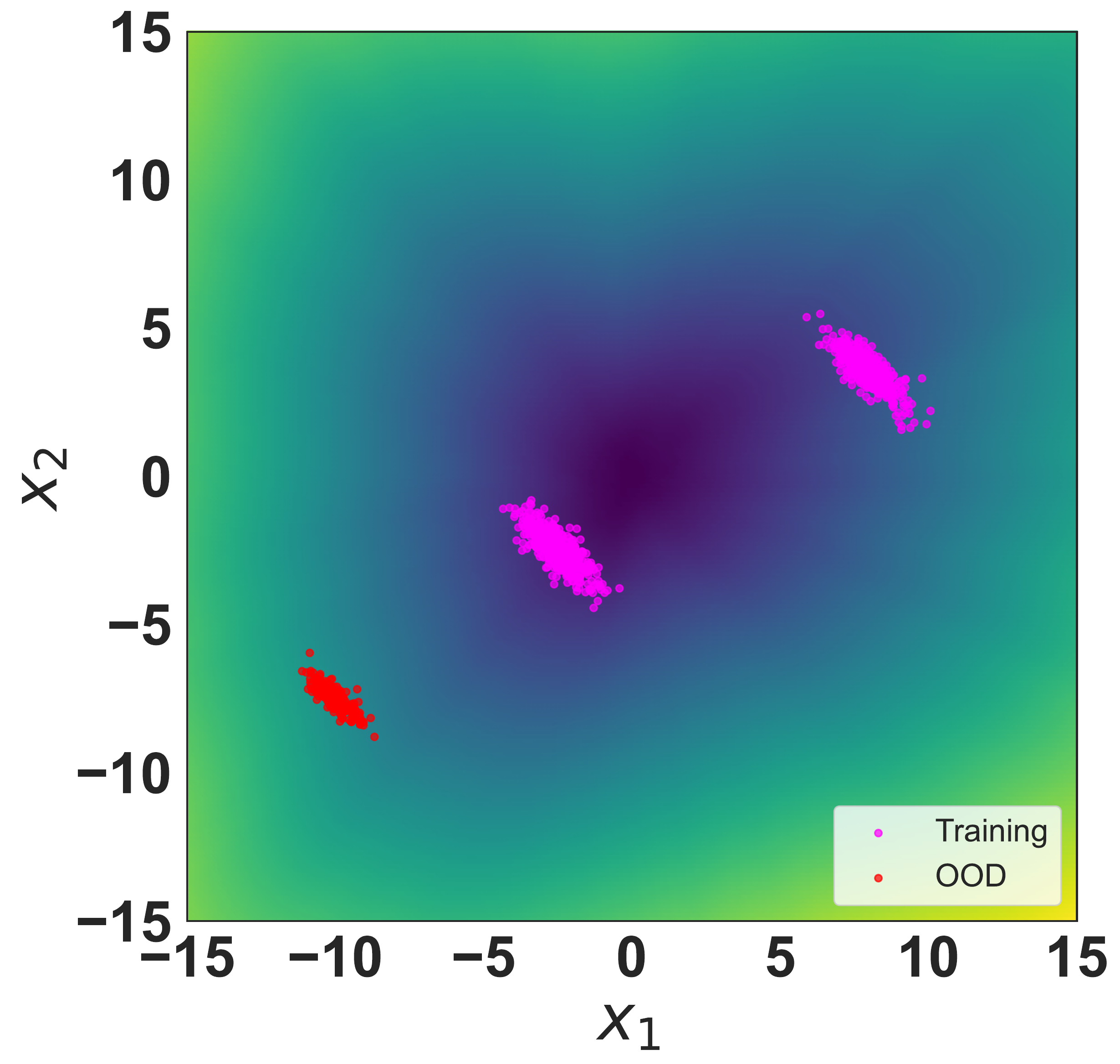}} 
    \subfigure[MC dropout]{\includegraphics[width=0.3\textwidth]{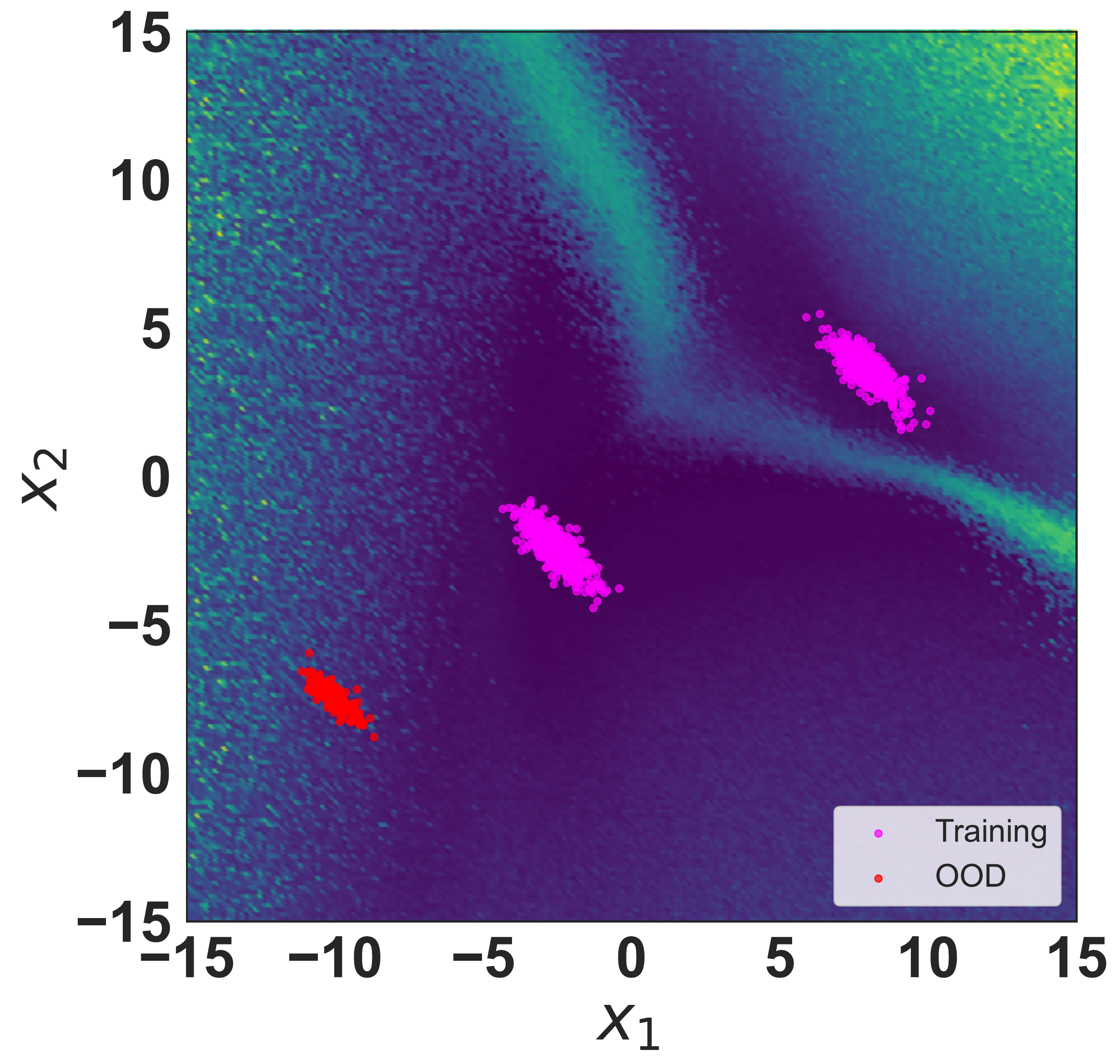}}
    
    \subfigure[Neural network ensemble]{\includegraphics[width=0.3\textwidth]{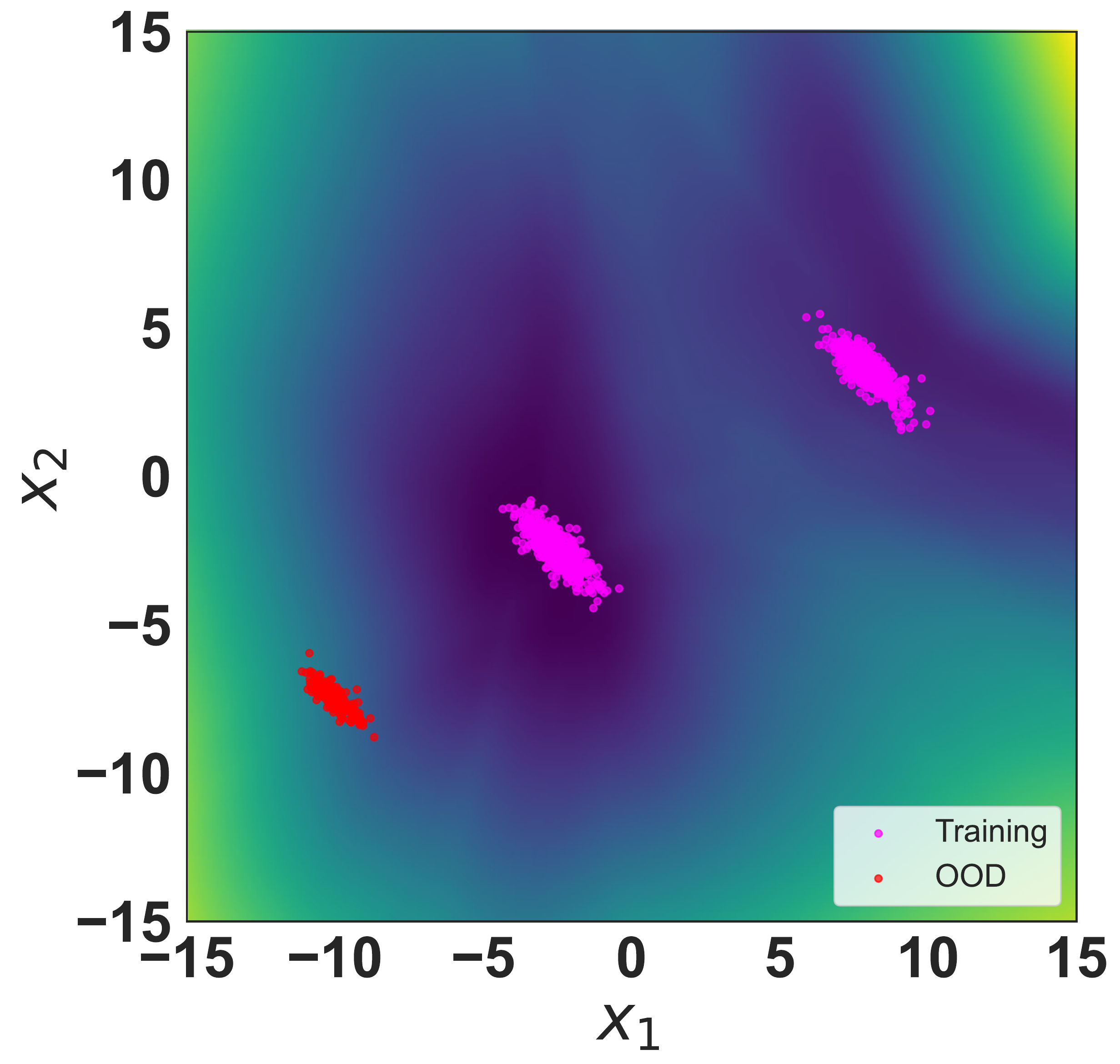}} 
    \subfigure[DNN-GPR]{\includegraphics[width=0.3\textwidth]{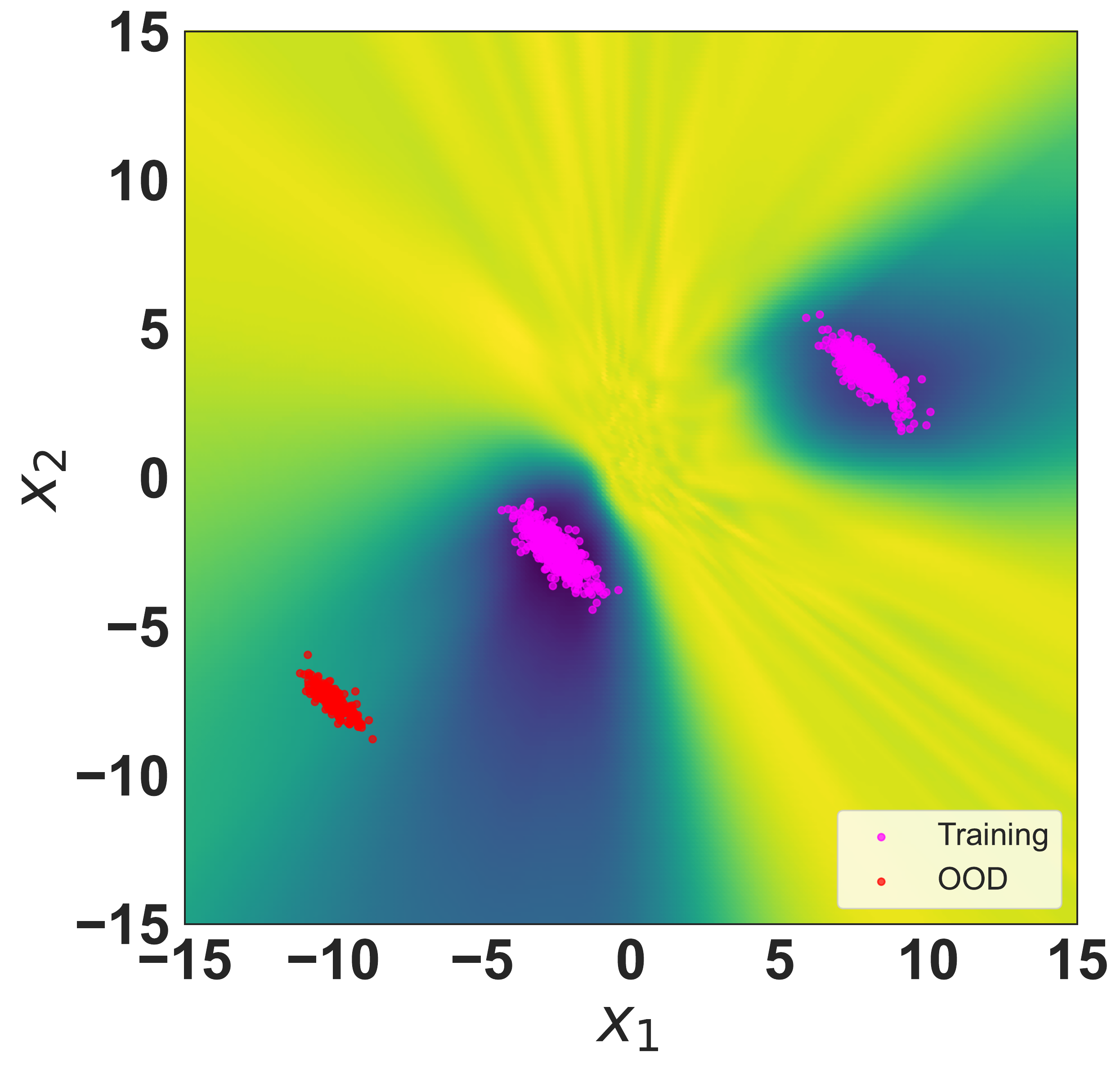}}
    \subfigure[SNGP]{\includegraphics[width=0.34\textwidth]{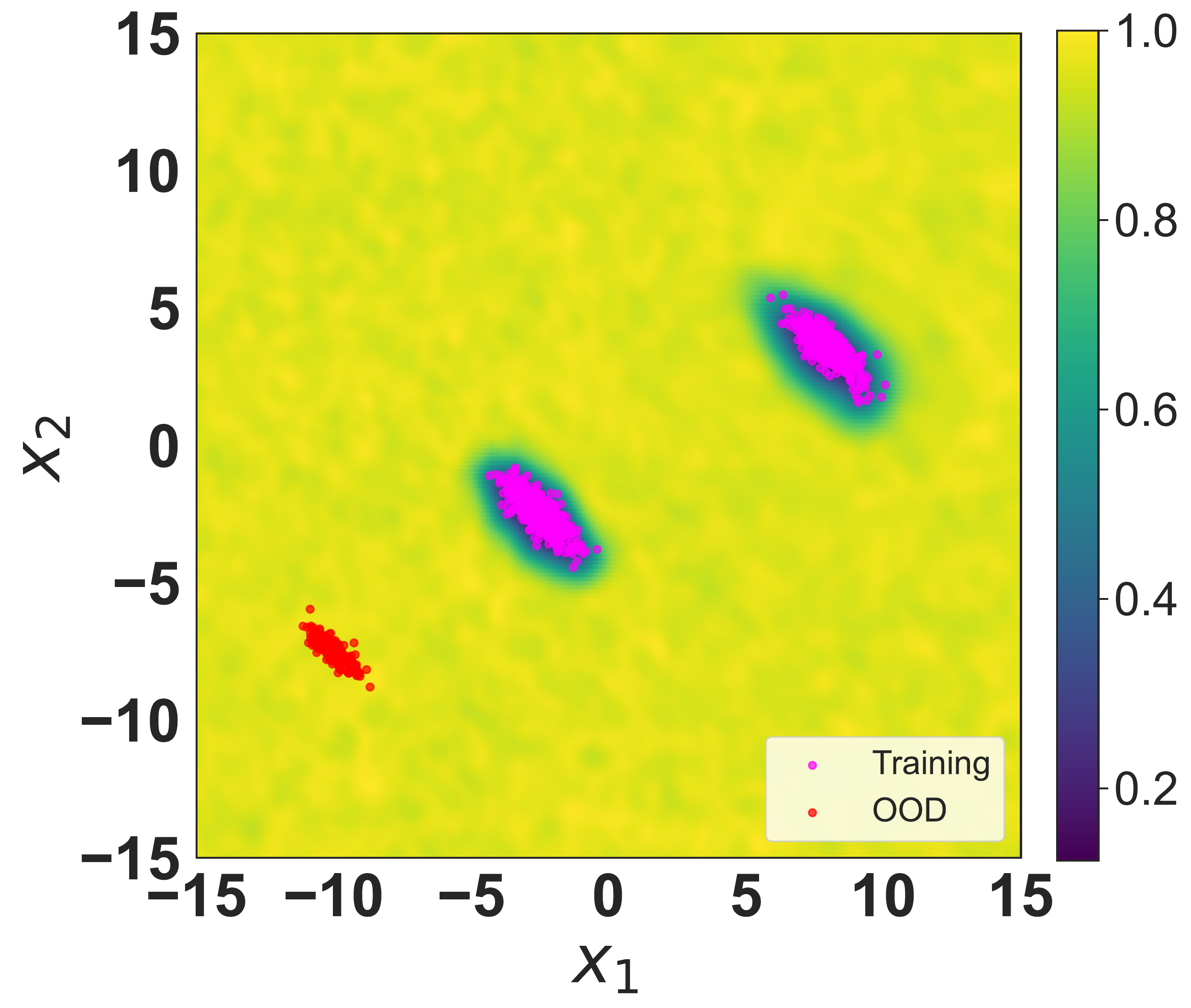}}
 
    \caption{The uncertainty maps by five different methods for UQ of ML models on the toy 2D regression problem. These methods are Gaussian process regression -- GPR (a), MFVI -- mean-field variational inference (b), Monte Carlo dropout -- MC dropout (c), neural network ensemble (d), deep neural network with Gaussian process regression -- DNN-GPR (e), Spectral-normalized Neural Gaussian Process -- SNGP (f). The two clusters colored in purple represent the training data, while the cluster colored in red indicates a cluster of OOD instances. The background in each 2D plot is color-coded according to the predictive uncertainty by the corresponding UQ method, with yellow (blue) indicating high (low) uncertainty.}
    \label{fig:uq_comparison}
\end{figure}

These training samples form two separate clusters with no overlap in between, as shown in Fig. \ref{fig:uq_comparison}. As can be observed in both Eq. (\ref{eq:bivariate_distributions}) and Fig. \ref{fig:uq_comparison}, the two clusters have an identical variance-covariance matrix and differ significantly only in the mean vector. We now apply the previously introduced UQ methods on the 800 training samples. For those methods requiring neural networks, the UQ methods are built on a backbone of similar residual neural network architectures with four 64-neuron residual layers. For example, in the case of neural network ensemble, a Gaussian layer is inserted at the end of a residual neural network; while in the case of MC dropout, dropout with a rate of 0.2 is applied at the end of each residual layer.


To test the UQ performance of different ML models, we generate a uniform meshgrid consisting of 40,000 ($=200\times200$) samples with $x_1$ and $x_2$ spanning in the range $\left[-15, 15 \right]$. Next, an uncertainty heap map is constructed to visualize the predictive uncertainty of each trained ML model within the domain. Figure~\ref{fig:uq_comparison} shows the uncertainty heat maps obtained by the five different UQ methods on this toy problem. At a quick glance, both GPR and SNGP exhibit a desirable behavior in producing high quality predictive uncertainty: the predictive uncertainty is quite low for samples in the proximity of the in-distribution/training data (dots in pink color). At the same time, both GPR and SNGP generate high predictive uncertainty when test sample point $[x_1,x_2]^\text{T}$ moves far away from the training data clusters. As a result, both GPR and SNGP successfully assigned high uncertainty to the 200 OOD samples (dots in red color at the bottom left of Fig.~\ref{fig:uq_comparison}) - which are randomly generated to test the OOD detection capability of different UQ techniques. 

Unlike GPR and SNGP, the other four UQ methods have a relatively poor performance in quantifying predictive uncertainty. As can be observed in Fig.~\ref{fig:uq_comparison} (c-e), MC dropout, deep ensemble, and DNN-GPR assign low uncertainty for samples that are quite far away from the training data. As a consequence, these three UQ techniques are likely to fail to detect the 200 OOD samples whose predictions are associated with relatively low uncertainty, as shown in the bottom-left corners of Fig.~\ref{fig:uq_comparison} (c), (d), and (e). Besides the lack of ability in OOD detection, these three UQ techniques share another feature in common: their uncertainty output is more sensitive to the (hypothetical) boundary that separates the two clusters of training data, while they exhibit a substantially faulty behavior when establishing the decision boundary (trustworthy vs. untrustworthy region) around each cluster of training data itself. More specifically, for a given test sample, the predictive uncertainty generated by these three UQ techniques has a low sensitivity to how distant is a test sample's distribution with respect to the training data. Regarding the mean-field variational inference (MFVI), its predictive uncertainty gets increased in accordance with the distance away from the two training clusters, however, MFVI assigns nearly an identical uncertainty for the data between the two training clusters as they are near the data, which contradicts with our anticipation. This suggests that MFVI suffers from the lack of in-between uncertainty due to the approximation to Bayesian inference, and such finding is also confirmed by~\citet{foong2020expressiveness}. Consequently, the predictive uncertainty by these UQ techniques is unprincipled because their quantified uncertainty does not match our expectation that uncertainty should clearly distinguish in-domain and out-domain data.

The significant difference in the uncertainty heat map across different UQ methods is primarily attributed to their distance awareness capability. MC dropout, deep ensemble, and DNN-GPR do not have the ability to properly quantify the distance of an input sample away from the training data manifold. Instead, the predictive uncertainty at an input sample quantified by MC dropout, deep ensemble, and DNN-GPR seems to be established upon the distance of the input sample from a decision boundary separating the two clusters of training data. Therefore, it is not surprising to see all these three UQ methods assign low uncertainty to the 200 OOD samples even though they are quite far from the training data. Distinct from MC dropout and deep ensemble, GPR, DNN-GPR, and SNGP are equipped with a good sense of awareness with respect to the distance between an input sample and the training data manifold. As a result, they are comparatively more principled in the sense that the uncertainty is much higher for the input sample that lies far from the training data. Finally, even though both DNN-GPR and SNGP have GPR as the output layer, DNN-GPR is free from determining what information to discard in the hidden space, while SNGP imposes a spectral normalization on the latent representation of the input sample, thus making the output layer distance sensitive in the hidden space. In a broad context, the sound UQ by GPR and SNGP substantially facilitates the identification of OOD samples, establishing a trustworthy region in the input space where ML predictions are reliable.

\subsection{Summary}

\begin{table}[!ht]\footnotesize
    \centering
    \caption{A qualitative comparison of state-of-the-art UQ approaches covered in this tutorial}
    \begin{threeparttable}
    \begin{tblr}{Q[l, m, 2.4cm]|Q[c, m, 1.5cm]|Q[c, m, 1.8cm]|Q[c, m, 1.7cm]|Q[c, m, 1.7cm]|Q[c, m, 1.7cm]|Q[c, m, 1.5cm]|Q[c, m, 1.5cm]}
    \hline \hline
    \SetCell[r=2,c=1]{c} Quantity of interest & \SetCell[r=2,c=1]{c} Gaussian process regression
     & \SetCell[c=3]{c} Bayesian neural network
     &  & & \SetCell[r=2,c=1]{c} Neural network ensemble & \SetCell[c=2]{c} Deterministic method \\
    \hline
     & & \SetCell[r=1,c=1]{c} MCMC & \SetCell[r=1,c=1]{c} {Variational  inference} & \SetCell[r=1,c=1]{c} {MC dropout} & & \SetCell[r=1,c=1]{c} DNN-GPR & \SetCell[r=1,c=1]{c} SNGP \\
    \hline
        Quality of UQ (e.g., measured by  calibration curve)  &	High  &	High-medium$^\text{a}$  & Medium  &	Medium-low & High & Medium & High \\
        \hline
        Computational cost (training)  &	High$^\text{b}$  &	High &	High-medium &	Low & Low & High & High\\
        \hline
        Computational efficiency (test)  &	High$^\text{b}$  & Low & High-medium	 &	Medium & Medium-low & Low & Low \\
        \hline
        Ability to detect OOD samples &	Strong & 	Weak &	Weak  & Weak &  Moderate & Strong-moderate & Strong \\
        \hline
        Scalability to high dimensions & Low & 	Low & Medium & High & High & High & High \\
        \hline
        Effort to convert a deterministic to a probabilistic model & Not applicable & High & High-medium & Low & Medium & High-medium & High-medium\\
        \hline
        Ability to distinguish aleatory and epistemic uncertainty & Yes & Yes & Yes & No & Yes & No & No\\
        \hline
        Basis of UQ & Analytical & Sampling & Sampling & Sampling & Hybrid & Analytical & Analytical\\
        \hline
        Stability of quantified uncertainty to parameter initialization & High & High & High & Low & Medium & High & High\\
    \hline \hline
    \end{tblr}
    \begin{tablenotes}
          \item[a] Accuracy is largely affected by the quality of the assumed prior. \\
          \item[b] Efficient only for problems of low dimensions (typically $<$ 10) and small training data (typically $<$ 5000 points). \\
    \end{tablenotes}
    \end{threeparttable}
    \label{tab:uq_of_ml_comp}
\end{table}

The numerical example in Sec.~\ref{sec:toy_example} demonstrates the performance difference among different UQ methods with an emphasis on OOD detection. Comprehensive comparison of these UQ methods may help better guide users to select appropriate UQ methods for specific ML applications. To this end, we construct a table (Table~\ref{tab:uq_of_ml_comp}) to qualitatively compare these methods along multiple dimensions, such as the quality of UQ, computational costs in training and test, etc. In the first place, regarding the calibration accuracy of these UQ methods, GPR and SNGP generally outperform other alternate UQ methods, which is also confirmed in the previous numerical example. For the computational cost associated with training an ML model, implementing a Bayesian neural network via MCMC or variational inference incurs a relatively higher computational cost than MC dropout, as MC dropout consumes nearly the same amount of computational time as training a regular neural network.
In terms of scalability, it is well-known that GPR suffers from the curse of high dimensionality, so training and testing GPR models may be computationally very expensive for high-dimensional problems. The other three UQ methods (neural network ensemble, DNN-GPR, and SNGP) are computationally cheaper than GPR, MCMC, and variational inference. We have similar findings regarding the computational burden of these UQ methods at test time.

An important function of UQ built atop the original deterministic ML model is to serve as a safeguard to detect OOD samples for the purpose of increasing the reliability of ML models. In this regard, SNGP achieves similar performance as the gold standard GPR, while the remaining UQ methods may perform poorly in detecting OOD samples. Besides strong OOD detection capability, SNGP also exhibits a desirable feature in scalability, while such a feature is missing in GP. However, compared to GP, SNGP requires an additional effort to turn a deterministic ML model into a probabilistic counterpart for UQ, while GPR is born with the capability of UQ. As for the uncertainty decomposition, GPR, Bayesian neural network, and neural network ensemble all have some capability to quantify aleatory and epistemic uncertainty separately, while such a capability may be lacking in the MC dropout version of Bayesian neural network as well as in DNN-GPR and SNGP. Next, both GPR and SNGP estimate the predictive uncertainty of ML models in an analytical form. In contrast, the other UQ methods draw Monte Carlo samples to approximate the uncertainty, which is a major performance barrier if critical applications require real-time inferences. 


\section{Evaluation of predictive uncertainty}
\label{sec:uncertainty_evaluation}
Let us now shift our focus to the performance evaluation of probabilistic ML models. A unique property of these models is that they do not simply produce a point estimate of $y$ and instead output a probability distribution of $y$, $p(y)$, that fully characterizes the predictive uncertainty. This unique property requires that the performance evaluation examines both the prediction accuracy, e.g., the RMSE or mean absolute error calculated based on the mean predictions for regression, and the quality of predictive uncertainty, e.g., how accurately the predictive uncertainty reflects the deviation of a model prediction from the actual observation. In what follows, we will discuss ways to assess the quality of predictive uncertainty.

\subsection{Calibration curves and metrics}
\label{sec:calibration}

A standard approach to assessing the quality of predictive uncertainty is creating a calibration curve, also called a reliability diagram \citep{degroot1983comparison,zadrozny2002transforming,niculescu2005predicting}. We will first give a detailed walkthrough of creating calibration curves for regression and classification and then present UQ performance metrics that can be derived from a calibration curve. 

\subsubsection{Calibration curves for regression}
\label{sec:calibration_regression}

Let us assume, in a regression setting, that we have a validation/test set of $N$ input-output pairs, $\mathcal{D} = \left\{ {\left( {{\mathbf{x}_1},{y_1}} \right),\left( {{\mathbf{x}_2},{y_2}} \right), \cdots ,\left( {{\mathbf{x}_N},{y_N}} \right)} \right\}$. Given a trained probabilistic ML model (e.g., an ensemble of probabilistic neural networks or simply called a neural network ensemble as discussed in Sec. \ref{sec:neuralnetworkensemble}) parameterized by $\bm{\uptheta}$, let $\widehat{y}_i = f \left( \mathbf{x}_i; \bm{\uptheta} \right)$ denote the predicted outcome for the $i$-th validation/test sample $\mathbf{x}_i$, $i=1, \cdots, N$. Without loss of generality, let us further assume that the probabilistic output $\widehat{y}_i$ follows a Gaussian distribution, characterized by a Gaussian probability density function, $p\left(\widehat{y}_i; \mu_{\bm{\uptheta}} \left(\mathbf{x}_i \right), \sigma_{\bm{\uptheta}} \left( \mathbf{x}_i \right) \right) = {\frac{1}{\sigma_{\bm{\uptheta}} \left( \mathbf{x}_i\right)}} \phi \left(\frac{\widehat{y}_i-\mu_{\bm{\uptheta}} \left(\mathbf{x}_i \right)}{\sigma_{\bm{\uptheta}} \left( \mathbf{x}_i\right)}\right)$, with the predicted mean $\mu_{\bm{\uptheta}} \left(\mathbf{x}_i \right)$ and standard deviation $\sigma_{\bm{\uptheta}} \left(\mathbf{x}_i \right)$. For a given confidence level $c \in [0, 1]$, we can easily derive a two-sided 100$c$\% confidence interval for the Gaussian random variable $\widehat y_{i}$ as:

\begin{equation}
CI_i^c = \left[ {{\mu _{\bm{\uptheta}} }\left( {{x_i}} \right) - z_{\frac{1+c}{2}} \sigma_{\bm{\uptheta}} \left(\mathbf{x}_i \right),{\mu _{\bm{\uptheta}} }\left( {{x_i}} \right) + z_{\frac{1+c}{2}} \sigma_{\bm{\uptheta}} \left(\mathbf{x}_i \right)} \right],
\end{equation}
where $z_{\frac{1+c}{2}}$ denotes the $\left(\frac{1+c}{2}\right)^\text{th}$ quantile of the standard normal distribution, i.e., $z_{\frac{1+c}{2}} = \Phi ^{{-1}}\left(\frac{1+c}{2}\right)$, with $\Phi(\cdot)$ denoting the cumulative distribution function (CDF) of the standard normal distribution. The probability of a random realization of $\widehat{y}_i$ falling into $CI_i^c$ equals $c$, expressed as
\begin{equation}
\int_{\mu_{\bm{\uptheta}} \left(x_i \right) - z_{\frac{1+c}{2}} \sigma_{\bm{\uptheta}} \left(\mathbf{x}_i \right)}^{\mu_{\bm{\uptheta}} \left(x_i \right) + z_{\frac{1+c}{2}} \sigma_{\bm{\uptheta}} \left(\mathbf{x}_i \right)} {p\left( {\widehat{y}_i;{\mu _{\bm{\uptheta }}}\left(\mathbf{x}_i \right),{\sigma _{\bm{\uptheta}} }\left( {\mathbf{x}_i} \right)} \right)} d\widehat{y}_i \underset{\tau \equiv {\left(\frac{\widehat{y}_i-\mu_{\bm{\uptheta}} \left(\mathbf{x}_i \right)}{\sigma_{\bm{\uptheta}} \left( \mathbf{x}_i\right)}\right)}} =  \int_{-z_{\frac{1+c}{2}}}^{z_{\frac{1+c}{2}}} {\phi \left(\tau\right)} d\tau = c.
\end{equation}

If we choose to use a CDF $P_i$ to characterize the probability distribution of $\widehat{y}_i$ that may not follow a Gaussian distribution, we can write out the 100$c$\% confidence interval for any arbitrary distribution type,
\begin{equation}\label{eq:confidence_estimation}
CI_i^c = \left[ {P_i^{ - 1}\left( {\frac{{1 - c}}{2}} \right),P_i^{ - 1}\left( {\frac{{1 + c}}{2}} \right)} \right],
\end{equation}
where $P_i^{ - 1}\left( c \right) = \inf \left( {{{\widehat y}_i}:P_i^{ - 1}\left( {{{\widehat y}_i}} \right) \ge c} \right)$. Here, $P_i^{ - 1}$ is an inverse of the CDF $P_i$, also called a quantile function, and becomes $\Phi ^{{-1}}$ for the standard normal distribution. Alternatively, we can derive a one-sided confidence interval $CI_i^c = \left[ { - \infty ,P_i^{ - 1}\left( c \right)} \right]$.

Ideally, the UQ of this ML model should yield a 100$c$\% confidence interval that contains the observed $y$ for approximately 100$c$\% of the time. For example, if $c$ = 0.95, then $y_i$ should fall into a 95\% confidence interval $CI^{0.95}_i$, one- or two-sided, for nearly 95\% of the time. In other words, we expect that approximately 95\% of the $N$ validation/test samples have their observed $y$ values fall into the respective 95\% confidence intervals. The fraction of validation/test samples for which the confidence intervals contain the observations can be called \textit{observed confidence} ($\hat{c}$) or sometimes \textit{accuracy}, expressed as $\widehat{c} = \frac{1}{N}\sum\limits_{i = 1}^N {\mathbb{I}\left( {{y_i} \in CI_i^c} \right)}$, where $\mathbb{I} \left(prop \right)$ is an indicator function that takes the value of 1 if the proposition $prop$ is true and 0 otherwise. If we plot observed confidence against \textit{expected confidence} ($c$) over $[0,1]$, we will create a calibration curve, sometimes called a reliability diagram (see an example in the right-most plot of Fig. \ref{fig:procedure_calibration_curve}). This calibration curve shows how well predictive uncertainty is quantified, and a perfect UQ should yield a calibration curve that overlaps with the diagonal line ($y = x$). If the observed confidence is higher than expected at some $c$ values, the model is said to be underconfident at these confidence levels; otherwise, the model is deemed overconfident. In predictive maintenance practices, reliability/maintenance engineers often prefer underconfident predictions over overconfident predictions, as overconfident predictions are more likely to trigger maintenance actions that are either unnecessarily early or too late. If 90\% or 95\% is chosen as the confidence level, it is preferred that the observed confidence (or accuracy) is very close to or slightly higher than 90\% or 95\%.

\begin{figure}[!ht]
    \centering
    \includegraphics[scale=1.2]{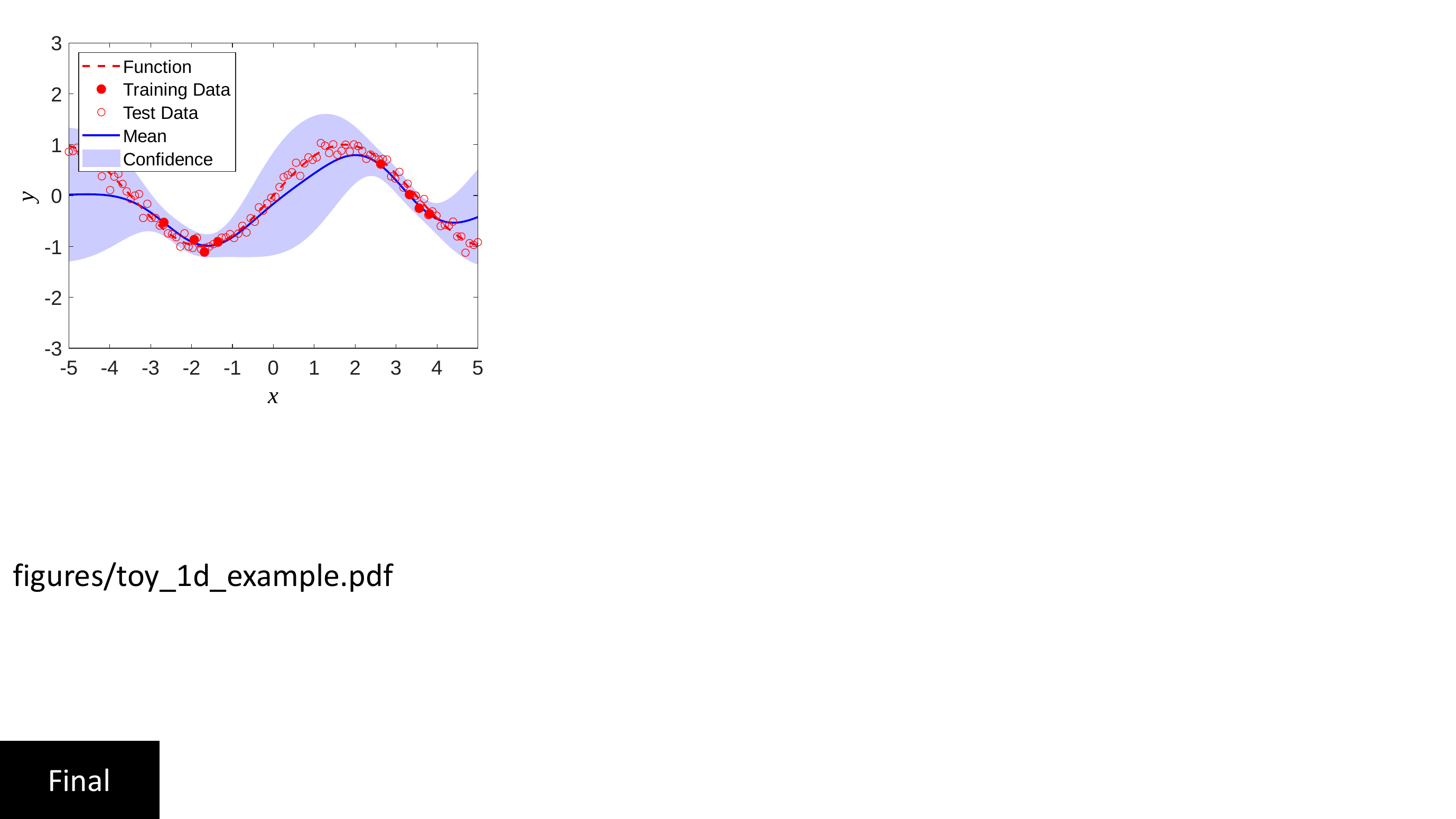}
    \caption{An example dataset with eight training samples (solid red circles) and 100 test samples (hollow red circles), plotted with the underlying one-dimensional function and fitted GPR model. Shown for the fitted GPR model is the poterior mean function (solid blue curve) and a collection of ~95\% confidence intervals (light blue shade) for the noisy observations ($\mathbf{y}_*$) at new/test points. These test points are equally spaced between -5 and 5 along the x-axis.}
    \label{fig:toy_1d_example}
\end{figure}

Let us now do a step-by-step walkthrough of how a calibration curve is created using a toy example. This example uses training and test data generated from the same 1D function and Gaussian observation model used to generate Figs. \ref{fig:samples_prior_posterior} and 
\ref{fig:hyperparameters_effect} in Sec. \ref{sec:basics_gpr}. The observation model consists of a sine function corrupted with a white Gaussian noise term, $y = \mathrm{sin}(0.9x) + \varepsilon
$ with $\varepsilon \sim \mathcal{N} \left(0, 0.1^2 \right)$. As shown in Fig.~\ref{fig:toy_1d_example}, we fit a GPR model to the eight training data points and test this model on 100 test points. It can be seen from the figure that the regressor reports high uncertainty at test points that fall outside of the $x$ ranges where training samples exist. If we compare the in-distribution test samples (i.e., whose $x$ values fall into $[-3,-1)$ or $[2,4)$) with the OOD samples (whose $x$ values lie within $[-5,-3)$, $[-1,2)$, or $[4,5)$), we observe higher predictive uncertainty on the OOD samples, where the model's predictions are more likely to be incorrect. Creating a calibration curve in this toy example consists of three steps.
\begin{enumerate}[label=\textbf{Step \arabic*:}, leftmargin=4.5em]
    \item We start by choosing $K$ confidence levels between 0 and 1, $0 \le c_1 < c_2 < \cdots < c_K \le 1$. In this example, we choose 11 ($K = 11$) confidence levels equally spaced between 0 and 1, i.e., 0, 0.1, $\cdots$, 0.9, 1 (see Step 1 in Fig.~\ref{fig:procedure_calibration_curve}). 
    
   \item  We then compute for each expected confidence level $c_j$ the observed confidence as:
   \begin{equation}
       {\widehat c_j} = \frac{1}{N}\sum\limits_{i = 1}^N {\mathbb{I}\left( {{y_i} \in CI_i^c} \right)}.
   \end{equation}
   As mentioned above, $CI_i^c = \left[ {P_i^{ - 1}\left( {\frac{{1 - c}}{2}} \right),P_i^{ - 1}\left( {\frac{{1 + c}}{2}} \right)} \right]$ for a two-sided confidence interval and $CI_i^c = \left[ { - \infty ,P_i^{ - 1}\left( c \right)} \right]$ for a one-sided confidence interval. Step 2 in Fig.~\ref{fig:procedure_calibration_curve} shows an example of how to implement Eq. (\ref{eq:confidence_estimation}) for $c_6=0.5$.
   
   \item We finally plot the $K$ pairs of expected vs. observed confidence, $\left\{ {\left( {{c_1},{{\widehat c}_1}} \right), \cdots ,\left( {{c_K},{{\widehat c}_K}} \right)} \right\}$, which gives rise to a calibration curve. In the toy example, we have 11 pairs of $\left(c_j, \widehat {c}_j \right)$ plotted to form a discrete calibration curve in Step 3 in Fig.~\ref{fig:procedure_calibration_curve}. 
\end{enumerate}

\begin{figure}[!ht]
    \centering
    \includegraphics[scale=0.84]{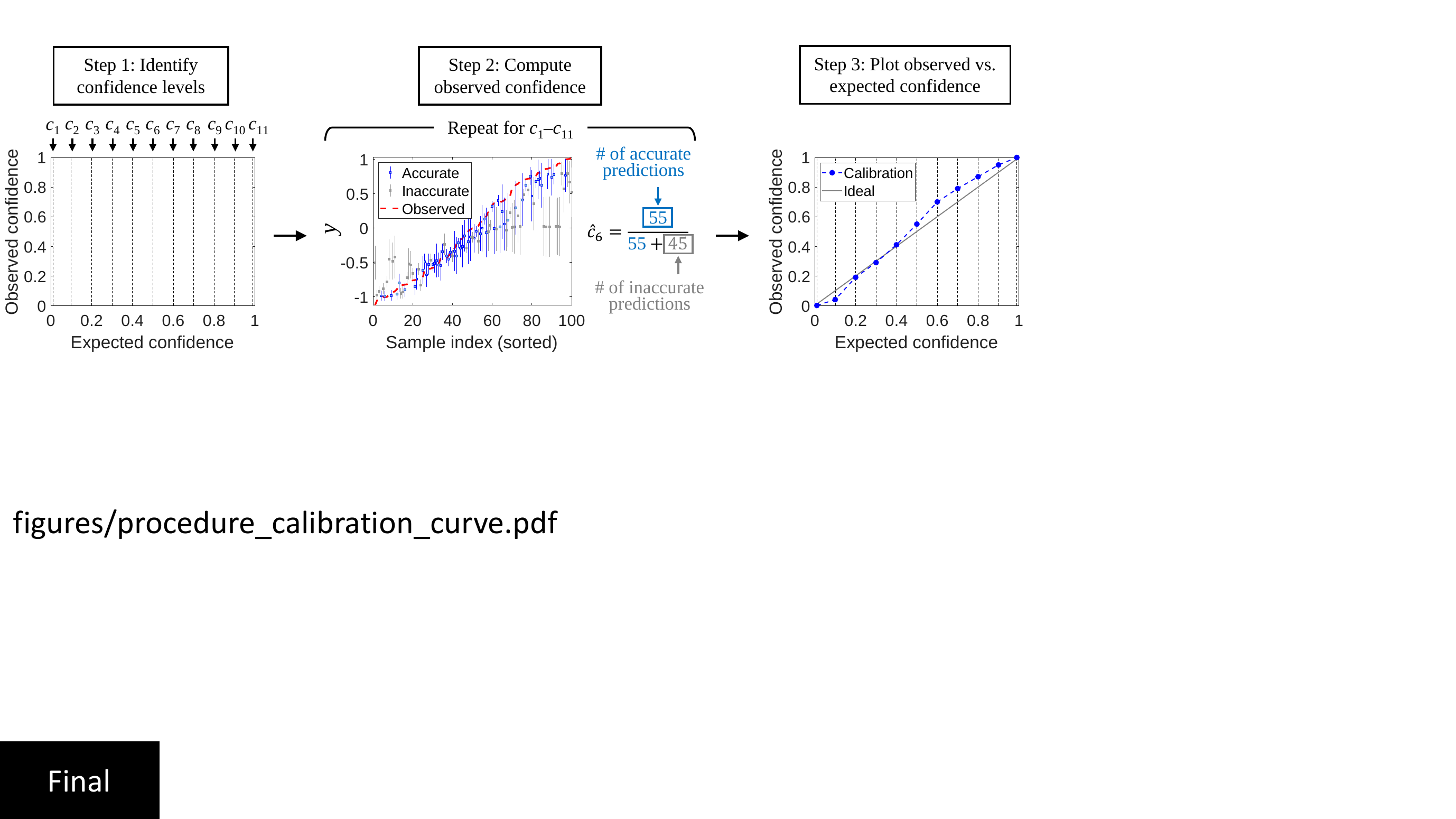}
    \caption{Illustration of three-step procedure to create a calibration curve for toy regression problem shown in Fig.~\ref{fig:toy_1d_example}.}
    \label{fig:procedure_calibration_curve}
\end{figure}

\begin{figure}[!ht]
    \centering
    \includegraphics[scale=1.2]{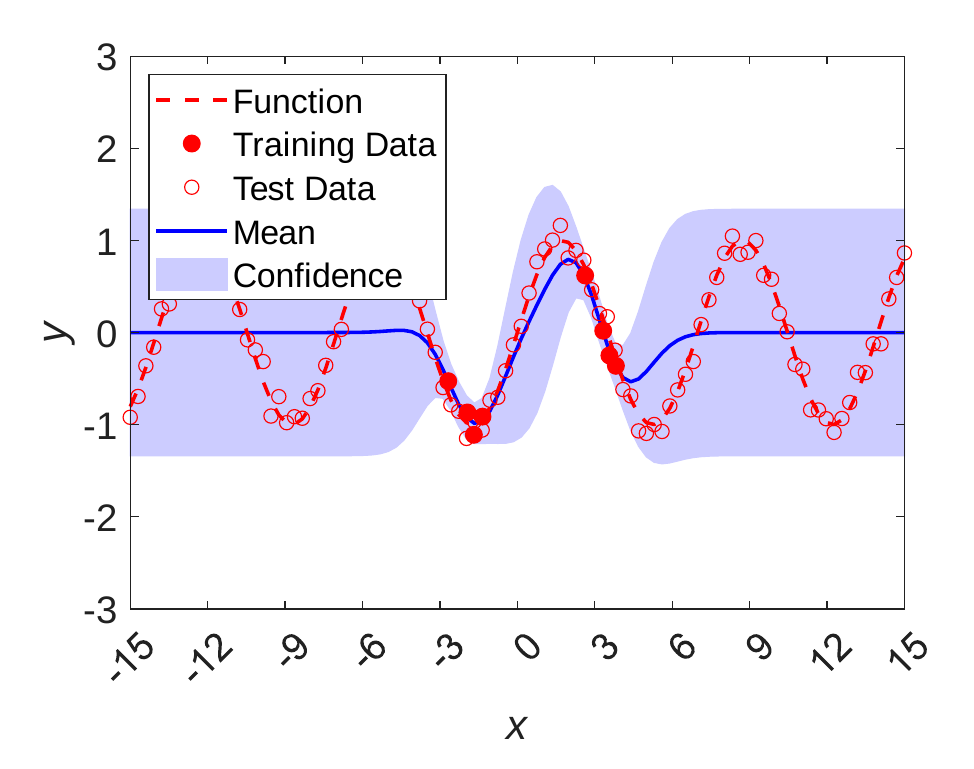}
    \caption{Toy example identical to the one in Fig. \ref{fig:toy_1d_example} but with an expanded range of $x$ on test data.}
    \label{fig:toy_1d_example_x_expanded}
\end{figure}

\begin{figure}[!ht]
    \centering
    \includegraphics[scale=1.2]{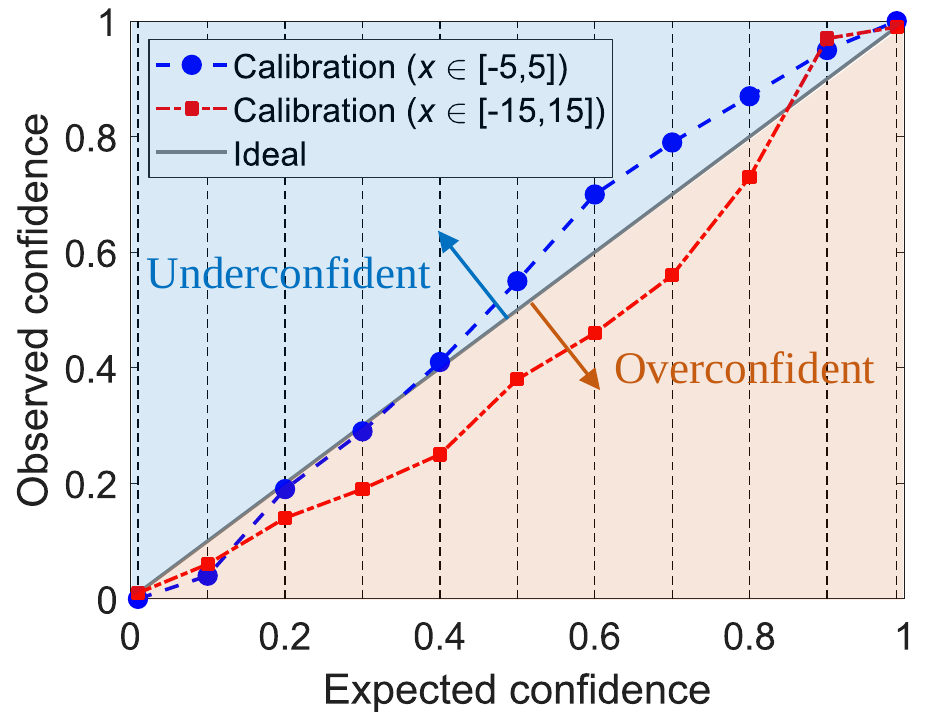}
    \caption{Comparison of calibration curves for two different ranges of test data for the toy 1D mathematical problem. Test samples are equally spaced between -5 and 5 (the same as Figs. \ref{fig:toy_1d_example} and \ref{fig:procedure_calibration_curve}) and between -15 and 15, respectively, for the two test ranges.}
    \label{fig:comparison_calibration_curves}
\end{figure}

Suppose we are interested in assessing the regression model’s UQ quality at the confidence level of 90\%. In that case, we can observe from the calibration curve drawn in Step 3 that the Gaussian process regressor tends to be underconfident, i.e., the confidence we expect the regressor to have ($c_{10}$ = 90\%) is lower than the observed (empirically estimated) confidence (${\widehat c}_{10} = 95\%$) or simply $c_{10} < {\widehat c}_{10}$. More specifically, the actual proportion of times that the model’s 90\% confidence interval contains the ground truth (i.e., the model is correct) is higher than the expected value (i.e., 90\%). Being underconfident also means that the model tends to produce higher-than-true uncertainty in its predictions, which is often more desirable in safety-critical applications than having an overconfident model.

To further understand how a calibration curve behaves as a test window varies, we expand the range of test data from $[-5, 5]$, as shown in Fig. \ref{fig:toy_1d_example}, to $[-15, 15]$, as shown in Fig. \ref{fig:toy_1d_example_x_expanded}, while keeping the same number of test samples (i.e., 100). As shown in Fig. \ref{fig:toy_1d_example_x_expanded}, the new test dataset includes much more OOD samples that fall outside the range of $[-5, 5]$. The calibration curve on this new dataset is plotted alongside the one on the original dataset in Fig. \ref{fig:comparison_calibration_curves}. Let us compare the new (red) and original (blue) calibration curves. We can observe that having more OOD test samples degrades the quality of UQ by moving the calibration curve further away from the ideal line. This observation is not surprising because high quality UQ (i.e., producing predictive uncertainty that accurately reflects prediction errors) is expected to be more challenging on OOD samples than in-distribution samples. Another interesting observation is that the GPR model appears more overconfident in making predictions on the new test dataset with more OOD samples. Our explanation for this observation is that as a test sample $x_i$ moves farther away from the training data, the prediction error may increase drastically (i.e., the model-predicted mean may deviate substantially more from the true observation), but the predictive uncertainty by a UQ method may start to saturate at a certain distance away from the training distribution (see, for example, the flat confidence bounds in Fig. \ref{fig:toy_1d_example_x_expanded} when $x_i \in [-15,-6] \cup [7,15]$), making it more difficult for a probabilistic prediction to be accurate (i.e., the predictive confidence interval at $x_i$ contains the ground truth ${y}_i$). Essentially, in some cases, the predictive uncertainty cannot catch up with the prediction error as a test sample moves further away from a training distribution. In that case, it is critically important to establish boundaries in the input space within which predictive uncertainty cannot be trusted. Very little effort has been devoted to trustworthy UQ, and more effort is urgently needed on this front.

\subsubsection{Calibration curves for classification}
\label{sec:calibration_classification}

Creating calibration curves for classification models involves a multi-step procedure that differs from that for regression models. Let us use a binary classifier as an example. Similar to the regression setting, we also have access to a validation/test set of $N$ input-output pairs, $\mathcal{D} = \left\{ {\left( {{\mathbf{x}_1},{y_1}} \right),\left( {{\mathbf{x}_2},{y_2}} \right), \cdots,\left( {{\mathbf{x}_N},{y_N}} \right)} \right\}$. In a binary classification setting, the output takes the value of either 0 or 1, i.e., $y \in \left\{ 0,1 \right\}$. Creating a calibration curve for this classification setting involves three steps. 

\begin{enumerate}[label=\textbf{Step \arabic*:}, leftmargin=4.5em]
    \item The first step is to discretize the observed confidence $c$ into some number ($K$) of bins of width $1/K$. For example, if $K$ = 10, we then have ten intervals of observed confidence, $[0, 0.1], (0.1, 0.2], \cdots, (0.9, 1. 0]$. 
    
    \item 	We then compute for each bin $B_j=\left(c_j-\frac{1}{{2K}},c_j+\frac{1}{{2K}} \right]$ the observed confidence as
    \begin{equation}
        {\widehat c_j} = \frac{{\sum\limits_{i = 1}^N {{y_i}\mathbb{I}\left( {{f_{\bm{\uptheta}}}\left( {\mathbf{x}_i} \right) \in {B_j}} \right)} }}{{\sum\limits_{i = 1}^N {\mathbb{I}\left( {{f_{\bm{\uptheta}}}\left( {\mathbf{x}_i} \right) \in {B_j}} \right)} }},
    \end{equation}
    where $f_{\bm{\uptheta}} \left(\mathbf{x}_i \right)$ outputs the probability of $y_i= 1$.
    
    \item 	The final step is to plot the predicted vs. the observed confidence for class 1 for each bin $B_j$. 
\end{enumerate}

\subsubsection{Calibration metrics}
\label{sec:calibration_metrics}

\begin{figure}[!ht]
    \centering
    \includegraphics[scale=0.86]{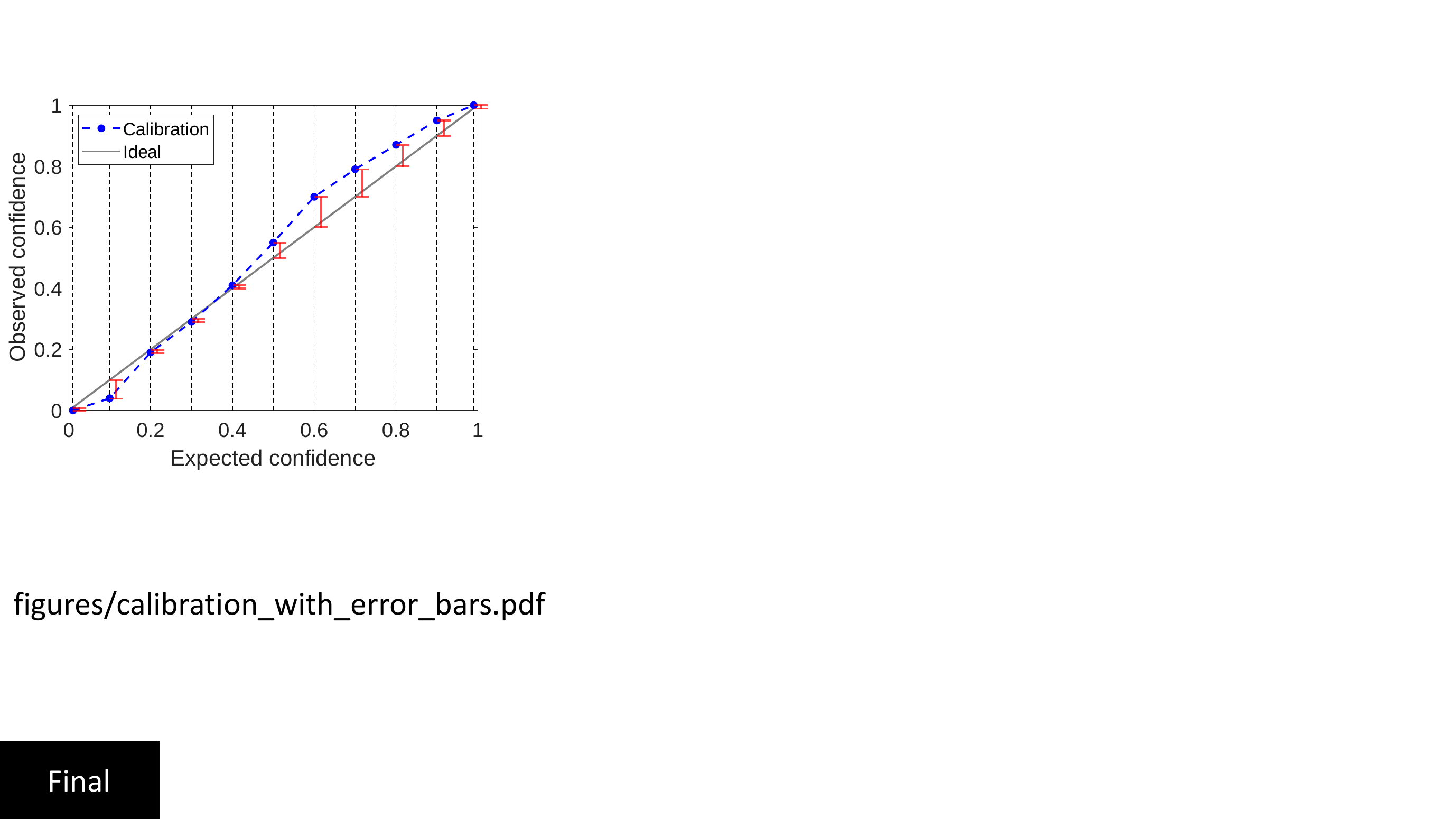}
    \caption{Calibration curve for the toy 1D mathematical problem shown in Fig.~\ref{fig:toy_1d_example}. This figure builds on the calibration curve shown in Step 3 of Fig.~\ref{fig:procedure_calibration_curve} and also includes the differences between calibrated and ideal (red error bars) used to calculate the ECE for this example.}
    \label{fig:calibration_with_error_bars}
\end{figure}

Several calibration metrics can be defined based on a calibration curve (see an example in Fig.~\ref{fig:procedure_calibration_curve}). For example, a simple metric can be the area between the calibration curve and the identity line, sometimes called the miscalibration area, which interestingly shares a similar concept with the area metric or u-pooling metric commonly used in the validation of computer simulation models \cite{liu2011toward}. Another calibration metric that is more widely used is the so-called expected calibration error (ECE), originally proposed for classification \citep{naeini2015obtaining} and later extended for regression \citep{kuleshov2018accurate}. Note though that the extension in \citep{kuleshov2018accurate} focused on deriving calibration curves and did not propose an ECE definition under regression settings. The ECE can be defined as the weighted average difference between a calibration curve and the ideal linear line, $ECE = \sum\limits_{j = 1}^K {{w_j}\left| {{{\widehat c}_j} - {c_j}} \right|} $, where the weight $w_j$ can be set as either a constant (i.e., $1/K$) or proportional to the number of samples falling into each bin, i.e., ${w_j} \propto \sum\limits_{i = 1}^N {\mathbb{I}\left( {{y_i} \in CI_i^C} \right)} $  for regression and ${w_j} \propto \sum\limits_{i = 1}^N {\mathbb{I}\left( {{f_{\bm{\uptheta }}}\left( {\mathbf{x}_i} \right) \in {B_j}} \right)} $ for binary classification \citep{kuleshov2018accurate}. Figure \ref{fig:calibration_with_error_bars} illustrates the calibration-ideal differences as error bars on the calibration curve obtained for the toy 1D mathematical problem shown in Fig.~\ref{fig:toy_1d_example}. Assuming equal weights (${w_1 = w_2 =,\cdots,= w_{11} = 1/11}$), the ECE for this calibration error is calculated to be 0.043, which means the observed confidence deviates from the expected confidence by 0.043 on average.

\subsubsection{Recalibration}
\label{sec:recalibration}
If the calibration curve deviates significantly from the identity function (perfect calibration), a recalibration may be needed to bring the calibration curve closer to the linear line. For example, this recalibration can be done by a parametric approach called Platt scaling, which modifies the non-probabilistic prediction of an ML binary classifier (e.g., a neural network or support vector classifier) using a two-parameter, simple linear regression model and optimizes the two model parameters by minimizing the NLL on a validation dataset \citep{platt1999probabilistic, niculescu2005predicting}. It is straightforward to extend Platt scaling to multi-class settings, for example, by expanding the simple linear regression model to a multivariate linear regression model \cite{guo2017calibration}. Another simple extension is temperature scaling, a single-parameter version of Platt scaling  \citep{guo2017calibration}, which was shown to be effective in re-calibrating deterministic neural networks capable of UQ \citep{mukhoti2021deterministic}. Another approach to recalibrating classification models is training an auxiliary regression model on top of the trained machine learning predictor, again using a validation dataset~\cite{kuleshov2018accurate}. A popular choice of the auxiliary regression model is an isotonic regression model, where a non-parametric isotonic (monotonically increasing) function maps probabilistic predictions to empirically observed values on a validation set. Recalibration using isotonic regression was originally proposed for classification \cite{zadrozny2002transforming,niculescu2005predicting} and then extended to regression \cite{kuleshov2018accurate}. It found recent applications in the PHM field, such as battery state-of-health estimation \citep{roman2021machine}. 

Both Platt scaling and isotonic regression require a separate validation dataset of a decent size (typically 20-50\% of the training dataset) to either optimize scaling parameters (Platt scaling) or build a non-parametric regression model (isotonic regression), while in reality, such a decent sized validation dataset may not be available. A comparative study of re-calibration approaches was performed in \citep{guo2017calibration}, where temperature scaling was found to be the most simple and effective.

\subsubsection{Connecting UQ calibration with model validation}
\label{sec:calibration_validation}

It is worth noting the connection between UQ calibration and the u-pooling method. U-pooling is a method for validating computer simulation models and has been well-established in the model validation community \citep{ferson2008model,liu2011toward}.  The u-pooling method aims to test whether all experimental observations, often made under multiple experimental conditions and sparse under each condition, come from the probability distributions predicted by a computer simulation model for the respective experimental conditions. If each observation comes from the corresponding predictive distribution, the CDF values of the experimental observations, ``pooled” together from all physical experiments, should follow a standard uniform distribution. Briefly, the u-pooling method first calculates the CDF value or \emph{$u$ value} of each observation, $u_i$, based on the predictive CDF by a computer simulation model, then plots the empirical CDF of $u$, where $u$ is along the x-axis and CDF is along the y-axis, and finally computes the area difference between the empirical CDF of $u$ and the CDF of the standard uniform distribution (diagonal line). The smaller the area difference, the more accurate (in a probabilistic sense) the computer simulation model. 

Plotting a UQ calibration curve like the ones in Fig. \ref{fig:comparison_calibration_curves} but for one-sided confidence intervals could also start by calculating the predictive CDF values ($u$ values in the u-pooling method) of all test samples, $u_1, \cdots, u_N$. Then, the observed confidence $\widehat{c}$ (y-axis) for any expected confidence $c$ (x-axis) can be calculated as the fraction of the CDF values that are smaller or equal to $c$, i.e., $\widehat{c} = \frac{1}{N}\sum\limits_{i = 1}^N {\mathbb{I}\left( {{u_i} \leq c} \right)}$. The differences are that (1) the empirical CDF plot in the u-pooling method shows $N$ eventually spaced empirical CDF values on the y-axis, while the number of expected confidence levels on the x-axis of a UQ calibration plot is manually selected; and (2) for each empirical CDF (y-axis for u-pooling) or expected confidence (x-axis for UQ calibration) value $c$, the u-pooling method plots the corresponding percentile of $u$, i.e., the ${100c}^\text{th}$ percentile of $u$ based on the dataset of $N$ $u$ values, while UQ calibration plots the corresponding fraction of the $u$ values that are no greater than $c$. Additionally, the u-pooling method strictly starts by looking at $u$ values. It then derives their empirical CDF values. In contrast, UQ calibration, to some degree, has a reverse process where it begins with manually choosing expected confidence levels and then calculates fractions of probabilistically accurate predictions (observed confidence values). However, the fraction calculation can use the $u$ values, as mentioned earlier. 

Before concluding on the connection between UQ calibration (ML community) and the u-pooling method (model validation community), we want to note that the u-pooling method could also be applied to assess the quality of the UQ of an ML model, with a different objective of measuring the degree to which each observation comes from the probability distribution predicted by the ML model, which differs from the objective of UQ calibration to test how underconfident or overconfident the ML model is. Similarly, the area metric or ``u-pooling” metric can be used to measure the mismatch between predictive distributions and observations in a global sense \citep{liu2011toward}. 

\subsection{Sparsification plots and metrics}
\label{sec:sparsification}

\begin{figure}[!ht]
    \centering
    \includegraphics[scale=0.86]{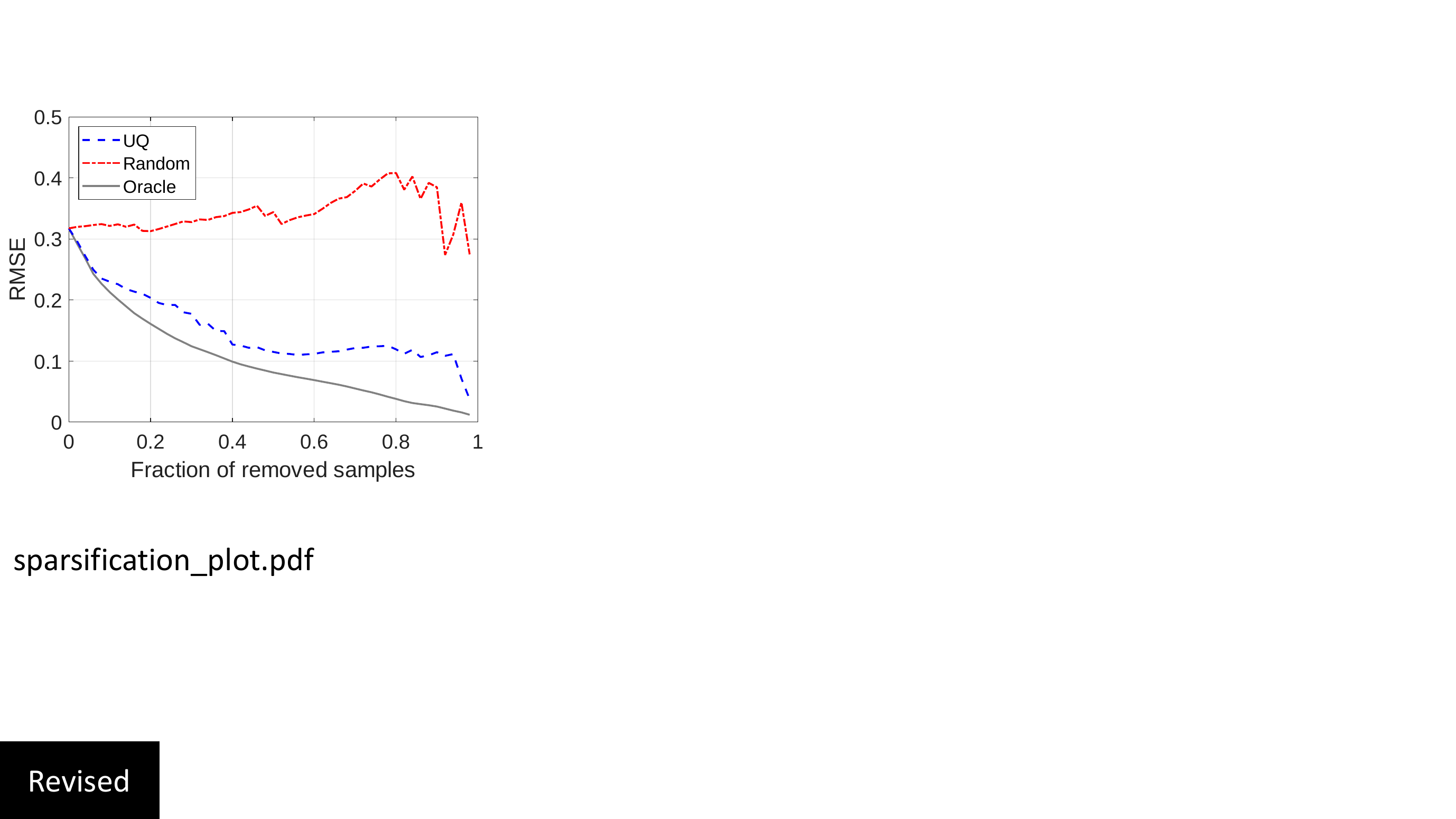}
    \caption{Sparsification curve and oracles for the toy 1D mathematical problem shown in Fig.~\ref{fig:toy_1d_example}.}
    \label{fig:sparsification_plot}
\end{figure}

Another method to assess the quality of predictive uncertainty is by creating the so-called \emph{sparsification plot} \citep{kondermann2008statistical}. A sparsification plot can be used to examine how well the predictive uncertainty of an ML model can serve as a proxy of the actual model prediction error, which is unknown without access to the ground truth. Creating a sparsification plot on a validation/test dataset consists of three steps. These three steps will be explained using the toy 1D regression problem from Sec. \ref{sec:calibration_regression} (see Fig.~\ref{fig:toy_1d_example}).
\begin{enumerate}[label=\textbf{Step \arabic*:}, leftmargin=4.5em]
    \item Given an uncertainty metric (e.g., variance for regression, entropy for classification), all samples in the validation/test dataset are sorted in descending order, starting with those with the highest predictive uncertainty. In the toy example, the 100 test samples are ranked according to the GPR model-predicted variance, with the first few samples having the largest predicted variances.
    \item A subset of samples (e.g., 2\% of the validation/test dataset) with the highest uncertainty is gradually removed, leaving an increasingly smaller dataset whose samples have lower predictive uncertainty than those removed. In the toy example, the sample removal process involves 50 iterations, each of which takes out 2\% of the remaining test samples with the highest predictive uncertainty. 
    \item Given an error metric (e.g., RMSE, mean absolute error), the prediction error is computed on the remaining samples each time a subset of high uncertainty samples is removed in Step 2. The toy example uses the RMSE as the error metric, computed by comparing the GPR model-predicted means with the actual (noisy) observations.
    \item The final step is to plot the error metric vs. fraction of removed samples for the combinations obtained in Steps 2 and 3. Figure \ref{fig:sparsification_plot} shows the sparsification plot (dashed blue curve) for the toy example.
\end{enumerate}

The resulting sparsification plot (see, for example, Fig. \ref{fig:sparsification_plot}) visualizes how the prediction error changes as a function of the fraction of removed samples. If predictive uncertainty is a good proxy for prediction error, the error metric on a sparsification plot should decrease monotonically with the fraction of removed high-uncertainty samples, as is the case in Fig. \ref{fig:sparsification_plot}. If ground truth is available, an ideal error curve (oracle) can be derived by ranking all samples in the validation/test dataset in descending order according to the actual prediction error. The oracle for the 1D toy regression problem is shown as a solid gray curve in Fig. \ref{fig:sparsification_plot}, where we can observe a small difference between the calculated and ideal error curves. If predictive uncertainty is a perfect representation of model prediction error, the calculated error curve and oracle will overlap on the sparsification plot. On the other extreme, random uncertainty estimates that do not reflect prediction error meaningfully would result in an almost constant error on the remaining samples, i.e., a (close to) flat error curve. An example of the sparsification curve under random uncertainty estimates is shown in Fig. \ref{fig:sparsification_plot} for the 1D toy regression problem (see the dash-dotted red curve). In this extreme case, a flat curve suggests that UQ provides little information about identifying problematic samples (e.g., OOD samples and those in regions of the input space with high measurement noise) whose model predictions may contain large errors.

Prior UQ studies in the ML community used plots similar to the sparsification plot to examine model accuracy as a function of model confidence \citep{lakshminarayanan2017simple,amini2020deep}. The only difference may be the label used for the x-axis, sometimes explicitly called \emph{confidence threshold} for classification \citep{lakshminarayanan2017simple} and regression \citep{amini2020deep}, instead of \emph{fraction of removed samples}. Per-sample model confidence was derived as the probability of the predicted label for classification \citep{lakshminarayanan2017simple} and the percentage of validation/test samples whose variances are higher than the validation/test sample of interest \citep{amini2020deep}. However, estimating the per-sample model confidence from the per-sample predictive uncertainty without access to the ground truth is difficult and remains an open research question. 

Since the model prediction error of one UQ approach on a validation/test sample most likely differs from that of a different approach, the ideal error curve (oracle) is likely to differ among UQ approaches. To compare these approaches, we can first calculate the difference between the sparsification and oracle for each fraction of removed samples, named \emph{sparcification error}. Then, we can compute two sparsification metrics: (1) the Area Under the Sparsification Error curve (AUSE), i.e., the area between the actual error curve and its oracle \citep{ilg2018uncertainty}, and (2) the Area Under the Random Gain curve (AURG), i.e., the area between the (close-to) flat random curve and the actual error curve. The lower the AUSE, the better the predictive uncertainty (derived from UQ) represents the actual prediction error (unknown). The higher the AURG (assuming the error curve shows a monotonically decreasing trend), the better UQ is compared to no UQ. 

\subsection{Negative log-likelihood}
\label{sec:nll}
Given a training dataset $\mathcal{D}$ and a validation/test data point $\bf{x}$, we can look at calculating the probability of observing its target value $y$ using the predictive probability density function of the target, expressed as $\hat{p}(y|\mathbf{x},\mathcal{D})$. We can repeat this process to get the probability of observing the target value for each sample in the validation/test dataset. Multiplying these predictive probabilities gives rise to a predictive likelihood. Taking a logarithmic transformation yields a predictive log-likelihood, which is a good measure of the goodness of fit of the probabilistic ML model to the validation/test data. The larger the log-likelihood, the better the model-data fit. Often, the negative counterpart of a log-likelihood, named NLL, is used in place of log-likelihood as the loss function or part of the loss function when training a probabilistic ML model. An example of the NLL has been given in Eq. (\ref{eq:ensemble_nn_loss}) as the loss function for training a prababilisitic neural network in a neural network ensemble, as discussed in Sec. \ref{sec:neuralnetwork_ensemble_aleatory}. It has been widely accepted that log-likelihood, or equivalently NLL, is a good measure of a probabilistic model's quality of fit \citep{hastie2009elements}. NLL can be viewed as an indirect measure of model calibration \citep{guo2017calibration} and is often used alongside calibration metrics to assess the quality of predictive uncertainty (see, for example, three recent methodological studies on UQ of ML models in \cite{d2021repulsive,liu2020simple,fortuin2021deep}).

\subsection{Accuracy vs. UQ quality}
\label{sec:accuracy_vs_UQquality}
An interesting finding about UQ of ML models was reported in \citep{guo2017calibration}, where NLL was found to behave inconsistently with traditional accuracy measures, such as the RMSE or mean absolute error for regression, during model training. It appeared that NLL and accuracy could become conflicting at some point during the training process when neural networks could learn to be more accurate at the cost of lower quality in UQ, as reported for classification problems in \citep{guo2017calibration}. This finding may help explain the observation in \cite{zhang2021understanding} that wide and deep neural networks trained with very limited regularization sometimes generalize surprisingly well \citep{guo2017calibration}. Specifically, the inconsistency between NLL and accuracy provides evidence that (1) these large-scale models exhibiting good generalization performance may still suffer from the common overfitting issue, and (2) overfitting occurs only for a probabilistic error metric (e.g., NLL), not a classification error metric (e.g., classification accuracy) for classification or an error metric calculated based on mean predictions (e.g., RMSE or mean absolute error) for regression. Nonetheless, it is still important to understand how well a model does probabilistically by looking at UQ quality metrics, such as calibration metrics (Sec. \ref{sec:calibration}), sparsification metrics (Sec. \ref{sec:sparsification}), and NLL (Sec. \ref{sec:nll}). Therefore, we strongly recommend academic researchers and industrial practitioners examine their ML models' performance in terms of both accuracy and UQ quality rather than focusing solely on accuracy metrics such as classification accuracy or RMSE. A seemingly highly accurate ML model may still have difficulties extrapolating to OOD samples, and it is crucial to estimate model confidence accurately through high quality UQ. We can now connect this discussion to an important statement in Sec. \ref{sec:reduction}, i.e., all models are wrong, but some are useful \cite{box1979all}.

\section{UQ of ML models in prognostics}
\label{sec:UQ_PHM}
As stated in Sec. \ref{sec:introduction}, our tutorial has an additional, secondary role, i.e., reviewing recent studies on engineering design and health prognostics applications of emerging UQ approaches.  To make this tutorial focused, we place our review of engineering design applications in \ref{sec:UQ_design} and only present the review of health prognostics applications in the main text of this tutorial (i.e., the present section). We believe such an arrangement will provide the additional benefit of creating a methodological transition into the two case studies in Sec. \ref{sec:case_studies} that are both related to health prognostics. 

\subsection{Uncertainty-aware ML for prognostics and health management}

\subsubsection{Prognostics and the role of UQ}
\label{sec:prognostics_UQ_role}
PHM is an engineering field that focuses on developing techniques and tools to establish effective maintenance strategies that balance system availability and performance with operational requirements and maintenance costs \cite{olgareview,biggio2020prognostics}. PHM comprises the main tasks of detecting the initiation of a fault (\emph{fault detection}), distinguishing between different types of fault and isolating the root cause (\emph{fault diagnostics}), and predicting the RUL (referred to as \emph{prognostics} \cite{olgareview,wang2019deep}). Notoriously, prognostics represents the most challenging task among the three main tasks of PHM \cite{thelen2022comprehensivepart1}. Effective prognostics enables \emph{just-in-time maintenance} \citep{lee2013recent, lee2013predictive}, which holds the promise of significantly reducing maintenance costs and system downtime while prolonging the lifetime of industrial and infrastructure assets, thereby increasing system availability. Besides its potential in terms of cost savings, effective prognostics also enables more environmentally sustainable operations of industrial and infrastructure assets by lowering the frequency of replacement and reducing the consumption of spare parts and resources \cite{biggio2020prognostics}. To be useful in mission- and safety-critical applications, successful prognostics approaches should be capable of not only predicting the RUL but also quantifying the associated uncertainty \cite{saxena2008metrics}. Knowledge of the associated uncertainty quantified in a principled manner allows users to conscientiously optimize the schedule of interventions and machine downtime with confidence rather than blindly relying upon the deterministic predictions of broadly applied black-box ML algorithms.
In reality, inaccurate predictions of the end of life or RUL due to low quality UQ can have catastrophic consequences in safety-critical applications. For example, when an ML model makes overconfident predictions, it could either over- or under-predict the end of life and RUL. Significant overpredictions can lead to unexpected safety failures, while substantial underpredictions can lead to a shortened useful lifespan of components. Ensuring reliable uncertainty estimates from data-driven algorithms is essential to mitigate these problems and optimize safety and cost-effectiveness in maintenance operations. This involves preventing disruptive events by avoiding delayed replacements and minimizing costs by preventing premature maintenance actions, such as replacements or repairs.

\subsubsection{The potential of DL for PHM}
\label{sec:potential_ML_prognostics}
Recently, deep learning (DL) methods have become more prevalent in PHM applications. One of the major advantages offered by DL techniques in PHM is the ability to automatically analyze sensor data, learn important features that characterize the system's health status, and track its changes over time until reaching the end of life. Industrial asset prognostics using DL can be implemented in two ways: directly predicting the RUL from sensor data or forecasting the future evolution of the system's health status until a pre-defined threshold is reached. The first approach, referred to as \emph{direct mapping} \citep{thelen2022comprehensivepart1}, requires a dataset that links sensor readings to corresponding RUL target labels and is treated as a regression task. The second approach, called \emph{time series forecasting} \citep{thelen2022comprehensivepart1}, involves identifying condition indicators that change in a predictable manner as the system deteriorates under different operational modes. These indicators may either be predetermined as strongly correlated with the machine's health and hence, interpretable, such as the internal resistance and capacity of a lithium-ion battery \citep{uqp13} or may be derived implicitly. A health indicator integrates several condition indicators into a single value, providing the user with information about the component's health status. The threshold for the health indicator, which may be subject to noise, also needs to be derived or learned. The importance of UQ in both approaches lies in the need to avoid unexpected safety-critical failures due to too-late replacements and to minimize costs by avoiding too-early replacements.  
UQ is, therefore, crucial to provide meaningful estimations and ensure accurate predictions in DL-based industrial asset prognostics. While quantifying the total predictive uncertainty (e.g., as a single variance value) already provides essential information for decision making, distinguishing between aleatory and epistemic uncertainty is equally important for prognostic applications. Particularly, considering that faults/failures are rare in safety-critical applications, epistemic uncertainty substantially impacts model performance due to the challenges in collecting representative run-to-failure datasets for  training. 

\subsubsection{Uncertainty-aware DL in prognostics}
\label{sec:Uncertainty_aware_ML_prognostics}

Modern DL techniques can often not be directly interpreted by humans. The black-box nature of DL models is clearly at odds with the need for trustworthy prognostic algorithms. UQ can remediate this drawback, and its integration in DNNs is the subject of an exciting - yet constantly evolving - research field in the DL community \cite{bdl1,bdl2,abdar2021review,bdl4,bdl5,Blundell2015,bdl7}, as discussed in Secs. \ref{sec:bnn}-\ref{sec:deterministic_uq}.  While a large number of research studies have focused on developing ML and DL approaches for providing point estimates of the RUL (\cite{olgareview,biggio2020prognostics, WANG202081} and the references therein), uncertainty-aware models, despite their great relevance, have not yet significantly impacted the research in this field.

In data-driven prognostics,  the models' predictions are inevitably affected by various sources of uncertainty. These sources of uncertainty include model-form uncertainty, insufficient representative historical data for model training, as well as errors in measurement and communication transmission, among others (refer to Table \ref{tab:epistemic uncertainty}). While ML and DL approaches have been increasingly applied for prognostics, most developed algorithms did not quantify the associated uncertainty. This limitation, among other factors, has prevented such approaches from being practically deployable in real mission- and safety-critical applications. UQ plays a vital role in enabling ML and DL to deliver high value in practical health prognostics applications. By instilling greater confidence in the predictions and streamlining the integration of the results into maintenance planning and scheduling, UQ reinforces user trust and enhances the effectiveness and safety of these applications \cite{rokhforoz2021multi, rokhforoz2023safe, olgareview, sankararaman2015significance, Zio2022PrognosticsAH, saxena2008metrics}.



\subsection{Uncertainty evaluation metrics for prognostics}\label{sec:uncertainty_metrics_prognostics}
While UQ for prognostics already significantly benefits from the standard UQ performance evaluation metrics commonly applied in other disciplines as well, such as NLL, the MSE, the RMSE, or the mean absolute percentage error (MAPE), the specificity of the prognostics problem often requires a set of customized metrics. One of the particularities of RUL predictions is for example that the closer the predictions progress to the end of life, the more certain the models should behave when making estimations about the predicted end of life. Therefore, the performance evaluation metrics should take such behavior into consideration and provide quantitative evaluation for it. 

Metrics, such as MSE or MAPE, do not take into account the statistical distribution of the RUL predictions around the ground-truth values. To account for such statistical deviations, a number of more informative probabilistic metrics have been introduced for applications in prognostics. Most of these metrics are built under the assumption that predicting the RUL at the initial time steps of the machine operation, is much harder and, as progressively additional information is acquired, the prediction task also gets simplified thanks to the fact that the severity of the fault increases and the corresponding symptoms tend to become more pronounced as the system approaches the end of life.

\begin{figure}[H]
    \centering
    \includegraphics[scale=0.26]{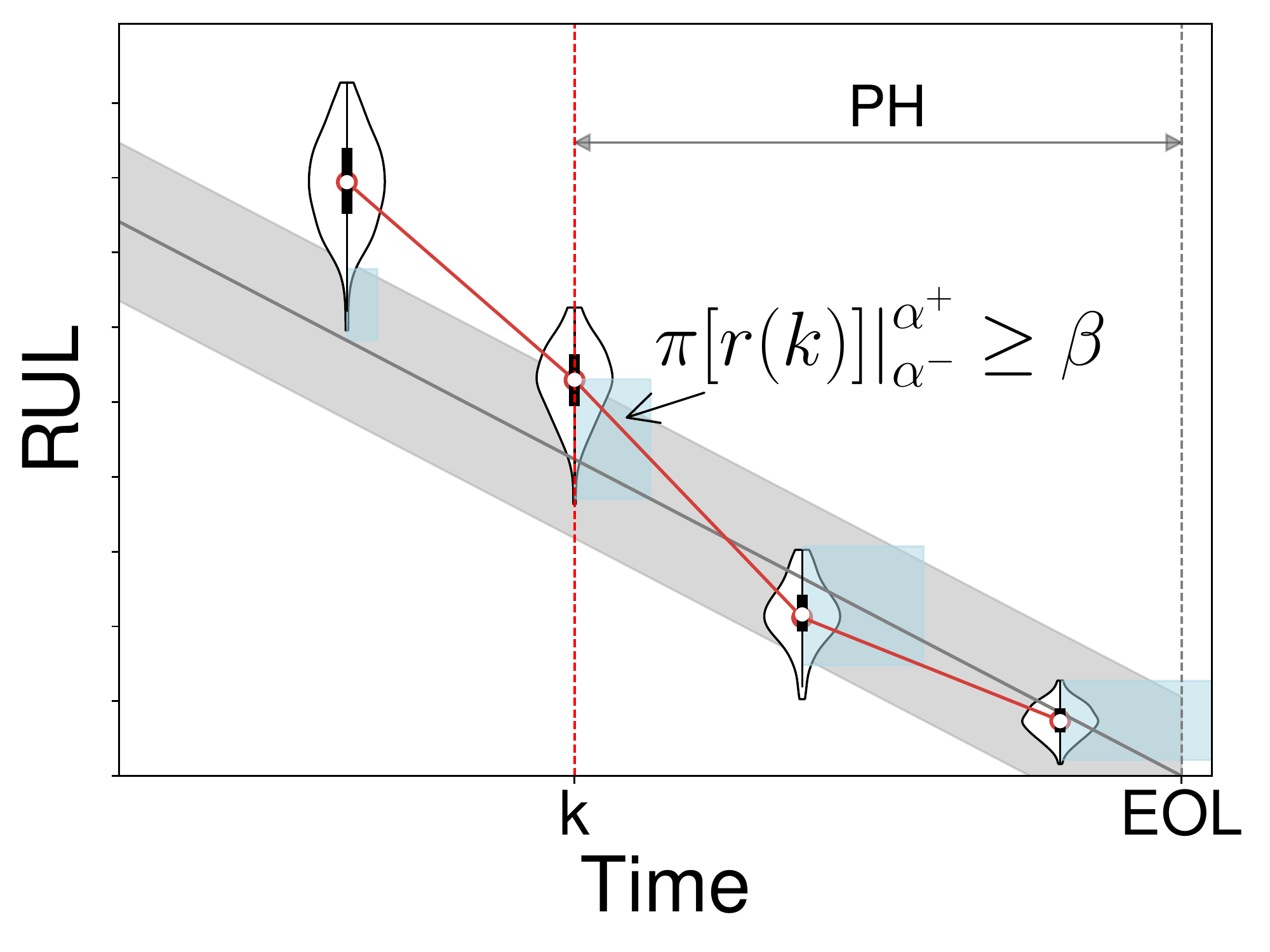}
    \includegraphics[scale=0.26]{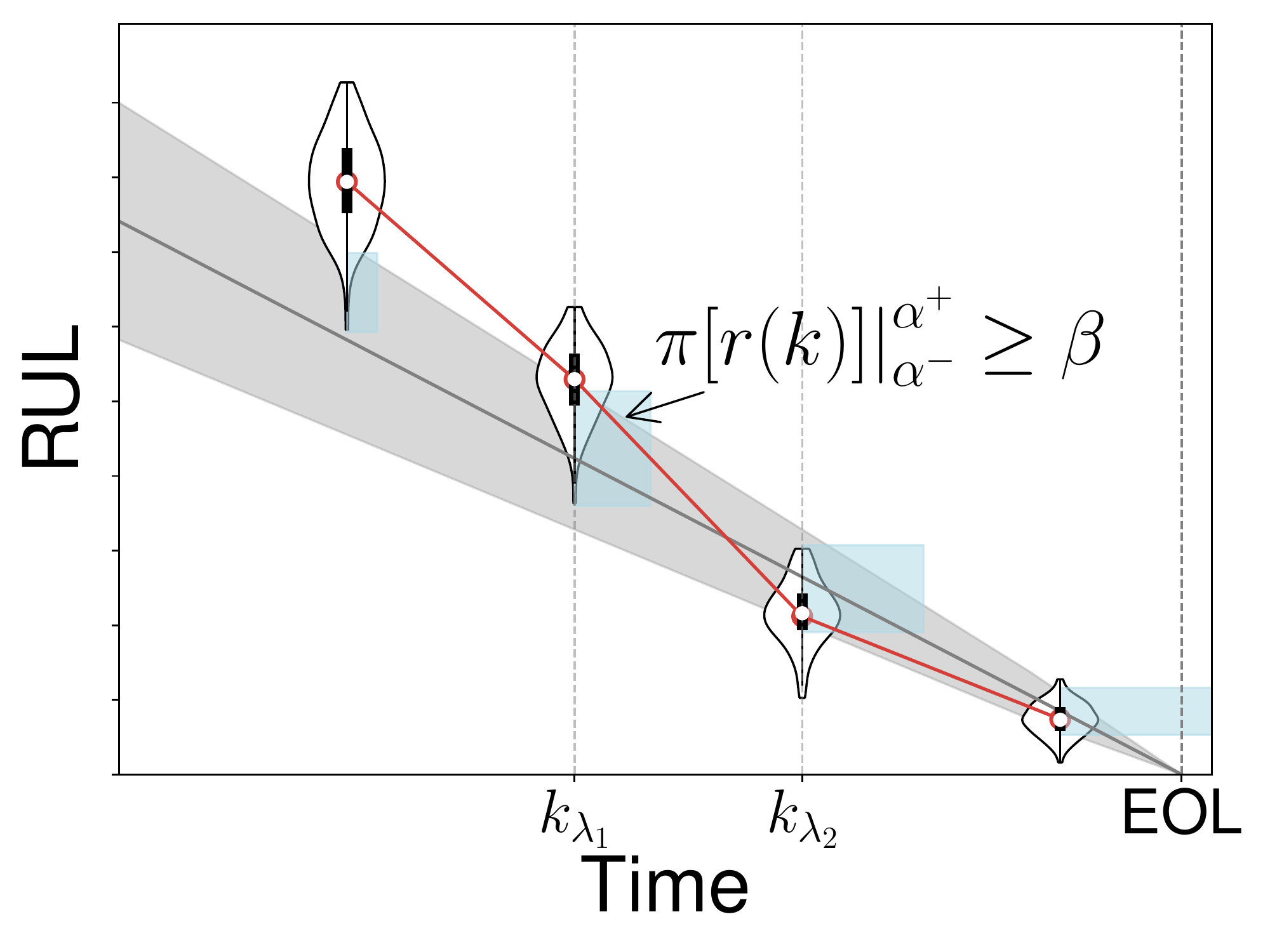}
    \includegraphics[scale=0.26]{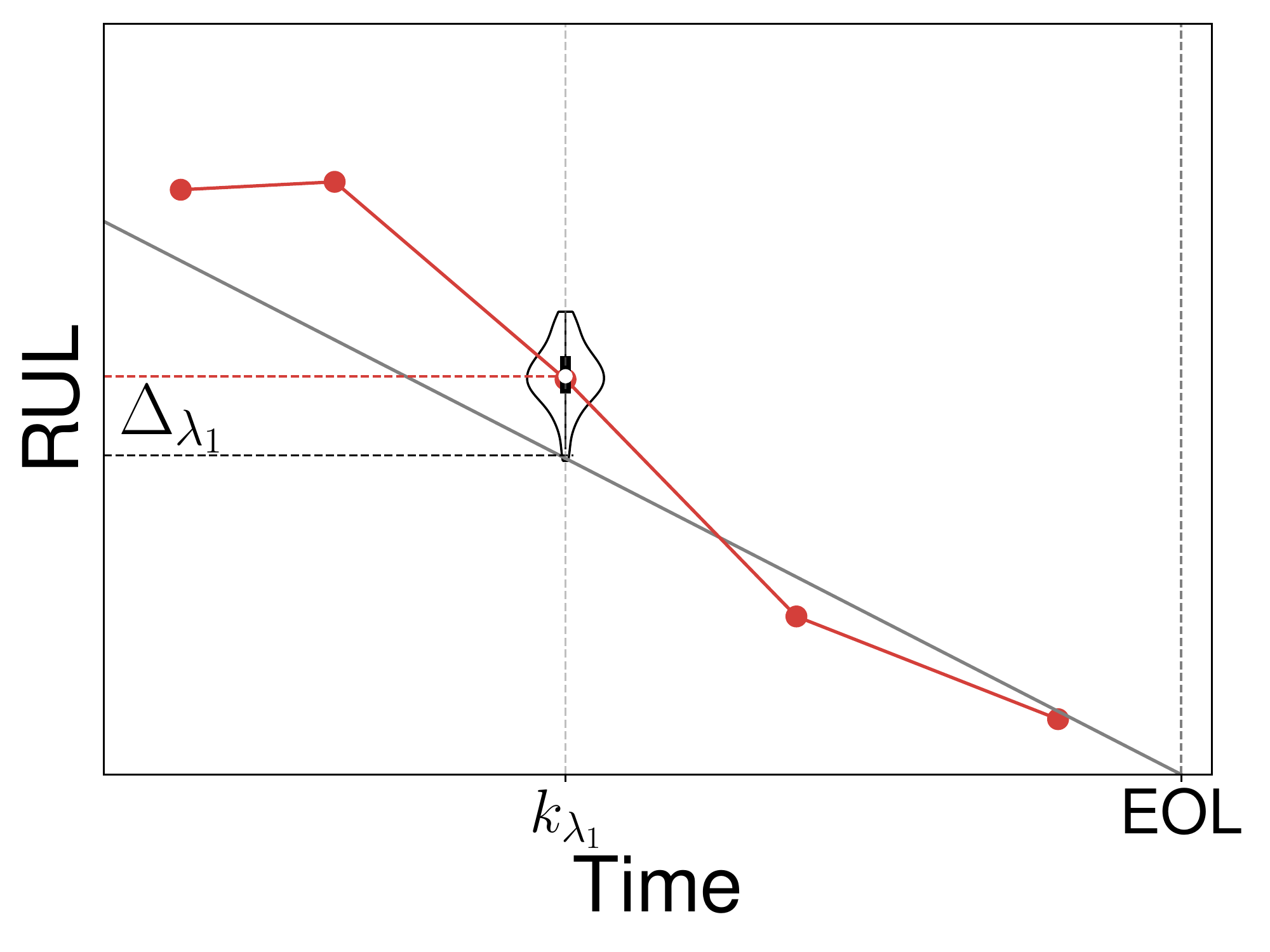}
    \caption{(Left) Prognostic Horizon (PH): here $[\pi(r(k))]^{\alpha^+}_{\alpha^-}$ indicates the probability that the distribution of the prediction $r$ at time $k$ falls within the confidence region $[r^*(k)-\alpha^-,r^*(k)+\alpha^+]$ (grey area), and $\beta$ is a pre-determined threshold; (Middle) $\alpha$-$\lambda$ metric calculated at $k_{\lambda_{1}}$ and $k_{\lambda_{2}}$: same notation as before, note that the confidence bounds around the ground-truth shrink as the end of life is approached; (Right) Relative Accuracy calculated at $k_{\lambda_{1}}$: $\Delta_{\lambda_{1}}$ indicates the difference between the median of the predictive distribution and the ground-truth value.}
    \label{fig:alpha_lambda}
\end{figure}

In the seminal work of \citet{saxena2010metrics}, the authors introduce four performance evaluation metrics for prognostics - meant to be measured sequentially - assessing different aspects of the RUL prediction problems, namely: the Prognostic Horizon, the $\alpha$-$\lambda$ performance, the Relative Accuracy, and the Convergence (Fig. \ref{fig:alpha_lambda}). While these performance evaluation metrics have mainly targeted physics-based prognostic methods, they are also applicable to DL-based UQ approaches. In the following, we briefly review their definitions and rationales. We refer interested readers to the original paper for more details. In essence, Prognostic Horizon is defined as the difference between the time step when the predicted RUL first meets the specified performance criteria and the time index for the end of life. The performance criteria are met if the predicted RUL value falls within an area determined by the ground-truth RUL value plus/minus a certain pre-selected confidence interval (called $\alpha$). The metric can be easily adapted to cases where the output of the model is probabilistic. In that case, the criterion is met if the probability of the predicted RUL falling within the previously defined area is larger than $\beta$, an additional parameter to be chosen a priori (Fig. \ref{fig:alpha_lambda}).

The $\alpha$-$\lambda$ metric is very similar to the Prediction Horizon but it differs in two aspects: first, it is binary, if the criterion is met at a certain time step, its value will be one, otherwise 0. Second, the confidence bounds around the ground-truth RUL are now a function of the predicted RUL and, as a result, will tend to shrink as the machine approaches the end of life. 

The relative accuracy is simply calculated as one minus the relative error of the model with respect to the ground truth at a certain time step. In particular, the relative error is computed by taking the ratio between the absolute difference between ground truth and a properly-chosen central tendency point estimate of the predicted RUL distribution, and the ground truth RUL value. The central tendency point estimate of the prediction distribution is arbitrary and depends on the statistical properties of the predictive distribution (Gaussian, mixture-of-Gaussians, multi-modal, etc.).  Finally, the Convergence acts as a meta-metric to measure how quickly each of the above metrics improves over time.


\subsection{Discussion}
\label{sec:phm_discussion}
Meaningful uncertainty estimates are crucial for ensuring the safe and reliable deployment of DL models in real-world applications, especially for safety-critical assets. This is essential to build trust in the models and ensure their effectiveness. This is because, in practice, decision making in the context of industrial applications involves a complicated trade-off between risky decisions and large potential economic benefits. DL has undoubtedly advanced the field by offering a valuable set of tools to efficiently learn from data and automate the entire prognostics process. Nevertheless, this is only one part - yet very significant - of the challenges arising in prognostics. ML and DL techniques need to be as trustworthy and reliable as possible, and for this reason, effective UQ and its integration into existing techniques remain an essential desideratum.

In previous research studies, MC dropout has been by far the most widely employed strategy for tackling UQ of neural networks, especially DNNs. There are likely two reasons for this: first, the interpretation of MC dropout is very intuitive; and second, it requires only a minimal modification to existing architectures, namely activating dropout layers at training time. Nevertheless, as shown in multiple studies \cite{lakshminarayanan2017simple,mcdp2,mcdp3}, the UQ performance of MC dropout is not always satisfactory, and more advanced solutions should be explored. Fortunately, the fields of UQ and Bayesian DL are constantly progressing, and applications of the resulting techniques to prognostics are an important research area to be further explored \cite{bdl1,bdl2,abdar2021review,bdl4,bdl5,Blundell2015,bdl7}. 

In addition, uncertainty-aware ML methods have been mainly used in the context of prognostics for RUL prediction. While this is arguably the most important end goal in this field, several other avenues could be investigated in the future. An example is, for instance, anomaly detection. In this setting, uncertainty can be used to detect abnormal health states in the machine operation by evaluating the level of confidence of the model corresponding to that time step. The assumption is that a high level of epistemic uncertainty associated with a certain input will be indicative of test data points that are less representative of the training data distribution. Hence, such data will probably correspond to unusual health states, assuming the training data are collected from a machine operating in a nominal regime. 

To conclude, a crucial criterion for any UQ technique used in prognostics is the ability to accurately disentangle aleatory and epistemic uncertainty. These two measures contain distinct  types of information and, therefore, must be interpreted separately to ensure appropriate analysis. 





\section{Case studies for benchmarking – Code Sharing on GitHub}
\label{sec:case_studies}
In this section, we benchmark the performance of several UQ methods in two engineering applications: (1) early life prediction of lithium-ion batteries and (2) RUL prediction of turbofan engines. In both case studies, we built UQ models with publicly available datasets and compared the models' performance. To ensure a fair comparison, these UQ models are built with nearly identical backbone architectures wherever applicable. These two case studies are widely used in the literature due to their broad significance in safety-critical applications and, therefore, a comprehensive understanding of the performance of different UQ methods helps to identify the right model to deploy in a particular application. A code walk-through is provided for the first case study to demonstrate the practical implementation of UQ methods. We acknowledge that there could be several other ways of implementing the same UQ models using different sets of libraries. In this discussion, we try to limit ourselves to using only TensorFlow and Keras libraries for building the neural network models. 

\subsection{Case study 1: Battery early life prediction}
In this section, we explore the utility of various UQ for ML model methods to tackle the early life prediction of lithium-ion batteries. The dataset used in this case study consists of run-to-failure data from 169 LFP/graphite APR18650M1A cells with a nominal capacity of 1.1Ah~\citep{severson2019data,attia2020closed}. The goal of this case study is to predict, with confidence, the remaining cycle life of lithium-ion cells based on data collected only in the first 100 cycles. This early life prediction is a challenging problem as most cells do not exhibit significant levels of degradation during the first 100 cycles. Therefore, it is important for researchers to associate each prediction with an uncertainty estimate.

\begin{figure}[!ht]
    \centering
    \includegraphics[scale=0.45]{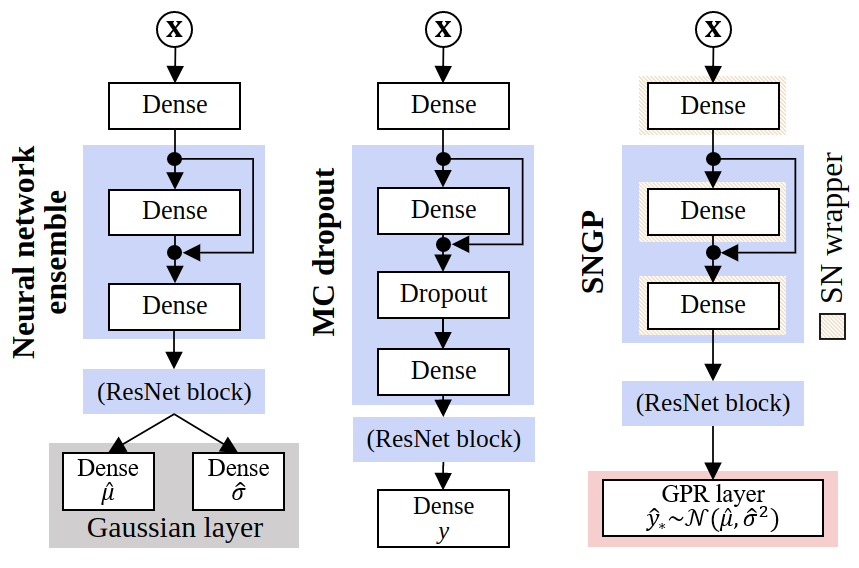}
    \caption{UQ model architectures with ResNet backbone used in case study 1. The ResNet block for each model is defined by the blue box.}
    \label{fig:cs2_architecture}
\end{figure}

The code for this case study can be found at our \href{https://github.com/VNemani14/UQ_ML_Review}{Github page}. In this section, we take the opportunity to provide a brief walk-through of the code while discussing the following UQ methods: (1) neural network ensemble, (2) MC dropout, (3) GPR, and (4) SNGP. The goal of this study is to compare several UQ methods with comparable prediction accuracy based on the current literature. The neural network-based models, namely neural network ensemble, MC dropout, and SNGP, are built on a ResNet with a similar backbone architecture as shown in Fig. \ref{fig:cs2_architecture}. 

\subsubsection{Dataset overview}
\begin{figure}[!ht]
    \centering
    \includegraphics[scale=0.5]{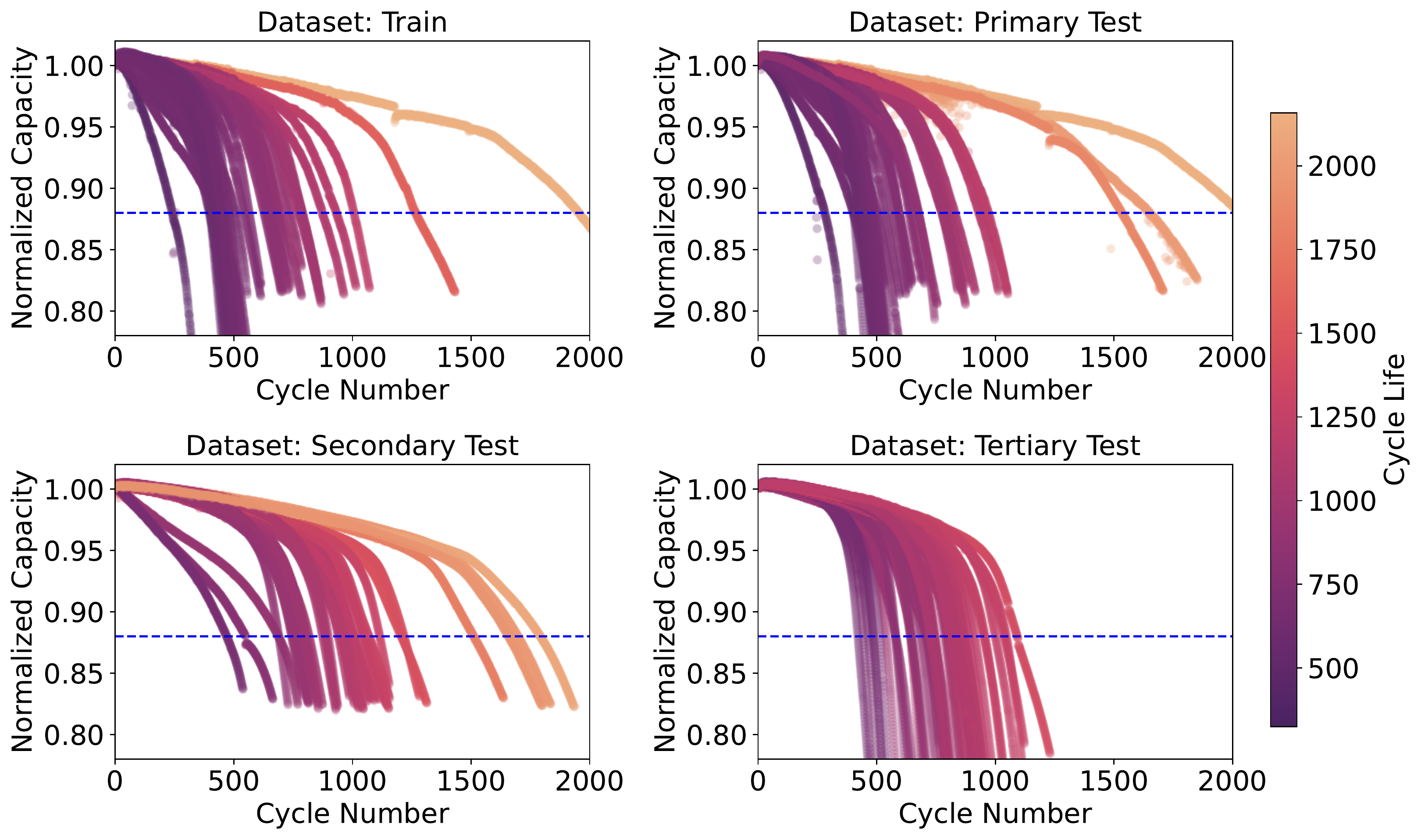}
    \caption{Normalized capacity curves for the four datasets mentioned in Table \ref{tab:LFP dataset}.}
    \label{fig:cs2_capacity_curve}
\end{figure}

The 169 LFP cell dataset is a combination of the 124-cell dataset provided by~\citet{severson2019data} and the 45-cell dataset provided by~\citet{attia2020closed}. These 169 cells are divided into three subsets as described in Table \ref{tab:LFP dataset}, where the partition for training, primary test, and secondary test datasets is consistent with that of \citet{severson2019data}, and the dataset from~\citet{attia2020closed} is used as the tertiary test dataset. The 169 LFP cells underwent different fast-charge protocols and storage time, but they had identical discharging conditions, which in turn led to a diverse set of capacity trajectories as illustrated in Fig. \ref{fig:cs2_capacity_curve}. Similar to the existing literature, we assume a cell to have reached the end of life when its capacity reaches 80\% of the nominal value (cutoff of 0.88Ah). A more detailed description of the battery cycling tests and raw data can be found at \url{https://data.matr.io/1/}.

\begin{table*}[!ht]
\centering
\caption{Summary of LFP battery dataset}
\begin{tabular}{p{5cm}|p{4cm}}
\hline \hline
{\bf{Type}}               & \bf{No. of cells}  \\ \hline 
Training & 41 \\ \hline
Primary test & 43 \\ \hline
Secondary test & 40 \\ \hline
Tertiary test & 45  \\ \hline \hline
\end{tabular}
\label{tab:LFP dataset}
\end{table*}


\begin{figure}[!ht]
    \centering
    \includegraphics[scale=0.54]{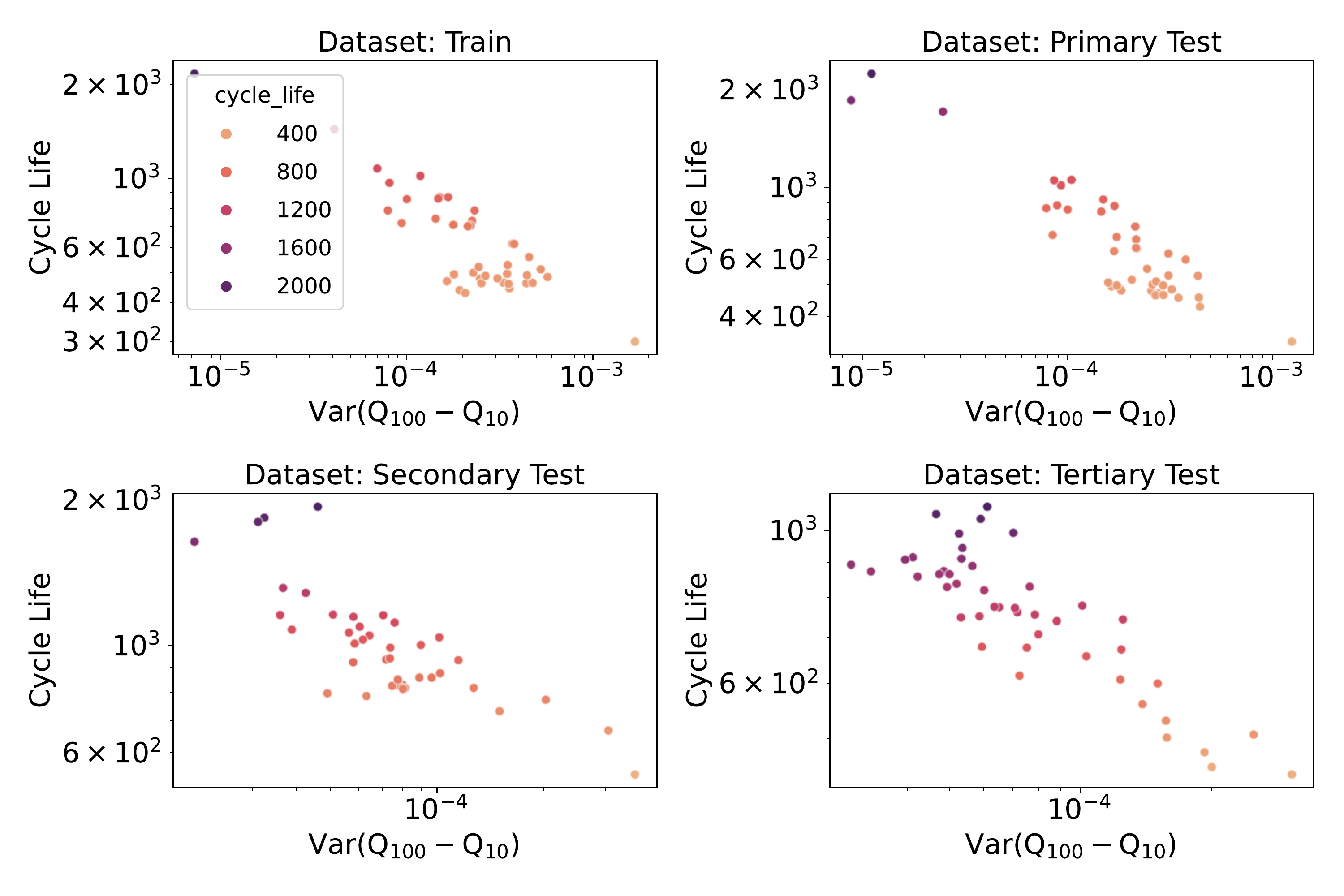}
    \caption{Correlating cycle life with $Var(\Delta Q_{100-10}(V))$.}
    \label{fig:cs2_vq_corr}
\end{figure}

The cycle-to-cycle evolution of voltage as a function of discharge capacity $V(Q)$ is often captured when conducting the experiments. However, the authors of the original dataset ~\citet{severson2019data} hypothesize and prove that the inverse relationship, where the discharge capacity as a function of voltage $Q(V)$ during the early cycles carries sufficient information to accurately predict the cycle life. We adopt a similar strategy of using $\Delta Q_{100-10}(V) = Q_{100}(V) - Q_{10}(V)$ as the input to our UQ models. Similar to~\citet{severson2019data}, we find that the cycle life is significantly correlated with $Var(\Delta Q_{100-10}(V))$ as shown in Fig. \ref{fig:cs2_vq_corr}.





\subsubsection{Neural Network Ensemble}
We first develop a neural network ensemble model (NNE) following the discussion from Sec. \ref{sec:neuralnetworkensemble}. Particularly, we develop a neural network learning framework following the work by \citet{lakshminarayanan2017simple}. Each individual model of the ensemble consists of a Gaussian layer as the final layer, and the Gaussian layer outputs a predicted mean $\mu$ and variance $\sigma^2$ for a given input $\textbf{x}$. Parameters $\bm{\uptheta}$ of the neural network are trained to minimize the NLL loss function defined in Eq. (\ref{eq:log_likelihood}) earlier, which corresponds to the implementation below:

\begin{lstlisting}[language=Python]
def custom_loss(variance):
    def nll_loss(y_true, y_pred):
        return tf.reduce_mean(0.5*tf.math.log((variance)) + 
                              0.5*tf.math.divide(tf.math.square(y_true - y_pred), 
                                                variance)) + 1e-6
    return nll_loss
\end{lstlisting}

\begin{tikzpicture}[remember picture,overlay,>=stealth']
\draw[<-,thin] (pic cs:line-texcode-1-end) +(12em,5ex) -| +(12,0.8) node[right,comment,thick] {NLL loss in Eq. (\ref{eq:log_likelihood})};
\end{tikzpicture}

In the code below, the Gaussian layer uses two kernels and biases to characterize $\mu$ and $\sigma$ by splitting the output of the previous layer (traditionally a fully connected layer with one dimension). Note that the kernel shape should be compatible with the number of hidden units in the previous dense layer. 


\begin{lstlisting}[language=Python]
class GaussianLayer(Layer):
    def build(self, input_shape):
        self.kernel_1 = self.add_weight(shape=(10, self.output_dim),...)
        self.kernel_2 = self.add_weight(shape=(10, self.output_dim),...)
        ... # (define bias_1 and bias_2)
    def call(self, x):
        output_mu  = K.dot(x, self.kernel_1) + self.bias_1
        output_var = K.dot(x, self.kernel_2) + self.bias_2
        output_var_pos = K.log(1 + K.exp(output_var)) + 1e-06  
        return [output_mu, output_var_pos]
\end{lstlisting}

\begin{tikzpicture}[remember picture,overlay,>=stealth']
\draw[<-,thin] (pic cs:line-texcode-1-end) +(36em,23.5ex) -| +(15.5,3.35) node[below left,comment,thin] {Two kernels + biases to split the output};
\draw[<-,thin] (pic cs:line-texcode-1-end) +(31em,8.2ex) -| +(15.5,1.8) node[above left,comment,thin] {Make variance positive};
\draw[<-,thin] (pic cs:line-texcode-3-end) +(21.5em,4.7ex) -| +(10.72,0.8) node[right,comment,thin] {Output mean and variance};
\end{tikzpicture}

Finally, a neural network model is constructed by appending the Gaussian layer to a simple ResNet model. The architecture for each individual model of the neural network ensemble is shown in Table \ref{tab:NN_model_arch}.

\begin{table*}[!ht]
\centering
\caption{Individual model of the neural network ensemble}
\begin{tabular}{p{5cm}|p{4cm}|p{4cm}}
\hline \hline
{\bf{Layer}}               & \bf{Output Shape}  &\bf{No. of Parameters}  \\ \hline 
Input & [(None, 1000)] & 0\\ \hline
Fully connected & (None, 100) & 100100\\ \hline
Fully connected & (None, 50) & 5050\\ \hline
Fully connected & (None, 50) & 2550\\ \hline
Fully connected & (None, 50) & 2550\\ \hline
Fully connected & (None, 50) & 2550\\ \hline
Fully connected & (None, 10) & 510\\ \hline
Gaussian layer & [(None, 1), (None, 1)]  & 22\\ \hline 
Total trainable parameters &  & 113332\\ \hline \hline
\end{tabular}
\label{tab:NN_model_arch}
\end{table*}

In total, we independently trained 15 models by randomizing the initialization of model weights in addition to shuffling the training samples. The size of the neural network ensemble is determined based on the elbow method - see Fig. \ref{fig:ensemble_elbow} for more details. Each individual model is trained for 300 epochs (based on validation split/validation loss to test for overfitting). 

\subsubsection{MC Dropout}
In this section, a simple MC dropout model is developed following the method described in Section \ref{sec:mcdropout}. The only differences between the implementation of the MC dropout and the neural network ensemble are (1) the inclusion of dropout layers with dropout being active during the prediction phase and (2) having a single deterministic output as the final output. Note that the dropout layer can also be introduced in other UQ methods, for example, in neural network ensembles, to mitigate overfitting. However, dropout is typically not activated during the prediction phase in such models. In the case of MC dropout, the output varies from one prediction run to another, where a certain percentage of neural network weights from the trained model are randomly dropped out at the prediction phase. The code snippet below showcases our implementation of the dropout layers within the ResNet block as shown in Fig. \ref{fig:cs2_architecture}.


\begin{lstlisting}[language=Python]
for _ in range(num_res_layers): # for each residual block
    x  = Dense(50, activation = actfn)(x)
    x1 = Dense(50, activation = actfn)(x)
    x = x1 + x
    x = Dropout(rate = 0.10)(x)
mu = Dense(1, activation = actfn)(x)
model = Model(feature_input, mu)
\end{lstlisting}
\begin{tikzpicture}[remember picture,overlay,>=stealth']
\draw[<-,thin] (pic cs:line-texcode-1-end) +(20em,11.5ex) -| +(9.6,2.2) node[above right,comment,thin] {Dropout within each ResNet block};

\draw[<-,thin] (pic cs:line-texcode-2-end) +(20em,8.2ex) -| +(12.0,1.4) node[right,comment,thin] {Single output (RUL)};
\end{tikzpicture}

The MC dropout model architecture and trainable parameters are similar to Table \ref{tab:NN_model_arch} except for the presence of dropout layers with a 10\% dropout rate. During the prediction phase, the trained MC dropout model is run 15 times with dropout enabled (the ensemble size was determined based on the elbow method - see description for Fig.~\ref{fig:ensemble_elbow}). An ensemble of all the individual deterministic RUL predictions produces the RUL prediction with uncertainty quantified.

\subsubsection{Spectral Normalization Gaussian Process (SNGP)}
Next, we implement the SNGP model discussed in Section \ref{sec:deterministic_uq} with the core idea of preserving distance awareness between training and test/OOD distributions when producing the uncertainty for each prediction. This is achieved by: (1) applying spectral normalization to the hidden layers of the neural network and (2) replacing the final layer with a Gaussian process layer. This is a single-model method with high performance in OOD detection. 

Following \citet{liu2020simple} and a corresponding \href{https://www.tensorflow.org/tutorials/understanding/sngp#the_sngp_model}{tutorial} of TensorFlow, as shown below,
we first define a model class {\it {FC\_SNGP}} inherited from the class of TensorFlow model. In this model class, we wrap some dense layers with the spectral normalization layer, where the normalization threshold has a constant value of {\it {spec\_norm\_bound}}. The {\it {RandomFeatureGaussianProcess}} layer with RBF kernel serves as the Gaussian process layer. 

\begin{lstlisting}[language=Python]
import official.nlp.modeling.layers as nlp_layers
class RN_SNGP(tf.keras.Model):
        ...
        self.dense_layers1 = nlp_layers.SpectralNormalization(
                self.make_dense_layer(100),norm_multiplier=self.spec_norm_bound)
        ...
    def make_output_layer(self, no_outputs):
        """Uses Gaussian process as the output layer."""
        return nlp_layers.RandomFeatureGaussianProcess(no_outputs,
                                  gp_cov_momentum=-1,**self.kwargs)
\end{lstlisting}

\begin{tikzpicture}[remember picture,overlay,>=stealth']
\draw[<-,thin] (pic cs:line-texcode-1-end) +(32em,22.0ex) -| +(13.0,4.0) node[above,comment,thin] {Spectral Normalization wrapper \\ applied to Dense layer};
\end{tikzpicture}

The value of {\it {gp\_cov\_momentum}} in the above figure decides if the calculated covariance is exact or approximated. A positive value of {\it {gp\_cov\_momentum}} updates the covariance across the batch using a momentum-based moving average technique, whereas a value of -1 calculates the exact covariance. Since the calculation of covariance could be affected by the batch size, it is recommended that the covariance matrix estimator be reset during each epoch. This can be done using Keras API to define a callback class and then appending it to {\it {FC\_SNGP}}. Finally, we train an SNGP model with the ReLU activation function and {\it {spec\_norm\_bound = 0.9}}.  

\begin{lstlisting}[language=Python]
class ResetCovarianceCallback(tf.keras.callbacks.Callback):
    def on_epoch_begin(self, epoch, logs=None):
        """Resets covariance matrix at the beginning of the epoch."""
        if epoch > 0:
            self.model.regressor.reset_covariance_matrix()
\end{lstlisting}

\subsubsection{Gaussian Process Regression}
At last, a standard GPR model with RBF kernel is trained using the scikit-learn Python package. The hyperparameters of the GPR models, such as length scale, are optimized using grid search during model fitting. 

\subsubsection{Evaluation/Results}
\begin{figure}[ht]
    \centering
    \includegraphics[scale=0.4]{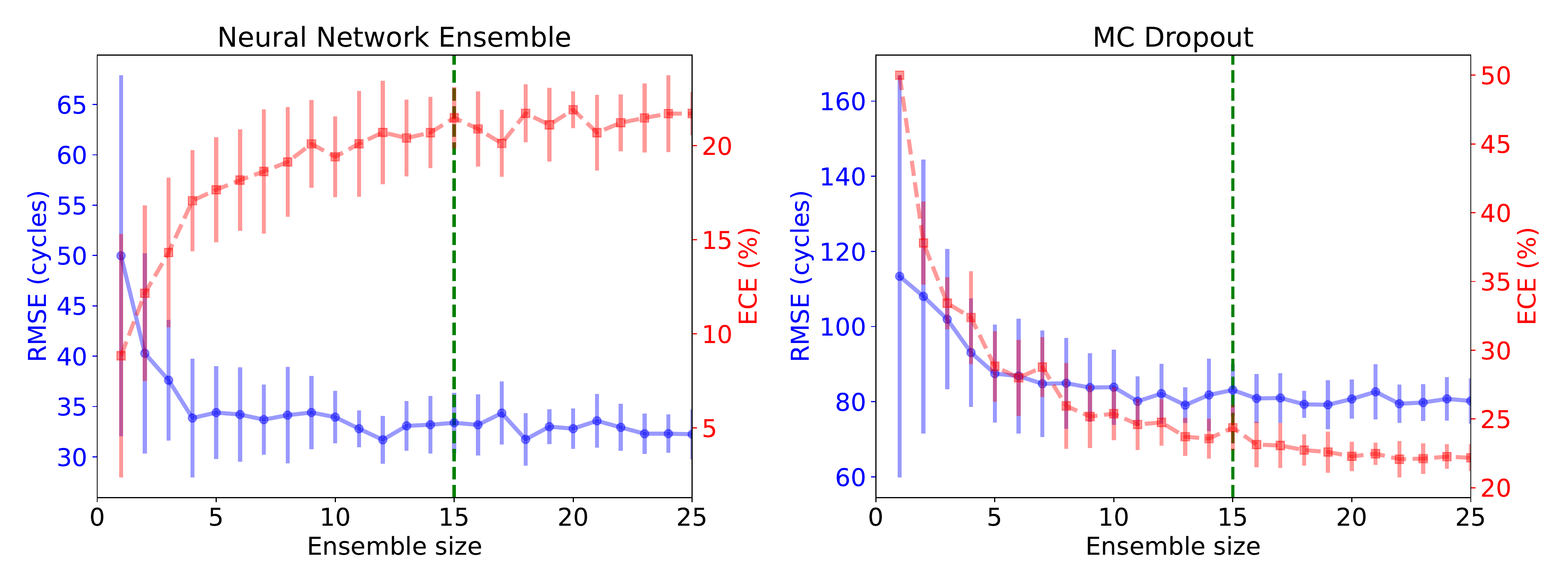}
    \caption{Determining the ensemble size for neural network ensemble and MC dropout. The selected ensemble size for this case study is determined by the green vertical line.}
    \label{fig:ensemble_elbow}
\end{figure}

In this section, we exploit the following metrics to quantitatively examine the uncertainty quantification performance of all the models: (1) root mean square error (RMSE), (2) average NLL defined in Eq. (\ref{eq:log_likelihood}), (3) expected calibration error (ECE) as defined in Section \ref{sec:calibration_metrics}, and (4) calibration curve introduced in Section \ref{sec:calibration_regression}. 
Since both neural network ensemble and MC dropout require an ensemble of individual models, it is essential to determine the ensemble size. Ideally, it is preferred that an ensemble has as many individual models as possible so that all the potential variations get manifested during the prediction stage. In other words, an ensemble benefits from models that undergo diverse learning paths and this would effectively capture the variations in predictions. However, beyond a certain ensemble size, the learning becomes increasingly less diverse and only trivially contributes to the ensemble at the expense of increased computational cost. Therefore, inspired by the elbow method, we systematically vary the ensemble size for constructing the neural network ensemble and MC dropout models while capturing the training RMSE and ECE as shown in Fig. \ref{fig:ensemble_elbow}. RMSE and ECE are chosen to strike a trade-off between accuracy and uncertainty quantification capabilities. Based on this study, we choose an ensemble size of 15 for both neural network ensemble and MC dropout.

\begin{table}[!ht]
\centering
\caption{Performance comparison across UQ methods for the 169 LFP cell dataset in terms of MSE $\pm$ standard deviation}
\begin{tabular}{c|cccc|}
\cline{2-5}
\textbf{}                              & \multicolumn{1}{c|}{\textbf{NNE}}   & \multicolumn{1}{c|}{\textbf{MC}}   & \multicolumn{1}{c|}{\textbf{SNGP}} & \textbf{GPR}                       \\ \hline
\multicolumn{1}{|c|}{\textbf{Dataset}} & \multicolumn{4}{c|}{\textbf{RMSE (cycles) $\boldsymbol{\downarrow}$}}                                                                                                      \\ \hline
\multicolumn{1}{|c|}{Train}            & 68.1±22.1                          & 69.4±16.8                           & 34.8±14.7                         & 0.0±0.0                           \\ \cline{1-1}
\multicolumn{1}{|c|}{Primary test}     & \cellcolor[HTML]{DCDCDC}137.3±20.9 & \cellcolor[HTML]{DCDCDC}149.9±18.4  & \cellcolor[HTML]{DCDCDC}148.1±16.2  & \cellcolor[HTML]{DCDCDC}141.1±0.0 \\ \cline{1-1}
\multicolumn{1}{|c|}{Secondary test}   & 205.1±27.4                         & 194.1±15.1                         & 249.3±33.6                        & 319.0±0.0                         \\ \cline{1-1}
\multicolumn{1}{|c|}{Tertiary test}    & \cellcolor[HTML]{DCDCDC}183.9±46.9	 & \cellcolor[HTML]{DCDCDC}195.0±29.1 & \cellcolor[HTML]{DCDCDC}258.9±60.3 & \cellcolor[HTML]{DCDCDC}406.5±0.0 \\ \hline
\multicolumn{1}{|l|}{}                 & \multicolumn{4}{c|}{\textbf{NLL} $\boldsymbol{\downarrow}$}                                                                                                                \\ \hline
\multicolumn{1}{|c|}{Train}            & 4.7±0.3                          & 8.6±2.6                            & 5.6±0.02                            & -3.8±0.0                          \\ \cline{1-1}
\multicolumn{1}{|c|}{Primary test}     & \cellcolor[HTML]{DCDCDC}5.4±0.2  & \cellcolor[HTML]{DCDCDC}14.3±6.5    & \cellcolor[HTML]{DCDCDC}5.7±0.03    & \cellcolor[HTML]{DCDCDC}5.7±0.0   \\ \cline{1-1}
\multicolumn{1}{|c|}{Secondary test}   & 5.7±0.2                         & 6.9±1.3                            & 6.1±0.2	                            & 6.0±0.0                           \\ \cline{1-1}
\multicolumn{1}{|c|}{Tertiary test}    & \cellcolor[HTML]{DCDCDC}5.7±0.1  & \cellcolor[HTML]{DCDCDC}9.2±1.7    & \cellcolor[HTML]{DCDCDC}5.9±0.1    & \cellcolor[HTML]{DCDCDC}6.4±0.0   \\ \hline
\multicolumn{1}{|l|}{}                 & \multicolumn{4}{c|}{\textbf{ECE (\%)} $\boldsymbol{\downarrow}$}                                                                                                           \\ \hline
\multicolumn{1}{|c|}{Train}            & 29.8±3.7                          & 15.2±6.8                            & 42.5±3.0                            & 49.9±0.0                          \\ \cline{1-1}
\multicolumn{1}{|c|}{Primary test}     & \cellcolor[HTML]{DCDCDC}10.5±5.0   & \cellcolor[HTML]{DCDCDC}24.4±5.3   & \cellcolor[HTML]{DCDCDC}21.5±2.3    & \cellcolor[HTML]{DCDCDC}6.9±0.0   \\ \cline{1-1}
\multicolumn{1}{|c|}{Secondary test}   & 13.5±5.7                           & 9.5±4.6                           & 12.7±4.6                           & 10.4±0.0                          \\ \cline{1-1}
\multicolumn{1}{|c|}{Tertiary test}    & \cellcolor[HTML]{DCDCDC}9.8±4.5   & \cellcolor[HTML]{DCDCDC}22.6±3.4   & \cellcolor[HTML]{DCDCDC}9.3±4.4   & \cellcolor[HTML]{DCDCDC}8.0±0.0   \\ \hline
\end{tabular}
\label{tab:cs2_error_metrics}
\end{table}

Table \ref{tab:cs2_error_metrics} reports the RMSE, NLL, and ECE across different UQ methods for the dataset described in Table \ref{tab:LFP dataset}. The variation in Table \ref{tab:cs2_error_metrics} results from 10 end-to-end independent runs. Note that the results may not be the best that each method could offer as all these methods are built on a backbone of a simple ResNet architecture except for GPR. It is likely that different UQ methods would require different architectures to obtain the best results. From Table~\ref{tab:cs2_error_metrics}, we observe that the GPR model perfectly fits the 41 training data points with an RMSE of zero and an extremely low NLL. However, GPR exhibits poor generalization when learning, as can be seen in the large RUL prediction error as well as high uncertainty at testing. In particular, for the secondary and tertiary test datasets that are known to be significantly different from the training dataset, the performance of GPR gets even worse. {\it {Secondly}}, the non-ensemble SNGP model performs much better in generalization when compared to GPR. The presence of neural network layers helps condense crucial information in the hidden space which is further enhanced by the spectral normalization wrapper.
But we generally found in this case study that SNGP tends to generate unnecessarily large uncertainty for each prediction, thus resulting in a large NLL and ECE. {\it {Third}}, among the two ensemble-like models, the neural network ensemble performs slightly better than MC dropout in terms of accuracy but exhibits a substantial advantage in UQ over MC dropout. We observe that the MC dropout predictions are generally overconfident with a low uncertainty estimate $\hat{\sigma}_{\text{RUL}}$ for each prediction. This low $\hat{\sigma}_{\text{RUL}}$ leads to large NLLs along with increased run-to-run variation. In the case that there is a larger $\hat{\sigma}_{\text{RUL}}$, small changes in $\hat{\mu}_{\text{RUL}}$ do not significantly affect the run-to-run variation. On the other hand, when $\hat{\sigma}_{\text{RUL}}$ is small, run-to-run variation of NLL becomes more sensitive to the changes in $\hat{\mu}_{\text{RUL}}$ around the true RUL. Note that the dropout rate hyperparameter of the MC dropout model significantly affects the model performance. A low dropout rate would lead to almost identical models within the ensemble, leading to very low predictive uncertainty and, thus, an overconfident model. On the contrary, a larger dropout rate could cause significant differences between different runs, thereby increasing uncertainty while compromising accuracy. {\it {Lastly}}, the better UQ ability of the neural network ensemble can be primarily attributed to the ability of each individual model within the ensemble to provide aleatory uncertainty, which during the ensemble process provides a more holistic picture of uncertainty.

\begin{figure}[H]
    \centering
    \includegraphics[scale=0.5]{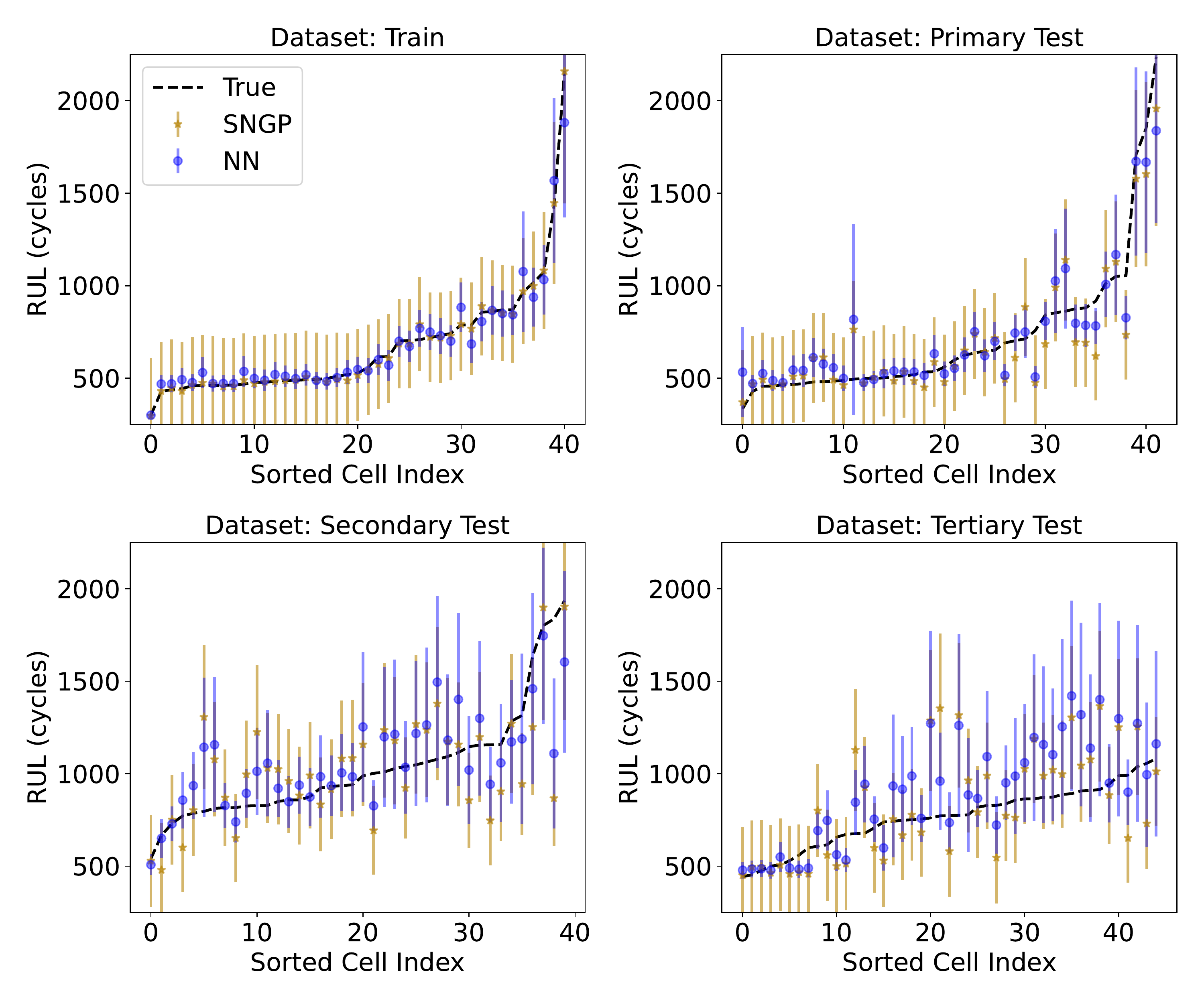}
    \caption{RUL prediction error curves with cells sorted based on true RUL values.}
    \label{fig:cs2_error_curves}
\end{figure}

Next, we visualize the prediction error with respect to a single end-to-end run for neural network ensemble and SNGP in Fig.~\ref{fig:cs2_error_curves}. To better depict prediction accuracy and the uncertainty estimate pertaining to each prediction, we plot the error curve associated with each cell in the dataset by their RUL in ascending order. As can be observed, regarding the training data, the mean RUL predictions of both SNGP and neural network ensemble models highly align with the true RUL prediction. In the case of the primary and secondary test datasets, a few instances of discrepancy between the mean RUL prediction and ground truth arise. However, these models fail to capture the true RULs of the tertiary test dataset, which is well known to be significantly different from the other three datasets. Another interesting observation across the first three considered datasets is that SNGP tends to yield a large uncertainty estimate for almost all predictions. As a result, SNGP is underconfident in most cases. In contrast, the neural network ensemble model produces significantly lower prediction uncertainty than SNGP. Only in the case of the tertiary test dataset, both neural network ensemble and SNGP associate large $\hat{\sigma}_{\text{RUL}}$ to most of the batteries.

\begin{figure}[!ht]
    \centering
    \includegraphics[scale=0.5]{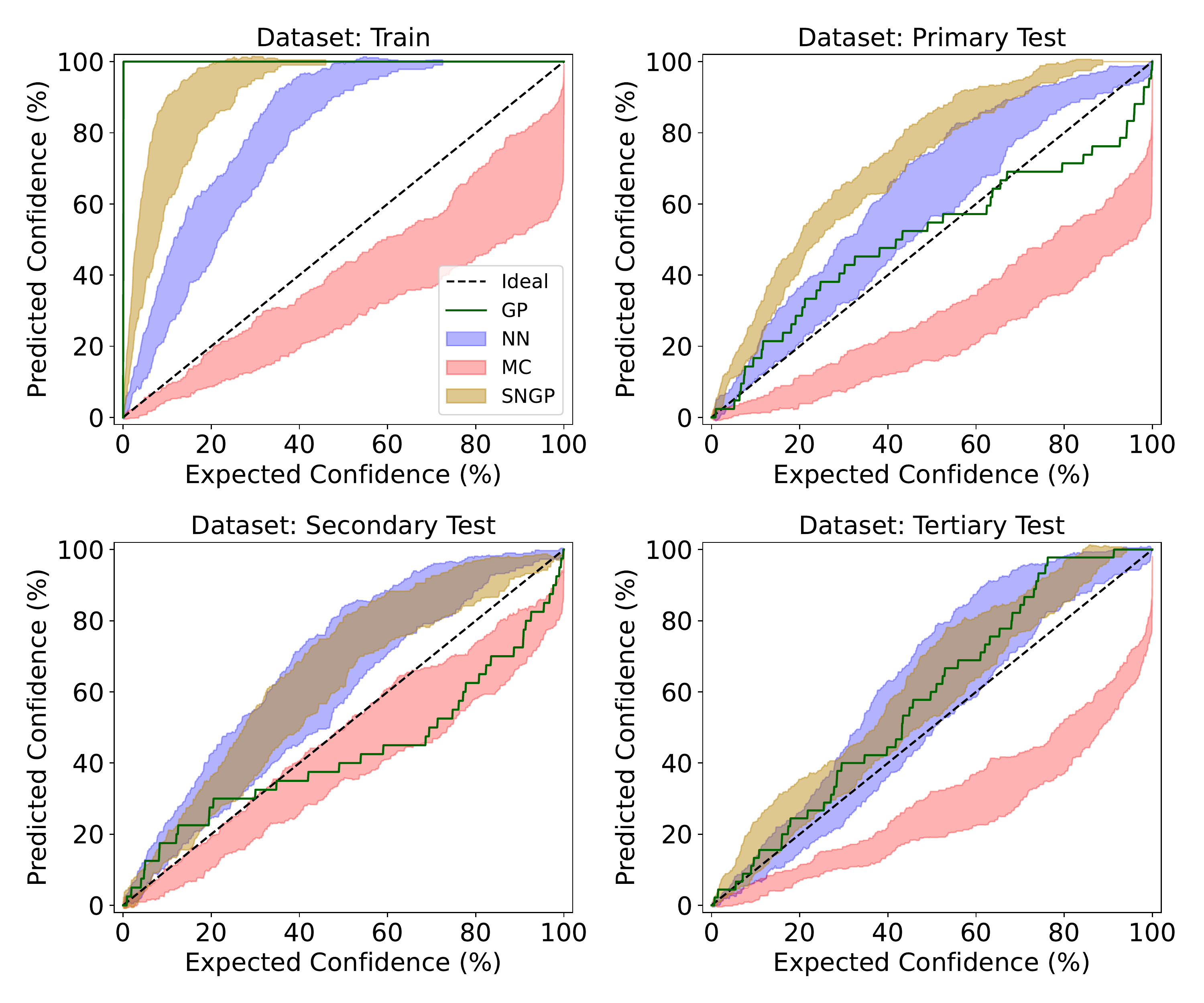}
    \caption{Calibration curves for the four models on all the datasets of the 169 LFP cell dataset. The shaded area captures the run-to-run variation of all the models.}
    \label{fig:cs2_cal_curves}
\end{figure}

In what follows, we construct the calibration curve based on each model's performance on the four datasets. As illustrated in Fig. \ref{fig:cs2_cal_curves}, the shaded area of each curve characterizes the run-to-run variation over 10 independent trials. {\it {First}}, since the GPR model fits the training data perfectly (zero RMSE), the observed confidence is 100\% and does not change with the expected confidence level. For the other datasets, GPR seems to be the closest to the expected line leading to the least ECE (see Table \ref{tab:cs2_error_metrics}). {\it {Next}}, we observe that both GPR and SNGP are relatively stable irrespective of model initialization leading to low run-to-run variation. On the other hand, models like neural network ensemble and MC dropout exhibit higher run-to-run variation (with MC dropout having the highest run-to-run variation), especially when considering OOD datasets like the tertiary dataset. These observations regarding model stability are in line with our qualitative comparison of UQ models summarized in Table \ref{tab:uq_of_ml_comp}. {\it {Lastly}}, MC dropout is generally overconfident across all the datasets, as reflected in the relatively low uncertainty associated with each RUL prediction. Different from MC dropout, neural network ensemble, and SNGP are consistently underconfident. Considering the safety-critical nature of early life prediction of batteries, underconfident models are desirable as they allow end users to stay on the safe side. 

\subsection{Case study 2: Turbofan engine prognostics}
In this section, similar to Case Study 1, we evaluate the performance of multiple UQ methods in predicting the RUL of nine turbofan engines that operate under varying conditions. To carry out our analysis, we utilize the New Commercial Modular Aero-Propulsion System Simulation (N-CMAPSS) prognostics dataset  \cite{arias2021aircraft}, which has been recently open-sourced. Specifically, we use the sub-dataset DS02, which has been used in several previous works, see Refs. \cite{chao2022fusing, tian2022real, song2022hierarchical}. Our objective is to predict the target RUL by employing a set of multivariate time series as inputs. In addition to providing a point estimate of the RUL, our aim is to quantify the uncertainty associated with the RUL prediction with the UQ methods surveyed in this paper. The code for this case study is available on our \href{https://github.com/VNemani14/UQ_ML_Review}{Github page}. The primary goal of this study is to pedagogically compare various UQ methods that exhibit similar prediction accuracy based on the current literature. We do not make any claims that the discussed methods outperform the existing literature's models.

\subsubsection{Dataset overview}

\begin{table}[!ht]
\caption[Table tb:CM]{Overview of the input variables. These condition monitoring signals include both scenario descriptors (first 6 rows) and measured physical properties (last 14 rows). The symbol used for each variable corresponds to its internal name in the CMAPSS dataset.}
\begin{center}
\begin{tabular}{cllc}
\hline
Variable No & Symbol        &  Description                       & Unit          \\ \hline
1    & alt          & Altitude                           & ft             \\
2    & XM           & Flight Mach number                 & -              \\
3    & TRA          & Throttle-resolver angle            & \%             \\
4    & T2           & Total temperature at fan inlet     & $^{\circ}$R   \\
5    & Nf           & Physical fan speed                 & rpm            \\
6    & Nc           & Physical core speed                & rpm            \\
7    & Wf           & Fuel flow                          & pps             \\
8    & T24          & Total temperature at LPC outlet    & $^{\circ}$R    \\
9    & T30          & Total temperature at HPC outlet    & $^{\circ}$R    \\
10   & T40          & Total temp. at burner outlet       & $^{\circ}$R    \\
11   & T48          & Total temperature at HPT outlet    & $^{\circ}$R    \\
12   & T50          & Total temperature at LPT outlet    & $^{\circ}$R    \\
13   & P15          & Total pressure in bypass-duct      & psia           \\
14   & P2           & Total pressure at fan inlet        & psia           \\
15   & P21          & Total pressure at fan outlet       & psia           \\
16   & P24          & Total pressure at LPC outlet       & psia           \\
17   & Ps30         & Static pressure at HPC outlet      & psia           \\
18   & P30          & Total pressure at HPC outlet       & psia           \\
19   & P40          & Total pressure at burner outlet    & psia           \\
20   & P50          & Total pressure at LPT outlet       & psia           \\ \hline
\end{tabular}
\label{tab:W_X_m2}
\end{center}
\end{table}

This case study comprises a collection of run-to-failure trajectories for a fleet of nine aircraft engines that operate under authentic flight conditions \cite{arias2021aircraft}. We use the open-source code presented in Ref. \cite{mo2022multi} to download and preprocess the data. For every RUL prediction time step, the input to the UQ model is a 20-dimensional vector that represents the measured physical properties of the engine as well as the scenario descriptors characterizing the engine's operating mode during the flight. At each time step, the UQ model produces RUL and its associated uncertainty as outputs. Table \ref{tab:W_X_m2} provides an overview of the input variables used in the model. As we adopted a purely data-driven approach, we did not utilize the virtual sensors or the calibration parameters that are available in the N-CMAPSS dataset \cite{arias2021aircraft, chao}. 

\begin{figure}[!ht]
\centering
\includegraphics[width=8cm]{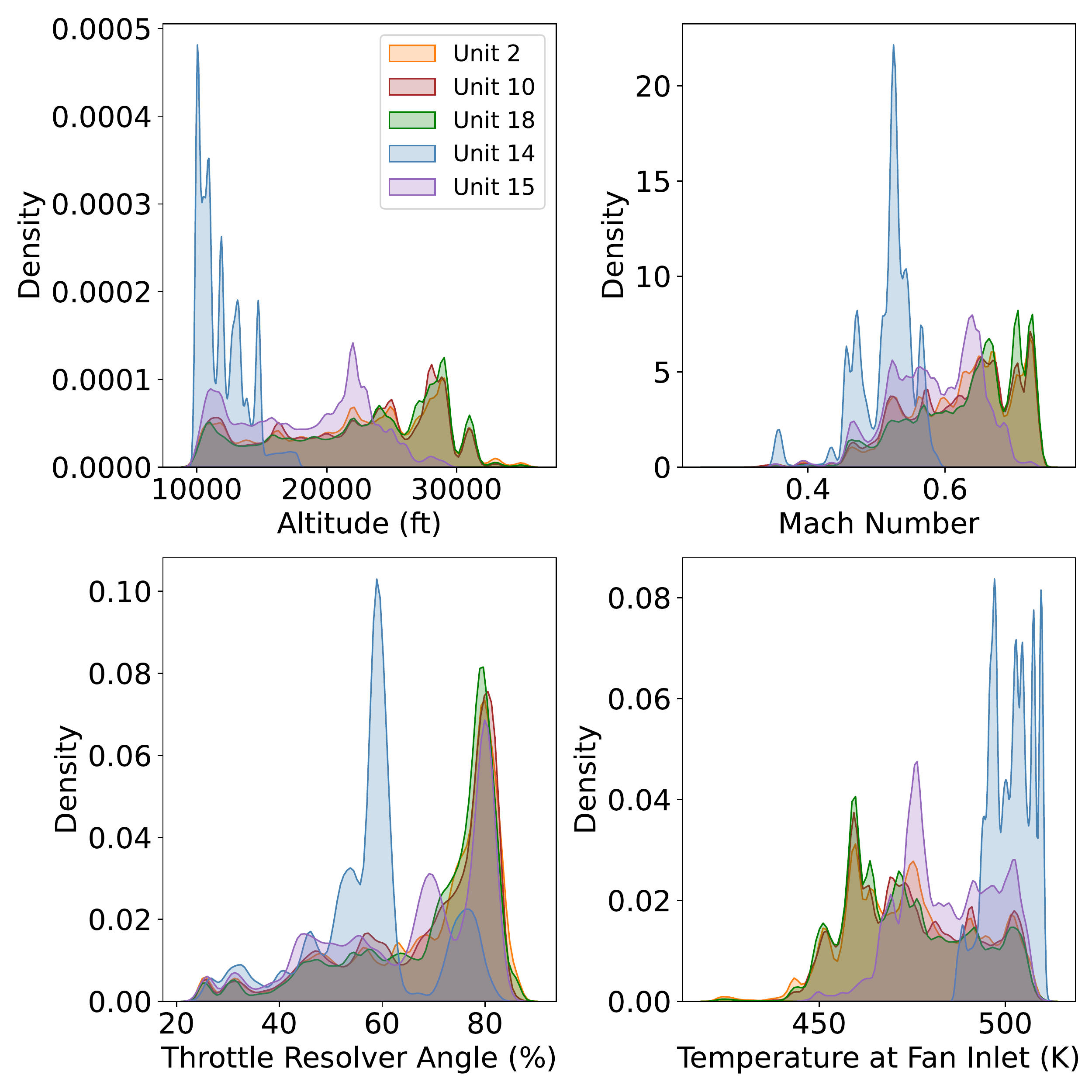}
\includegraphics[width=8cm]{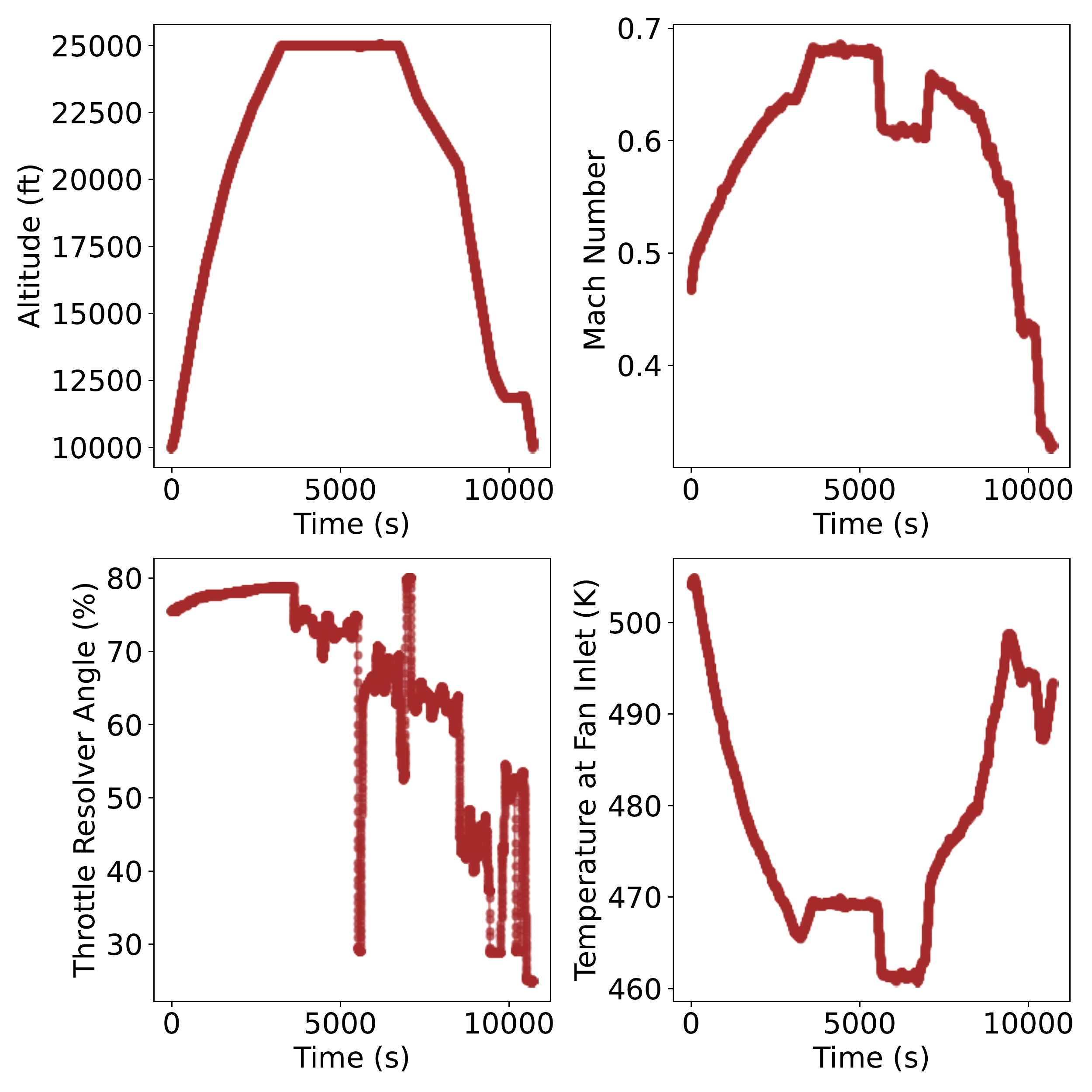}
\caption{(Left) The flight envelopes simulated for climb, cruise, and descend conditions were estimated using kernel density estimation based on measurements of altitude, flight Mach number, throttle-resolver angle, and total temperature at the fan inlet. The densities of these measurements are shown for three representative training units ($u=$ 2, 10, and 18) and two test units ($u=$ 14 and 15). (Right) A typical flight cycle for unit 10 with traces of the scenario-descriptor variables depicting the climb, cruise, and descend phases of the flight, covering different flight routes operated by the aircraft, where altitude was above 10,000 ft.}
\label{fig:operation_hist}
\end{figure}

Consistent with Ref. \cite{chao}, we split the entire dataset into a training dataset, which comprises the time-to-failure trajectories of six units (i.e., units 2, 5, 10, 16, 18 and 20), and a testing dataset, which includes the trajectories of three units (i.e., units 11, 14 and 15). Figure  \ref{fig:operation_hist} illustrates the distributions of the flight conditions across all units and provides an example of a flight cycle obtained by traces of the scenario-descriptor variables for unit 10. Finally, to address the memory consumption concerns associated with the size of the dataset, we downsampled the data by a factor of 500 by using the code from Ref. \cite{mo2022multi}, thus resulting in a sampling frequency of 0.002 Hz.

\subsubsection{Evaluation/Results}

For the sake of clarity and consistency, in this case study, we have used the same code structure/functions from the previous case study. However, we have excluded GPR from our evaluation due to the large size of the dataset and the well-known scaling issues associated with this UQ method. For further implementation details, we refer the reader to the detailed descriptions in the previous case study or to the code implementation on GitHub.

\begin{figure}[!ht]
    \centering
    \includegraphics[scale=0.37]{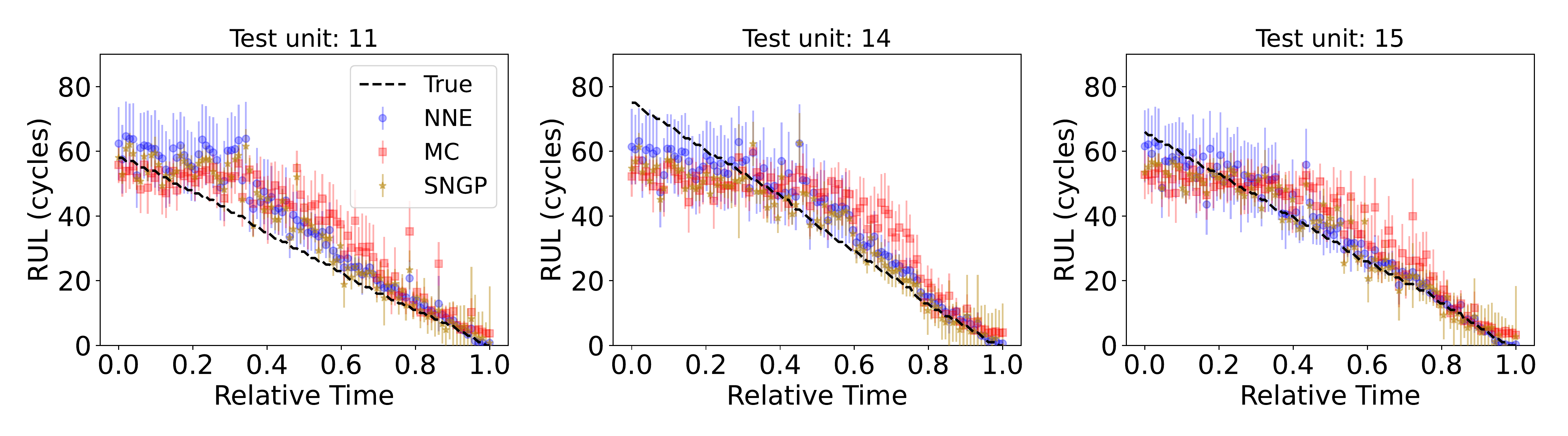}
    \caption{RUL prediction error curves for the N-CMAPSS dataset.}
    \label{fig:cs3_error_curves}
\end{figure}

The performance of NNE, MC, and SNGP on the three test units is compared in Table \ref{tab:cs3_error_metrics} using RMSE, NLL, and ECE metrics. Overall, NNE seems to outperform MC and SNGP in terms of all the metrics considered, with SNGP providing slightly better performance than MC. Figure \ref{fig:cs3_error_curves} shows that all the three models are able to capture the decreasing trend of the RUL over time, but they encounter difficulties at the beginning of the trajectory, i.e., at the onset of degradation. Interestingly, NNE appears to address this issue by assigning higher uncertainty corresponding to such points.


\begin{table}[!ht]
\caption{Comparison of the error metrics across different UQ methods on the N-CMAPSS dataset}
\centering
\begin{tabular}{c|ccc|}
\cline{2-4}
\textbf{}                              & \multicolumn{1}{c|}{\textbf{NNE}}   & \multicolumn{1}{c|}{\textbf{MC}}   & \multicolumn{1}{c|}{\textbf{SNGP}}         \\ \hline
\multicolumn{1}{|c|}{\textbf{Dataset}} & \multicolumn{3}{c|}{\textbf{RMSE (cycles) $\boldsymbol{\downarrow}$}}                                                                                                      \\ \hline
\multicolumn{1}{|c|}{Train}            & 7.1±0.1                          & 10.2±0.1                           & 8.7±0.7                                                \\ \cline{1-1}

\multicolumn{1}{|c|}{Unit 11}     & \cellcolor[HTML]{DCDCDC}8.5±0.5 & \cellcolor[HTML]{DCDCDC}10.0±0.3  & \cellcolor[HTML]{DCDCDC}8.9±1.8   \\ \cline{1-1}

\multicolumn{1}{|c|}{Unit 14}   & 7.4±0.2                        & 11.5±0.1                         & 9.3±1.4                                                \\ \cline{1-1}

\multicolumn{1}{|c|}{Unit 15}    & \cellcolor[HTML]{DCDCDC}4.8±0.3 & \cellcolor[HTML]{DCDCDC}8.2±0.2 & \cellcolor[HTML]{DCDCDC}6.8±1.2  \\ \hline
\multicolumn{1}{|l|}{}                 & \multicolumn{3}{c|}{\textbf{NLL} $\boldsymbol{\downarrow}$}                                                                                                                \\ \hline
\multicolumn{1}{|c|}{Train}            &  2.0±0.0                            & 3.7±0.1                            & 4.4±0.7                                                     \\ \cline{1-1}

\multicolumn{1}{|c|}{Unit 11}     & \cellcolor[HTML]{DCDCDC}2.3±0.1    & \cellcolor[HTML]{DCDCDC}3.0±0.1    & \cellcolor[HTML]{DCDCDC}4.8±1.8       \\ \cline{1-1}

\multicolumn{1}{|c|}{Unit 14}   & 2.2±0.0                           & 4.2±0.2                           & 4.4±1.3                                                      \\ \cline{1-1}

\multicolumn{1}{|c|}{Unit 15}    & \cellcolor[HTML]{DCDCDC}1.8±0.0    & \cellcolor[HTML]{DCDCDC}2.8±0.1    & \cellcolor[HTML]{DCDCDC}3.1±0.6       \\ \hline
\multicolumn{1}{|l|}{}                 & \multicolumn{3}{c|}{\textbf{ECE (\%)} $\boldsymbol{\downarrow}$}                                                                                                           \\ \hline

\multicolumn{1}{|c|}{Train}            & 6.2±0.8                          & 12.8±1.2                            & 9.6±2.7                                                    \\ \cline{1-1}

\multicolumn{1}{|c|}{Unit 11}     & \cellcolor[HTML]{DCDCDC}15.1±2.5   & \cellcolor[HTML]{DCDCDC}19.6±1.5   & \cellcolor[HTML]{DCDCDC}15.9±7.3       \\ \cline{1-1}

\multicolumn{1}{|c|}{Unit 14}   & 5.8±1.0                           & 25.1±1.2                           & 13.0±3.5                                                   \\ \cline{1-1}

\multicolumn{1}{|c|}{Unit 15}    & \cellcolor[HTML]{DCDCDC}14.9±2.7   & \cellcolor[HTML]{DCDCDC}11.5±1.6   & \cellcolor[HTML]{DCDCDC}8.5±3.0     \\ \hline
\end{tabular}
\label{tab:cs3_error_metrics}
\end{table}

\begin{figure}[!ht]
    \centering
    \includegraphics[scale=0.37]{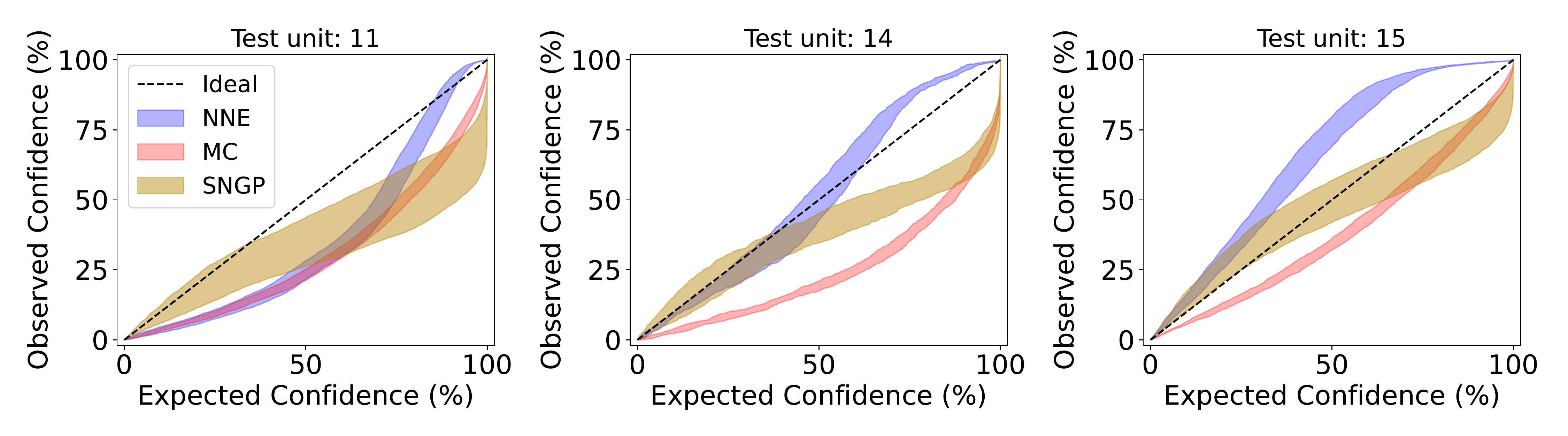}
    \caption{Calibration curves for the three models on all the datasets. The shaded area captures the run-to-run variation of all the models.}
    \label{fig:cs3_cal_curves}
\end{figure}

The calibration curves presented in Fig. \ref{fig:cs3_cal_curves} suggest that the methods used in this study tend to produce over-confident predictions, particularly for unit 11. This overconfidence can have serious implications for safety in prognostics. While MC exhibits overconfidence across all test units, NNE performs best on unit 14 and SNGP on unit 15, displaying a calibration curve that is closer to the ideal. Overall, NNE generally outperforms other UQ models as demonstrated by its accurate predictions (i.e., low RMSE and NLL scores). Furthermore, NNE's calibration curve is more closely aligned with the ideal leading to low ECE values.


As a final remark, we would like to acknowledge that the present results could be improved by optimizing the hyperparameters of each model individually, i.e., the number of layers and nodes, the dropout rate, the number of ensemble components, and the type of activation functions. However, the present study serves as a solid foundation for investigating the UQ capabilities of the analyzed methods in challenging and realistic case studies.

\section{Other topics related to UQ of ML models}

\subsection{Physics-informed ML and its synergy with UQ and probabilistic ML}

Physics-informed ML, and more broadly methods of scientific ML, has been developed to alleviate the challenge of training data scarcity and to improve the predictive capability of ML models by combining \emph{physics}-based and \emph{data}-driven modeling. Such a hybrid strategy is especially valuable for domains where training data is difficult or expensive to obtain, 
and where the modeling and downstream decision-making consequences are high (e.g., pertaining to health, safety, and security). In essence, physics-informed ML develops techniques to enable 
a seamless combination of physics-based models and observation data, or 
the embedding of physical and domain knowledge into data-driven ML models. 
Prior work on physics-informed ML can be broadly grouped into seven classes~\citep{thelen2022comprehensivepart1}: (1) impose physical knowledge as soft constraints in the loss functions of an ML model such as neural networks, 
for example the works of
physics-informed neural networks (PINNs) \cite{lagaris1998artificial,cursi2007physically,raissi2019physics}; (2) combine first-principle simulation data with experimental data to construct an augmented training dataset~\cite{ritto2021digital,oviedo2019fast}; (3) train an ML model with first-principle simulation data, then fine-tune the trained ML model with experimental data~\cite{huang2022transfer,kapusuzoglu2020physics}, which is often referred to as transfer learning; (4) build an ML model in parallel with a physics-based model, and using the ML model to learn missing/unmodeled physics from experimental data \citep{yucesan2020physics, jiang2022model}; (5) use ML models to enhance physics-based models such as in delta or residual learning \cite{thompson1994modeling,wang2017physics,thelen2022integrating} and reduced-order modeling for building models with lower complexity and degrees of freedom for rapid and reliable model evaluations \cite{azzi2023acceleration,chen2021physics,gong2022data};
(6) use neural networks to predict the input or parameters of a physics-based model \citep{yucesan2022hybrid,ramadesigan2011parameter, downey2019physics, lui2021physics}; and (7) enforcing physical models in the architecture design of neural networks, such as architectures dedicated to specific physics and engineering problems \cite{ramuhalli2005finite,darbon2021some} and utilizing a large amount of simulation data to emulate the dynamics of physical systems, such as deep operator networks \cite{lu2021learning} and Fourier neural operator \cite{li2020fourier}. A more detailed summary of these seven physics-informed ML categories can be found in Part 1 of our recent review on digital twins \citep{thelen2022comprehensivepart1}. As mentioned in this review, the above list of seven categories is not exhaustive by any means, and 
many other approaches for combining data and physics have been developed over the past decade.   
Comprehensive reviews dedicated to physics-informed ML are also available in Refs. \cite{karniadakis2021physics,cai2021physics}.

Regardless of the specific means of incorporating physical knowledge into ML modeling, parameter and model-form  uncertainty inevitably persist due to the imperfect knowledge of physics, and assumptions and approximations made to simplify the problem setup during the modeling process. In the case of uncertainty of physical parameters (e.g., uncertain parameters in a PDE), the corresponding probability distribution of solution variables can be generated with 
those parameters as inputs to 
neural network representations of the solution field \cite{zhu2019physics} or utilizing generative adversarial networks~\cite{yang2020physics}. However, these approaches do not consider the uncertainty induced by the use of physics-informed ML model itself (e.g., uncertainty due to the use of a neural network). For neural networks, the commonly used MC dropout helps increase robustness of training associated with randomization of the network architecture, while BNNs more directly seek to quantify the parameter uncertainty of the neural network (e.g., for its weight and bias terms). Moreover, physics-constrained BNNs \cite{sun2020physics,malashkhia2022physics} have been developed to address the uncertainty in PINNs. We direct interested readers to two recent review papers for a more comprehensive, in-depth discussion on UQ for physics-informed ML~\cite{cuomo2022scientific,psaros2023uncertainty}, with emphasis on PINNs~\cite{cuomo2022scientific,psaros2023uncertainty} and deep operator networks~\cite{psaros2023uncertainty}.

\subsection{Probabilistic Learning on Manifolds (PLoM)}

Another ML approach that naturally captures the uncertainty of data while simultaneously performing dimension reduction is the Probabilistic Learning on Manifolds (PLoM)~\cite{Soize2019}. PLoM builds a generative model from an initial set of data samples by identifying a manifold where the unknown probability measure concentrates.
The learning procedure starts by scaling the training data via principal component analysis (PCA) followed by performing a density estimation (e.g., Gaussian kernel density estimation) on the training data in the PCA space. Then, an It\^{o} stochastic differential equation is established as a sample-generating mechanism whose invariant distribution matches the probability density just estimated. In order to ensure the generated samples coalesce around a low-dimensional manifold, additional structure is injected by forming a reduced-order ``diffusion-maps basis'' induced by an isotropic diffusion kernel to help constrain the sample coordinates. Putting everything together, new samples consistent with the training data distribution can be generated on a low-dimensional manifold by numerically solving the It\^{o} equations through a discretization scheme.

With its ability to find low-dimensional manifolds, PLoM is particularly suitable for dimension reduction of high-dimensional datasets. Its strength and focus thus differs from ML constructs such as GPR and BNN that are more directly concerned with function approximation and regression tasks. To be effective, PLoM generally requires a sufficiently large quantity of data samples that can reasonably reveal the underlying distribution geometry. 
This also differs from GPR and BNN that, by design, engender a larger degree of uncertainty in the model (e.g., by falling back towards their prior uncertainty) when less data is available. Nonetheless, PLoM has been demonstrated to work well even in settings with relatively small datasets, especially if additional constraining from relevant governing PDEs is available~\cite{Soize2021}. Lastly, the generative model resulting from PLoM can be highly versatile and used for a range of applications beyond sample generation and surrogate modeling, such as density estimates of statistics of interest~\cite{Soize2019a}, optimization under uncertainty~\cite{Ghanem2019}, and design using digital twins~\cite{Ghanem2022}, as some examples.

\subsection{Interpretability of ML models for dynamic systems}

Data-driven system identification plays a vital role in structural health monitoring, system failure prognostics, design and control as well as risk assessment of dynamic systems. In the past decades, various approaches have been developed to accomplish this task. Some representative examples include autoregressive models, autoregressive–moving-average models, nonlinear autoregressive moving average with exogenous inputs models, the Volterra series gray-box tooling method, and ML-based methods emerging in recent years~\cite{thelen2022comprehensive, thelen2023comprehensive}. While these black-box or grey-box models show promising performance in various applications, they are often criticized for their lack of interpretability.

As introduced earlier, significant efforts have been devoted to addressing the challenge of interpretability in ML models. Among the techniques that stand out are SHAP, Grad-CAM, and other methods. Notably, over the past decade, there has been a remarkable stride in enhancing the interpretability of ML models through the integration of data, genetic programming, and sparsity. This fusion has led to the formulation of evolution equations that are not only simplistic but also parsimonious. Several approaches have been proposed to construct interpretable ML models, particularly symbolic regression, which has been applied with different techniques \cite{angelis2023artificial}. A pivotal advancement in this realm is the emergence of the Sparse Identification of Nonlinear Dynamics (SINDy) technique, which has become a cornerstone in addressing this issue. Initially proposed by~\citet{brunton2016discovering}, SINDy aims to uncover the underlying partial differential equations governing nonlinear dynamic systems. This discovery is accomplished even in the presence of noisy measurement data~\cite{rudy2017data,kaheman2020sindy}.

What sets SINDy apart is its ability to exploit the dominance of only a handful of terms in shaping the behavior of nonlinear dynamic systems. This is achieved by encouraging sparsity in the data-driven identification of governing equations, leveraging an extensive library of potential function bases. From the model interpretation point of view, the sparsity promoting the discovery of governing equations of dynamic systems results in parsimonious and interpretable models that strike a sound balance between regression accuracy with model complexity. In particular, the parsimonious model is achieved by employing sparsity-promoting regularization techniques~\cite{brunton2016discovering,hirsh2022sparsifying}, such as LASSO regression, also known as L1 regularization, using sparsifying priors, hard thresholding with Pareto analysis. The resulting parsimonious representations through sparsity lead to interpretable models with good generalization to unseen data. Besides, the sparsity in the resulting function basis offers valuable insights into the management of model selection uncertainty in the context of hybrid dynamical systems~\cite{mangan2019model}. For instance, hybrid SINDy employed the Akaike information criterion score on out-of-sample validation data to match the SINDy model with a specific regime in a hybrid dynamical system, from which the switching point of the hybrid system can be found~\cite{mangan2019model}.

The elegance and clarity inherent in the models derived through SINDy are of particular importance when considering ML model interpretability. Building upon the foundational work of Brunton and Kutz, a multitude of SINDy variants have emerged, finding applications even in UQ contexts. A remarkable instance worth highlighting is the approach introduced by~\citet{hirsh2022sparsifying}, wherein the SINDy approach is extended into a Bayesian probabilistic framework. This novel approach, termed Uncertainty Quantification SINDy (UQ-SINDy), accounts for uncertainties in SINDy coefficients arising from observation errors and limited data. The central innovation lies in the integration of sparsifying priors, specifically the spike and slab prior and the regularized horseshoe prior, into the Bayesian inference of SINDy coefficients. By unifying UQ with SINDy variants, this approach not only heightens the interpretability of ML models but also facilitates the quantification of the prediction's confidence level.

\subsection{PCE and its relationship with GPR and connection with ML}
A key role for both GPR (see Sec. \ref{sec:introduction} and \ref{sec:UQ_design}) and polynomial chaos expansion (PCE) is building surrogate models for solving engineering design problems. The need for surrogate modeling stems from the multi-query nature of uncertainty propagation and design optimization, which often require many repeated simulation runs (e.g., $10^3-10^6$) to assess the behavior of output responses under different realizations of input design variables and simulation model parameters. This process may become prohibitively expensive for high-fidelity models where each simulation may require hours to days. One strategy to accelerate these computations, as explained in \ref{sec:needs_ML}, is to build a cheap-to-evaluate surrogate of the computationally expensive simulation model---i.e. to trade model fidelity for speed. The surrogate model, sometimes called metamodel or response surface, is often an explicit mathematical function (e.g., as in GPR and PCE), allowing for rapid predictions at different input realizations.

Having presented GPR in detail in Sec. \ref{sec:introduction} and \ref{sec:UQ_design}, we briefly introduce PCE here. PCE was originally proposed in the 1930s to model stochastic processes using a spectral expansion of multivariate Hermite polynomials of Gaussian random variables \citep{wiener1938homogeneous}. These Hermite polynomial basis functions are orthogonal with respect to the joint probability distribution of the respective Gaussian variables. PCE was later applied to solve physics and engineering problems \citep{ghanem2003stochastic} and extended to non-Gaussian probability distributions, giving rise to the \emph{generalized PCE} \citep{xiu2002wiener}. Since the input variables of a PCE are naturally formulated to follow certain probability distributions, PCE has been a convenient and popular tool for conducting UQ. However, PCE has not been employed much for UQ of ML models, since most ML models are already relatively inexpensive to evaluate; rather, PCE brings more value for enabling UQ of expensive computer simulation models. 
In that case, a PCE surrogate model is built to approximate the original simulation model, where the PCE's expansion coefficients can be computed, for example, non-intrusively by projection (numerical integration via quadrature or simulation) \citep{le2002stochastic} or regression (least squares minimization of the fitting error) \citep{berveiller2006stochastic}.

One major challenge faced by PCE is the curse of dimensionality, where the number of model parameters (and in turn training samples of the simulation model) increases exponentially with the input dimension (i.e., the number of input random variables). Several algorithmic techniques have been developed to alleviate this issue
through truncation schemes that can identify a sparse set of important polynomials to be included. Two notable methods for introducing sparsity are the Smolyak sparse constructions (and their adaptive versions) \citep{Smolyak1963,Constantine2012,Conrad2013}, and variants of compressive sensing (such as least angle regression and LASSO) \citep{blatman2011adaptive,hampton2015compressive, tsilifis2019compressive}. 
Such effort has been made in the context of surrogate modeling \citep{blatman2010adaptive, blatman2011adaptive, hampton2015compressive, tsilifis2019compressive} and reliability analysis \citep{hu2011adaptive,pan2017sliced,xu2018cubature,bhattacharyya2021structural}. 
A comprehensive review of sparse PCE is provided in Ref. \citep{luthen2021sparse}.

Historically, PCE and GPR (or kriging) have been studied separately and mostly in isolation, although both methods have produced many success stories in surrogate modeling. Recently, attempts have been made to combine PCE and kriging, resulting in PCE-kriging hybrids \citep{schobi2015polynomial}. The basic idea is to use PCE to represent the mean function $m(\mathbf{x})$ of the Gaussian process prior (see Eq. (\ref{eq:priormean})) that captures the global trend of the computer simulation model (i.e., $f(\mathbf{x})$). The GPR formulation with a non-zero, non-constant mean function is called universal kriging, which differs from ordinary kriging where the mean function is set as a constant (e.g., zero). When combined with kriging in this manner, PCE serves the purpose of a \emph{deterministic} (non-probabilistic) mean (trend) function. 
Such PCE-kriging hybrids have found applications to uncertainty propagation in computational dosimetry \citep{kersaudy2015new} and damage quantification in structural health monitoring \citep{pavlack2022polynomial}. 
More broadly, while PCE is typically not used for UQ of ML models, it may be combined with other ML techniques (e.g., kriging \citep{schobi2015polynomial} and radial basis functions \citep{shang2021efficient}) to produce hybrid PCE-ML models with improved prediction accuracy over standalone PCE surrogates. 
On a final note, although PCE is typically not categorized as an ML technique, it was reported to offer surrogate modeling accuracy on par with state-of-the-art ML techniques such as regression tree, neural network, and support vector machine \citep{torre2019data}.

\section{Conclusion and outlook}
\label{sec:conclusion_outlook}
This tutorial aims to cover the fundamental role of UQ in ML, particularly focusing on a detailed introduction of state-of-the-art UQ methods for neural networks and a brief review of applications in engineering design and PHM. It possesses four salient characteristics: (1) classification of uncertainty types (aleatory vs. epistemic), sources, and causes pertaining to ML models; (2) tutorial-style descriptions of emerging UQ techniques; (3) quantitative metrics for evaluation and calibration of predictive uncertainty; and (4) easily accessible source codes for implementing and comparing several state-of-the-art UQ techniques in engineering design and PHM applications. Two case studies are developed to demonstrate the implementation of UQ methods and benchmark their performance in predicting battery life using early-life data (case study 1) and turbofan engine RUL using online-accessible measurements (case study 2). Our rigorous examination of the state-of-the-art techniques for UQ, calibration, and evaluation and the two case studies offers a holistic lens on pressing issues that need to be tackled along the future development of UQ techniques in terms of scalability, principleness, and decomposition given the increasing importance of UQ in safeguarding the usage of ML models in high-stakes applications. 

It is important to note that the case studies presented in this paper are not optimized in terms of their hyperparameters, and it is reasonable to expect that optimizing them would yield even better performance results. The primary objective of this paper is to offer a user-friendly platform for individuals seeking to comprehend the analyzed methods and to encourage them to enhance and suggest new ones.

Essentially, UQ acts as a layer of safety assurance on top of ML models, enabling rigorous and quantitative risk assessment and management of ML solutions in high-stakes applications. As UQ methods for ML models continue to mature, they are anticipated to play a crucial role in creating safe, reliable, and trustworthy ML solutions by safeguarding against various risks such as OOD, adversarial attacks, and spurious correlations. From this perspective, the development of UQ methods is of paramount significance in expanding the adoption of ML models in breadth and depth. The accurate, sound, and principled quantification of uncertainty in ML model prediction has great potential to fundamentally tackle the safety assurance problem that haunts ML's development. Towards this end, several long-standing challenges encompassing the UQ development need to be addressed by the research community: 
\begin{enumerate}
    \item The need for a unified and well-acknowledged testbed to comprehensively examine the performance of the diverse and expanding set of UQ methods in uncertainty quantification, calibration (and recalibration), decomposition, attribution, and interpretation. Although some recent efforts were devoted to developing standardized benchmarks for UQ~\cite{nado2021uncertainty}, most of these efforts primarily emphasized conventional performance metrics, such as prediction accuracy metrics and UQ calibration errors. However, other key performance aspects (e.g., uncertainty decomposition and uncertainty attribution) essential to ensuring high quality UQ have rarely been investigated. The lack of these key elements emerges as a significant challenge to the sound development of the UQ ecosystem. Hence, there is an imperative demand calling for establishing UQ testbeds with community-acknowledged standards to facilitate comprehensive testing and verification of the behavior of uncertainty generated by different UQ methods, especially on edge cases. Establishing such testbeds with the support of synthetic data generation is expected to tremendously benefit the long-term and sustainable development of UQ methods for ML models.
    
    \item The need for principled, scalable, and computationally efficient UQ methods to enable high quality and large-scale UQ. As summarized in Table~\ref{tab:uq_of_ml_comp}, each method covered in this tutorial has its own strengths and shortcomings. Although numerous efforts have been made to elevate the soundness and principleness of UQ methods of ML models, the existing methods still suffer from a common but critical deficiency: a lack of (limited) theoretical guarantee in detecting OOD instances. It is thus imperative to investigate further along this direction to fill the loophole. Emerging deterministic methods such as SNGP exhibit a strong OOD detection capability due to distance awareness. In addition, the computational efficiency of UQ methods needs to be further improved to satisfy the need for real-time or near real-time decision making in a broad range of safety-critical applications (e.g., autonomous driving and aviation). Thus, more research efforts need to be invested in enabling three key essential features of high quality UQ: principleness, scalability, and efficiency.



    
    \item ML models have shown promising potential in addressing long-standing engineering design problems in recent years. Especially for GPR, its applications in engineering design have resulted in a family of adaptive surrogate modeling methods for reliability-based design optimization, robust design, and design optimization in general. These ML-based design methods have revolutionized engineering design in various applications, including but not limited to design and discovery of new materials, design for additive manufacturing, and topology optimization. Despite these revolutionary advances, extending these methods to larger-scale and more complicated problems becomes increasingly urgent. To this end, various DNN-based methods have been investigated in engineering design to overcome the limitations of classical ML methods, such as the GPR-based approaches. Even though the emerging DNN-based methods show promise in addressing computational challenges in high-dimensional engineering design problems, their potential as efficient surrogates or accelerated optimizers has not yet been fully realized. The UQ methods for ML models presented in this paper will play a key role in fully releasing the power of DNNs in engineering design by enabling adaptive DNNs in the context of active learning to reduce the required quantity of training data without sacrificing the accuracy in surrogate modeling, reliability analysis, and optimization, (2) accelerated design optimization for large-scale systems, and (3) efficient and accurate UQ in engineering design accounting for various sources of aleatory and epistemic uncertainty (e.g., input-dependent aleatory uncertainty).
    
    \item The PHM community has long recognized the importance of estimating the predictive uncertainty of prognostic models. These prognostic models can be built based on supervised ML or more traditional state-space models (see, for example, the Bayes filter in one of the earliest studies on battery prognostics \citep{saha2008prognostics}). As discussed in Sec. \ref{sec:phm_discussion}, in the PHM field, UQ of ML models has been predominantly applied to the task of predicting the RUL of a system or component. The focus of UQ in this context is to provide a probability distribution of the RUL rather than a single point estimate. While UQ in the PHM field has primarily been focused on RUL prediction, there is a growing interest in applying UQ to other tasks, such as anomaly detection, fault detection and classification, and health estimation. Many of the UQ methods discussed in detail in Sec. \ref{sec:UQ_methods} can also be readily applied to these classification and regression tasks in the PHM field. Looking ahead, we identify three research directions along which positive and significant impacts could be made on the PHM field surrounding UQ of ML models. First, decomposing the total predictive uncertainty into its aleatoric and epistemic components is highly desirable and sometimes essential, as noted in Sec. \ref{sec:phm_discussion}. Such a decomposition has several benefits, for example, highlighting the need for improved sensing solutions with lower measurement noise to reduce aleatory uncertainty and identifying areas where further data collection or model refinement efforts may be necessary to reduce epistemic uncertainty. More work is needed to develop UQ methods with built-in uncertainty decomposition capability and create procedures to assess the accuracy of uncertainty decomposition. Second, prognostic studies involving UQ mostly evaluate UQ quality subjectively and qualitatively by looking at whether a two-sided 95\% confidence interval of the RUL estimate gets narrows with time and contains the true RUL, especially toward the end of life. As discussed in a general context in Sec. \ref{sec:accuracy_vs_UQquality}, we call for consistent effort among PHM researchers and practitioners to quantitatively evaluate their ML models’ UQ quality using some of the metrics introduced in Sec. \ref{sec:uncertainty_evaluation}, such as calibration metrics (Sec. \ref{sec:calibration}), sparsification metrics (Sec. \ref{sec:sparsification}), and NLL (Sec. \ref{sec:nll}). Ideally, UQ quality assessment should also become standard practice when building and deploying ML models in PHM applications, just as prediction accuracy assessment is currently standard practice. Third, both UQ and interpretation serve the purpose of improving model transparency and trustworthiness, as noted in Sec. \ref{sec:introduction}. An under-explored question is whether UQ capability can help improve interpretability and vice versa. For example, interpretability can provide insights into the most important input features for making predictions. Such an understanding could allow distance-aware UQ models to define their distance measures based only on highly important features, potentially improving the UQ quality.
    
    \item Model uncertainty quantification for label-free learning is another future research direction. Obtaining labels by solving implicit engineering physics models is usually costly. Label-free machine learning embeds physics models in a cost function or as constraints in the model training process without solving them. As a result, labels are not required. Physics-informed neural network (PINN) is one such label-free method ~\cite{raissi2019physics,karniadakis2021physics}. This method has gained much attention because it makes the regression task feasible without solving the true label. In addition, the physical constraints prevent the regression from severe overfitting in conventional neural networks, especially when data are limited. Since labels are not available, the quantification of prediction uncertainty of the machine learning model is extremely difficult. Even the prediction errors at the training points are unknown. Due to this reason, the GPR method has not been used for label-free learning since the prediction of a GPR model requires labels at the training points. A proof-of-concept study has been conducted for quantifying epistemic uncertainty for physics-based label-free regression~\cite{li2022uncertainty}. This method integrates neural networks and GPR models and can produce both systematic error (represented by a mean) and random error (represented by a standard deviation) for a model prediction. The method, however, has not been extended to time- and space-dependent problems where partially different equations are involved. There is a need to develop generic uncertainty quantification methods for label-free learning.
    
\end{enumerate}

\section*{Authors' contributions}
All the authors read and approved the final manuscript. Hu, C. and Zhang, X. devised the original concept of the tutorial paper. Hu, Z., Hu, C., Du, X., Wang, Y., and Huan, X. were responsible for the classification of types and sources of uncertainty pertaining to ML models. Hu, C. and Tran, A. were responsible for GPR. Huan, X. was responsible for implementing BNN via the means of MCMC and variational inference. Zhang, X. and X. Huan were responsible for MC dropout. Zhang, X. and Hu, C. were responsible for neural network ensemble. Hu, C. was responsible for deterministic methods for UQ of neural networks. Zhang, X. and Nemani, V., were responsible for the toy example to compare the predictive uncertainty produced by different UQ methods. Zhang, X. and Hu, C. were responsible for the summary of the qualitative comparison of different UQ methods. Hu, C. and Nemani, V. were responsible for the evaluation of predictive uncertainty. Hu, Z., Zhang, X., Hu., C., and Tran, A. were responsible for the review on UQ of ML models in engineering design. Biggio, L., and Fink, O. were responsible for the review of UQ of ML models in prognostics. Nemani, V. and Hu, C. were responsible for case study 1 -- battery early life prediction. Biggio L. and Fink O. were responsible for case study 2 -- turbofan engine prognostics. Zhang, X., Hu, C., and Hu, Z. were responsible for the conclusion and outlook. All authors participated in manuscript writing, review, and editing. All correspondence should be addressed to Xiaoge
Zhang (e-mail: xiaoge.zhang@polyu.edu.hk) and Chao Hu (e-mails: chao.hu@uconn.edu; huchaostu@gmail.com).

\section*{Acknowledgements}
Xiaoping Du at Indiana University–Purdue University Indianapolis contributed to this manuscript by providing helpful inputs on Section 2 surrounding the classification of types and sources of uncertainty pertaining to ML models. Luca Biggio acknowledges the financial support from the CSEM Data Program fund. Xun Huan acknowledges the financial support provided by the U.S. Department of Energy, Office of Science, Office of Advanced Scientific Computing Research (ASCR), under Award Number DE-SC0021397. Zhen Hu acknowledges financial support from the United States Army Corps of Engineers through the US Army Engineer Research and Development Center Research Cooperative Agreement W9132T-22-2-20014, the U.S. Army CCDC Ground Vehicle Systems Center (GVSC) through the Automotive Research Center (ARC) in accordance with Cooperative Agreement W56HZV-19-2-0001, and the U.S. National Science Foundation
under Grant CMMI-2301012. Olga Fink acknowledges the financial support from the Swiss National Science Foundation under the Grant Number 	$200021\textunderscore200461$. Yan Wang received financial support from the U.S. National Science Foundation under Grant Nos. CMMI-1306996 and CMMI-1663227, as well as the George W. Woodruff Faculty Fellowship at the Georgia Institute of Technology. Xiaoge Zhang was supported by a grant from the Research Grants Council of the Hong Kong Special Administrative Region, China (Project No. PolyU 25206422) and the Research Committee of The Hong Kong Polytechnic University under project code G-UAMR. He was also partly supported by the Centre for Advances in Reliability and Safety (CAiRS), admitted under AIR@InnoHK Research Cluster. Chao Hu received financial support from the U.S. National Science Foundation under Grant No. ECCS-2015710. The opinions, findings, and conclusions presented in this article are solely those of the authors and do not necessarily reflect the views of the sponsors that provided funding support for this research. 

Sandia National Laboratories is a multimission laboratory managed and operated by National Technology and Engineering Solutions of Sandia, LLC., a wholly owned subsidiary of Honeywell International, Inc., for the U.S. Department of Energy's National Nuclear Security Administration under contract DE-NA-0003525. 

\bibliographystyle{elsarticle-num-names}
\bibliography{ref}

\begin{thebibliography}{388}
\providecommand{\natexlab}[1]{#1}
\providecommand{\url}[1]{\texttt{#1}}
\providecommand{\urlprefix}{URL }
\expandafter\ifx\csname urlstyle\endcsname\relax
  \providecommand{\doi}[1]{doi:\discretionary{}{}{}#1}\else
  \providecommand{\doi}[1]{doi:\discretionary{}{}{}\begingroup
  \urlstyle{rm}\url{#1}\endgroup}\fi
\providecommand{\bibinfo}[2]{#2}

\bibitem[{LeCun et~al.(1998)LeCun, Bottou, Bengio, and
  Haffner}]{lecun1998gradient}
\bibinfo{author}{Y.~LeCun}, \bibinfo{author}{L.~Bottou},
  \bibinfo{author}{Y.~Bengio}, \bibinfo{author}{P.~Haffner},
  \bibinfo{title}{Gradient-based learning applied to document recognition},
  \bibinfo{journal}{Proceedings of the IEEE}
  \bibinfo{volume}{86}~(\bibinfo{number}{11}) (\bibinfo{year}{1998})
  \bibinfo{pages}{2278--2324}, \bibinfo{note}{doi:
  \url{http://dx.doi.org/10.1109/5.726791}}.

\bibitem[{Deng et~al.(2009)Deng, Dong, Socher, Li, Li, and
  Fei-Fei}]{deng2009imagenet}
\bibinfo{author}{J.~Deng}, \bibinfo{author}{W.~Dong},
  \bibinfo{author}{R.~Socher}, \bibinfo{author}{L.-J. Li},
  \bibinfo{author}{K.~Li}, \bibinfo{author}{L.~Fei-Fei},
  \bibinfo{title}{{ImageNet}: A large-scale hierarchical image database}, in:
  \bibinfo{booktitle}{2009 IEEE Conference on Computer Vision and Pattern
  Recognition}, \bibinfo{organization}{IEEE}, \bibinfo{pages}{248--255},
  \bibinfo{note}{doi: \url{http://dx.doi.org/10.1109/CVPR.2009.5206848}},
  \bibinfo{year}{2009}.

\bibitem[{Zhou et~al.(2014)Zhou, Lapedriza, Xiao, Torralba, and
  Oliva}]{zhou2014learning}
\bibinfo{author}{B.~Zhou}, \bibinfo{author}{A.~Lapedriza},
  \bibinfo{author}{J.~Xiao}, \bibinfo{author}{A.~Torralba},
  \bibinfo{author}{A.~Oliva}, \bibinfo{title}{Learning deep features for scene
  recognition using places database}, \bibinfo{journal}{Advances in Neural
  Information Processing Systems} \bibinfo{volume}{27}.

\bibitem[{Lin et~al.(2014)Lin, Maire, Belongie, Hays, Perona, Ramanan,
  Doll{\'a}r, and Zitnick}]{lin2014microsoft}
\bibinfo{author}{T.-Y. Lin}, \bibinfo{author}{M.~Maire},
  \bibinfo{author}{S.~Belongie}, \bibinfo{author}{J.~Hays},
  \bibinfo{author}{P.~Perona}, \bibinfo{author}{D.~Ramanan},
  \bibinfo{author}{P.~Doll{\'a}r}, \bibinfo{author}{C.~L. Zitnick},
  \bibinfo{title}{Microsoft {COCO}: Common objects in context}, in:
  \bibinfo{booktitle}{European Conference on Computer Vision},
  \bibinfo{organization}{Springer}, \bibinfo{pages}{740--755},
  \bibinfo{year}{2014}.

\bibitem[{Blitzer et~al.(2007)Blitzer, Dredze, and
  Pereira}]{blitzer2007biographies}
\bibinfo{author}{J.~Blitzer}, \bibinfo{author}{M.~Dredze},
  \bibinfo{author}{F.~Pereira}, \bibinfo{title}{Biographies, bollywood,
  boom-boxes and blenders: Domain adaptation for sentiment classification}, in:
  \bibinfo{booktitle}{Proceedings of the 45th Annual Meeting of the Association
  of Computational Linguistics}, \bibinfo{pages}{440--447},
  \bibinfo{year}{2007}.

\bibitem[{Glorot et~al.(2011)Glorot, Bordes, and Bengio}]{glorot2011domain}
\bibinfo{author}{X.~Glorot}, \bibinfo{author}{A.~Bordes},
  \bibinfo{author}{Y.~Bengio}, \bibinfo{title}{Domain adaptation for
  large-scale sentiment classification: A deep learning approach}, in:
  \bibinfo{booktitle}{Proceedings of the 28th International Conference on
  Machine Learning (ICML-11)}, \bibinfo{pages}{513--520}, \bibinfo{year}{2011}.

\bibitem[{Li et~al.(2021{\natexlab{a}})Li, Shen, Chen, and
  Zhu}]{li2021knowledge}
\bibinfo{author}{Q.~Li}, \bibinfo{author}{C.~Shen}, \bibinfo{author}{L.~Chen},
  \bibinfo{author}{Z.~Zhu}, \bibinfo{title}{Knowledge mapping-based adversarial
  domain adaptation: A novel fault diagnosis method with high generalizability
  under variable working conditions}, \bibinfo{journal}{Mechanical Systems and
  Signal Processing} \bibinfo{volume}{147} (\bibinfo{year}{2021}{\natexlab{a}})
  \bibinfo{pages}{107095}, \bibinfo{note}{doi:
  \url{https://doi.org/10.1016/j.ymssp.2020.107095}}.

\bibitem[{Lundberg and Lee(2017)}]{lundberg2017unified}
\bibinfo{author}{S.~M. Lundberg}, \bibinfo{author}{S.-I. Lee},
  \bibinfo{title}{A unified approach to interpreting model predictions},
  \bibinfo{journal}{Advances in Neural Information Processing Systems}
  \bibinfo{volume}{30}.

\bibitem[{Selvaraju et~al.(2017)Selvaraju, Cogswell, Das, Vedantam, Parikh, and
  Batra}]{selvaraju2017grad}
\bibinfo{author}{R.~R. Selvaraju}, \bibinfo{author}{M.~Cogswell},
  \bibinfo{author}{A.~Das}, \bibinfo{author}{R.~Vedantam},
  \bibinfo{author}{D.~Parikh}, \bibinfo{author}{D.~Batra},
  \bibinfo{title}{Grad-{CAM}: Visual explanations from deep networks via
  gradient-based localization}, in: \bibinfo{booktitle}{Proceedings of the IEEE
  International Conference on Computer Vision}, \bibinfo{pages}{618--626},
  \bibinfo{note}{doi: \url{http://dx.doi.org/10.1109/ICCV.2017.74}},
  \bibinfo{year}{2017}.

\bibitem[{Molnar(2020)}]{molnar2020interpretable}
\bibinfo{author}{C.~Molnar}, \bibinfo{title}{Interpretable machine learning},
  \bibinfo{publisher}{Lulu. com}, \bibinfo{year}{2020}.

\bibitem[{Jim{\'e}nez-Luna et~al.(2020)Jim{\'e}nez-Luna, Grisoni, and
  Schneider}]{jimenez2020drug}
\bibinfo{author}{J.~Jim{\'e}nez-Luna}, \bibinfo{author}{F.~Grisoni},
  \bibinfo{author}{G.~Schneider}, \bibinfo{title}{Drug discovery with
  explainable artificial intelligence}, \bibinfo{journal}{Nature Machine
  Intelligence} \bibinfo{volume}{2}~(\bibinfo{number}{10})
  (\bibinfo{year}{2020}) \bibinfo{pages}{573--584}, \bibinfo{note}{doi:
  \url{https://doi.org/10.1038/s42256-020-00236-4}}.

\bibitem[{Guo et~al.(2022)Guo, Ding, Li, Feng, Xiong, Su, and
  Feng}]{guo2022hierarchical}
\bibinfo{author}{S.~Guo}, \bibinfo{author}{H.~Ding}, \bibinfo{author}{Y.~Li},
  \bibinfo{author}{H.~Feng}, \bibinfo{author}{X.~Xiong},
  \bibinfo{author}{Z.~Su}, \bibinfo{author}{W.~Feng}, \bibinfo{title}{A
  hierarchical deep convolutional regression framework with sensor network
  fail-safe adaptation for acoustic-emission-based structural health
  monitoring}, \bibinfo{journal}{Mechanical Systems and Signal Processing}
  \bibinfo{volume}{181} (\bibinfo{year}{2022}) \bibinfo{pages}{109508},
  \bibinfo{note}{doi: \url{https://doi.org/10.1016/j.ymssp.2022.109508}}.

\bibitem[{Khan and Yairi(2018)}]{khan2018review}
\bibinfo{author}{S.~Khan}, \bibinfo{author}{T.~Yairi}, \bibinfo{title}{A review
  on the application of deep learning in system health management},
  \bibinfo{journal}{Mechanical Systems and Signal Processing}
  \bibinfo{volume}{107} (\bibinfo{year}{2018}) \bibinfo{pages}{241--265},
  \bibinfo{note}{doi: \url{https://doi.org/10.1016/j.ymssp.2017.11.024}}.

\bibitem[{Thelen et~al.(2022{\natexlab{a}})Thelen, Zhang, Fink, Lu, Ghosh,
  Youn, Todd, Mahadevan, Hu, and Hu}]{thelen2022comprehensivepart1}
\bibinfo{author}{A.~Thelen}, \bibinfo{author}{X.~Zhang},
  \bibinfo{author}{O.~Fink}, \bibinfo{author}{Y.~Lu},
  \bibinfo{author}{S.~Ghosh}, \bibinfo{author}{B.~D. Youn},
  \bibinfo{author}{M.~D. Todd}, \bibinfo{author}{S.~Mahadevan},
  \bibinfo{author}{C.~Hu}, \bibinfo{author}{Z.~Hu}, \bibinfo{title}{A
  comprehensive review of digital twin—part 1: modeling and twinning enabling
  technologies}, \bibinfo{journal}{Structural and Multidisciplinary
  Optimization} \bibinfo{volume}{65}~(\bibinfo{number}{12})
  (\bibinfo{year}{2022}{\natexlab{a}}) \bibinfo{pages}{1--55},
  \bibinfo{note}{doi: \url{https://doi.org/10.1007/s00158-022-03425-4}}.

\bibitem[{Begoli et~al.(2019)Begoli, Bhattacharya, and
  Kusnezov}]{begoli2019need}
\bibinfo{author}{E.~Begoli}, \bibinfo{author}{T.~Bhattacharya},
  \bibinfo{author}{D.~Kusnezov}, \bibinfo{title}{The need for uncertainty
  quantification in machine-assisted medical decision making},
  \bibinfo{journal}{Nature Machine Intelligence}
  \bibinfo{volume}{1}~(\bibinfo{number}{1}) (\bibinfo{year}{2019})
  \bibinfo{pages}{20--23}, \bibinfo{note}{doi:
  \url{https://doi.org/10.1038/s42256-018-0004-1}}.

\bibitem[{Rudin(2019)}]{rudin2019stop}
\bibinfo{author}{C.~Rudin}, \bibinfo{title}{Stop explaining black box machine
  learning models for high stakes decisions and use interpretable models
  instead}, \bibinfo{journal}{Nature Machine Intelligence}
  \bibinfo{volume}{1}~(\bibinfo{number}{5}) (\bibinfo{year}{2019})
  \bibinfo{pages}{206--215}, \bibinfo{note}{doi:
  \url{https://doi.org/10.1038/s42256-019-0048-x}}.

\bibitem[{Sensoy et~al.(2018)Sensoy, Kaplan, and
  Kandemir}]{sensoy2018evidential}
\bibinfo{author}{M.~Sensoy}, \bibinfo{author}{L.~Kaplan},
  \bibinfo{author}{M.~Kandemir}, \bibinfo{title}{Evidential deep learning to
  quantify classification uncertainty}, \bibinfo{journal}{Advances in Neural
  Information Processing Systems} \bibinfo{volume}{31}, \bibinfo{note}{doi:
  \url{https://doi.org/10.48550/arXiv.1806.01768}}.

\bibitem[{Zhang et~al.(2022{\natexlab{a}})Zhang, Zhong, and
  Mahadevan}]{zhang2022airport}
\bibinfo{author}{X.~Zhang}, \bibinfo{author}{S.~Zhong},
  \bibinfo{author}{S.~Mahadevan}, \bibinfo{title}{Airport surface movement
  prediction and safety assessment with spatial--temporal graph convolutional
  neural network}, \bibinfo{journal}{Transportation Research Part C: Emerging
  Technologies} \bibinfo{volume}{144} (\bibinfo{year}{2022}{\natexlab{a}})
  \bibinfo{pages}{103873}, \bibinfo{note}{doi:
  \url{https://doi.org/10.1038/s42256-019-0048-x}}.

\bibitem[{H{\"u}llermeier and Waegeman(2021)}]{hullermeier2021aleatoric}
\bibinfo{author}{E.~H{\"u}llermeier}, \bibinfo{author}{W.~Waegeman},
  \bibinfo{title}{Aleatoric and epistemic uncertainty in machine learning: An
  introduction to concepts and methods}, \bibinfo{journal}{Machine Learning}
  \bibinfo{volume}{110}~(\bibinfo{number}{3}) (\bibinfo{year}{2021})
  \bibinfo{pages}{457--506}, \bibinfo{note}{doi:
  \url{https://doi.org/10.1007/s10994-021-05946-3}}.

\bibitem[{Szegedy et~al.(2013)Szegedy, Zaremba, Sutskever, Bruna, Erhan,
  Goodfellow, and Fergus}]{szegedy2013intriguing}
\bibinfo{author}{C.~Szegedy}, \bibinfo{author}{W.~Zaremba},
  \bibinfo{author}{I.~Sutskever}, \bibinfo{author}{J.~Bruna},
  \bibinfo{author}{D.~Erhan}, \bibinfo{author}{I.~Goodfellow},
  \bibinfo{author}{R.~Fergus}, \bibinfo{title}{Intriguing properties of neural
  networks} \bibinfo{note}{{doi}:
  \url{https://doi.org/10.48550/arXiv.1312.6199}}.

\bibitem[{Gal and Ghahramani(2016{\natexlab{a}})}]{gal2016dropout}
\bibinfo{author}{Y.~Gal}, \bibinfo{author}{Z.~Ghahramani},
  \bibinfo{title}{Dropout as a {B}ayesian approximation: Representing model
  uncertainty in deep learning}, in: \bibinfo{booktitle}{International
  Conference on Machine Learning}, \bibinfo{organization}{PMLR},
  \bibinfo{pages}{1050--1059}, \bibinfo{note}{doi:
  \url{https://doi.org/10.48550/arXiv.1506.02142}},
  \bibinfo{year}{2016}{\natexlab{a}}.

\bibitem[{Lakshminarayanan et~al.(2017)Lakshminarayanan, Pritzel, and
  Blundell}]{lakshminarayanan2017simple}
\bibinfo{author}{B.~Lakshminarayanan}, \bibinfo{author}{A.~Pritzel},
  \bibinfo{author}{C.~Blundell}, \bibinfo{title}{Simple and scalable predictive
  uncertainty estimation using deep ensembles}, \bibinfo{journal}{Advances in
  Neural Information Processing Systems} \bibinfo{volume}{30}.

\bibitem[{Kendall and Gal(2017)}]{kendall2017uncertainties}
\bibinfo{author}{A.~Kendall}, \bibinfo{author}{Y.~Gal}, \bibinfo{title}{What
  uncertainties do we need in {B}ayesian deep learning for computer vision?},
  \bibinfo{journal}{Advances in Neural Information Processing Systems}
  \bibinfo{volume}{30}, \bibinfo{note}{doi:
  \url{https://doi.org/10.48550/arXiv.1703.04977}}.

\bibitem[{Jin et~al.(2001)Jin, Chen, and Simpson}]{jin2001comparative}
\bibinfo{author}{R.~Jin}, \bibinfo{author}{W.~Chen}, \bibinfo{author}{T.~W.
  Simpson}, \bibinfo{title}{Comparative studies of metamodelling techniques
  under multiple modelling criteria}, \bibinfo{journal}{Structural and
  Multidisciplinary Optimization} \bibinfo{volume}{23}~(\bibinfo{number}{1})
  (\bibinfo{year}{2001}) \bibinfo{pages}{1--13}, \bibinfo{note}{doi:
  \url{https://doi.org/10.1007/s00158-001-0160-4}}.

\bibitem[{Queipo et~al.(2005)Queipo, Haftka, Shyy, Goel, Vaidyanathan, and
  Tucker}]{queipo2005surrogate}
\bibinfo{author}{N.~V. Queipo}, \bibinfo{author}{R.~T. Haftka},
  \bibinfo{author}{W.~Shyy}, \bibinfo{author}{T.~Goel},
  \bibinfo{author}{R.~Vaidyanathan}, \bibinfo{author}{P.~K. Tucker},
  \bibinfo{title}{Surrogate-based analysis and optimization},
  \bibinfo{journal}{Progress in Aerospace Sciences}
  \bibinfo{volume}{41}~(\bibinfo{number}{1}) (\bibinfo{year}{2005})
  \bibinfo{pages}{1--28}, \bibinfo{note}{doi:
  \url{https://doi.org/10.1016/j.paerosci.2005.02.001}}.

\bibitem[{Wang and Shan(2006)}]{wang2007review}
\bibinfo{author}{G.~G. Wang}, \bibinfo{author}{S.~Shan}, \bibinfo{title}{Review
  of metamodeling techniques in support of engineering design optimization},
  in: \bibinfo{booktitle}{International Design Engineering Technical
  Conferences and Computers and Information in Engineering Conference}, vol.
  \bibinfo{volume}{4255}, \bibinfo{pages}{415--426}, \bibinfo{note}{doi:
  \url{https://doi.org/10.1115/1.2429697}}, \bibinfo{year}{2006}.

\bibitem[{Jin et~al.(2002)Jin, Chen, and Sudjianto}]{jin2002sequential}
\bibinfo{author}{R.~Jin}, \bibinfo{author}{W.~Chen},
  \bibinfo{author}{A.~Sudjianto}, \bibinfo{title}{On sequential sampling for
  global metamodeling in engineering design}, in:
  \bibinfo{booktitle}{International Design Engineering Technical Conferences
  and Computers and Information in Engineering Conference}, vol.
  \bibinfo{volume}{36223}, \bibinfo{pages}{539--548}, \bibinfo{note}{doi:
  \url{https://doi.org/10.1115/DETC2002/DAC-34092}}, \bibinfo{year}{2002}.

\bibitem[{Bichon et~al.(2008)Bichon, Eldred, Swiler, Mahadevan, and
  McFarland}]{bichon2008efficient}
\bibinfo{author}{B.~J. Bichon}, \bibinfo{author}{M.~S. Eldred},
  \bibinfo{author}{L.~P. Swiler}, \bibinfo{author}{S.~Mahadevan},
  \bibinfo{author}{J.~M. McFarland}, \bibinfo{title}{Efficient global
  reliability analysis for nonlinear implicit performance functions},
  \bibinfo{journal}{AIAA Journal} \bibinfo{volume}{46}~(\bibinfo{number}{10})
  (\bibinfo{year}{2008}) \bibinfo{pages}{2459--2468}, \bibinfo{note}{doi:
  \url{http://dx.doi.org/10.2514/1.34321}}.

\bibitem[{Echard et~al.(2011)Echard, Gayton, and Lemaire}]{echard2011ak}
\bibinfo{author}{B.~Echard}, \bibinfo{author}{N.~Gayton},
  \bibinfo{author}{M.~Lemaire}, \bibinfo{title}{AK-MCS: an active learning
  reliability method combining Kriging and Monte Carlo simulation},
  \bibinfo{journal}{Structural Safety}
  \bibinfo{volume}{33}~(\bibinfo{number}{2}) (\bibinfo{year}{2011})
  \bibinfo{pages}{145--154}, \bibinfo{note}{doi:
  \url{https://doi.org/10.1016/j.strusafe.2011.01.002}}.

\bibitem[{Jones et~al.(1998)Jones, Schonlau, and Welch}]{jones1998efficient}
\bibinfo{author}{D.~R. Jones}, \bibinfo{author}{M.~Schonlau},
  \bibinfo{author}{W.~J. Welch}, \bibinfo{title}{Efficient global optimization
  of expensive black-box functions}, \bibinfo{journal}{Journal of Global
  Optimization} \bibinfo{volume}{13}~(\bibinfo{number}{4})
  (\bibinfo{year}{1998}) \bibinfo{pages}{455--492}, \bibinfo{note}{doi:
  \url{https://doi.org/10.1023/A:1008306431147}}.

\bibitem[{Shahriari et~al.(2016)Shahriari, Swersky, Wang, Adams, and
  de~Freitas}]{shahriari2016taking}
\bibinfo{author}{B.~Shahriari}, \bibinfo{author}{K.~Swersky},
  \bibinfo{author}{Z.~Wang}, \bibinfo{author}{R.~P. Adams},
  \bibinfo{author}{N.~de~Freitas}, \bibinfo{title}{Taking the human out of the
  loop: A review of {B}ayesian optimization}, \bibinfo{journal}{Proceedings of
  the IEEE} \bibinfo{volume}{104}~(\bibinfo{number}{1}) (\bibinfo{year}{2016})
  \bibinfo{pages}{148--175}, \bibinfo{note}{doi:
  \url{https://doi.org/10.1109/JPROC.2015.2494218}}.

\bibitem[{Lee and Chen(2009)}]{lee2009comparative}
\bibinfo{author}{S.~H. Lee}, \bibinfo{author}{W.~Chen}, \bibinfo{title}{A
  comparative study of uncertainty propagation methods for black-box-type
  problems}, \bibinfo{journal}{Structural and Multidisciplinary Optimization}
  \bibinfo{volume}{37}~(\bibinfo{number}{3}) (\bibinfo{year}{2009})
  \bibinfo{pages}{239--253}, \bibinfo{note}{doi:
  \url{https://doi.org/10.1007/s00158-008-0234-7}}.

\bibitem[{Chakraborty(2020)}]{chakraborty2020simulation}
\bibinfo{author}{S.~Chakraborty}, \bibinfo{title}{Simulation free reliability
  analysis: A physics-informed deep learning based approach}
  \bibinfo{note}{{doi}: \url{https://doi.org/10.48550/arXiv.2005.01302}}.

\bibitem[{Li and Wang(2020)}]{li2020deep}
\bibinfo{author}{M.~Li}, \bibinfo{author}{Z.~Wang}, \bibinfo{title}{Deep
  learning for high-dimensional reliability analysis},
  \bibinfo{journal}{Mechanical Systems and Signal Processing}
  \bibinfo{volume}{139} (\bibinfo{year}{2020}) \bibinfo{pages}{106399},
  \bibinfo{note}{doi: \url{https://doi.org/10.1016/j.ymssp.2019.106399}}.

\bibitem[{Zhang and Shafieezadeh(2022)}]{zhang2022simulation}
\bibinfo{author}{C.~Zhang}, \bibinfo{author}{A.~Shafieezadeh},
  \bibinfo{title}{Simulation-free reliability analysis with active learning and
  Physics-Informed Neural Network}, \bibinfo{journal}{Reliability Engineering
  \& System Safety} \bibinfo{volume}{226} (\bibinfo{year}{2022})
  \bibinfo{pages}{108716}, \bibinfo{note}{doi:
  \url{https://doi.org/10.1016/j.ress.2022.108716}}.

\bibitem[{Coble and Hines(2008)}]{coble2008prognostic}
\bibinfo{author}{J.~B. Coble}, \bibinfo{author}{J.~W. Hines},
  \bibinfo{title}{Prognostic algorithm categorization with PHM challenge
  application}, in: \bibinfo{booktitle}{2008 International Conference on
  Prognostics and Health Management}, \bibinfo{organization}{IEEE},
  \bibinfo{pages}{1--11}, \bibinfo{note}{doi:
  \url{https://doi.org/10.1109/PHM.2008.4711456}}, \bibinfo{year}{2008}.

\bibitem[{Tipping(2001)}]{tipping2001sparse}
\bibinfo{author}{M.~E. Tipping}, \bibinfo{title}{Sparse Bayesian learning and
  the relevance vector machine}, \bibinfo{journal}{Journal of Machine Learning
  Research} \bibinfo{volume}{1}~(\bibinfo{number}{Jun}) (\bibinfo{year}{2001})
  \bibinfo{pages}{211--244}, \bibinfo{note}{doi:
  \url{https://doi.org/10.1162/15324430152748236}}.

\bibitem[{Saha et~al.(2008)Saha, Goebel, Poll, and
  Christophersen}]{saha2008prognostics}
\bibinfo{author}{B.~Saha}, \bibinfo{author}{K.~Goebel},
  \bibinfo{author}{S.~Poll}, \bibinfo{author}{J.~Christophersen},
  \bibinfo{title}{Prognostics methods for battery health monitoring using a
  Bayesian framework}, \bibinfo{journal}{IEEE Transactions on Instrumentation
  and Measurement} \bibinfo{volume}{58}~(\bibinfo{number}{2})
  (\bibinfo{year}{2008}) \bibinfo{pages}{291--296}, \bibinfo{note}{doi:
  \url{https://doi.org/10.1109/TIM.2008.2005965}}.

\bibitem[{Wang et~al.(2013)Wang, Miao, and Pecht}]{wang2013prognostics}
\bibinfo{author}{D.~Wang}, \bibinfo{author}{Q.~Miao},
  \bibinfo{author}{M.~Pecht}, \bibinfo{title}{Prognostics of lithium-ion
  batteries based on relevance vectors and a conditional three-parameter
  capacity degradation model}, \bibinfo{journal}{Journal of Power Sources}
  \bibinfo{volume}{239} (\bibinfo{year}{2013}) \bibinfo{pages}{253--264},
  \bibinfo{note}{doi: \url{https://doi.org/10.1016/j.jpowsour.2013.03.129}}.

\bibitem[{Chang et~al.(2022)Chang, Zou, Fan, Peng, and Fang}]{CHANG2022109166}
\bibinfo{author}{Y.~Chang}, \bibinfo{author}{J.~Zou}, \bibinfo{author}{S.~Fan},
  \bibinfo{author}{C.~Peng}, \bibinfo{author}{H.~Fang},
  \bibinfo{title}{Remaining useful life prediction of degraded system with the
  capability of uncertainty management}, \bibinfo{journal}{Mechanical Systems
  and Signal Processing} \bibinfo{volume}{177} (\bibinfo{year}{2022})
  \bibinfo{pages}{109166}, \bibinfo{note}{doi:
  \url{https://doi.org/10.1016/j.ymssp.2022.109166}}.

\bibitem[{Wang et~al.(2012)Wang, Youn, and Hu}]{wang2012generic}
\bibinfo{author}{P.~Wang}, \bibinfo{author}{B.~D. Youn},
  \bibinfo{author}{C.~Hu}, \bibinfo{title}{A generic probabilistic framework
  for structural health prognostics and uncertainty management},
  \bibinfo{journal}{Mechanical Systems and Signal Processing}
  \bibinfo{volume}{28} (\bibinfo{year}{2012}) \bibinfo{pages}{622--637},
  \bibinfo{note}{doi: \url{https://doi.org/10.1016/j.ymssp.2011.10.019}}.

\bibitem[{Hu et~al.(2012)Hu, Youn, Wang, and Yoon}]{hu2012ensemble}
\bibinfo{author}{C.~Hu}, \bibinfo{author}{B.~D. Youn},
  \bibinfo{author}{P.~Wang}, \bibinfo{author}{J.~T. Yoon},
  \bibinfo{title}{Ensemble of data-driven prognostic algorithms for robust
  prediction of remaining useful life}, \bibinfo{journal}{Reliability
  Engineering \& System Safety} \bibinfo{volume}{103} (\bibinfo{year}{2012})
  \bibinfo{pages}{120--135}, \bibinfo{note}{doi:
  \url{https://doi.org/10.1016/j.ress.2012.03.008}}.

\bibitem[{Liu et~al.(2013)Liu, Pang, Zhou, Peng, and
  Pecht}]{liu2013prognostics}
\bibinfo{author}{D.~Liu}, \bibinfo{author}{J.~Pang}, \bibinfo{author}{J.~Zhou},
  \bibinfo{author}{Y.~Peng}, \bibinfo{author}{M.~Pecht},
  \bibinfo{title}{Prognostics for state of health estimation of lithium-ion
  batteries based on combination Gaussian process functional regression},
  \bibinfo{journal}{Microelectronics Reliability}
  \bibinfo{volume}{53}~(\bibinfo{number}{6}) (\bibinfo{year}{2013})
  \bibinfo{pages}{832--839}, \bibinfo{note}{{doi}:
  \url{https://doi.org/10.1016/j.microrel.2013.03.010}}.

\bibitem[{Richardson et~al.(2017)Richardson, Osborne, and
  Howey}]{richardson2017gaussian}
\bibinfo{author}{R.~R. Richardson}, \bibinfo{author}{M.~A. Osborne},
  \bibinfo{author}{D.~A. Howey}, \bibinfo{title}{Gaussian process regression
  for forecasting battery state of health}, \bibinfo{journal}{Journal of Power
  Sources} \bibinfo{volume}{357} (\bibinfo{year}{2017})
  \bibinfo{pages}{209--219}, \bibinfo{note}{doi:
  \url{https://doi.org/10.1016/j.jpowsour.2017.05.004}}.

\bibitem[{Thelen et~al.(2022{\natexlab{b}})Thelen, Li, Hu, Bekyarova, Kalinin,
  and Sanghadasa}]{thelen2022augmented}
\bibinfo{author}{A.~Thelen}, \bibinfo{author}{M.~Li}, \bibinfo{author}{C.~Hu},
  \bibinfo{author}{E.~Bekyarova}, \bibinfo{author}{S.~Kalinin},
  \bibinfo{author}{M.~Sanghadasa}, \bibinfo{title}{Augmented model-based
  framework for battery remaining useful life prediction},
  \bibinfo{journal}{Applied Energy} \bibinfo{volume}{324}
  (\bibinfo{year}{2022}{\natexlab{b}}) \bibinfo{pages}{119624},
  \bibinfo{note}{doi: \url{https://doi.org/10.1016/j.apenergy.2022.119624}}.

\bibitem[{Sankararaman(2015)}]{sankararaman2015significance}
\bibinfo{author}{S.~Sankararaman}, \bibinfo{title}{Significance,
  interpretation, and quantification of uncertainty in prognostics and
  remaining useful life prediction}, \bibinfo{journal}{Mechanical Systems and
  Signal Processing} \bibinfo{volume}{52} (\bibinfo{year}{2015})
  \bibinfo{pages}{228--247}, \bibinfo{note}{doi:
  \url{https://doi.org/10.1016/j.ymssp.2014.05.029}}.

\bibitem[{Sankararaman and Goebel(2015)}]{sankararaman2015uncertainty}
\bibinfo{author}{S.~Sankararaman}, \bibinfo{author}{K.~Goebel},
  \bibinfo{title}{Uncertainty in prognostics and systems health management},
  \bibinfo{journal}{International Journal of Prognostics and Health Management}
  \bibinfo{volume}{6}~(\bibinfo{number}{4}), \bibinfo{note}{doi:
  \url{https://doi.org/10.36001/ijphm.2015.v6i4.2319}}.

\bibitem[{Abdar et~al.(2021)Abdar, Pourpanah, Hussain, Rezazadegan, Liu,
  Ghavamzadeh, Fieguth, Cao, Khosravi, Acharya et~al.}]{abdar2021review}
\bibinfo{author}{M.~Abdar}, \bibinfo{author}{F.~Pourpanah},
  \bibinfo{author}{S.~Hussain}, \bibinfo{author}{D.~Rezazadegan},
  \bibinfo{author}{L.~Liu}, \bibinfo{author}{M.~Ghavamzadeh},
  \bibinfo{author}{P.~Fieguth}, \bibinfo{author}{X.~Cao},
  \bibinfo{author}{A.~Khosravi}, \bibinfo{author}{U.~R. Acharya}, et~al.,
  \bibinfo{title}{A review of uncertainty quantification in deep learning:
  Techniques, applications and challenges}, \bibinfo{journal}{Information
  Fusion} \bibinfo{volume}{76} (\bibinfo{year}{2021})
  \bibinfo{pages}{243--297}, \bibinfo{note}{doi:
  \url{https://doi.org/10.1016/j.inffus.2021.05.008}}.

\bibitem[{Bhatt et~al.(2021)Bhatt, Antor{\'a}n, Zhang, Liao, Sattigeri,
  Fogliato, Melan{\c{c}}on, Krishnan, Stanley, Tickoo
  et~al.}]{bhatt2021uncertainty}
\bibinfo{author}{U.~Bhatt}, \bibinfo{author}{J.~Antor{\'a}n},
  \bibinfo{author}{Y.~Zhang}, \bibinfo{author}{Q.~V. Liao},
  \bibinfo{author}{P.~Sattigeri}, \bibinfo{author}{R.~Fogliato},
  \bibinfo{author}{G.~Melan{\c{c}}on}, \bibinfo{author}{R.~Krishnan},
  \bibinfo{author}{J.~Stanley}, \bibinfo{author}{O.~Tickoo}, et~al.,
  \bibinfo{title}{Uncertainty as a form of transparency: Measuring,
  communicating, and using uncertainty}, in: \bibinfo{booktitle}{Proceedings of
  the 2021 AAAI/ACM Conference on AI, Ethics, and Society},
  \bibinfo{pages}{401--413}, \bibinfo{note}{doi:
  \url{https://doi.org/10.48550/arXiv.2011.07586}}, \bibinfo{year}{2021}.

\bibitem[{Gawlikowski et~al.(2021)Gawlikowski, Tassi, Ali, Lee, Humt, Feng,
  Kruspe, Triebel, Jung, Roscher et~al.}]{gawlikowski2021survey}
\bibinfo{author}{J.~Gawlikowski}, \bibinfo{author}{C.~R.~N. Tassi},
  \bibinfo{author}{M.~Ali}, \bibinfo{author}{J.~Lee},
  \bibinfo{author}{M.~Humt}, \bibinfo{author}{J.~Feng},
  \bibinfo{author}{A.~Kruspe}, \bibinfo{author}{R.~Triebel},
  \bibinfo{author}{P.~Jung}, \bibinfo{author}{R.~Roscher}, et~al.,
  \bibinfo{title}{A survey of uncertainty in deep neural networks}
  \bibinfo{note}{{doi}: \url{https://doi.org/10.48550/arXiv.2107.03342}}.

\bibitem[{Psaros et~al.(2023)Psaros, Meng, Zou, Guo, and
  Karniadakis}]{psaros2023uncertainty}
\bibinfo{author}{A.~F. Psaros}, \bibinfo{author}{X.~Meng},
  \bibinfo{author}{Z.~Zou}, \bibinfo{author}{L.~Guo}, \bibinfo{author}{G.~E.
  Karniadakis}, \bibinfo{title}{Uncertainty quantification in scientific
  machine learning: Methods, metrics, and comparisons},
  \bibinfo{journal}{Journal of Computational Physics} \bibinfo{volume}{477}
  (\bibinfo{year}{2023}) \bibinfo{pages}{111902}, \bibinfo{note}{doi:
  \url{https://doi.org/10.1016/j.jcp.2022.111902}}.

\bibitem[{Zhang et~al.(2019{\natexlab{a}})Zhang, Yang, Jiang, Nigam, Yamakawa,
  Furuhata, Shimada, and Kara}]{zhang20193d}
\bibinfo{author}{W.~Zhang}, \bibinfo{author}{Z.~Yang},
  \bibinfo{author}{H.~Jiang}, \bibinfo{author}{S.~Nigam},
  \bibinfo{author}{S.~Yamakawa}, \bibinfo{author}{T.~Furuhata},
  \bibinfo{author}{K.~Shimada}, \bibinfo{author}{L.~B. Kara},
  \bibinfo{title}{3D shape synthesis for conceptual design and optimization
  using variational autoencoders}, in: \bibinfo{booktitle}{International Design
  Engineering Technical Conferences and Computers and Information in
  Engineering Conference}, vol. \bibinfo{volume}{59186},
  \bibinfo{organization}{American Society of Mechanical Engineers},
  \bibinfo{pages}{V02AT03A017}, \bibinfo{note}{doi:
  \url{https://doi.org/10.1115/DETC2019-98525}},
  \bibinfo{year}{2019}{\natexlab{a}}.

\bibitem[{Chen and Fuge(2018)}]{chen2018b}
\bibinfo{author}{W.~Chen}, \bibinfo{author}{M.~Fuge},
  \bibinfo{title}{B\'ezierGAN: Automatic Generation of Smooth Curves from
  Interpretable Low-Dimensional Parameters} \bibinfo{note}{{doi}:
  \url{https://doi.org/10.48550/arXiv.1808.08871}}.

\bibitem[{Chen and Fuge(2019)}]{chen2019synthesizing}
\bibinfo{author}{W.~Chen}, \bibinfo{author}{M.~Fuge},
  \bibinfo{title}{Synthesizing designs with interpart dependencies using
  hierarchical generative adversarial networks}, \bibinfo{journal}{Journal of
  Mechanical Design} \bibinfo{volume}{141}~(\bibinfo{number}{11})
  (\bibinfo{year}{2019}) \bibinfo{pages}{111403}, \bibinfo{note}{doi:
  \url{https://doi.org/10.1115/1.4044076}}.

\bibitem[{He and He(2017)}]{he2017deep}
\bibinfo{author}{M.~He}, \bibinfo{author}{D.~He}, \bibinfo{title}{Deep learning
  based approach for bearing fault diagnosis}, \bibinfo{journal}{IEEE
  Transactions on Industry Applications}
  \bibinfo{volume}{53}~(\bibinfo{number}{3}) (\bibinfo{year}{2017})
  \bibinfo{pages}{3057--3065}, \bibinfo{note}{doi:
  \url{https://doi.org/10.1109/TIA.2017.2661250}}.

\bibitem[{Hoang and Kang(2019)}]{hoang2019survey}
\bibinfo{author}{D.-T. Hoang}, \bibinfo{author}{H.-J. Kang}, \bibinfo{title}{A
  survey on deep learning based bearing fault diagnosis},
  \bibinfo{journal}{Neurocomputing} \bibinfo{volume}{335}
  (\bibinfo{year}{2019}) \bibinfo{pages}{327--335}, \bibinfo{note}{doi:
  \url{https://doi.org/10.1016/j.neucom.2018.06.078}}.

\bibitem[{Lu et~al.(2023)Lu, Nemani, Barzegar, Allen, Hu, Laflamme, Sarkar, and
  Zimmerman}]{lu2023physics}
\bibinfo{author}{H.~Lu}, \bibinfo{author}{V.~P. Nemani},
  \bibinfo{author}{V.~Barzegar}, \bibinfo{author}{C.~Allen},
  \bibinfo{author}{C.~Hu}, \bibinfo{author}{S.~Laflamme},
  \bibinfo{author}{S.~Sarkar}, \bibinfo{author}{A.~T. Zimmerman},
  \bibinfo{title}{A physics-informed feature weighting method for bearing fault
  diagnostics}, \bibinfo{journal}{Mechanical Systems and Signal Processing}
  \bibinfo{volume}{191} (\bibinfo{year}{2023}) \bibinfo{pages}{110171},
  \bibinfo{note}{doi: \url{https://doi.org/10.1016/j.ymssp.2023.110171}}.

\bibitem[{Hou et~al.(2022)Hou, Wang, Chen, Wang, Peng, and
  Tsui}]{hou2022interpretable}
\bibinfo{author}{B.~Hou}, \bibinfo{author}{D.~Wang}, \bibinfo{author}{Y.~Chen},
  \bibinfo{author}{H.~Wang}, \bibinfo{author}{Z.~Peng}, \bibinfo{author}{K.-L.
  Tsui}, \bibinfo{title}{Interpretable online updated weights: Optimized square
  envelope spectrum for machine condition monitoring and fault diagnosis},
  \bibinfo{journal}{Mechanical Systems and Signal Processing}
  \bibinfo{volume}{169} (\bibinfo{year}{2022}) \bibinfo{pages}{108779},
  \bibinfo{note}{doi: \url{https://doi.org/10.1016/j.ymssp.2021.108779}}.

\bibitem[{Sinitsin et~al.(2022)Sinitsin, Ibryaeva, Sakovskaya, and
  Eremeeva}]{sinitsin2022intelligent}
\bibinfo{author}{V.~Sinitsin}, \bibinfo{author}{O.~Ibryaeva},
  \bibinfo{author}{V.~Sakovskaya}, \bibinfo{author}{V.~Eremeeva},
  \bibinfo{title}{Intelligent bearing fault diagnosis method combining mixed
  input and hybrid CNN-MLP model}, \bibinfo{journal}{Mechanical Systems and
  Signal Processing} \bibinfo{volume}{180} (\bibinfo{year}{2022})
  \bibinfo{pages}{109454}, \bibinfo{note}{doi:
  \url{https://doi.org/10.1016/j.ymssp.2022.109454}}.

\bibitem[{Deutsch and He(2017)}]{deutsch2017using}
\bibinfo{author}{J.~Deutsch}, \bibinfo{author}{D.~He}, \bibinfo{title}{Using
  deep learning-based approach to predict remaining useful life of rotating
  components}, \bibinfo{journal}{IEEE Transactions on Systems, Man, and
  Cybernetics: Systems} \bibinfo{volume}{48}~(\bibinfo{number}{1})
  (\bibinfo{year}{2017}) \bibinfo{pages}{11--20}, \bibinfo{note}{doi:
  \url{https://doi.org/10.1109/TSMC.2017.2697842}}.

\bibitem[{Yu et~al.(2019)Yu, Kim, and Mechefske}]{yu2019remaining}
\bibinfo{author}{W.~Yu}, \bibinfo{author}{I.~Y. Kim},
  \bibinfo{author}{C.~Mechefske}, \bibinfo{title}{Remaining useful life
  estimation using a bidirectional recurrent neural network based autoencoder
  scheme}, \bibinfo{journal}{Mechanical Systems and Signal Processing}
  \bibinfo{volume}{129} (\bibinfo{year}{2019}) \bibinfo{pages}{764--780},
  \bibinfo{note}{doi: \url{https://doi.org/10.1016/j.ymssp.2019.05.005}}.

\bibitem[{Li et~al.(2019)Li, Zhang, and Ding}]{li2019deep}
\bibinfo{author}{X.~Li}, \bibinfo{author}{W.~Zhang}, \bibinfo{author}{Q.~Ding},
  \bibinfo{title}{Deep learning-based remaining useful life estimation of
  bearings using multi-scale feature extraction}, \bibinfo{journal}{Reliability
  Engineering \& System Safety} \bibinfo{volume}{182} (\bibinfo{year}{2019})
  \bibinfo{pages}{208--218}, \bibinfo{note}{doi:
  \url{https://doi.org/10.1016/j.ress.2018.11.011}}.

\bibitem[{Der~Kiureghian and Ditlevsen(2009)}]{der2009aleatory}
\bibinfo{author}{A.~Der~Kiureghian}, \bibinfo{author}{O.~Ditlevsen},
  \bibinfo{title}{Aleatory or epistemic? Does it matter?},
  \bibinfo{journal}{Structural Safety}
  \bibinfo{volume}{31}~(\bibinfo{number}{2}) (\bibinfo{year}{2009})
  \bibinfo{pages}{105--112}, \bibinfo{note}{doi:
  \url{https://doi.org/10.1016/j.strusafe.2008.06.020}}.

\bibitem[{Gal et~al.(2017)Gal, Hron, and Kendall}]{gal2017concrete}
\bibinfo{author}{Y.~Gal}, \bibinfo{author}{J.~Hron},
  \bibinfo{author}{A.~Kendall}, \bibinfo{title}{Concrete dropout},
  \bibinfo{journal}{Advances in Neural Information Processing Systems}
  \bibinfo{volume}{30}.

\bibitem[{Sanjay and Sriram(2022)}]{fidelity2022}
\bibinfo{author}{R.~Sanjay}, \bibinfo{author}{R.~Sriram}, \bibinfo{title}{Data
  Fidelity and Latency: All things Clinical}, \bibinfo{journal}{Innovaccer}
  \bibinfo{volume}{1} (\bibinfo{year}{2022})
  \bibinfo{pages}{\url{https://innovaccer.com/resources/blogs/data--fidelity--and--latency--all--things--clinical}}.

\bibitem[{Saltelli et~al.(2004)Saltelli, Tarantola, Campolongo, Ratto
  et~al.}]{saltelli2004sensitivity}
\bibinfo{author}{A.~Saltelli}, \bibinfo{author}{S.~Tarantola},
  \bibinfo{author}{F.~Campolongo}, \bibinfo{author}{M.~Ratto}, et~al.,
  \bibinfo{title}{Sensitivity analysis in practice: a guide to assessing
  scientific models}, \bibinfo{journal}{Chichester, England}
  \bibinfo{note}{Doi: \url{https://doi.org/10.1111/j.1467-985X.2005.358_16.x}}.

\bibitem[{Sobol'(1990)}]{sobol1990sensitivity}
\bibinfo{author}{I.~M. Sobol'}, \bibinfo{title}{On sensitivity estimation for
  nonlinear mathematical models}, \bibinfo{journal}{Matematicheskoe
  Modelirovanie} \bibinfo{volume}{2}~(\bibinfo{number}{1})
  (\bibinfo{year}{1990}) \bibinfo{pages}{112--118}.

\bibitem[{Sobol(2001)}]{sobol2001global}
\bibinfo{author}{I.~M. Sobol}, \bibinfo{title}{Global sensitivity indices for
  nonlinear mathematical models and their Monte Carlo estimates},
  \bibinfo{journal}{Mathematics and Computers in Simulation}
  \bibinfo{volume}{55}~(\bibinfo{number}{1-3}) (\bibinfo{year}{2001})
  \bibinfo{pages}{271--280}, \bibinfo{note}{doi:
  \url{https://doi.org/10.1016/S0378-4754(00)00270-6}}.

\bibitem[{Gal(2016)}]{gal2016uncertainty}
\bibinfo{author}{Y.~Gal}, \bibinfo{title}{Uncertainty in deep learning}, Ph.D.
  thesis, \bibinfo{school}{PhD thesis, University of Cambridge},
  \bibinfo{year}{2016}.

\bibitem[{Depeweg et~al.(2018)Depeweg, Hernandez-Lobato, Doshi-Velez, and
  Udluft}]{depeweg2018decomposition}
\bibinfo{author}{S.~Depeweg}, \bibinfo{author}{J.-M. Hernandez-Lobato},
  \bibinfo{author}{F.~Doshi-Velez}, \bibinfo{author}{S.~Udluft},
  \bibinfo{title}{Decomposition of uncertainty in Bayesian deep learning for
  efficient and risk-sensitive learning}, in: \bibinfo{booktitle}{International
  Conference on Machine Learning}, \bibinfo{organization}{PMLR},
  \bibinfo{pages}{1184--1193}, \bibinfo{year}{2018}.

\bibitem[{Smith and Gal(2018)}]{smith2018understanding}
\bibinfo{author}{L.~Smith}, \bibinfo{author}{Y.~Gal},
  \bibinfo{title}{Understanding measures of uncertainty for adversarial example
  detection} \bibinfo{note}{{doi}:
  \url{https://doi.org/10.48550/arXiv.1803.08533}}.

\bibitem[{Malinin and Gales(2018)}]{malinin2018predictive}
\bibinfo{author}{A.~Malinin}, \bibinfo{author}{M.~Gales},
  \bibinfo{title}{Predictive uncertainty estimation via prior networks},
  \bibinfo{journal}{Advances in Neural Information Processing Systems}
  \bibinfo{volume}{31}, \bibinfo{note}{doi:
  \url{https://doi.org/10.48550/arXiv.1802.10501}}.

\bibitem[{Murphy(2022)}]{murphy2022probabilistic}
\bibinfo{author}{K.~P. Murphy}, \bibinfo{title}{Probabilistic machine learning:
  an introduction}, \bibinfo{publisher}{MIT press}, \bibinfo{year}{2022}.

\bibitem[{Saltelli et~al.(2010)Saltelli, Annoni, Azzini, Campolongo, Ratto, and
  Tarantola}]{saltelli2010variance}
\bibinfo{author}{A.~Saltelli}, \bibinfo{author}{P.~Annoni},
  \bibinfo{author}{I.~Azzini}, \bibinfo{author}{F.~Campolongo},
  \bibinfo{author}{M.~Ratto}, \bibinfo{author}{S.~Tarantola},
  \bibinfo{title}{Variance based sensitivity analysis of model output. Design
  and estimator for the total sensitivity index}, \bibinfo{journal}{Computer
  Physics Communications} \bibinfo{volume}{181}~(\bibinfo{number}{2})
  (\bibinfo{year}{2010}) \bibinfo{pages}{259--270}, \bibinfo{note}{doi:
  \url{https://doi.org/10.1016/j.cpc.2009.09.018}}.

\bibitem[{Shorten and Khoshgoftaar(2019)}]{shorten2019survey}
\bibinfo{author}{C.~Shorten}, \bibinfo{author}{T.~M. Khoshgoftaar},
  \bibinfo{title}{A survey on image data augmentation for deep learning},
  \bibinfo{journal}{Journal of Big Data}
  \bibinfo{volume}{6}~(\bibinfo{number}{1}) (\bibinfo{year}{2019})
  \bibinfo{pages}{1--48}, \bibinfo{note}{doi:
  \url{https://doi.org/10.1186/s40537-019-0197-0}}.

\bibitem[{Karniadakis et~al.(2021)Karniadakis, Kevrekidis, Lu, Perdikaris,
  Wang, and Yang}]{karniadakis2021physics}
\bibinfo{author}{G.~E. Karniadakis}, \bibinfo{author}{I.~G. Kevrekidis},
  \bibinfo{author}{L.~Lu}, \bibinfo{author}{P.~Perdikaris},
  \bibinfo{author}{S.~Wang}, \bibinfo{author}{L.~Yang},
  \bibinfo{title}{Physics-informed machine learning}, \bibinfo{journal}{Nature
  Reviews Physics} \bibinfo{volume}{3}~(\bibinfo{number}{6})
  (\bibinfo{year}{2021}) \bibinfo{pages}{422--440}.

\bibitem[{Thelen et~al.(2023{\natexlab{a}})Thelen, Zhang, Fink, Lu, Ghosh,
  Youn, Todd, Mahadevan, Hu, and Hu}]{thelen2022comprehensive}
\bibinfo{author}{A.~Thelen}, \bibinfo{author}{X.~Zhang},
  \bibinfo{author}{O.~Fink}, \bibinfo{author}{Y.~Lu},
  \bibinfo{author}{S.~Ghosh}, \bibinfo{author}{B.~D. Youn},
  \bibinfo{author}{M.~D. Todd}, \bibinfo{author}{S.~Mahadevan},
  \bibinfo{author}{C.~Hu}, \bibinfo{author}{Z.~Hu}, \bibinfo{title}{A
  Comprehensive Review of Digital Twin--Part 2: Roles of Uncertainty
  Quantification and Optimization, a Battery Digital Twin, and Perspectives},
  \bibinfo{journal}{Structural and Multidisciplinary Optimization}
  \bibinfo{volume}{66}~(\bibinfo{number}{1})
  (\bibinfo{year}{2023}{\natexlab{a}}) \bibinfo{pages}{1--43},
  \bibinfo{note}{doi: \url{https://doi.org/10.1007/s00158-022-03410-x}}.

\bibitem[{Xu et~al.(2022)Xu, Kohtz, Boakye, Gardoni, and Wang}]{xu2022physics}
\bibinfo{author}{Y.~Xu}, \bibinfo{author}{S.~Kohtz},
  \bibinfo{author}{J.~Boakye}, \bibinfo{author}{P.~Gardoni},
  \bibinfo{author}{P.~Wang}, \bibinfo{title}{Physics-informed machine learning
  for reliability and systems safety applications: State of the art and
  challenges}, \bibinfo{journal}{Reliability Engineering \& System Safety}
  \bibinfo{volume}{230} (\bibinfo{year}{2022}) \bibinfo{pages}{108900},
  \bibinfo{note}{doi: \url{https://doi.org/10.1016/j.ress.2022.108900}}.

\bibitem[{Hu et~al.(2023)Hu, Goebel, Howey, Peng, Wang, Wang, and
  Youn}]{hu2023special}
\bibinfo{author}{C.~Hu}, \bibinfo{author}{K.~Goebel},
  \bibinfo{author}{D.~Howey}, \bibinfo{author}{Z.~Peng},
  \bibinfo{author}{D.~Wang}, \bibinfo{author}{P.~Wang}, \bibinfo{author}{B.~D.
  Youn}, \bibinfo{title}{Special issue on Physics-informed machine learning
  enabling fault feature extraction and robust failure prognosis},
  \bibinfo{journal}{Mechanical Systems and Signal Processing}
  \bibinfo{volume}{192} (\bibinfo{year}{2023}) \bibinfo{pages}{110219},
  \bibinfo{note}{doi: \url{https://doi.org/10.1016/j.ymssp.2023.110219}}.

\bibitem[{Wang and Coit(4 18)}]{pinn_si}
\bibinfo{author}{P.~Wang}, \bibinfo{author}{D.~Coit},
  \bibinfo{title}{Physics-Informed Machine Learning for Reliability and
  Safety},
  \urlprefix\url{https://www.sciencedirect.com/journal/reliability-engineering-and-system-safety/special-issue/1084PD0CV5B},
  \bibinfo{year}{2023 ({A}ccessed on 2023-04-18)}.

\bibitem[{Malashkhia et~al.(2023)Malashkhia, Liu, Lu, and
  Wang}]{malashkhia2022physics}
\bibinfo{author}{L.~Malashkhia}, \bibinfo{author}{D.~Liu},
  \bibinfo{author}{Y.~Lu}, \bibinfo{author}{Y.~Wang},
  \bibinfo{title}{Physics-Constrained Bayesian Neural Network for Bias and
  Variance Reduction}, \bibinfo{journal}{Journal of Computing and Information
  Science in Engineering} \bibinfo{volume}{23}~(\bibinfo{number}{1})
  (\bibinfo{year}{2023}) \bibinfo{pages}{011012}, \bibinfo{note}{doi:
  \url{https://doi.org/10.1115/1.4055924}}.

\bibitem[{Deng(2020)}]{deng2020multifidelity}
\bibinfo{author}{Y.~Deng}, \bibinfo{title}{Multifidelity Data Fusion via
  Gradient-Enhanced Gaussian Process Regression},
  \bibinfo{journal}{Communications in Computational Physics}
  \bibinfo{volume}{28}~(\bibinfo{number}{5}) (\bibinfo{year}{2020})
  \bibinfo{pages}{1812--1837}, \bibinfo{note}{doi:
  \url{https://doi.org/10.4208/cicp.OA-2020-0151}}.

\bibitem[{Plumlee and Joseph(2018)}]{plumlee2018orthogonal}
\bibinfo{author}{M.~Plumlee}, \bibinfo{author}{V.~R. Joseph},
  \bibinfo{title}{Orthogonal Gaussian process models},
  \bibinfo{journal}{Statistica Sinica}  (\bibinfo{year}{2018})
  \bibinfo{pages}{601--619}\bibinfo{note}{Doi:
  \url{https://doi.org/10.5705/ss.202015.0404}}.

\bibitem[{Tran et~al.(2023)Tran, Maupin, and Rodgers}]{tran2023monotonic}
\bibinfo{author}{A.~Tran}, \bibinfo{author}{K.~Maupin},
  \bibinfo{author}{T.~Rodgers}, \bibinfo{title}{Monotonic Gaussian process for
  physics-constrained machine learning with materials science applications},
  \bibinfo{journal}{Journal of Computing and Information Science in
  Engineering} \bibinfo{volume}{23}~(\bibinfo{number}{1})
  (\bibinfo{year}{2023}) \bibinfo{pages}{011011}, \bibinfo{note}{doi:
  \url{https://doi.org/10.1115/1.4055852}}.

\bibitem[{Raghu et~al.(2019)Raghu, Zhang, Kleinberg, and
  Bengio}]{raghu2019transfusion}
\bibinfo{author}{M.~Raghu}, \bibinfo{author}{C.~Zhang},
  \bibinfo{author}{J.~Kleinberg}, \bibinfo{author}{S.~Bengio},
  \bibinfo{title}{Transfusion: Understanding transfer learning for medical
  imaging}, \bibinfo{journal}{Advances in Neural Information Processing
  Systems} \bibinfo{volume}{32}.

\bibitem[{Bottou(2012)}]{bottou2012stochastic}
\bibinfo{author}{L.~Bottou}, \bibinfo{title}{Stochastic gradient descent
  tricks}, in: \bibinfo{booktitle}{Neural networks: Tricks of the trade},
  \bibinfo{publisher}{Springer}, \bibinfo{pages}{421--436}, \bibinfo{note}{doi:
  \url{https://doi.org/10.1007/978-3-642-35289-8}}, \bibinfo{year}{2012}.

\bibitem[{Liu and Wang(2021)}]{liu2021dual}
\bibinfo{author}{D.~Liu}, \bibinfo{author}{Y.~Wang}, \bibinfo{title}{A
  Dual-Dimer method for training physics-constrained neural networks with
  minimax architecture}, \bibinfo{journal}{Neural Networks}
  \bibinfo{volume}{136} (\bibinfo{year}{2021}) \bibinfo{pages}{112--125},
  \bibinfo{note}{doi: \url{https://doi.org/10.1016/j.neunet.2020.12.028}}.

\bibitem[{Cai et~al.(2018)Cai, Luo, Wang, and Yang}]{cai2018feature}
\bibinfo{author}{J.~Cai}, \bibinfo{author}{J.~Luo}, \bibinfo{author}{S.~Wang},
  \bibinfo{author}{S.~Yang}, \bibinfo{title}{Feature selection in machine
  learning: A new perspective}, \bibinfo{journal}{Neurocomputing}
  \bibinfo{volume}{300} (\bibinfo{year}{2018}) \bibinfo{pages}{70--79},
  \bibinfo{note}{doi: \url{https://doi.org/10.1016/j.neucom.2017.11.077}}.

\bibitem[{Chandrashekar and Sahin(2014)}]{chandrashekar2014survey}
\bibinfo{author}{G.~Chandrashekar}, \bibinfo{author}{F.~Sahin},
  \bibinfo{title}{A survey on feature selection methods},
  \bibinfo{journal}{Computers \& Electrical Engineering}
  \bibinfo{volume}{40}~(\bibinfo{number}{1}) (\bibinfo{year}{2014})
  \bibinfo{pages}{16--28}, \bibinfo{note}{doi:
  \url{https://doi.org/10.1016/j.compeleceng.2013.11.024}}.

\bibitem[{Box(1979)}]{box1979all}
\bibinfo{author}{G.~Box}, \bibinfo{title}{All models are wrong, but some are
  useful}, \bibinfo{journal}{Robustness in Statistics}
  \bibinfo{volume}{202}~(\bibinfo{number}{1979}) (\bibinfo{year}{1979})
  \bibinfo{pages}{549}, \bibinfo{note}{doi:
  \url{https://doi.org/10.1007/s10815-020-01895-3}}.

\bibitem[{Tran et~al.(2020)Tran, Tranchida, Wildey, and
  Thompson}]{tran2020multi}
\bibinfo{author}{A.~Tran}, \bibinfo{author}{J.~Tranchida},
  \bibinfo{author}{T.~Wildey}, \bibinfo{author}{A.~P. Thompson},
  \bibinfo{title}{Multi-fidelity machine-learning with uncertainty
  quantification and Bayesian optimization for materials design: Application to
  ternary random alloys}, \bibinfo{journal}{The Journal of Chemical Physics}
  \bibinfo{volume}{153}~(\bibinfo{number}{7}) (\bibinfo{year}{2020})
  \bibinfo{pages}{074705}, \bibinfo{note}{doi:
  \url{https://doi.org/10.1063/5.0015672}}.

\bibitem[{Pilania et~al.(2017)Pilania, Gubernatis, and
  Lookman}]{pilania2017multi}
\bibinfo{author}{G.~Pilania}, \bibinfo{author}{J.~E. Gubernatis},
  \bibinfo{author}{T.~Lookman}, \bibinfo{title}{Multi-fidelity machine learning
  models for accurate bandgap predictions of solids},
  \bibinfo{journal}{Computational Materials Science} \bibinfo{volume}{129}
  (\bibinfo{year}{2017}) \bibinfo{pages}{156--163}, \bibinfo{note}{doi:
  \url{https://doi.org/10.1016/j.commatsci.2016.12.004}}.

\bibitem[{Liu and Wang(2019)}]{liu2019multi}
\bibinfo{author}{D.~Liu}, \bibinfo{author}{Y.~Wang},
  \bibinfo{title}{Multi-fidelity physics-constrained neural network and its
  application in materials modeling}, \bibinfo{journal}{Journal of Mechanical
  Design} \bibinfo{volume}{141}~(\bibinfo{number}{12}) (\bibinfo{year}{2019})
  \bibinfo{pages}{121403}, \bibinfo{note}{{doi}:
  \url{https://doi.org/10.1115/1.4044400}}.

\bibitem[{Liu et~al.(2023)Liu, Pusarla, and Wang}]{liu2023multi}
\bibinfo{author}{D.~Liu}, \bibinfo{author}{P.~Pusarla},
  \bibinfo{author}{Y.~Wang}, \bibinfo{title}{Multi-Fidelity Physics-Constrained
  Neural Networks with Minimax Architecture}, \bibinfo{journal}{Journal of
  Computing and Information Science in Engineering}
  \bibinfo{volume}{23}~(\bibinfo{number}{3}) (\bibinfo{year}{2023})
  \bibinfo{pages}{031008}, \bibinfo{note}{{doi}:
  \url{https://doi.org/10.1115/1.4055316}}.

\bibitem[{Huang et~al.(2022)Huang, Xie, Wang, Chen, Zhou, and
  Hu}]{huang2022transfer}
\bibinfo{author}{X.~Huang}, \bibinfo{author}{T.~Xie},
  \bibinfo{author}{Z.~Wang}, \bibinfo{author}{L.~Chen},
  \bibinfo{author}{Q.~Zhou}, \bibinfo{author}{Z.~Hu}, \bibinfo{title}{A
  transfer learning-based multi-fidelity point-cloud neural network approach
  for melt pool modeling in additive manufacturing},
  \bibinfo{journal}{ASCE-ASME Journal of Risk and Uncertainty in Engineering
  Systems, Part B: Mechanical Engineering}
  \bibinfo{volume}{8}~(\bibinfo{number}{1}), \bibinfo{note}{doi:
  \url{https://doi.org/10.1115/1.4051749}}.

\bibitem[{LeCun et~al.(2015)LeCun, Bengio, and Hinton}]{lecun2015deep}
\bibinfo{author}{Y.~LeCun}, \bibinfo{author}{Y.~Bengio},
  \bibinfo{author}{G.~Hinton}, \bibinfo{title}{Deep learning},
  \bibinfo{journal}{Nature} \bibinfo{volume}{521}~(\bibinfo{number}{7553})
  (\bibinfo{year}{2015}) \bibinfo{pages}{436--444}, \bibinfo{note}{doi:
  \url{https://doi.org/10.1038/nature14539}}.

\bibitem[{Zhang et~al.(2022{\natexlab{b}})Zhang, Chan, Yan, and
  Bose}]{zhang2022towards}
\bibinfo{author}{X.~Zhang}, \bibinfo{author}{F.~T. Chan},
  \bibinfo{author}{C.~Yan}, \bibinfo{author}{I.~Bose}, \bibinfo{title}{Towards
  risk-aware artificial intelligence and machine learning systems: An
  overview}, \bibinfo{journal}{Decision Support Systems} \bibinfo{volume}{159}
  (\bibinfo{year}{2022}{\natexlab{b}}) \bibinfo{pages}{113800},
  \bibinfo{note}{{doi}: \url{https://doi.org/10.1016/j.dss.2022.113800}}.

\bibitem[{He et~al.(2016)He, Zhang, Ren, and Sun}]{he2016deep}
\bibinfo{author}{K.~He}, \bibinfo{author}{X.~Zhang}, \bibinfo{author}{S.~Ren},
  \bibinfo{author}{J.~Sun}, \bibinfo{title}{Deep residual learning for image
  recognition}, in: \bibinfo{booktitle}{Proceedings of the IEEE Conference on
  Computer Vision and Pattern Recognition}, \bibinfo{pages}{770--778},
  \bibinfo{note}{doi: \url{https://doi.org/10.1109/CVPR.2016.90}},
  \bibinfo{year}{2016}.

\bibitem[{Zhang and Mahadevan(2020)}]{zhang2020bayesian_DSS}
\bibinfo{author}{X.~Zhang}, \bibinfo{author}{S.~Mahadevan},
  \bibinfo{title}{Bayesian neural networks for flight trajectory prediction and
  safety assessment}, \bibinfo{journal}{Decision Support Systems}
  \bibinfo{volume}{131} (\bibinfo{year}{2020}) \bibinfo{pages}{113246},
  \bibinfo{note}{doi: \url{https://doi.org/10.1016/j.dss.2020.113246}}.

\bibitem[{Zhang et~al.(2022{\natexlab{c}})Zhang, Chan, and
  Mahadevan}]{zhang2022explainable}
\bibinfo{author}{X.~Zhang}, \bibinfo{author}{F.~T. Chan},
  \bibinfo{author}{S.~Mahadevan}, \bibinfo{title}{Explainable machine learning
  in image classification models: An uncertainty quantification perspective},
  \bibinfo{journal}{Knowledge-Based Systems} \bibinfo{volume}{243}
  (\bibinfo{year}{2022}{\natexlab{c}}) \bibinfo{pages}{108418},
  \bibinfo{note}{doi: \url{https://doi.org/10.1016/j.knosys.2022.108418}}.

\bibitem[{Cheng et~al.(2020{\natexlab{a}})Cheng, Yang, Brear, and
  Frenklach}]{cheng2020quantifying}
\bibinfo{author}{S.~Cheng}, \bibinfo{author}{Y.~Yang}, \bibinfo{author}{M.~J.
  Brear}, \bibinfo{author}{M.~Frenklach}, \bibinfo{title}{Quantifying
  uncertainty in kinetic simulation of engine autoignition},
  \bibinfo{journal}{Combustion and Flame} \bibinfo{volume}{216}
  (\bibinfo{year}{2020}{\natexlab{a}}) \bibinfo{pages}{174--184},
  \bibinfo{note}{doi:
  \url{https://doi.org/10.1016/j.combustflame.2020.02.025}}.

\bibitem[{M{\aa}rtensson et~al.(2020)M{\aa}rtensson, Ferreira, Granberg,
  Cavallin, Oppedal, Padovani, Rektorova, Bonanni, Pardini, Kramberger
  et~al.}]{maartensson2020reliability}
\bibinfo{author}{G.~M{\aa}rtensson}, \bibinfo{author}{D.~Ferreira},
  \bibinfo{author}{T.~Granberg}, \bibinfo{author}{L.~Cavallin},
  \bibinfo{author}{K.~Oppedal}, \bibinfo{author}{A.~Padovani},
  \bibinfo{author}{I.~Rektorova}, \bibinfo{author}{L.~Bonanni},
  \bibinfo{author}{M.~Pardini}, \bibinfo{author}{M.~G. Kramberger}, et~al.,
  \bibinfo{title}{The reliability of a deep learning model in clinical
  out-of-distribution MRI data: a multicohort study}, \bibinfo{journal}{Medical
  Image Analysis} \bibinfo{volume}{66} (\bibinfo{year}{2020})
  \bibinfo{pages}{101714}, \bibinfo{note}{doi:
  \url{https://doi.org/10.1016/j.media.2020.101714}}.

\bibitem[{Tagasovska and Lopez-Paz(2019)}]{tagasovska2019single}
\bibinfo{author}{N.~Tagasovska}, \bibinfo{author}{D.~Lopez-Paz},
  \bibinfo{title}{Single-model uncertainties for deep learning},
  \bibinfo{journal}{Advances in Neural Information Processing Systems}
  \bibinfo{volume}{32}.

\bibitem[{Osawa et~al.(2019)Osawa, Swaroop, Khan, Jain, Eschenhagen, Turner,
  and Yokota}]{osawa2019practical}
\bibinfo{author}{K.~Osawa}, \bibinfo{author}{S.~Swaroop},
  \bibinfo{author}{M.~E.~E. Khan}, \bibinfo{author}{A.~Jain},
  \bibinfo{author}{R.~Eschenhagen}, \bibinfo{author}{R.~E. Turner},
  \bibinfo{author}{R.~Yokota}, \bibinfo{title}{Practical deep learning with
  Bayesian principles}, \bibinfo{journal}{Advances in Neural Information
  Processing Systems} \bibinfo{volume}{32}.

\bibitem[{Rasmussen(2006)}]{rasmussen2006gaussian}
\bibinfo{author}{C.~E. Rasmussen}, \bibinfo{title}{Gaussian processes in
  machine learning}, \bibinfo{publisher}{{MIT Press}}, \bibinfo{note}{doi:
  \url{https://doi.org/10.1007/978-3-540-28650-9_4}}, \bibinfo{year}{2006}.

\bibitem[{Brochu et~al.(2010)Brochu, Cora, and de~Freitas}]{brochu2010tutorial}
\bibinfo{author}{E.~Brochu}, \bibinfo{author}{V.~M. Cora},
  \bibinfo{author}{N.~de~Freitas}, \bibinfo{title}{A tutorial on {B}ayesian
  optimization of expensive cost functions, with application to active user
  modeling and hierarchical reinforcement learning}, \bibinfo{journal}{arXiv
  preprint arXiv:1012.2599} \bibinfo{note}{Doi:
  \url{https://doi.org/10.48550/arXiv.1012.2599}}.

\bibitem[{Neal(2012)}]{neal2012bayesian}
\bibinfo{author}{R.~M. Neal}, \bibinfo{title}{Bayesian learning for neural
  networks}, vol. \bibinfo{volume}{118}, \bibinfo{publisher}{Springer Science
  \& Business Media}, \bibinfo{year}{2012}.

\bibitem[{Furrer et~al.(2006)Furrer, Genton, and Nychka}]{furrer2006covariance}
\bibinfo{author}{R.~Furrer}, \bibinfo{author}{M.~G. Genton},
  \bibinfo{author}{D.~Nychka}, \bibinfo{title}{Covariance tapering for
  interpolation of large spatial datasets}, \bibinfo{journal}{Journal of
  Computational and Graphical Statistics}
  \bibinfo{volume}{15}~(\bibinfo{number}{3}) (\bibinfo{year}{2006})
  \bibinfo{pages}{502--523}, \bibinfo{note}{doi:
  \url{https://doi.org/10.1198/106186006X132178}}.

\bibitem[{Kaufman et~al.(2008)Kaufman, Schervish, and
  Nychka}]{kaufman2008covariance}
\bibinfo{author}{C.~G. Kaufman}, \bibinfo{author}{M.~J. Schervish},
  \bibinfo{author}{D.~W. Nychka}, \bibinfo{title}{Covariance tapering for
  likelihood-based estimation in large spatial data sets},
  \bibinfo{journal}{Journal of the American Statistical Association}
  \bibinfo{volume}{103}~(\bibinfo{number}{484}) (\bibinfo{year}{2008})
  \bibinfo{pages}{1545--1555}, \bibinfo{note}{doi:
  \url{https://doi.org/10.1198/016214508000000959}}.

\bibitem[{Cressie and Johannesson(2008)}]{cressie2008fixed}
\bibinfo{author}{N.~Cressie}, \bibinfo{author}{G.~Johannesson},
  \bibinfo{title}{Fixed rank kriging for very large spatial data sets},
  \bibinfo{journal}{Journal of the Royal Statistical Society: Series B
  (Statistical Methodology)} \bibinfo{volume}{70}~(\bibinfo{number}{1})
  (\bibinfo{year}{2008}) \bibinfo{pages}{209--226}, \bibinfo{note}{doi:
  \url{https://doi.org/10.1111/j.1467-9868.2007.00633.x}}.

\bibitem[{Banerjee et~al.(2008)Banerjee, Gelfand, Finley, and
  Sang}]{banerjee2008gaussian}
\bibinfo{author}{S.~Banerjee}, \bibinfo{author}{A.~E. Gelfand},
  \bibinfo{author}{A.~O. Finley}, \bibinfo{author}{H.~Sang},
  \bibinfo{title}{Gaussian predictive process models for large spatial data
  sets}, \bibinfo{journal}{Journal of the Royal Statistical Society: Series B
  (Statistical Methodology)} \bibinfo{volume}{70}~(\bibinfo{number}{4})
  (\bibinfo{year}{2008}) \bibinfo{pages}{825--848}, \bibinfo{note}{doi:
  \url{https://doi.org/10.1111/j.1467-9868.2008.00663.x}}.

\bibitem[{Neal(1997)}]{neal1997monte}
\bibinfo{author}{R.~M. Neal}, \bibinfo{title}{Monte Carlo implementation of
  Gaussian process models for Bayesian regression and classification},
  \bibinfo{journal}{arXiv preprint physics/9701026} .

\bibitem[{Andrianakis and Challenor(2012)}]{andrianakis2012effect}
\bibinfo{author}{I.~Andrianakis}, \bibinfo{author}{P.~G. Challenor},
  \bibinfo{title}{The effect of the nugget on Gaussian process emulators of
  computer models}, \bibinfo{journal}{Computational Statistics \& Data
  Analysis} \bibinfo{volume}{56}~(\bibinfo{number}{12}) (\bibinfo{year}{2012})
  \bibinfo{pages}{4215--4228}, \bibinfo{note}{doi:
  \url{https://doi.org/10.1016/j.csda.2012.04.020}}.

\bibitem[{Le~Gratiet et~al.(2014)Le~Gratiet, Cannamela, and
  Iooss}]{le2014bayesian}
\bibinfo{author}{L.~Le~Gratiet}, \bibinfo{author}{C.~Cannamela},
  \bibinfo{author}{B.~Iooss}, \bibinfo{title}{A Bayesian approach for global
  sensitivity analysis of (multifidelity) computer codes},
  \bibinfo{journal}{SIAM/ASA Journal on Uncertainty Quantification}
  \bibinfo{volume}{2}~(\bibinfo{number}{1}) (\bibinfo{year}{2014})
  \bibinfo{pages}{336--363}, \bibinfo{note}{doi:
  \url{https://doi.org/10.1137/130926869}}.

\bibitem[{Menz et~al.(2021)Menz, Dubreuil, Morio, Gogu, Bartoli, and
  Chiron}]{menz2021variance}
\bibinfo{author}{M.~Menz}, \bibinfo{author}{S.~Dubreuil},
  \bibinfo{author}{J.~Morio}, \bibinfo{author}{C.~Gogu},
  \bibinfo{author}{N.~Bartoli}, \bibinfo{author}{M.~Chiron},
  \bibinfo{title}{Variance based sensitivity analysis for Monte Carlo and
  importance sampling reliability assessment with Gaussian processes},
  \bibinfo{journal}{Structural Safety} \bibinfo{volume}{93}
  (\bibinfo{year}{2021}) \bibinfo{pages}{102116}, \bibinfo{note}{doi:
  \url{https://doi.org/10.1016/j.strusafe.2021.102116}}.

\bibitem[{Wei et~al.(2023)Wei, Zheng, Fu, Xu, and Gao}]{wei2023expected}
\bibinfo{author}{P.~Wei}, \bibinfo{author}{Y.~Zheng}, \bibinfo{author}{J.~Fu},
  \bibinfo{author}{Y.~Xu}, \bibinfo{author}{W.~Gao}, \bibinfo{title}{An
  expected integrated error reduction function for accelerating Bayesian active
  learning of failure probability}, \bibinfo{journal}{Reliability Engineering
  \& System Safety} \bibinfo{volume}{231} (\bibinfo{year}{2023})
  \bibinfo{pages}{108971}, \bibinfo{note}{doi:
  \url{https://doi.org/10.1016/j.ress.2022.108971}}.

\bibitem[{Le et~al.(2005)Le, Smola, and Canu}]{le2005heteroscedastic}
\bibinfo{author}{Q.~V. Le}, \bibinfo{author}{A.~J. Smola},
  \bibinfo{author}{S.~Canu}, \bibinfo{title}{Heteroscedastic {G}aussian process
  regression}, in: \bibinfo{booktitle}{Proceedings of the 22nd International
  Conference on Machine learning}, \bibinfo{pages}{489--496},
  \bibinfo{year}{2005}.

\bibitem[{Stein(1999)}]{stein1999interpolation}
\bibinfo{author}{M.~L. Stein}, \bibinfo{title}{Interpolation of spatial data:
  some theory for kriging}, \bibinfo{publisher}{Springer Science \& Business
  Media}, \bibinfo{year}{1999}.

\bibitem[{Liu et~al.(2020{\natexlab{a}})Liu, Ong, Shen, and
  Cai}]{liu2020gaussian}
\bibinfo{author}{H.~Liu}, \bibinfo{author}{Y.-S. Ong},
  \bibinfo{author}{X.~Shen}, \bibinfo{author}{J.~Cai}, \bibinfo{title}{When
  Gaussian process meets big data: A review of scalable GPs},
  \bibinfo{journal}{IEEE Transactions on Neural Networks and Learning Systems}
  \bibinfo{volume}{31}~(\bibinfo{number}{11})
  (\bibinfo{year}{2020}{\natexlab{a}}) \bibinfo{pages}{4405--4423},
  \bibinfo{note}{doi: \url{https://doi.org/10.1109/TNNLS.2019.2957109}}.

\bibitem[{Wang et~al.(2016)Wang, Hutter, Zoghi, Matheson, and
  De~Feitas}]{wang2016bayesian}
\bibinfo{author}{Z.~Wang}, \bibinfo{author}{F.~Hutter},
  \bibinfo{author}{M.~Zoghi}, \bibinfo{author}{D.~Matheson},
  \bibinfo{author}{N.~De~Feitas}, \bibinfo{title}{Bayesian optimization in a
  billion dimensions via random embeddings}, \bibinfo{journal}{Journal of
  Artificial Intelligence Research} \bibinfo{volume}{55} (\bibinfo{year}{2016})
  \bibinfo{pages}{361--387}, \bibinfo{note}{doi:
  \url{https://doi.org/10.1613/jair.4806}}.

\bibitem[{Tripathy et~al.(2016)Tripathy, Bilionis, and
  Gonzalez}]{tripathy2016gaussian}
\bibinfo{author}{R.~Tripathy}, \bibinfo{author}{I.~Bilionis},
  \bibinfo{author}{M.~Gonzalez}, \bibinfo{title}{Gaussian processes with
  built-in dimensionality reduction: Applications to high-dimensional
  uncertainty propagation}, \bibinfo{journal}{Journal of Computational Physics}
  \bibinfo{volume}{321} (\bibinfo{year}{2016}) \bibinfo{pages}{191--223},
  \bibinfo{note}{doi: \url{https://doi.org/10.1016/j.jcp.2016.05.039}}.

\bibitem[{Bouhlel et~al.(2016)Bouhlel, Bartoli, Otsmane, and
  Morlier}]{bouhlel2016improving}
\bibinfo{author}{M.~A. Bouhlel}, \bibinfo{author}{N.~Bartoli},
  \bibinfo{author}{A.~Otsmane}, \bibinfo{author}{J.~Morlier},
  \bibinfo{title}{Improving kriging surrogates of high-dimensional design
  models by Partial Least Squares dimension reduction},
  \bibinfo{journal}{Structural and Multidisciplinary Optimization}
  \bibinfo{volume}{53} (\bibinfo{year}{2016}) \bibinfo{pages}{935--952},
  \bibinfo{note}{doi: \url{https://doi.org/10.1007/s00158-015-1395-9}}.

\bibitem[{Durrande et~al.(2012)Durrande, Ginsbourger, and
  Roustant}]{durrande2012additive}
\bibinfo{author}{N.~Durrande}, \bibinfo{author}{D.~Ginsbourger},
  \bibinfo{author}{O.~Roustant}, \bibinfo{title}{Additive covariance kernels
  for high-dimensional Gaussian process modeling}, in:
  \bibinfo{booktitle}{Annales de la Facult{\'e} des sciences de Toulouse:
  Math{\'e}matiques}, vol.~\bibinfo{volume}{21}, \bibinfo{pages}{481--499},
  \bibinfo{year}{2012}.

\bibitem[{Binois and Wycoff(2022)}]{binois2022survey}
\bibinfo{author}{M.~Binois}, \bibinfo{author}{N.~Wycoff}, \bibinfo{title}{A
  survey on high-dimensional Gaussian process modeling with application to
  Bayesian optimization}, \bibinfo{journal}{ACM Transactions on Evolutionary
  Learning and Optimization} \bibinfo{volume}{2}~(\bibinfo{number}{2})
  (\bibinfo{year}{2022}) \bibinfo{pages}{1--26}, \bibinfo{note}{doi:
  \url{https://doi.org/10.1145/3545611}}.

\bibitem[{Rumelhart et~al.(1986)Rumelhart, Hinton, and
  Williams}]{Rumelhart1986}
\bibinfo{author}{D.~E. Rumelhart}, \bibinfo{author}{G.~E. Hinton},
  \bibinfo{author}{R.~J. Williams}, \bibinfo{title}{{Learning representations
  by back-propagating errors}}, \bibinfo{journal}{Nature}
  \bibinfo{volume}{323}~(\bibinfo{number}{6088}) (\bibinfo{year}{1986})
  \bibinfo{pages}{533--536}, \bibinfo{note}{doi:
  \url{https://doi.org/10.1038/323533a0}}.

\bibitem[{Robbins and Monro(1951)}]{Robbins1951}
\bibinfo{author}{H.~Robbins}, \bibinfo{author}{S.~Monro}, \bibinfo{title}{{A
  Stochastic Approximation Method}}, \bibinfo{journal}{The Annals of
  Mathematical Statistics} \bibinfo{volume}{22}~(\bibinfo{number}{3})
  (\bibinfo{year}{1951}) \bibinfo{pages}{400--407}, \bibinfo{note}{doi:
  \url{https://doi.org/10.1214/aoms/1177729586}}.

\bibitem[{LeCun et~al.(2012)LeCun, Bottou, Orr, and M{\"{u}}ller}]{LeCun2012}
\bibinfo{author}{Y.~A. LeCun}, \bibinfo{author}{L.~Bottou},
  \bibinfo{author}{G.~B. Orr}, \bibinfo{author}{K.-R. M{\"{u}}ller},
  \bibinfo{title}{{Efficient BackProp}}, in: \bibinfo{editor}{G.~Montavon},
  \bibinfo{editor}{G.~B. Orr}, \bibinfo{editor}{K.-R. M{\"{u}}ller} (Eds.),
  \bibinfo{booktitle}{Neural Networks: Tricks of the Trade},
  \bibinfo{publisher}{Springer-Verlag Berlin Heidelberg},
  \bibinfo{pages}{9--48},
  \bibinfo{note}{doi:\url{https://doi.org/10.1007/978-3-642-35289-8_3}},
  \bibinfo{year}{2012}.

\bibitem[{Berger(1985)}]{Berger1985}
\bibinfo{author}{J.~O. Berger}, \bibinfo{title}{{Statistical Decision Theory
  and Bayesian Analysis}}, Springer Series in Statistics,
  \bibinfo{publisher}{Springer New York}, \bibinfo{address}{New York, NY}, ISBN
  \bibinfo{isbn}{978-1-4419-3074-3}, \bibinfo{note}{doi:
  \url{https://doi.org/10.1007/978-1-4757-4286-2}}, \bibinfo{year}{1985}.

\bibitem[{Bernardo and Smith(2000)}]{Bernardo2000}
\bibinfo{author}{J.~M. Bernardo}, \bibinfo{author}{A.~F.~M. Smith},
  \bibinfo{title}{{Bayesian Theory}}, \bibinfo{publisher}{John Wiley \& Sons},
  \bibinfo{address}{New York, NY}, \bibinfo{year}{2000}.

\bibitem[{Sivia and Skilling(2006)}]{Sivia2006}
\bibinfo{author}{D.~S. Sivia}, \bibinfo{author}{J.~Skilling},
  \bibinfo{title}{{Data Analysis: A Bayesian Tutorial}},
  \bibinfo{publisher}{Oxford University Press}, \bibinfo{address}{New York,
  NY}, \bibinfo{edition}{2nd} edn., \bibinfo{year}{2006}.

\bibitem[{MacKay(1992)}]{MacKay1992}
\bibinfo{author}{D.~J.~C. MacKay}, \bibinfo{title}{{A Practical Bayesian
  Framework for Backpropagation Networks}}, \bibinfo{journal}{Neural
  Computation} \bibinfo{volume}{4}~(\bibinfo{number}{3}) (\bibinfo{year}{1992})
  \bibinfo{pages}{448--472}, \bibinfo{note}{doi:
  \url{https://doi.org/10.1162/neco.1992.4.3.448}}.

\bibitem[{Graves(2011)}]{Graves2011}
\bibinfo{author}{A.~Graves}, \bibinfo{title}{{Practical Variational Inference
  for Neural Networks}}, in: \bibinfo{booktitle}{Advances in Neural Information
  Processing Systems 24 (NIPS 2011)}, \bibinfo{address}{Granada, Spain},
  \bibinfo{pages}{2348--2356}, \bibinfo{year}{2011}.

\bibitem[{Blundell et~al.(2015)Blundell, Cornebise, Kavukcuoglu, and
  Wierstra}]{Blundell2015}
\bibinfo{author}{C.~Blundell}, \bibinfo{author}{J.~Cornebise},
  \bibinfo{author}{K.~Kavukcuoglu}, \bibinfo{author}{D.~Wierstra},
  \bibinfo{title}{{Weight Uncertainty in Neural Networks}}, in:
  \bibinfo{booktitle}{Proceedings of the 32nd International Conference on
  Machine Learning}, vol.~\bibinfo{volume}{37}, \bibinfo{pages}{1613--1622},
  \bibinfo{year}{2015}.

\bibitem[{O'Hagan et~al.(2006)O'Hagan, Buck, Daneshkhah, Eiser, Garthwaite,
  Jenkinson, Oakley, and Rakow}]{OHagan2006a}
\bibinfo{author}{A.~O'Hagan}, \bibinfo{author}{C.~E. Buck},
  \bibinfo{author}{A.~Daneshkhah}, \bibinfo{author}{J.~R. Eiser},
  \bibinfo{author}{P.~H. Garthwaite}, \bibinfo{author}{D.~J. Jenkinson},
  \bibinfo{author}{J.~E. Oakley}, \bibinfo{author}{T.~Rakow},
  \bibinfo{title}{{Uncertain Judgements: Eliciting Experts' Probabilities}},
  \bibinfo{publisher}{John Wiley \& Sons, Ltd}, \bibinfo{address}{Chichester,
  UK}, \bibinfo{note}{doi: \url{https://doi.org/10.1002/0470033312}},
  \bibinfo{year}{2006}.

\bibitem[{Jeffreys(1946)}]{Jeffreys1946}
\bibinfo{author}{H.~Jeffreys}, \bibinfo{title}{{An invariant form for the prior
  probability in estimation problems}}, \bibinfo{journal}{Proceedings of the
  Royal Society of London. Series A. Mathematical and Physical Sciences}
  \bibinfo{volume}{186}~(\bibinfo{number}{1007}) (\bibinfo{year}{1946})
  \bibinfo{pages}{453--461}, \bibinfo{note}{doi:
  \url{https://doi.org/10.1098/rspa.1946.0056}}.

\bibitem[{Jaynes(1968)}]{Jaynes1968}
\bibinfo{author}{E.~T. Jaynes}, \bibinfo{title}{{Prior Probabilities}},
  \bibinfo{journal}{IEEE Transactions on Systems Science and Cybernetics}
  \bibinfo{volume}{4}~(\bibinfo{number}{3}) (\bibinfo{year}{1968})
  \bibinfo{pages}{227--241}, \bibinfo{note}{doi:
  \url{https://doi.org/10.1109/TSSC.1968.300117}}.

\bibitem[{Fortuin(2022)}]{Fortuin2022}
\bibinfo{author}{V.~Fortuin}, \bibinfo{title}{{Priors in Bayesian Deep
  Learning: A Review}}, \bibinfo{journal}{International Statistical Review}
  \bibinfo{volume}{90}~(\bibinfo{number}{3}) (\bibinfo{year}{2022})
  \bibinfo{pages}{563--591}, \bibinfo{note}{doi:
  \url{https://doi.org/10.1111/insr.12502}}.

\bibitem[{Andrieu et~al.(2003)Andrieu, de~Freitas, Doucet, and
  Jordan}]{Andrieu2003}
\bibinfo{author}{C.~Andrieu}, \bibinfo{author}{N.~de~Freitas},
  \bibinfo{author}{A.~Doucet}, \bibinfo{author}{M.~I. Jordan},
  \bibinfo{title}{{An Introduction to MCMC for Machine Learning}},
  \bibinfo{journal}{Machine Learning} \bibinfo{volume}{50}
  (\bibinfo{year}{2003}) \bibinfo{pages}{5--43}, \bibinfo{note}{doi:
  \url{https://doi.org/10.1023/A:1020281327116}}.

\bibitem[{Brooks et~al.(2011)Brooks, Gelman, Jones, and Meng}]{Various2011}
\bibinfo{editor}{S.~Brooks}, \bibinfo{editor}{A.~Gelman},
  \bibinfo{editor}{G.~Jones}, \bibinfo{editor}{X.-L. Meng} (Eds.),
  \bibinfo{title}{{Handbook of Markov Chain Monte Carlo}},
  \bibinfo{publisher}{Chapman \& Hall/CRC}, \bibinfo{note}{doi:
  \url{https://doi.org/10.1201/b10905}}, \bibinfo{year}{2011}.

\bibitem[{Metropolis et~al.(1953)Metropolis, Rosenbluth, Rosenbluth, Teller,
  and Teller}]{Metropolis1953a}
\bibinfo{author}{N.~Metropolis}, \bibinfo{author}{A.~W. Rosenbluth},
  \bibinfo{author}{M.~N. Rosenbluth}, \bibinfo{author}{A.~H. Teller},
  \bibinfo{author}{E.~Teller}, \bibinfo{title}{{Equation of State Calculations
  by Fast Computing Machines}}, \bibinfo{journal}{The Journal of Chemical
  Physics} \bibinfo{volume}{21}~(\bibinfo{number}{6}) (\bibinfo{year}{1953})
  \bibinfo{pages}{1087--1092}, \bibinfo{note}{doi:
  \url{https://doi.org/10.1063/1.1699114}}.

\bibitem[{Hastings(1970)}]{Hastings1970a}
\bibinfo{author}{W.~K. Hastings}, \bibinfo{title}{{Monte Carlo sampling methods
  using Markov chains and their applications}}, \bibinfo{journal}{Biometrika}
  \bibinfo{volume}{57}~(\bibinfo{number}{1}) (\bibinfo{year}{1970})
  \bibinfo{pages}{97--109}, \bibinfo{note}{doi:
  \url{https://doi.org/10.1093/biomet/57.1.97}}.

\bibitem[{Neal(2011)}]{Neal2011a}
\bibinfo{author}{R.~M. Neal}, \bibinfo{title}{{MCMC Using Hamiltonian
  Dynamics}}, in: \bibinfo{booktitle}{Handbook of Markov Chain Monte Carlo},
  \bibinfo{pages}{113--162}, \bibinfo{note}{doi:
  \url{https://doi.org/10.1201/b10905-6}}, \bibinfo{year}{2011}.

\bibitem[{Betancourt(2017)}]{Betancourt2017}
\bibinfo{author}{M.~Betancourt}, \bibinfo{title}{{A Conceptual Introduction to
  Hamiltonian Monte Carlo}}, \bibinfo{journal}{arXiv preprint arXiv:1701.02434}
  \bibinfo{note}{{doi}: \url{https://doi.org/10.48550/arXiv.1701.02434}}.

\bibitem[{Chen et~al.(2014)Chen, Fox, and Guestrin}]{Chen2014}
\bibinfo{author}{T.~Chen}, \bibinfo{author}{E.~B. Fox},
  \bibinfo{author}{C.~Guestrin}, \bibinfo{title}{{Stochastic Gradient
  Hamiltonian Monte Carlo}}, in: \bibinfo{booktitle}{Proceedings of the 31st
  International Conference on Machine Learning}, vol.~\bibinfo{volume}{32},
  \bibinfo{address}{Beijing}, \bibinfo{pages}{1683--1691},
  \bibinfo{year}{2014}.

\bibitem[{Zhang et~al.(2018)Zhang, Shahbaba, and Zhao}]{Zhang2018}
\bibinfo{author}{C.~Zhang}, \bibinfo{author}{B.~Shahbaba},
  \bibinfo{author}{H.~Zhao}, \bibinfo{title}{{Variational Hamiltonian Monte
  Carlo via Score Matching}}, \bibinfo{journal}{Bayesian Analysis}
  \bibinfo{volume}{13}~(\bibinfo{number}{2}) (\bibinfo{year}{2018})
  \bibinfo{pages}{485--506}, \bibinfo{note}{doi:
  \url{https://doi.org/10.1214/17-BA1060}}.

\bibitem[{Blei et~al.(2017)Blei, Kucukelbir, and McAuliffe}]{Blei2017}
\bibinfo{author}{D.~M. Blei}, \bibinfo{author}{A.~Kucukelbir},
  \bibinfo{author}{J.~D. McAuliffe}, \bibinfo{title}{{Variational Inference: A
  Review for Statisticians}}, \bibinfo{journal}{Journal of the American
  Statistical Association} \bibinfo{volume}{112}~(\bibinfo{number}{518})
  (\bibinfo{year}{2017}) \bibinfo{pages}{859--877}, \bibinfo{note}{doi:
  \url{https://doi.org/10.1080/01621459.2017.1285773}}.

\bibitem[{Zhang et~al.(2019{\natexlab{b}})Zhang, Butepage, Kjellstrom, and
  Mandt}]{Zhang2019}
\bibinfo{author}{C.~Zhang}, \bibinfo{author}{J.~Butepage},
  \bibinfo{author}{H.~Kjellstrom}, \bibinfo{author}{S.~Mandt},
  \bibinfo{title}{{Advances in Variational Inference}}, \bibinfo{journal}{IEEE
  Transactions on Pattern Analysis and Machine Intelligence}
  \bibinfo{volume}{41}~(\bibinfo{number}{8})
  (\bibinfo{year}{2019}{\natexlab{b}}) \bibinfo{pages}{2008--2026},
  \bibinfo{note}{doi: \url{https://doi.org/10.1109/TPAMI.2018.2889774}}.

\bibitem[{Rezende and Mohamed(2015)}]{Rezende2015}
\bibinfo{author}{D.~J. Rezende}, \bibinfo{author}{S.~Mohamed},
  \bibinfo{title}{{Variational inference with normalizing flows}}, in:
  \bibinfo{booktitle}{32nd International Conference on Machine Learning, ICML
  2015}, vol.~\bibinfo{volume}{2}, \bibinfo{pages}{1530--1538},
  \bibinfo{year}{2015}.

\bibitem[{Marzouk et~al.(2016)Marzouk, Moselhy, Parno, and
  Spantini}]{Marzouk2016}
\bibinfo{author}{Y.~Marzouk}, \bibinfo{author}{T.~Moselhy},
  \bibinfo{author}{M.~Parno}, \bibinfo{author}{A.~Spantini},
  \bibinfo{title}{{Sampling via Measure Transport: An Introduction}}, in:
  \bibinfo{booktitle}{Handbook of Uncertainty Quantification},
  \bibinfo{publisher}{Springer International Publishing},
  \bibinfo{address}{Cham}, \bibinfo{pages}{1--41}, \bibinfo{note}{doi:
  \url{https://doi.org/10.1007/978-3-319-11259-6_23-1}}, \bibinfo{year}{2016}.

\bibitem[{Liu and Wang(2016)}]{Liu2016}
\bibinfo{author}{Q.~Liu}, \bibinfo{author}{D.~Wang}, \bibinfo{title}{{Stein
  Variational Gradient Descent: A General Purpose Bayesian Inference
  Algorithm}}, in: \bibinfo{booktitle}{Advances in Neural Information
  Processing Systems 29 (NIPS 2016)}, \bibinfo{address}{Barcelona, Spain},
  \bibinfo{pages}{2378--2386}, \bibinfo{year}{2016}.

\bibitem[{Detommaso et~al.(2018)Detommaso, Cui, Spantini, Marzouk, and
  Scheichl}]{Detommaso2018}
\bibinfo{author}{G.~Detommaso}, \bibinfo{author}{T.~Cui},
  \bibinfo{author}{A.~Spantini}, \bibinfo{author}{Y.~Marzouk},
  \bibinfo{author}{R.~Scheichl}, \bibinfo{title}{{A Stein variational Newton
  method}}, in: \bibinfo{booktitle}{Advances in Neural Information Processing
  Systems}, \bibinfo{pages}{9169--9179}, \bibinfo{note}{doi:
  \url{https://doi.org/10.48550/arXiv.1806.03085}}, \bibinfo{year}{2018}.

\bibitem[{Leviyev et~al.(2022)Leviyev, Chen, Wang, Ghattas, and
  Zimmerman}]{Leviyev2022}
\bibinfo{author}{A.~Leviyev}, \bibinfo{author}{J.~Chen},
  \bibinfo{author}{Y.~Wang}, \bibinfo{author}{O.~Ghattas},
  \bibinfo{author}{A.~Zimmerman}, \bibinfo{title}{{A stochastic Stein
  Variational Newton method}}, \bibinfo{journal}{arXiv preprint
  arXiv:2204.09039} ~(\bibinfo{number}{2016}) (\bibinfo{year}{2022})
  \bibinfo{pages}{1--17}, \bibinfo{note}{doi:
  \url{https://doi.org/10.48550/arXiv.2204.09039}}.

\bibitem[{Chen and Ghattas(2020)}]{Chen2020}
\bibinfo{author}{P.~Chen}, \bibinfo{author}{O.~Ghattas},
  \bibinfo{title}{{Projected stein variational gradient descent}}, in:
  \bibinfo{booktitle}{Advances in Neural Information Processing Systems},
  \bibinfo{note}{doi: \url{https://doi.org/10.48550/arXiv.2002.03469}},
  \bibinfo{year}{2020}.

\bibitem[{Minka(2001)}]{minka2001expectation}
\bibinfo{author}{T.~P. Minka}, \bibinfo{title}{Expectation propagation for
  approximate {B}ayesian inference}, in: \bibinfo{booktitle}{Proceedings of the
  Seventeenth Conference on Uncertainty in Artificial Intelligence}, UAI’01,
  \bibinfo{publisher}{AUAI Press}, \bibinfo{address}{Seattle, Washington, USA},
  \bibinfo{pages}{362--369}, \bibinfo{note}{doi:
  \url{https://doi.org/10.48550/arXiv.1301.2294}}, \bibinfo{year}{2001}.

\bibitem[{Lauritzen(1992)}]{lauritzen1992propagation}
\bibinfo{author}{S.~L. Lauritzen}, \bibinfo{title}{Propagation of
  probabilities, means, and variances in mixed graphical association models},
  \bibinfo{journal}{Journal of the American Statistical Association}
  \bibinfo{volume}{87}~(\bibinfo{number}{420}) (\bibinfo{year}{1992})
  \bibinfo{pages}{1098--1108}, \bibinfo{note}{doi:
  \url{https://doi.org/10.2307/2290647}}.

\bibitem[{Opper and Winther(1999)}]{opper1999bayesian}
\bibinfo{author}{M.~Opper}, \bibinfo{author}{O.~Winther}, \bibinfo{title}{A
  {B}ayesian approach to on-line learning} \bibinfo{note}{{doi}:
  \url{https://doi.org/10.2277/0521652634}}.

\bibitem[{Shen et~al.(2021)Shen, Chen, and Deng}]{shen2021variational}
\bibinfo{author}{G.~Shen}, \bibinfo{author}{X.~Chen},
  \bibinfo{author}{Z.~Deng}, \bibinfo{title}{Variational learning of Bayesian
  neural networks via Bayesian dark knowledge}, in:
  \bibinfo{booktitle}{Proceedings of the Twenty-Ninth International Conference
  on International Joint Conferences on Artificial Intelligence},
  \bibinfo{pages}{2037--2043}, \bibinfo{note}{doi:
  \url{https://doi.org/10.24963/ijcai.2020/282}}, \bibinfo{year}{2021}.

\bibitem[{Srivastava et~al.(2014)Srivastava, Hinton, Krizhevsky, Sutskever, and
  Salakhutdinov}]{srivastava2014dropout}
\bibinfo{author}{N.~Srivastava}, \bibinfo{author}{G.~Hinton},
  \bibinfo{author}{A.~Krizhevsky}, \bibinfo{author}{I.~Sutskever},
  \bibinfo{author}{R.~Salakhutdinov}, \bibinfo{title}{Dropout: a simple way to
  prevent neural networks from overfitting}, \bibinfo{journal}{The Journal of
  Machine Learning Research} \bibinfo{volume}{15}~(\bibinfo{number}{1})
  (\bibinfo{year}{2014}) \bibinfo{pages}{1929--1958}.

\bibitem[{Gal and Ghahramani(2016{\natexlab{b}})}]{gal2016theoretically}
\bibinfo{author}{Y.~Gal}, \bibinfo{author}{Z.~Ghahramani}, \bibinfo{title}{A
  theoretically grounded application of dropout in recurrent neural networks},
  in: \bibinfo{booktitle}{Advances in Neural Information Processing Systems},
  \bibinfo{pages}{1019--1027}, \bibinfo{note}{doi:
  \url{https://doi.org/10.48550/arXiv.1512.05287}},
  \bibinfo{year}{2016}{\natexlab{b}}.

\bibitem[{Gal and Ghahramani(2015)}]{gal2015bayesian}
\bibinfo{author}{Y.~Gal}, \bibinfo{author}{Z.~Ghahramani},
  \bibinfo{title}{Bayesian convolutional neural networks with Bernoulli
  approximate variational inference}, \bibinfo{journal}{arXiv preprint
  arXiv:1506.02158} \bibinfo{note}{Doi:
  \url{https://doi.org/10.48550/arXiv.1506.02158}}.

\bibitem[{Osband(2016)}]{osband2016risk}
\bibinfo{author}{I.~Osband}, \bibinfo{title}{Risk versus uncertainty in deep
  learning: Bayes, bootstrap and the dangers of dropout}, in:
  \bibinfo{booktitle}{NIPS Workshop on {B}ayesian Deep Learning}, vol.
  \bibinfo{volume}{192}, \bibinfo{year}{2016}.

\bibitem[{Alarab et~al.(2021)Alarab, Prakoonwit, and
  Nacer}]{alarab2021illustrative}
\bibinfo{author}{I.~Alarab}, \bibinfo{author}{S.~Prakoonwit},
  \bibinfo{author}{M.~I. Nacer}, \bibinfo{title}{Illustrative discussion of
  mc-dropout in general dataset: Uncertainty estimation in bitcoin},
  \bibinfo{journal}{Neural Processing Letters}
  \bibinfo{volume}{53}~(\bibinfo{number}{2}) (\bibinfo{year}{2021})
  \bibinfo{pages}{1001--1011}, \bibinfo{note}{doi:
  \url{https://doi.org/10.1007/s11063-021-10424-x}}.

\bibitem[{Caldeira and Nord(2020)}]{caldeira2020deeply}
\bibinfo{author}{J.~Caldeira}, \bibinfo{author}{B.~Nord},
  \bibinfo{title}{Deeply uncertain: comparing methods of uncertainty
  quantification in deep learning algorithms}, \bibinfo{journal}{Machine
  Learning: Science and Technology} \bibinfo{volume}{2}~(\bibinfo{number}{1})
  (\bibinfo{year}{2020}) \bibinfo{pages}{015002}, \bibinfo{note}{doi:
  \url{https://doi.org/10.1088/2632-2153/aba6f3}}.

\bibitem[{Foong et~al.(2020)Foong, Burt, Li, and
  Turner}]{foong2020expressiveness}
\bibinfo{author}{A.~Foong}, \bibinfo{author}{D.~Burt}, \bibinfo{author}{Y.~Li},
  \bibinfo{author}{R.~Turner}, \bibinfo{title}{On the expressiveness of
  approximate inference in {B}ayesian neural networks},
  \bibinfo{journal}{Advances in Neural Information Processing Systems}
  \bibinfo{volume}{33} (\bibinfo{year}{2020}) \bibinfo{pages}{15897--15908},
  \bibinfo{note}{doi: \url{https://doi.org/10.48550/arXiv.1909.00719}}.

\bibitem[{Verdoja and Kyrki(2020)}]{verdoja2020notes}
\bibinfo{author}{F.~Verdoja}, \bibinfo{author}{V.~Kyrki}, \bibinfo{title}{Notes
  on the behavior of {MC} dropout} \bibinfo{note}{{doi}:
  \url{https://doi.org/10.48550/arXiv.2008.02627}}.

\bibitem[{Opitz and Maclin(1999)}]{opitz1999popular}
\bibinfo{author}{D.~Opitz}, \bibinfo{author}{R.~Maclin},
  \bibinfo{title}{Popular ensemble methods: An empirical study},
  \bibinfo{journal}{Journal of Artificial Intelligence Research}
  \bibinfo{volume}{11} (\bibinfo{year}{1999}) \bibinfo{pages}{169--198},
  \bibinfo{note}{doi: \url{https://doi.org/10.1613/jair.614}}.

\bibitem[{Dietterich(2000)}]{dietterich2000ensemble}
\bibinfo{author}{T.~G. Dietterich}, \bibinfo{title}{Ensemble methods in machine
  learning}, in: \bibinfo{booktitle}{International Workshop on Multiple
  Classifier Systems}, \bibinfo{organization}{Springer},
  \bibinfo{pages}{1--15}, \bibinfo{note}{doi:
  \url{https://doi.org/10.1007/3-540-45014-9_1}}, \bibinfo{year}{2000}.

\bibitem[{Breiman(1996)}]{breiman1996bagging}
\bibinfo{author}{L.~Breiman}, \bibinfo{title}{Bagging predictors},
  \bibinfo{journal}{Machine Learning}
  \bibinfo{volume}{24}~(\bibinfo{number}{2}) (\bibinfo{year}{1996})
  \bibinfo{pages}{123--140}, \bibinfo{note}{doi:
  \url{https://doi.org/10.1007/BF00058655}}.

\bibitem[{Schapire and Freund(2013)}]{schapire2013boosting}
\bibinfo{author}{R.~E. Schapire}, \bibinfo{author}{Y.~Freund},
  \bibinfo{title}{Boosting: Foundations and algorithms},
  \bibinfo{journal}{Kybernetes} \bibinfo{note}{{doi}:
  \url{https://doi.org/10.7551/mitpress/8291.001.0001}}.

\bibitem[{Zhang and Mahadevan(2019)}]{zhang2019ensemble}
\bibinfo{author}{X.~Zhang}, \bibinfo{author}{S.~Mahadevan},
  \bibinfo{title}{Ensemble machine learning models for aviation incident risk
  prediction}, \bibinfo{journal}{Decision Support Systems}
  \bibinfo{volume}{116} (\bibinfo{year}{2019}) \bibinfo{pages}{48--63},
  \bibinfo{note}{doi: \url{https://doi.org/10.1016/j.dss.2018.10.009}}.

\bibitem[{Ovadia et~al.(2019)Ovadia, Fertig, Ren, Nado, Sculley, Nowozin,
  Dillon, Lakshminarayanan, and Snoek}]{ovadia2019can}
\bibinfo{author}{Y.~Ovadia}, \bibinfo{author}{E.~Fertig},
  \bibinfo{author}{J.~Ren}, \bibinfo{author}{Z.~Nado},
  \bibinfo{author}{D.~Sculley}, \bibinfo{author}{S.~Nowozin},
  \bibinfo{author}{J.~Dillon}, \bibinfo{author}{B.~Lakshminarayanan},
  \bibinfo{author}{J.~Snoek}, \bibinfo{title}{Can you trust your model's
  uncertainty? evaluating predictive uncertainty under dataset shift},
  \bibinfo{journal}{Advances in Neural Information Processing Systems}
  \bibinfo{volume}{32}, \bibinfo{note}{doi:
  \url{https://doi.org/10.48550/arXiv.1906.02530}}.

\bibitem[{Nix and Weigend(1994)}]{nix1994estimating}
\bibinfo{author}{D.~A. Nix}, \bibinfo{author}{A.~S. Weigend},
  \bibinfo{title}{Estimating the mean and variance of the target probability
  distribution}, in: \bibinfo{booktitle}{Proceedings of 1994 IEEE International
  Conference on Neural Networks (ICNN'94)}, vol.~\bibinfo{volume}{1},
  \bibinfo{organization}{IEEE}, \bibinfo{pages}{55--60}, \bibinfo{note}{doi:
  \url{https://doi.org/10.1109/ICNN.1994.374138}}, \bibinfo{year}{1994}.

\bibitem[{Fort et~al.(2019)Fort, Hu, and Lakshminarayanan}]{fort2019deep}
\bibinfo{author}{S.~Fort}, \bibinfo{author}{H.~Hu},
  \bibinfo{author}{B.~Lakshminarayanan}, \bibinfo{title}{Deep ensembles: A loss
  landscape perspective} \bibinfo{note}{{doi}:
  \url{https://doi.org/10.48550/arXiv.1912.02757}}.

\bibitem[{Dodson et~al.(2022)Dodson, Downey, Laflamme, Todd, Moura, Wang, Mao,
  Avitabile, and Blasch}]{dodson2022high}
\bibinfo{author}{J.~Dodson}, \bibinfo{author}{A.~Downey},
  \bibinfo{author}{S.~Laflamme}, \bibinfo{author}{M.~D. Todd},
  \bibinfo{author}{A.~G. Moura}, \bibinfo{author}{Y.~Wang},
  \bibinfo{author}{Z.~Mao}, \bibinfo{author}{P.~Avitabile},
  \bibinfo{author}{E.~Blasch}, \bibinfo{title}{High-rate structural health
  monitoring and prognostics: An overview}, \bibinfo{journal}{Data Science in
  Engineering, Volume 9}  (\bibinfo{year}{2022})
  \bibinfo{pages}{213--217}\bibinfo{note}{{doi}:
  \url{https://doi.org/10.1007/978-3-030-76004-5_23}}.

\bibitem[{Kiran et~al.(2021)Kiran, Sobh, Talpaert, Mannion, Al~Sallab,
  Yogamani, and P{\'e}rez}]{kiran2021deep}
\bibinfo{author}{B.~R. Kiran}, \bibinfo{author}{I.~Sobh},
  \bibinfo{author}{V.~Talpaert}, \bibinfo{author}{P.~Mannion},
  \bibinfo{author}{A.~A. Al~Sallab}, \bibinfo{author}{S.~Yogamani},
  \bibinfo{author}{P.~P{\'e}rez}, \bibinfo{title}{Deep reinforcement learning
  for autonomous driving: A survey}, \bibinfo{journal}{IEEE Transactions on
  Intelligent Transportation Systems}
  \bibinfo{volume}{23}~(\bibinfo{number}{6}) (\bibinfo{year}{2021})
  \bibinfo{pages}{4909--4926}, \bibinfo{note}{doi:
  \url{https://doi.org/10.1109/TITS.2021.3054625}}.

\bibitem[{Van~Amersfoort et~al.(2020)Van~Amersfoort, Smith, Teh, and
  Gal}]{van2020uncertainty}
\bibinfo{author}{J.~Van~Amersfoort}, \bibinfo{author}{L.~Smith},
  \bibinfo{author}{Y.~W. Teh}, \bibinfo{author}{Y.~Gal},
  \bibinfo{title}{Uncertainty estimation using a single deep deterministic
  neural network}, in: \bibinfo{booktitle}{International Conference on Machine
  Learning}, \bibinfo{organization}{PMLR}, \bibinfo{pages}{9690--9700},
  \bibinfo{note}{doi: \url{https://doi.org/10.48550/arXiv.2003.02037}},
  \bibinfo{year}{2020}.

\bibitem[{Mukhoti et~al.(2021)Mukhoti, Kirsch, van Amersfoort, Torr, and
  Gal}]{mukhoti2021deterministic}
\bibinfo{author}{J.~Mukhoti}, \bibinfo{author}{A.~Kirsch},
  \bibinfo{author}{J.~van Amersfoort}, \bibinfo{author}{P.~H. Torr},
  \bibinfo{author}{Y.~Gal}, \bibinfo{title}{Deterministic neural networks with
  appropriate inductive biases capture epistemic and aleatoric uncertainty},
  \bibinfo{journal}{arXiv preprint arXiv:2102.11582} .

\bibitem[{van Amersfoort et~al.(2021)van Amersfoort, Smith, Jesson, Key, and
  Gal}]{van2021feature}
\bibinfo{author}{J.~van Amersfoort}, \bibinfo{author}{L.~Smith},
  \bibinfo{author}{A.~Jesson}, \bibinfo{author}{O.~Key},
  \bibinfo{author}{Y.~Gal}, \bibinfo{title}{On feature collapse and deep kernel
  learning for single forward pass uncertainty} \bibinfo{note}{{doi}:
  \url{https://doi.org/10.48550/arXiv.2102.11409}}.

\bibitem[{Liu et~al.(2020{\natexlab{b}})Liu, Lin, Padhy, Tran, Bedrax~Weiss,
  and Lakshminarayanan}]{liu2020simple}
\bibinfo{author}{J.~Liu}, \bibinfo{author}{Z.~Lin}, \bibinfo{author}{S.~Padhy},
  \bibinfo{author}{D.~Tran}, \bibinfo{author}{T.~Bedrax~Weiss},
  \bibinfo{author}{B.~Lakshminarayanan}, \bibinfo{title}{Simple and principled
  uncertainty estimation with deterministic deep learning via distance
  awareness}, \bibinfo{journal}{Advances in Neural Information Processing
  Systems} \bibinfo{volume}{33} (\bibinfo{year}{2020}{\natexlab{b}})
  \bibinfo{pages}{7498--7512}, \bibinfo{note}{doi:
  \url{https://doi.org/10.48550/arXiv.2006.10108}}.

\bibitem[{Fortuin et~al.(2021)Fortuin, Collier, Wenzel, Allingham, Liu, Tran,
  Lakshminarayanan, Berent, Jenatton, and Kokiopoulou}]{fortuin2021deep}
\bibinfo{author}{V.~Fortuin}, \bibinfo{author}{M.~Collier},
  \bibinfo{author}{F.~Wenzel}, \bibinfo{author}{J.~Allingham},
  \bibinfo{author}{J.~Liu}, \bibinfo{author}{D.~Tran},
  \bibinfo{author}{B.~Lakshminarayanan}, \bibinfo{author}{J.~Berent},
  \bibinfo{author}{R.~Jenatton}, \bibinfo{author}{E.~Kokiopoulou},
  \bibinfo{title}{Deep classifiers with label noise modeling and distance
  awareness} \bibinfo{note}{{doi}:
  \url{https://doi.org/10.48550/arXiv.2110.02609}}.

\bibitem[{Gulrajani et~al.(2017)Gulrajani, Ahmed, Arjovsky, Dumoulin, and
  Courville}]{gulrajani2017improved}
\bibinfo{author}{I.~Gulrajani}, \bibinfo{author}{F.~Ahmed},
  \bibinfo{author}{M.~Arjovsky}, \bibinfo{author}{V.~Dumoulin},
  \bibinfo{author}{A.~C. Courville}, \bibinfo{title}{Improved training of
  wasserstein gans}, \bibinfo{journal}{Advances in Neural Information
  Processing Systems} \bibinfo{volume}{30}, \bibinfo{note}{doi:
  \url{https://doi.org/10.48550/arXiv.1704.00028}}.

\bibitem[{Miyato et~al.(2018)Miyato, Kataoka, Koyama, and
  Yoshida}]{miyato2018spectral}
\bibinfo{author}{T.~Miyato}, \bibinfo{author}{T.~Kataoka},
  \bibinfo{author}{M.~Koyama}, \bibinfo{author}{Y.~Yoshida},
  \bibinfo{title}{Spectral normalization for generative adversarial networks}
  \bibinfo{note}{{doi}: \url{https://doi.org/10.48550/arXiv.1802.05957}}.

\bibitem[{Postels et~al.(2021)Postels, Segu, Sun, Van~Gool, Yu, and
  Tombari}]{postels2021practicality}
\bibinfo{author}{J.~Postels}, \bibinfo{author}{M.~Segu},
  \bibinfo{author}{T.~Sun}, \bibinfo{author}{L.~Van~Gool},
  \bibinfo{author}{F.~Yu}, \bibinfo{author}{F.~Tombari}, \bibinfo{title}{On the
  practicality of deterministic epistemic uncertainty} \bibinfo{note}{{doi}:
  \url{https://doi.org/10.48550/arXiv.2107.00649}}.

\bibitem[{Van~Landeghem et~al.(2022)Van~Landeghem, Blaschko, Anckaert, and
  Moens}]{van2022benchmarking}
\bibinfo{author}{J.~Van~Landeghem}, \bibinfo{author}{M.~Blaschko},
  \bibinfo{author}{B.~Anckaert}, \bibinfo{author}{M.-F. Moens},
  \bibinfo{title}{Benchmarking scalable predictive uncertainty in text
  classification}, \bibinfo{journal}{IEEE Access} \bibinfo{volume}{10}
  (\bibinfo{year}{2022}) \bibinfo{pages}{43703--43737}, \bibinfo{note}{doi:
  \url{https://doi.org/10.1109/ACCESS.2022.3168734}}.

\bibitem[{DeGroot and Fienberg(1983)}]{degroot1983comparison}
\bibinfo{author}{M.~H. DeGroot}, \bibinfo{author}{S.~E. Fienberg},
  \bibinfo{title}{The comparison and evaluation of forecasters},
  \bibinfo{journal}{Journal of the Royal Statistical Society: Series D (The
  Statistician)} \bibinfo{volume}{32}~(\bibinfo{number}{1-2})
  (\bibinfo{year}{1983}) \bibinfo{pages}{12--22}, \bibinfo{note}{doi:
  \url{https://doi.org/10.2307/2987588}}.

\bibitem[{Zadrozny and Elkan(2002)}]{zadrozny2002transforming}
\bibinfo{author}{B.~Zadrozny}, \bibinfo{author}{C.~Elkan},
  \bibinfo{title}{Transforming classifier scores into accurate multiclass
  probability estimates}, in: \bibinfo{booktitle}{Proceedings of the Eighth ACM
  SIGKDD International Conference on Knowledge discovery and Data Mining},
  \bibinfo{pages}{694--699}, \bibinfo{note}{doi:
  \url{https://doi.org/10.1145/775047.775151}}, \bibinfo{year}{2002}.

\bibitem[{Niculescu-Mizil and Caruana(2005)}]{niculescu2005predicting}
\bibinfo{author}{A.~Niculescu-Mizil}, \bibinfo{author}{R.~Caruana},
  \bibinfo{title}{Predicting good probabilities with supervised learning}, in:
  \bibinfo{booktitle}{Proceedings of the 22nd International Conference on
  Machine Learning}, \bibinfo{pages}{625--632}, \bibinfo{note}{doi:
  \url{https://doi.org/10.1145/1102351.1102430}}, \bibinfo{year}{2005}.

\bibitem[{Liu et~al.(2011)Liu, Chen, Arendt, and Huang}]{liu2011toward}
\bibinfo{author}{Y.~Liu}, \bibinfo{author}{W.~Chen},
  \bibinfo{author}{P.~Arendt}, \bibinfo{author}{H.-Z. Huang},
  \bibinfo{title}{Toward a better understanding of model validation metrics},
  \bibinfo{journal}{Journal of Mechanical Design}
  \bibinfo{volume}{133}~(\bibinfo{number}{7}), \bibinfo{note}{doi:
  \url{https://doi.org/10.1115/1.4004223}}.

\bibitem[{Naeini et~al.(2015)Naeini, Cooper, and
  Hauskrecht}]{naeini2015obtaining}
\bibinfo{author}{M.~P. Naeini}, \bibinfo{author}{G.~Cooper},
  \bibinfo{author}{M.~Hauskrecht}, \bibinfo{title}{Obtaining well calibrated
  probabilities using {B}ayesian binning}, in: \bibinfo{booktitle}{Twenty-Ninth
  AAAI Conference on Artificial Intelligence}, \bibinfo{note}{doi:
  \url{https://doi.org/10.1609/aaai.v29i1.9602}}, \bibinfo{year}{2015}.

\bibitem[{Kuleshov et~al.(2018)Kuleshov, Fenner, and
  Ermon}]{kuleshov2018accurate}
\bibinfo{author}{V.~Kuleshov}, \bibinfo{author}{N.~Fenner},
  \bibinfo{author}{S.~Ermon}, \bibinfo{title}{Accurate uncertainties for deep
  learning using calibrated regression}, in: \bibinfo{booktitle}{International
  Conference on Machine Learning}, \bibinfo{organization}{PMLR},
  \bibinfo{pages}{2796--2804}, \bibinfo{note}{doi:
  \url{https://doi.org/10.48550/arXiv.1807.00263}}, \bibinfo{year}{2018}.

\bibitem[{Platt et~al.(1999)}]{platt1999probabilistic}
\bibinfo{author}{J.~Platt}, et~al., \bibinfo{title}{Probabilistic outputs for
  support vector machines and comparisons to regularized likelihood methods},
  \bibinfo{journal}{Advances in Large Margin Classifiers}
  \bibinfo{volume}{10}~(\bibinfo{number}{3}) (\bibinfo{year}{1999})
  \bibinfo{pages}{61--74}.

\bibitem[{Guo et~al.(2017)Guo, Pleiss, Sun, and
  Weinberger}]{guo2017calibration}
\bibinfo{author}{C.~Guo}, \bibinfo{author}{G.~Pleiss},
  \bibinfo{author}{Y.~Sun}, \bibinfo{author}{K.~Q. Weinberger},
  \bibinfo{title}{On calibration of modern neural networks}, in:
  \bibinfo{booktitle}{International Conference on Machine Learning},
  \bibinfo{organization}{PMLR}, \bibinfo{pages}{1321--1330},
  \bibinfo{note}{doi: \url{https://doi.org/10.48550/arXiv.1706.04599}},
  \bibinfo{year}{2017}.

\bibitem[{Roman et~al.(2021)Roman, Saxena, Robu, Pecht, and
  Flynn}]{roman2021machine}
\bibinfo{author}{D.~Roman}, \bibinfo{author}{S.~Saxena},
  \bibinfo{author}{V.~Robu}, \bibinfo{author}{M.~Pecht},
  \bibinfo{author}{D.~Flynn}, \bibinfo{title}{Machine learning pipeline for
  battery state-of-health estimation}, \bibinfo{journal}{Nature Machine
  Intelligence} \bibinfo{volume}{3}~(\bibinfo{number}{5})
  (\bibinfo{year}{2021}) \bibinfo{pages}{447--456}, \bibinfo{note}{doi:
  \url{https://doi.org/10.1038/s42256-021-00312-3}}.

\bibitem[{Ferson et~al.(2008)Ferson, Oberkampf, and Ginzburg}]{ferson2008model}
\bibinfo{author}{S.~Ferson}, \bibinfo{author}{W.~L. Oberkampf},
  \bibinfo{author}{L.~Ginzburg}, \bibinfo{title}{Model validation and
  predictive capability for the thermal challenge problem},
  \bibinfo{journal}{Computer Methods in Applied Mechanics and Engineering}
  \bibinfo{volume}{197}~(\bibinfo{number}{29-32}) (\bibinfo{year}{2008})
  \bibinfo{pages}{2408--2430}, \bibinfo{note}{doi:
  \url{https://doi.org/10.1016/j.cma.2007.07.030}}.

\bibitem[{Kondermann et~al.(2008)Kondermann, Mester, and
  Garbe}]{kondermann2008statistical}
\bibinfo{author}{C.~Kondermann}, \bibinfo{author}{R.~Mester},
  \bibinfo{author}{C.~Garbe}, \bibinfo{title}{A statistical confidence measure
  for optical flows}, in: \bibinfo{booktitle}{European Conference on Computer
  Vision}, \bibinfo{organization}{Springer}, \bibinfo{pages}{290--301},
  \bibinfo{note}{doi: \url{https://doi.org/10.1007/978-3-540-88690-7_22}},
  \bibinfo{year}{2008}.

\bibitem[{Amini et~al.(2020)Amini, Schwarting, Soleimany, and
  Rus}]{amini2020deep}
\bibinfo{author}{A.~Amini}, \bibinfo{author}{W.~Schwarting},
  \bibinfo{author}{A.~Soleimany}, \bibinfo{author}{D.~Rus},
  \bibinfo{title}{Deep evidential regression}, \bibinfo{journal}{Advances in
  Neural Information Processing Systems} \bibinfo{volume}{33}
  (\bibinfo{year}{2020}) \bibinfo{pages}{14927--14937}, \bibinfo{note}{doi:
  \url{https://doi.org/10.48550/arXiv.1910.02600}}.

\bibitem[{Ilg et~al.(2018)Ilg, Cicek, Galesso, Klein, Makansi, Hutter, and
  Brox}]{ilg2018uncertainty}
\bibinfo{author}{E.~Ilg}, \bibinfo{author}{O.~Cicek},
  \bibinfo{author}{S.~Galesso}, \bibinfo{author}{A.~Klein},
  \bibinfo{author}{O.~Makansi}, \bibinfo{author}{F.~Hutter},
  \bibinfo{author}{T.~Brox}, \bibinfo{title}{Uncertainty estimates and
  multi-hypotheses networks for optical flow}, in:
  \bibinfo{booktitle}{Proceedings of the European Conference on Computer Vision
  (ECCV)}, \bibinfo{pages}{652--667}, \bibinfo{note}{doi:
  \url{https://doi.org/10.1007/978-3-030-01234-2_40}}, \bibinfo{year}{2018}.

\bibitem[{Hastie et~al.(2009)Hastie, Tibshirani, Friedman, and
  Friedman}]{hastie2009elements}
\bibinfo{author}{T.~Hastie}, \bibinfo{author}{R.~Tibshirani},
  \bibinfo{author}{J.~H. Friedman}, \bibinfo{author}{J.~H. Friedman},
  \bibinfo{title}{The elements of statistical learning: data mining, inference,
  and prediction}, vol.~\bibinfo{volume}{2}, \bibinfo{publisher}{Springer},
  \bibinfo{note}{doi: \url{https://doi.org/10.1007/978-0-387-84858-7}},
  \bibinfo{year}{2009}.

\bibitem[{D'Angelo and Fortuin(2021)}]{d2021repulsive}
\bibinfo{author}{F.~D'Angelo}, \bibinfo{author}{V.~Fortuin},
  \bibinfo{title}{Repulsive deep ensembles are {B}ayesian},
  \bibinfo{journal}{Advances in Neural Information Processing Systems}
  \bibinfo{volume}{34} (\bibinfo{year}{2021}) \bibinfo{pages}{3451--3465},
  \bibinfo{note}{doi: \url{https://doi.org/10.48550/arXiv.2106.11642}}.

\bibitem[{Zhang et~al.(2021)Zhang, Bengio, Hardt, Recht, and
  Vinyals}]{zhang2021understanding}
\bibinfo{author}{C.~Zhang}, \bibinfo{author}{S.~Bengio},
  \bibinfo{author}{M.~Hardt}, \bibinfo{author}{B.~Recht},
  \bibinfo{author}{O.~Vinyals}, \bibinfo{title}{Understanding deep learning
  (still) requires rethinking generalization}, \bibinfo{journal}{Communications
  of the ACM} \bibinfo{volume}{64}~(\bibinfo{number}{3}) (\bibinfo{year}{2021})
  \bibinfo{pages}{107--115}, \bibinfo{note}{doi:
  \url{https://doi.org/10.1145/3446776}}.

\bibitem[{Fink et~al.(2020)Fink, Wang, Svensén, Dersin, Lee, and
  Ducoffe}]{olgareview}
\bibinfo{author}{O.~Fink}, \bibinfo{author}{Q.~Wang},
  \bibinfo{author}{M.~Svensén}, \bibinfo{author}{P.~Dersin},
  \bibinfo{author}{W.-J. Lee}, \bibinfo{author}{M.~Ducoffe},
  \bibinfo{title}{Potential, Challenges and Future Directions for Deep Learning
  in Prognostics and Health Management Applications}, \bibinfo{note}{doi:
  \url{https://doi.org/10.1016/j.engappai.2020.103678}}, \bibinfo{year}{2020}.

\bibitem[{Biggio and Kastanis(2020)}]{biggio2020prognostics}
\bibinfo{author}{L.~Biggio}, \bibinfo{author}{I.~Kastanis},
  \bibinfo{title}{Prognostics and health management of industrial assets:
  Current progress and road ahead}, \bibinfo{journal}{Frontiers in Artificial
  Intelligence} \bibinfo{volume}{3} (\bibinfo{year}{2020})
  \bibinfo{pages}{578613}, \bibinfo{note}{doi:
  \url{https://doi.org/10.3389/frai.2020.578613}}.

\bibitem[{Wang et~al.(2019)Wang, Lei, Li, and Yan}]{wang2019deep}
\bibinfo{author}{B.~Wang}, \bibinfo{author}{Y.~Lei}, \bibinfo{author}{N.~Li},
  \bibinfo{author}{T.~Yan}, \bibinfo{title}{Deep separable convolutional
  network for remaining useful life prediction of machinery},
  \bibinfo{journal}{Mechanical systems and signal processing}
  \bibinfo{volume}{134} (\bibinfo{year}{2019}) \bibinfo{pages}{106330},
  \bibinfo{note}{doi: \url{https://doi.org/10.1016/j.ymssp.2019.106330}}.

\bibitem[{Lee et~al.(2013{\natexlab{a}})Lee, Lapira, Bagheri, and
  Kao}]{lee2013recent}
\bibinfo{author}{J.~Lee}, \bibinfo{author}{E.~Lapira},
  \bibinfo{author}{B.~Bagheri}, \bibinfo{author}{H.-a. Kao},
  \bibinfo{title}{Recent advances and trends in predictive manufacturing
  systems in big data environment}, \bibinfo{journal}{Manufacturing letters}
  \bibinfo{volume}{1}~(\bibinfo{number}{1})
  (\bibinfo{year}{2013}{\natexlab{a}}) \bibinfo{pages}{38--41},
  \bibinfo{note}{doi: \url{https://doi.org/10.1016/j.mfglet.2013.09.005}}.

\bibitem[{Lee et~al.(2013{\natexlab{b}})Lee, Lapira, Yang, and
  Kao}]{lee2013predictive}
\bibinfo{author}{J.~Lee}, \bibinfo{author}{E.~Lapira},
  \bibinfo{author}{S.~Yang}, \bibinfo{author}{A.~Kao},
  \bibinfo{title}{Predictive manufacturing system-Trends of next-generation
  production systems}, \bibinfo{journal}{{IFAC} proceedings volumes}
  \bibinfo{volume}{46}~(\bibinfo{number}{7})
  (\bibinfo{year}{2013}{\natexlab{b}}) \bibinfo{pages}{150--156},
  \bibinfo{note}{doi: \url{https://doi.org/10.3182/20130522-3-BR-4036.00107}}.

\bibitem[{Saxena et~al.(2008)Saxena, Celaya, Balaban, Goebel, Saha, Saha, and
  Schwabacher}]{saxena2008metrics}
\bibinfo{author}{A.~Saxena}, \bibinfo{author}{J.~Celaya},
  \bibinfo{author}{E.~Balaban}, \bibinfo{author}{K.~Goebel},
  \bibinfo{author}{B.~Saha}, \bibinfo{author}{S.~Saha},
  \bibinfo{author}{M.~Schwabacher}, \bibinfo{title}{Metrics for evaluating
  performance of prognostic techniques}, in: \bibinfo{booktitle}{2008
  International Conference on Prognostics and Health Management},
  \bibinfo{organization}{IEEE}, \bibinfo{pages}{1--17}, \bibinfo{note}{doi:
  \url{https://doi.org/10.1109/PHM.2008.4711436}}, \bibinfo{year}{2008}.

\bibitem[{Biggio et~al.(2022)Biggio, Bendinelli, Kulkarni, and Fink}]{uqp13}
\bibinfo{author}{L.~Biggio}, \bibinfo{author}{T.~Bendinelli},
  \bibinfo{author}{C.~Kulkarni}, \bibinfo{author}{O.~Fink},
  \bibinfo{title}{Dynaformer: A Deep Learning Model for Ageing-aware Battery
  Discharge Prediction} \bibinfo{note}{{doi}:
  \url{https://doi.org/10.48550/arXiv.2206.02555}}.

\bibitem[{Daxberger et~al.(2021)Daxberger, Kristiadi, Immer, Eschenhagen,
  Bauer, and Hennig}]{bdl1}
\bibinfo{author}{E.~Daxberger}, \bibinfo{author}{A.~Kristiadi},
  \bibinfo{author}{A.~Immer}, \bibinfo{author}{R.~Eschenhagen},
  \bibinfo{author}{M.~Bauer}, \bibinfo{author}{P.~Hennig},
  \bibinfo{title}{Laplace redux-effortless {B}ayesian deep learning},
  \bibinfo{journal}{Advances in Neural Information Processing Systems}
  \bibinfo{volume}{34} (\bibinfo{year}{2021}) \bibinfo{pages}{20089--20103},
  \bibinfo{note}{doi: \url{https://doi.org/10.48550/arXiv.2106.14806}}.

\bibitem[{Wilson(2020)}]{bdl2}
\bibinfo{author}{A.~G. Wilson}, \bibinfo{title}{The Case for Bayesian Deep
  Learning}, \bibinfo{note}{doi:
  \url{https://doi.org/10.48550/arXiv.2001.10995}}, \bibinfo{year}{2020}.

\bibitem[{Jospin et~al.(2022)Jospin, Laga, Boussaid, Buntine, and
  Bennamoun}]{bdl4}
\bibinfo{author}{L.~V. Jospin}, \bibinfo{author}{H.~Laga},
  \bibinfo{author}{F.~Boussaid}, \bibinfo{author}{W.~Buntine},
  \bibinfo{author}{M.~Bennamoun}, \bibinfo{title}{Hands-On Bayesian Neural
  Networks{\textemdash}A Tutorial for Deep Learning Users},
  \bibinfo{journal}{{IEEE} Computational Intelligence Magazine}
  \bibinfo{volume}{17}~(\bibinfo{number}{2}) (\bibinfo{year}{2022})
  \bibinfo{pages}{29--48}, \bibinfo{note}{doi:
  \url{https://doi.org/10.1109/MCI.2022.3155327}}.

\bibitem[{Teye et~al.(2018)Teye, Azizpour, and Smith}]{bdl5}
\bibinfo{author}{M.~Teye}, \bibinfo{author}{H.~Azizpour},
  \bibinfo{author}{K.~Smith}, \bibinfo{title}{Bayesian Uncertainty Estimation
  for Batch Normalized Deep Networks}, \bibinfo{note}{doi:
  \url{https://doi.org/10.48550/arXiv.1802.06455}}, \bibinfo{year}{2018}.

\bibitem[{Ritter et~al.(2018)Ritter, Botev, and Barber}]{bdl7}
\bibinfo{author}{H.~Ritter}, \bibinfo{author}{A.~Botev},
  \bibinfo{author}{D.~Barber}, \bibinfo{title}{A Scalable Laplace Approximation
  for Neural Networks}, in: \bibinfo{booktitle}{International Conference on
  Learning Representations}, \bibinfo{year}{2018}.

\bibitem[{Wang et~al.(2020)Wang, Zhao, and Addepalli}]{WANG202081}
\bibinfo{author}{Y.~Wang}, \bibinfo{author}{Y.~Zhao},
  \bibinfo{author}{S.~Addepalli}, \bibinfo{title}{Remaining useful life
  prediction using deep learning approaches: A review},
  \bibinfo{journal}{Procedia Manufacturing} \bibinfo{volume}{49}
  (\bibinfo{year}{2020}) \bibinfo{pages}{81--88}, \bibinfo{note}{doi:
  \url{https://doi.org/10.1016/j.promfg.2020.06.015}}.

\bibitem[{Rokhforoz et~al.(2021)Rokhforoz, Gjorgiev, Sansavini, and
  Fink}]{rokhforoz2021multi}
\bibinfo{author}{P.~Rokhforoz}, \bibinfo{author}{B.~Gjorgiev},
  \bibinfo{author}{G.~Sansavini}, \bibinfo{author}{O.~Fink},
  \bibinfo{title}{Multi-agent maintenance scheduling based on the coordination
  between central operator and decentralized producers in an electricity
  market}, \bibinfo{journal}{Reliability Engineering \& System Safety}
  \bibinfo{volume}{210} (\bibinfo{year}{2021}) \bibinfo{pages}{107495},
  \bibinfo{note}{doi: \url{https://doi.org/10.1016/j.ress.2021.107495}}.

\bibitem[{Rokhforoz et~al.(2023)Rokhforoz, Montazeri, and
  Fink}]{rokhforoz2023safe}
\bibinfo{author}{P.~Rokhforoz}, \bibinfo{author}{M.~Montazeri},
  \bibinfo{author}{O.~Fink}, \bibinfo{title}{Safe multi-agent deep
  reinforcement learning for joint bidding and maintenance scheduling of
  generation units}, \bibinfo{journal}{Reliability Engineering \& System
  Safety} \bibinfo{volume}{232} (\bibinfo{year}{2023}) \bibinfo{pages}{109081},
  \bibinfo{note}{doi: \url{https://doi.org/10.48550/arXiv.2112.10459}}.

\bibitem[{Zio(2022)}]{Zio2022PrognosticsAH}
\bibinfo{author}{E.~Zio}, \bibinfo{title}{Prognostics and Health Management
  (PHM): Where are we and where do we (need to) go in theory and practice},
  \bibinfo{journal}{Reliability Engineering \& System Safety}
  \bibinfo{volume}{218} (\bibinfo{year}{2022}) \bibinfo{pages}{108119},
  \bibinfo{note}{doi: \url{https://doi.org/10.1016/j.ress.2021.108119}}.

\bibitem[{Saxena et~al.(2010)Saxena, Celaya, Saha, Saha, and
  Goebel}]{saxena2010metrics}
\bibinfo{author}{A.~Saxena}, \bibinfo{author}{J.~Celaya},
  \bibinfo{author}{B.~Saha}, \bibinfo{author}{S.~Saha},
  \bibinfo{author}{K.~Goebel}, \bibinfo{title}{Metrics for Offline Evaluation
  of Prognostic Performance}, \bibinfo{journal}{International Journal of
  Prognostics and Health Management} \bibinfo{volume}{1}~(\bibinfo{number}{1}),
  \bibinfo{note}{doi: \url{https://doi.org/10.36001/ijphm.2010.v1i1.1336}}.

\bibitem[{Louizos and Welling(2017)}]{mcdp2}
\bibinfo{author}{C.~Louizos}, \bibinfo{author}{M.~Welling},
  \bibinfo{title}{Multiplicative Normalizing Flows for Variational Bayesian
  Neural Networks}, \urlprefix\url{https://arxiv.org/abs/1703.01961},
  \bibinfo{year}{2017}.

\bibitem[{Folgoc et~al.(2021)Folgoc, Baltatzis, Desai, Devaraj, Ellis,
  Manzanera, Nair, Qiu, Schnabel, and Glocker}]{mcdp3}
\bibinfo{author}{L.~L. Folgoc}, \bibinfo{author}{V.~Baltatzis},
  \bibinfo{author}{S.~Desai}, \bibinfo{author}{A.~Devaraj},
  \bibinfo{author}{S.~Ellis}, \bibinfo{author}{O.~E.~M. Manzanera},
  \bibinfo{author}{A.~Nair}, \bibinfo{author}{H.~Qiu},
  \bibinfo{author}{J.~Schnabel}, \bibinfo{author}{B.~Glocker},
  \bibinfo{title}{Is MC Dropout Bayesian?}, \bibinfo{note}{doi:
  \url{https://doi.org/10.48550/arXiv.2110.04286}}, \bibinfo{year}{2021}.

\bibitem[{Severson et~al.(2019)Severson, Attia, Jin, Perkins, Jiang, Yang,
  Chen, Aykol, Herring, Fraggedakis et~al.}]{severson2019data}
\bibinfo{author}{K.~A. Severson}, \bibinfo{author}{P.~M. Attia},
  \bibinfo{author}{N.~Jin}, \bibinfo{author}{N.~Perkins},
  \bibinfo{author}{B.~Jiang}, \bibinfo{author}{Z.~Yang}, \bibinfo{author}{M.~H.
  Chen}, \bibinfo{author}{M.~Aykol}, \bibinfo{author}{P.~K. Herring},
  \bibinfo{author}{D.~Fraggedakis}, et~al., \bibinfo{title}{Data-driven
  prediction of battery cycle life before capacity degradation},
  \bibinfo{journal}{Nature Energy} \bibinfo{volume}{4}~(\bibinfo{number}{5})
  (\bibinfo{year}{2019}) \bibinfo{pages}{383--391}, \bibinfo{note}{doi:
  \url{https://doi.org/10.1038/s41560-019-0356-8}}.

\bibitem[{Attia et~al.(2020)Attia, Grover, Jin, Severson, Markov, Liao, Chen,
  Cheong, Perkins, Yang et~al.}]{attia2020closed}
\bibinfo{author}{P.~M. Attia}, \bibinfo{author}{A.~Grover},
  \bibinfo{author}{N.~Jin}, \bibinfo{author}{K.~A. Severson},
  \bibinfo{author}{T.~M. Markov}, \bibinfo{author}{Y.-H. Liao},
  \bibinfo{author}{M.~H. Chen}, \bibinfo{author}{B.~Cheong},
  \bibinfo{author}{N.~Perkins}, \bibinfo{author}{Z.~Yang}, et~al.,
  \bibinfo{title}{Closed-loop optimization of fast-charging protocols for
  batteries with machine learning}, \bibinfo{journal}{Nature}
  \bibinfo{volume}{578}~(\bibinfo{number}{7795}) (\bibinfo{year}{2020})
  \bibinfo{pages}{397--402}, \bibinfo{note}{doi:
  \url{https://doi.org/10.1038/s41586-020-1994-5}}.

\bibitem[{Arias~Chao et~al.(2021)Arias~Chao, Kulkarni, Goebel, and
  Fink}]{arias2021aircraft}
\bibinfo{author}{M.~Arias~Chao}, \bibinfo{author}{C.~Kulkarni},
  \bibinfo{author}{K.~Goebel}, \bibinfo{author}{O.~Fink},
  \bibinfo{title}{Aircraft engine run-to-failure dataset under real flight
  conditions for prognostics and diagnostics}, \bibinfo{journal}{Data}
  \bibinfo{volume}{6}~(\bibinfo{number}{1}) (\bibinfo{year}{2021})
  \bibinfo{pages}{5}, \bibinfo{note}{doi:
  \url{https://doi.org/10.3390/data6010005}}.

\bibitem[{Chao et~al.(2022{\natexlab{a}})Chao, Kulkarni, Goebel, and
  Fink}]{chao2022fusing}
\bibinfo{author}{M.~A. Chao}, \bibinfo{author}{C.~Kulkarni},
  \bibinfo{author}{K.~Goebel}, \bibinfo{author}{O.~Fink},
  \bibinfo{title}{Fusing physics-based and deep learning models for
  prognostics}, \bibinfo{journal}{Reliability Engineering \& System Safety}
  \bibinfo{volume}{217} (\bibinfo{year}{2022}{\natexlab{a}})
  \bibinfo{pages}{107961}, \bibinfo{note}{doi:
  \url{https://doi.org/10.1016/j.ress.2021.107961}}.

\bibitem[{Tian et~al.(2022)Tian, Chao, Kulkarni, Goebel, and
  Fink}]{tian2022real}
\bibinfo{author}{Y.~Tian}, \bibinfo{author}{M.~A. Chao},
  \bibinfo{author}{C.~Kulkarni}, \bibinfo{author}{K.~Goebel},
  \bibinfo{author}{O.~Fink}, \bibinfo{title}{Real-time model calibration with
  deep reinforcement learning}, \bibinfo{journal}{Mechanical Systems and Signal
  Processing} \bibinfo{volume}{165} (\bibinfo{year}{2022})
  \bibinfo{pages}{108284}, \bibinfo{note}{doi:
  \url{https://doi.org/10.1016/j.ymssp.2021.108284}}.

\bibitem[{Song et~al.(2022)Song, Liu, Wu, Jin, and
  Jiang}]{song2022hierarchical}
\bibinfo{author}{T.~Song}, \bibinfo{author}{C.~Liu}, \bibinfo{author}{R.~Wu},
  \bibinfo{author}{Y.~Jin}, \bibinfo{author}{D.~Jiang}, \bibinfo{title}{A
  hierarchical scheme for remaining useful life prediction with long short-term
  memory networks}, \bibinfo{journal}{Neurocomputing} \bibinfo{volume}{487}
  (\bibinfo{year}{2022}) \bibinfo{pages}{22--33}, \bibinfo{note}{doi:
  \url{https://doi.org/10.1016/j.neucom.2022.02.032}}.

\bibitem[{Mo and Iacca(2022)}]{mo2022multi}
\bibinfo{author}{H.~Mo}, \bibinfo{author}{G.~Iacca},
  \bibinfo{title}{Multi-Objective Optimization of Extreme Learning Machine for
  Remaining Useful Life Prediction}, in: \bibinfo{booktitle}{International
  Conference on the Applications of Evolutionary Computation (Part of
  EvoStar)}, \bibinfo{organization}{Springer}, \bibinfo{pages}{191--206},
  \bibinfo{note}{doi: \url{https://doi.org/10.1007/978-3-031-02462-7_13}},
  \bibinfo{year}{2022}.

\bibitem[{Chao et~al.(2022{\natexlab{b}})Chao, Kulkarni, Goebel, and
  Fink}]{chao}
\bibinfo{author}{M.~A. Chao}, \bibinfo{author}{C.~Kulkarni},
  \bibinfo{author}{K.~Goebel}, \bibinfo{author}{O.~Fink},
  \bibinfo{title}{Fusing physics-based and deep learning models for
  prognostics}, \bibinfo{journal}{Reliability Engineering \& System Safety}
  \bibinfo{volume}{217} (\bibinfo{year}{2022}{\natexlab{b}})
  \bibinfo{pages}{107961}, \bibinfo{note}{doi:
  \url{https://doi.org/10.1016/j.ress.2021.107961}}.

\bibitem[{Lagaris et~al.(1998)Lagaris, Likas, and
  Fotiadis}]{lagaris1998artificial}
\bibinfo{author}{I.~E. Lagaris}, \bibinfo{author}{A.~Likas},
  \bibinfo{author}{D.~I. Fotiadis}, \bibinfo{title}{Artificial neural networks
  for solving ordinary and partial differential equations},
  \bibinfo{journal}{IEEE transactions on neural networks}
  \bibinfo{volume}{9}~(\bibinfo{number}{5}) (\bibinfo{year}{1998})
  \bibinfo{pages}{987--1000}.

\bibitem[{Cursi and Koscianski(2007)}]{cursi2007physically}
\bibinfo{author}{J.~Cursi}, \bibinfo{author}{A.~Koscianski},
  \bibinfo{title}{Physically constrained neural network models for simulation},
  in: \bibinfo{booktitle}{Advances and Innovations in Systems, Computing
  Sciences and Software Engineering}, \bibinfo{publisher}{Springer},
  \bibinfo{pages}{567--572}, \bibinfo{year}{2007}.

\bibitem[{Raissi et~al.(2019)Raissi, Perdikaris, and
  Karniadakis}]{raissi2019physics}
\bibinfo{author}{M.~Raissi}, \bibinfo{author}{P.~Perdikaris},
  \bibinfo{author}{G.~E. Karniadakis}, \bibinfo{title}{Physics-informed neural
  networks: A deep learning framework for solving forward and inverse problems
  involving nonlinear partial differential equations},
  \bibinfo{journal}{Journal of Computational Physics} \bibinfo{volume}{378}
  (\bibinfo{year}{2019}) \bibinfo{pages}{686--707}, \bibinfo{note}{doi:
  \url{https://doi.org/10.1016/j.jcp.2018.10.045}}.

\bibitem[{Ritto and Rochinha(2021)}]{ritto2021digital}
\bibinfo{author}{T.~Ritto}, \bibinfo{author}{F.~Rochinha},
  \bibinfo{title}{Digital twin, physics-based model, and machine learning
  applied to damage detection in structures}, \bibinfo{journal}{Mechanical
  Systems and Signal Processing} \bibinfo{volume}{155} (\bibinfo{year}{2021})
  \bibinfo{pages}{107614}, \bibinfo{note}{doi:
  \url{https://doi.org/10.1016/j.ymssp.2021.107614}}.

\bibitem[{Oviedo et~al.(2019)Oviedo, Ren, Sun, Settens, Liu, Hartono, Ramasamy,
  DeCost, Tian, Romano et~al.}]{oviedo2019fast}
\bibinfo{author}{F.~Oviedo}, \bibinfo{author}{Z.~Ren},
  \bibinfo{author}{S.~Sun}, \bibinfo{author}{C.~Settens},
  \bibinfo{author}{Z.~Liu}, \bibinfo{author}{N.~T.~P. Hartono},
  \bibinfo{author}{S.~Ramasamy}, \bibinfo{author}{B.~L. DeCost},
  \bibinfo{author}{S.~I. Tian}, \bibinfo{author}{G.~Romano}, et~al.,
  \bibinfo{title}{Fast and interpretable classification of small X-ray
  diffraction datasets using data augmentation and deep neural networks},
  \bibinfo{journal}{npj Computational Materials}
  \bibinfo{volume}{5}~(\bibinfo{number}{1}) (\bibinfo{year}{2019})
  \bibinfo{pages}{60}.

\bibitem[{Kapusuzoglu and Mahadevan(2020)}]{kapusuzoglu2020physics}
\bibinfo{author}{B.~Kapusuzoglu}, \bibinfo{author}{S.~Mahadevan},
  \bibinfo{title}{Physics-informed and hybrid machine learning in additive
  manufacturing: application to fused filament fabrication},
  \bibinfo{journal}{Jom} \bibinfo{volume}{72}~(\bibinfo{number}{12})
  (\bibinfo{year}{2020}) \bibinfo{pages}{4695--4705}, \bibinfo{note}{doi:
  \url{https://doi.org/10.1007/s11837-020-04438-4}}.

\bibitem[{Yucesan and Viana(2020)}]{yucesan2020physics}
\bibinfo{author}{Y.~A. Yucesan}, \bibinfo{author}{F.~A. Viana},
  \bibinfo{title}{A physics-informed neural network for wind turbine main
  bearing fatigue}, \bibinfo{journal}{International Journal of Prognostics and
  Health Management} \bibinfo{volume}{11}~(\bibinfo{number}{1}),
  \bibinfo{note}{doi: \url{https://doi.org/10.36001/ijphm.2020.v11i1.2594}}.

\bibitem[{Jiang et~al.(2022)Jiang, Vega, Todd, and Hu}]{jiang2022model}
\bibinfo{author}{C.~Jiang}, \bibinfo{author}{M.~A. Vega},
  \bibinfo{author}{M.~D. Todd}, \bibinfo{author}{Z.~Hu}, \bibinfo{title}{Model
  correction and updating of a stochastic degradation model for failure
  prognostics of miter gates}, \bibinfo{journal}{Reliability Engineering \&
  System Safety} \bibinfo{volume}{218} (\bibinfo{year}{2022})
  \bibinfo{pages}{108203}, \bibinfo{note}{doi:
  \url{https://doi.org/10.1016/j.ress.2021.108203}}.

\bibitem[{Thompson and Kramer(1994)}]{thompson1994modeling}
\bibinfo{author}{M.~L. Thompson}, \bibinfo{author}{M.~A. Kramer},
  \bibinfo{title}{Modeling chemical processes using prior knowledge and neural
  networks}, \bibinfo{journal}{AIChE Journal}
  \bibinfo{volume}{40}~(\bibinfo{number}{8}) (\bibinfo{year}{1994})
  \bibinfo{pages}{1328--1340}, \bibinfo{note}{doi:
  \url{https://doi.org/10.1002/aic.690400806}}.

\bibitem[{Wang et~al.(2017)Wang, Wu, and Xiao}]{wang2017physics}
\bibinfo{author}{J.-X. Wang}, \bibinfo{author}{J.-L. Wu},
  \bibinfo{author}{H.~Xiao}, \bibinfo{title}{Physics-informed machine learning
  approach for reconstructing Reynolds stress modeling discrepancies based on
  DNS data}, \bibinfo{journal}{Physical Review Fluids}
  \bibinfo{volume}{2}~(\bibinfo{number}{3}) (\bibinfo{year}{2017})
  \bibinfo{pages}{034603}, \bibinfo{note}{doi:
  \url{https://doi.org/10.1103/PhysRevFluids.2.034603}}.

\bibitem[{Thelen et~al.(2022{\natexlab{c}})Thelen, Lui, Shen, Laflamme, Hu, Ye,
  and Hu}]{thelen2022integrating}
\bibinfo{author}{A.~Thelen}, \bibinfo{author}{Y.~H. Lui},
  \bibinfo{author}{S.~Shen}, \bibinfo{author}{S.~Laflamme},
  \bibinfo{author}{S.~Hu}, \bibinfo{author}{H.~Ye}, \bibinfo{author}{C.~Hu},
  \bibinfo{title}{Integrating physics-based modeling and machine learning for
  degradation diagnostics of lithium-ion batteries}, \bibinfo{journal}{Energy
  Storage Materials} \bibinfo{volume}{50} (\bibinfo{year}{2022}{\natexlab{c}})
  \bibinfo{pages}{668--695}, \bibinfo{note}{doi:
  \url{https://doi.org/10.1016/j.ensm.2022.05.047}}.

\bibitem[{Azzi et~al.(2023)Azzi, Ghnatios, Avery, and
  Farhat}]{azzi2023acceleration}
\bibinfo{author}{M.-J. Azzi}, \bibinfo{author}{C.~Ghnatios},
  \bibinfo{author}{P.~Avery}, \bibinfo{author}{C.~Farhat},
  \bibinfo{title}{Acceleration of a Physics-Based Machine Learning Approach for
  Modeling and Quantifying Model-Form Uncertainties and Performing Model
  Updating}, \bibinfo{journal}{Journal of Computing and Information Science in
  Engineering} \bibinfo{volume}{23}~(\bibinfo{number}{1})
  (\bibinfo{year}{2023}) \bibinfo{pages}{011009}, \bibinfo{note}{doi:
  \url{https://doi.org/10.1115/1.4055546}}.

\bibitem[{Chen et~al.(2021)Chen, Wang, Hesthaven, and Zhang}]{chen2021physics}
\bibinfo{author}{W.~Chen}, \bibinfo{author}{Q.~Wang}, \bibinfo{author}{J.~S.
  Hesthaven}, \bibinfo{author}{C.~Zhang}, \bibinfo{title}{Physics-informed
  machine learning for reduced-order modeling of nonlinear problems},
  \bibinfo{journal}{Journal of Computational Physics} \bibinfo{volume}{446}
  (\bibinfo{year}{2021}) \bibinfo{pages}{110666}, \bibinfo{note}{doi:
  \url{https://doi.org/10.1016/j.jcp.2021.110666}}.

\bibitem[{Gong et~al.(2022)Gong, Cheng, Chen, and Li}]{gong2022data}
\bibinfo{author}{H.~Gong}, \bibinfo{author}{S.~Cheng},
  \bibinfo{author}{Z.~Chen}, \bibinfo{author}{Q.~Li},
  \bibinfo{title}{Data-enabled physics-informed machine learning for
  reduced-order modeling digital twin: application to nuclear reactor physics},
  \bibinfo{journal}{Nuclear Science and Engineering}
  \bibinfo{volume}{196}~(\bibinfo{number}{6}) (\bibinfo{year}{2022})
  \bibinfo{pages}{668--693}, \bibinfo{note}{doi:
  \url{https://doi.org/10.1080/00295639.2021.2014752}}.

\bibitem[{Yucesan and Viana(2022)}]{yucesan2022hybrid}
\bibinfo{author}{Y.~A. Yucesan}, \bibinfo{author}{F.~A. Viana},
  \bibinfo{title}{A hybrid physics-informed neural network for main bearing
  fatigue prognosis under grease quality variation},
  \bibinfo{journal}{Mechanical Systems and Signal Processing}
  \bibinfo{volume}{171} (\bibinfo{year}{2022}) \bibinfo{pages}{108875},
  \bibinfo{note}{doi: \url{https://doi.org/10.1016/j.ymssp.2022.108875}}.

\bibitem[{Ramadesigan et~al.(2011)Ramadesigan, Chen, Burns, Boovaragavan,
  Braatz, and Subramanian}]{ramadesigan2011parameter}
\bibinfo{author}{V.~Ramadesigan}, \bibinfo{author}{K.~Chen},
  \bibinfo{author}{N.~A. Burns}, \bibinfo{author}{V.~Boovaragavan},
  \bibinfo{author}{R.~D. Braatz}, \bibinfo{author}{V.~R. Subramanian},
  \bibinfo{title}{Parameter estimation and capacity fade analysis of
  lithium-ion batteries using reformulated models}, \bibinfo{journal}{Journal
  of the Electrochemical Society} \bibinfo{volume}{158}~(\bibinfo{number}{9})
  (\bibinfo{year}{2011}) \bibinfo{pages}{A1048}, \bibinfo{note}{doi:
  \url{https://doi.org/10.1149/1.3609926}}.

\bibitem[{Downey et~al.(2019)Downey, Lui, Hu, Laflamme, and
  Hu}]{downey2019physics}
\bibinfo{author}{A.~Downey}, \bibinfo{author}{Y.-H. Lui},
  \bibinfo{author}{C.~Hu}, \bibinfo{author}{S.~Laflamme},
  \bibinfo{author}{S.~Hu}, \bibinfo{title}{Physics-based prognostics of
  lithium-ion battery using non-linear least squares with dynamic bounds},
  \bibinfo{journal}{Reliability Engineering \& System Safety}
  \bibinfo{volume}{182} (\bibinfo{year}{2019}) \bibinfo{pages}{1--12},
  \bibinfo{note}{doi: \url{https://doi.org/10.1016/j.ress.2018.09.018}}.

\bibitem[{Lui et~al.(2021)Lui, Li, Downey, Shen, Nemani, Ye, VanElzen, Jain,
  Hu, Laflamme et~al.}]{lui2021physics}
\bibinfo{author}{Y.~H. Lui}, \bibinfo{author}{M.~Li},
  \bibinfo{author}{A.~Downey}, \bibinfo{author}{S.~Shen},
  \bibinfo{author}{V.~P. Nemani}, \bibinfo{author}{H.~Ye},
  \bibinfo{author}{C.~VanElzen}, \bibinfo{author}{G.~Jain},
  \bibinfo{author}{S.~Hu}, \bibinfo{author}{S.~Laflamme}, et~al.,
  \bibinfo{title}{Physics-based prognostics of implantable-grade lithium-ion
  battery for remaining useful life prediction}, \bibinfo{journal}{Journal of
  Power Sources} \bibinfo{volume}{485} (\bibinfo{year}{2021})
  \bibinfo{pages}{229327}, \bibinfo{note}{doi:
  \url{https://doi.org/10.1016/j.jpowsour.2020.229327}}.

\bibitem[{Ramuhalli et~al.(2005)Ramuhalli, Udpa, and
  Udpa}]{ramuhalli2005finite}
\bibinfo{author}{P.~Ramuhalli}, \bibinfo{author}{L.~Udpa},
  \bibinfo{author}{S.~S. Udpa}, \bibinfo{title}{Finite-element neural networks
  for solving differential equations}, \bibinfo{journal}{IEEE Transactions on
  Neural Networks} \bibinfo{volume}{16}~(\bibinfo{number}{6})
  (\bibinfo{year}{2005}) \bibinfo{pages}{1381--1392}, \bibinfo{note}{doi:
  \url{https://doi.org/10.1109/TNN.2005.857945}}.

\bibitem[{Darbon and Meng(2021)}]{darbon2021some}
\bibinfo{author}{J.~Darbon}, \bibinfo{author}{T.~Meng}, \bibinfo{title}{On some
  neural network architectures that can represent viscosity solutions of
  certain high dimensional Hamilton--Jacobi partial differential equations},
  \bibinfo{journal}{Journal of Computational Physics} \bibinfo{volume}{425}
  (\bibinfo{year}{2021}) \bibinfo{pages}{109907}, \bibinfo{note}{doi:
  \url{https://doi.org/10.1016/j.jcp.2020.109907}}.

\bibitem[{Lu et~al.(2021)Lu, Jin, Pang, Zhang, and
  Karniadakis}]{lu2021learning}
\bibinfo{author}{L.~Lu}, \bibinfo{author}{P.~Jin}, \bibinfo{author}{G.~Pang},
  \bibinfo{author}{Z.~Zhang}, \bibinfo{author}{G.~E. Karniadakis},
  \bibinfo{title}{Learning nonlinear operators via DeepONet based on the
  universal approximation theorem of operators}, \bibinfo{journal}{Nature
  Machine Intelligence} \bibinfo{volume}{3}~(\bibinfo{number}{3})
  (\bibinfo{year}{2021}) \bibinfo{pages}{218--229}, \bibinfo{note}{doi:
  \url{https://doi.org/10.1038/s42256-021-00302-5}}.

\bibitem[{Li et~al.(2020{\natexlab{a}})Li, Kovachki, Azizzadenesheli, Liu,
  Bhattacharya, Stuart, and Anandkumar}]{li2020fourier}
\bibinfo{author}{Z.~Li}, \bibinfo{author}{N.~Kovachki},
  \bibinfo{author}{K.~Azizzadenesheli}, \bibinfo{author}{B.~Liu},
  \bibinfo{author}{K.~Bhattacharya}, \bibinfo{author}{A.~Stuart},
  \bibinfo{author}{A.~Anandkumar}, \bibinfo{title}{Fourier neural operator for
  parametric partial differential equations}, \bibinfo{journal}{arXiv preprint
  arXiv:2010.08895} .

\bibitem[{Cai et~al.(2021)Cai, Mao, Wang, Yin, and
  Karniadakis}]{cai2021physics}
\bibinfo{author}{S.~Cai}, \bibinfo{author}{Z.~Mao}, \bibinfo{author}{Z.~Wang},
  \bibinfo{author}{M.~Yin}, \bibinfo{author}{G.~E. Karniadakis},
  \bibinfo{title}{Physics-informed neural networks (PINNs) for fluid mechanics:
  A review}, \bibinfo{journal}{Acta Mechanica Sinica}
  \bibinfo{volume}{37}~(\bibinfo{number}{12}) (\bibinfo{year}{2021})
  \bibinfo{pages}{1727--1738}.

\bibitem[{Zhu et~al.(2019)Zhu, Zabaras, Koutsourelakis, and
  Perdikaris}]{zhu2019physics}
\bibinfo{author}{Y.~Zhu}, \bibinfo{author}{N.~Zabaras}, \bibinfo{author}{P.-S.
  Koutsourelakis}, \bibinfo{author}{P.~Perdikaris},
  \bibinfo{title}{Physics-constrained deep learning for high-dimensional
  surrogate modeling and uncertainty quantification without labeled data},
  \bibinfo{journal}{Journal of Computational Physics} \bibinfo{volume}{394}
  (\bibinfo{year}{2019}) \bibinfo{pages}{56--81}, \bibinfo{note}{doi:
  \url{https://doi.org/10.1016/j.jcp.2019.05.024}}.

\bibitem[{Yang et~al.(2020)Yang, Zhang, and Karniadakis}]{yang2020physics}
\bibinfo{author}{L.~Yang}, \bibinfo{author}{D.~Zhang}, \bibinfo{author}{G.~E.
  Karniadakis}, \bibinfo{title}{Physics-informed generative adversarial
  networks for stochastic differential equations}, \bibinfo{journal}{SIAM
  Journal on Scientific Computing} \bibinfo{volume}{42}~(\bibinfo{number}{1})
  (\bibinfo{year}{2020}) \bibinfo{pages}{A292--A317}, \bibinfo{note}{doi:
  \url{https://doi.org/10.1137/18M1225409}}.

\bibitem[{Sun and Wang(2020)}]{sun2020physics}
\bibinfo{author}{L.~Sun}, \bibinfo{author}{J.-X. Wang},
  \bibinfo{title}{Physics-constrained bayesian neural network for fluid flow
  reconstruction with sparse and noisy data}, \bibinfo{journal}{Theoretical and
  Applied Mechanics Letters} \bibinfo{volume}{10}~(\bibinfo{number}{3})
  (\bibinfo{year}{2020}) \bibinfo{pages}{161--169}, \bibinfo{note}{doi:
  \url{https://doi.org/10.1016/j.taml.2020.01.031}}.

\bibitem[{Cuomo et~al.(2022)Cuomo, Di~Cola, Giampaolo, Rozza, Raissi, and
  Piccialli}]{cuomo2022scientific}
\bibinfo{author}{S.~Cuomo}, \bibinfo{author}{V.~S. Di~Cola},
  \bibinfo{author}{F.~Giampaolo}, \bibinfo{author}{G.~Rozza},
  \bibinfo{author}{M.~Raissi}, \bibinfo{author}{F.~Piccialli},
  \bibinfo{title}{Scientific machine learning through physics--informed neural
  networks: Where we are and what’s next}, \bibinfo{journal}{Journal of
  Scientific Computing} \bibinfo{volume}{92}~(\bibinfo{number}{3})
  (\bibinfo{year}{2022}) \bibinfo{pages}{88}, \bibinfo{note}{doi:
  \url{https://doi.org/10.1007/s10915-022-01939-z}}.

\bibitem[{Soize et~al.(2019{\natexlab{a}})Soize, Ghanem, Safta, Huan, Vane,
  Oefelein, Lacaze, Najm, Tang, and Chen}]{Soize2019}
\bibinfo{author}{C.~Soize}, \bibinfo{author}{R.~G. Ghanem},
  \bibinfo{author}{C.~Safta}, \bibinfo{author}{X.~Huan}, \bibinfo{author}{Z.~P.
  Vane}, \bibinfo{author}{J.~C. Oefelein}, \bibinfo{author}{G.~Lacaze},
  \bibinfo{author}{H.~N. Najm}, \bibinfo{author}{Q.~Tang},
  \bibinfo{author}{X.~Chen}, \bibinfo{title}{{Entropy-based closure for
  probabilistic learning on manifolds}}, \bibinfo{journal}{Journal of
  Computational Physics} \bibinfo{volume}{388}~(\bibinfo{number}{1})
  (\bibinfo{year}{2019}{\natexlab{a}}) \bibinfo{pages}{518--533},
  \bibinfo{note}{doi: \url{https://doi.org/10.1016/j.jcp.2018.12.029}}.

\bibitem[{Soize and Ghanem(2021)}]{Soize2021}
\bibinfo{author}{C.~Soize}, \bibinfo{author}{R.~Ghanem},
  \bibinfo{title}{{Probabilistic learning on manifolds constrained by nonlinear
  partial differential equations for small datasets}},
  \bibinfo{journal}{Computer Methods in Applied Mechanics and Engineering}
  \bibinfo{volume}{380} (\bibinfo{year}{2021}) \bibinfo{pages}{113777},
  \bibinfo{note}{doi: \url{https://doi.org/10.1016/j.cma.2021.113777}}.

\bibitem[{Soize et~al.(2019{\natexlab{b}})Soize, Ghanem, Safta, Huan, Vane,
  Oefelein, Lacaze, and Najm}]{Soize2019a}
\bibinfo{author}{C.~Soize}, \bibinfo{author}{R.~Ghanem},
  \bibinfo{author}{C.~Safta}, \bibinfo{author}{X.~Huan}, \bibinfo{author}{Z.~P.
  Vane}, \bibinfo{author}{J.~C. Oefelein}, \bibinfo{author}{G.~Lacaze},
  \bibinfo{author}{H.~N. Najm}, \bibinfo{title}{{Enhancing Model Predictability
  for a Scramjet Using Probabilistic Learning on Manifolds}},
  \bibinfo{journal}{AIAA Journal} \bibinfo{volume}{57}~(\bibinfo{number}{1})
  (\bibinfo{year}{2019}{\natexlab{b}}) \bibinfo{pages}{365--378},
  \bibinfo{note}{doi: \url{https://doi.org/10.2514/1.J057069}}.

\bibitem[{Ghanem et~al.(2019)Ghanem, Soize, Safta, Huan, Lacaze, Oefelein, and
  Najm}]{Ghanem2019}
\bibinfo{author}{R.~G. Ghanem}, \bibinfo{author}{C.~Soize},
  \bibinfo{author}{C.~Safta}, \bibinfo{author}{X.~Huan},
  \bibinfo{author}{G.~Lacaze}, \bibinfo{author}{J.~C. Oefelein},
  \bibinfo{author}{H.~N. Najm}, \bibinfo{title}{{Design optimization of a
  scramjet under uncertainty using probabilistic learning on manifolds}},
  \bibinfo{journal}{Journal of Computational Physics} \bibinfo{volume}{399}
  (\bibinfo{year}{2019}) \bibinfo{pages}{108930}, \bibinfo{note}{doi:
  \url{https://doi.org/10.1016/j.jcp.2019.108930}}.

\bibitem[{Ghanem et~al.(2022)Ghanem, Soize, Mehrez, and Aitharaju}]{Ghanem2022}
\bibinfo{author}{R.~Ghanem}, \bibinfo{author}{C.~Soize},
  \bibinfo{author}{L.~Mehrez}, \bibinfo{author}{V.~Aitharaju},
  \bibinfo{title}{{Probabilistic learning and updating of a digital twin for
  composite material systems}}, \bibinfo{journal}{International Journal for
  Numerical Methods in Engineering}
  \bibinfo{volume}{123}~(\bibinfo{number}{13}) (\bibinfo{year}{2022})
  \bibinfo{pages}{3004--3020}, \bibinfo{note}{doi:
  \url{https://doi.org/10.1002/nme.6430}}.

\bibitem[{Thelen et~al.(2023{\natexlab{b}})Thelen, Zhang, Fink, Lu, Ghosh,
  Youn, Todd, Mahadevan, Hu, and Hu}]{thelen2023comprehensive}
\bibinfo{author}{A.~Thelen}, \bibinfo{author}{X.~Zhang},
  \bibinfo{author}{O.~Fink}, \bibinfo{author}{Y.~Lu},
  \bibinfo{author}{S.~Ghosh}, \bibinfo{author}{B.~D. Youn},
  \bibinfo{author}{M.~D. Todd}, \bibinfo{author}{S.~Mahadevan},
  \bibinfo{author}{C.~Hu}, \bibinfo{author}{Z.~Hu}, \bibinfo{title}{A
  comprehensive review of digital twin—part 2: roles of uncertainty
  quantification and optimization, a battery digital twin, and perspectives},
  \bibinfo{journal}{Structural and Multidisciplinary Optimization}
  \bibinfo{volume}{66}~(\bibinfo{number}{1})
  (\bibinfo{year}{2023}{\natexlab{b}}) \bibinfo{pages}{1}, \bibinfo{note}{doi:
  \url{https://doi.org/10.1007/s00158-022-03476-7}}.

\bibitem[{Angelis et~al.(2023)Angelis, Sofos, and
  Karakasidis}]{angelis2023artificial}
\bibinfo{author}{D.~Angelis}, \bibinfo{author}{F.~Sofos},
  \bibinfo{author}{T.~E. Karakasidis}, \bibinfo{title}{Artificial Intelligence
  in Physical Sciences: Symbolic Regression Trends and Perspectives},
  \bibinfo{journal}{Archives of Computational Methods in Engineering}
  (\bibinfo{year}{2023}) \bibinfo{pages}{1--21}\bibinfo{note}{Doi:
  \url{https://doi.org/10.1007/s11831-023-09922-z}}.

\bibitem[{Brunton et~al.(2016)Brunton, Proctor, and
  Kutz}]{brunton2016discovering}
\bibinfo{author}{S.~L. Brunton}, \bibinfo{author}{J.~L. Proctor},
  \bibinfo{author}{J.~N. Kutz}, \bibinfo{title}{Discovering governing equations
  from data by sparse identification of nonlinear dynamical systems},
  \bibinfo{journal}{Proceedings of the National Academy of Sciences}
  \bibinfo{volume}{113}~(\bibinfo{number}{15}) (\bibinfo{year}{2016})
  \bibinfo{pages}{3932--3937}, \bibinfo{note}{doi:
  \url{https://doi.org/10.1073/pnas.1517384113}}.

\bibitem[{Rudy et~al.(2017)Rudy, Brunton, Proctor, and Kutz}]{rudy2017data}
\bibinfo{author}{S.~H. Rudy}, \bibinfo{author}{S.~L. Brunton},
  \bibinfo{author}{J.~L. Proctor}, \bibinfo{author}{J.~N. Kutz},
  \bibinfo{title}{Data-driven discovery of partial differential equations},
  \bibinfo{journal}{Science Advances} \bibinfo{volume}{3}~(\bibinfo{number}{4})
  (\bibinfo{year}{2017}) \bibinfo{pages}{e1602614}, \bibinfo{note}{doi:
  \url{https://doi.org/10.1098/10.1126/sciadv.1602614}}.

\bibitem[{Kaheman et~al.(2020)Kaheman, Kutz, and Brunton}]{kaheman2020sindy}
\bibinfo{author}{K.~Kaheman}, \bibinfo{author}{J.~N. Kutz},
  \bibinfo{author}{S.~L. Brunton}, \bibinfo{title}{SINDy-PI: a robust algorithm
  for parallel implicit sparse identification of nonlinear dynamics},
  \bibinfo{journal}{Proceedings of the Royal Society A}
  \bibinfo{volume}{476}~(\bibinfo{number}{2242}) (\bibinfo{year}{2020})
  \bibinfo{pages}{20200279}, \bibinfo{note}{doi:
  \url{https://doi.org/10.1098/rspa.2020.0279}}.

\bibitem[{Hirsh et~al.(2022)Hirsh, Barajas-Solano, and
  Kutz}]{hirsh2022sparsifying}
\bibinfo{author}{S.~M. Hirsh}, \bibinfo{author}{D.~A. Barajas-Solano},
  \bibinfo{author}{J.~N. Kutz}, \bibinfo{title}{Sparsifying priors for Bayesian
  uncertainty quantification in model discovery}, \bibinfo{journal}{Royal
  Society Open Science} \bibinfo{volume}{9}~(\bibinfo{number}{2})
  (\bibinfo{year}{2022}) \bibinfo{pages}{211823}, \bibinfo{note}{doi:
  \url{https://doi.org/10.1098/rsos.211823}}.

\bibitem[{Mangan et~al.(2019)Mangan, Askham, Brunton, Kutz, and
  Proctor}]{mangan2019model}
\bibinfo{author}{N.~M. Mangan}, \bibinfo{author}{T.~Askham},
  \bibinfo{author}{S.~L. Brunton}, \bibinfo{author}{J.~N. Kutz},
  \bibinfo{author}{J.~L. Proctor}, \bibinfo{title}{Model selection for hybrid
  dynamical systems via sparse regression}, \bibinfo{journal}{Proceedings of
  the Royal Society A} \bibinfo{volume}{475}~(\bibinfo{number}{2223})
  (\bibinfo{year}{2019}) \bibinfo{pages}{20180534}, \bibinfo{note}{doi:
  \url{https://doi.org/10.1098/rspa.2018.0534}}.

\bibitem[{Wiener(1938)}]{wiener1938homogeneous}
\bibinfo{author}{N.~Wiener}, \bibinfo{title}{The homogeneous chaos},
  \bibinfo{journal}{American Journal of Mathematics}
  \bibinfo{volume}{60}~(\bibinfo{number}{4}) (\bibinfo{year}{1938})
  \bibinfo{pages}{897--936}, \bibinfo{note}{doi:
  \url{https://doi.org/10.2307/2371268}}.

\bibitem[{Ghanem and Spanos(2003)}]{ghanem2003stochastic}
\bibinfo{author}{R.~G. Ghanem}, \bibinfo{author}{P.~D. Spanos},
  \bibinfo{title}{Stochastic finite elements: a spectral approach},
  \bibinfo{publisher}{Courier Corporation}, \bibinfo{year}{2003}.

\bibitem[{Xiu and Karniadakis(2002)}]{xiu2002wiener}
\bibinfo{author}{D.~Xiu}, \bibinfo{author}{G.~E. Karniadakis},
  \bibinfo{title}{The Wiener--Askey polynomial chaos for stochastic
  differential equations}, \bibinfo{journal}{SIAM Journal on Scientific
  Computing} \bibinfo{volume}{24}~(\bibinfo{number}{2}) (\bibinfo{year}{2002})
  \bibinfo{pages}{619--644}, \bibinfo{note}{doi:
  \url{https://doi.org/10.1137/S1064827501387826}}.

\bibitem[{Le~Ma{\i}tre et~al.(2002)Le~Ma{\i}tre, Reagan, Najm, Ghanem, and
  Knio}]{le2002stochastic}
\bibinfo{author}{O.~P. Le~Ma{\i}tre}, \bibinfo{author}{M.~T. Reagan},
  \bibinfo{author}{H.~N. Najm}, \bibinfo{author}{R.~G. Ghanem},
  \bibinfo{author}{O.~M. Knio}, \bibinfo{title}{A stochastic projection method
  for fluid flow: II. Random process}, \bibinfo{journal}{Journal of
  Computational Physics} \bibinfo{volume}{181}~(\bibinfo{number}{1})
  (\bibinfo{year}{2002}) \bibinfo{pages}{9--44}, \bibinfo{note}{doi:
  \url{https://doi.org/10.1006/jcph.2002.7104}}.

\bibitem[{Berveiller et~al.(2006)Berveiller, Sudret, and
  Lemaire}]{berveiller2006stochastic}
\bibinfo{author}{M.~Berveiller}, \bibinfo{author}{B.~Sudret},
  \bibinfo{author}{M.~Lemaire}, \bibinfo{title}{Stochastic finite element: a
  non intrusive approach by regression}, \bibinfo{journal}{European Journal of
  Computational Mechanics/Revue Europ{\'e}enne de M{\'e}canique Num{\'e}rique}
  \bibinfo{volume}{15}~(\bibinfo{number}{1-3}) (\bibinfo{year}{2006})
  \bibinfo{pages}{81--92}, \bibinfo{note}{doi:
  \url{https://doi.org/10.3166/remn.15.81-92}}.

\bibitem[{Smolyak(1963)}]{Smolyak1963}
\bibinfo{author}{S.~Smolyak}, \bibinfo{title}{{Quadrature and interpolation
  formulas for tensor products of certain classes of functions}},
  \bibinfo{journal}{Dokl. Akad. Nauk SSSR}
  \bibinfo{volume}{148}~(\bibinfo{number}{5}) (\bibinfo{year}{1963})
  \bibinfo{pages}{1042--1045}.

\bibitem[{Constantine et~al.(2012)Constantine, Eldred, and
  Phipps}]{Constantine2012}
\bibinfo{author}{P.~G. Constantine}, \bibinfo{author}{M.~S. Eldred},
  \bibinfo{author}{E.~T. Phipps}, \bibinfo{title}{{Sparse pseudospectral
  approximation method}}, \bibinfo{journal}{Computer Methods in Applied
  Mechanics and Engineering} \bibinfo{volume}{229-232} (\bibinfo{year}{2012})
  \bibinfo{pages}{1--12}, ISSN \bibinfo{issn}{00457825}, \bibinfo{note}{doi:
  \url{https://doi.org/10.1016/j.cma.2012.03.019}}.

\bibitem[{Conrad and Marzouk(2013)}]{Conrad2013}
\bibinfo{author}{P.~R. Conrad}, \bibinfo{author}{Y.~M. Marzouk},
  \bibinfo{title}{{Adaptive Smolyak Pseudospectral Approximations}},
  \bibinfo{journal}{SIAM Journal on Scientific Computing}
  \bibinfo{volume}{35}~(\bibinfo{number}{6}) (\bibinfo{year}{2013})
  \bibinfo{pages}{A2643--A2670}, ISSN \bibinfo{issn}{1064-8275},
  \bibinfo{note}{doi: \url{https://doi.org/10.1137/120890715}}.

\bibitem[{Blatman and Sudret(2011)}]{blatman2011adaptive}
\bibinfo{author}{G.~Blatman}, \bibinfo{author}{B.~Sudret},
  \bibinfo{title}{Adaptive sparse polynomial chaos expansion based on least
  angle regression}, \bibinfo{journal}{Journal of Computational Physics}
  \bibinfo{volume}{230}~(\bibinfo{number}{6}) (\bibinfo{year}{2011})
  \bibinfo{pages}{2345--2367}, \bibinfo{note}{doi:
  \url{https://doi.org/10.1016/j.jcp.2010.12.021}}.

\bibitem[{Hampton and Doostan(2015)}]{hampton2015compressive}
\bibinfo{author}{J.~Hampton}, \bibinfo{author}{A.~Doostan},
  \bibinfo{title}{Compressive sampling of polynomial chaos expansions:
  Convergence analysis and sampling strategies}, \bibinfo{journal}{Journal of
  Computational Physics} \bibinfo{volume}{280} (\bibinfo{year}{2015})
  \bibinfo{pages}{363--386}, \bibinfo{note}{doi:
  \url{https://doi.org/10.1016/j.jcp.2014.09.019}}.

\bibitem[{Tsilifis et~al.(2019)Tsilifis, Huan, Safta, Sargsyan, Lacaze,
  Oefelein, Najm, and Ghanem}]{tsilifis2019compressive}
\bibinfo{author}{P.~Tsilifis}, \bibinfo{author}{X.~Huan},
  \bibinfo{author}{C.~Safta}, \bibinfo{author}{K.~Sargsyan},
  \bibinfo{author}{G.~Lacaze}, \bibinfo{author}{J.~C. Oefelein},
  \bibinfo{author}{H.~N. Najm}, \bibinfo{author}{R.~G. Ghanem},
  \bibinfo{title}{Compressive sensing adaptation for polynomial chaos
  expansions}, \bibinfo{journal}{Journal of Computational Physics}
  \bibinfo{volume}{380} (\bibinfo{year}{2019}) \bibinfo{pages}{29--47},
  \bibinfo{note}{doi: \url{https://doi.org/10.1016/j.jcp.2018.12.010}}.

\bibitem[{Blatman and Sudret(2010)}]{blatman2010adaptive}
\bibinfo{author}{G.~Blatman}, \bibinfo{author}{B.~Sudret}, \bibinfo{title}{An
  adaptive algorithm to build up sparse polynomial chaos expansions for
  stochastic finite element analysis}, \bibinfo{journal}{Probabilistic
  Engineering Mechanics} \bibinfo{volume}{25}~(\bibinfo{number}{2})
  (\bibinfo{year}{2010}) \bibinfo{pages}{183--197}, \bibinfo{note}{doi:
  \url{https://doi.org/10.1016/j.probengmech.2009.10.003}}.

\bibitem[{Hu and Youn(2011)}]{hu2011adaptive}
\bibinfo{author}{C.~Hu}, \bibinfo{author}{B.~D. Youn},
  \bibinfo{title}{Adaptive-sparse polynomial chaos expansion for reliability
  analysis and design of complex engineering systems},
  \bibinfo{journal}{Structural and Multidisciplinary Optimization}
  \bibinfo{volume}{43} (\bibinfo{year}{2011}) \bibinfo{pages}{419--442},
  \bibinfo{note}{doi: \url{https://doi.org/10.1007/s00158-010-0568-9}}.

\bibitem[{Pan and Dias(2017)}]{pan2017sliced}
\bibinfo{author}{Q.~Pan}, \bibinfo{author}{D.~Dias}, \bibinfo{title}{Sliced
  inverse regression-based sparse polynomial chaos expansions for reliability
  analysis in high dimensions}, \bibinfo{journal}{Reliability Engineering \&
  System Safety} \bibinfo{volume}{167} (\bibinfo{year}{2017})
  \bibinfo{pages}{484--493}, \bibinfo{note}{doi:
  \url{https://doi.org/10.1016/j.ress.2017.06.026}}.

\bibitem[{Xu and Kong(2018)}]{xu2018cubature}
\bibinfo{author}{J.~Xu}, \bibinfo{author}{F.~Kong}, \bibinfo{title}{A cubature
  collocation based sparse polynomial chaos expansion for efficient structural
  reliability analysis}, \bibinfo{journal}{Structural Safety}
  \bibinfo{volume}{74} (\bibinfo{year}{2018}) \bibinfo{pages}{24--31},
  \bibinfo{note}{doi: \url{https://doi.org/10.1016/j.strusafe.2018.04.001}}.

\bibitem[{Bhattacharyya(2021)}]{bhattacharyya2021structural}
\bibinfo{author}{B.~Bhattacharyya}, \bibinfo{title}{Structural reliability
  analysis by a Bayesian sparse polynomial chaos expansion},
  \bibinfo{journal}{Structural Safety} \bibinfo{volume}{90}
  (\bibinfo{year}{2021}) \bibinfo{pages}{102074}, \bibinfo{note}{doi:
  \url{https://doi.org/10.1016/j.strusafe.2020.102074}}.

\bibitem[{L\"{u}then et~al.(2021)L\"{u}then, Marelli, and
  Sudret}]{luthen2021sparse}
\bibinfo{author}{N.~L\"{u}then}, \bibinfo{author}{S.~Marelli},
  \bibinfo{author}{B.~Sudret}, \bibinfo{title}{Sparse polynomial chaos
  expansions: Literature survey and benchmark}, \bibinfo{journal}{SIAM/ASA
  Journal on Uncertainty Quantification}
  \bibinfo{volume}{9}~(\bibinfo{number}{2}) (\bibinfo{year}{2021})
  \bibinfo{pages}{593--649}, \bibinfo{note}{doi:
  \url{https://doi.org/10.1137/20M1315774}}.

\bibitem[{Schobi et~al.(2015)Schobi, Sudret, and Wiart}]{schobi2015polynomial}
\bibinfo{author}{R.~Schobi}, \bibinfo{author}{B.~Sudret},
  \bibinfo{author}{J.~Wiart}, \bibinfo{title}{Polynomial-chaos-based Kriging},
  \bibinfo{journal}{International Journal for Uncertainty Quantification}
  \bibinfo{volume}{5}~(\bibinfo{number}{2}), \bibinfo{note}{doi:
  \url{https://doi.org/10.1615/Int.J.UncertaintyQuantification.2015012467}}.

\bibitem[{Kersaudy et~al.(2015)Kersaudy, Sudret, Varsier, Picon, and
  Wiart}]{kersaudy2015new}
\bibinfo{author}{P.~Kersaudy}, \bibinfo{author}{B.~Sudret},
  \bibinfo{author}{N.~Varsier}, \bibinfo{author}{O.~Picon},
  \bibinfo{author}{J.~Wiart}, \bibinfo{title}{A new surrogate modeling
  technique combining Kriging and polynomial chaos expansions--Application to
  uncertainty analysis in computational dosimetry}, \bibinfo{journal}{Journal
  of Computational Physics} \bibinfo{volume}{286} (\bibinfo{year}{2015})
  \bibinfo{pages}{103--117}, \bibinfo{note}{doi:
  \url{https://doi.org/10.1016/j.jcp.2015.01.034}}.

\bibitem[{Pavlack et~al.(2022)Pavlack, Paix{\~a}o, Da~Silva, Cunha~Jr, and
  Garcia~Cava}]{pavlack2022polynomial}
\bibinfo{author}{B.~Pavlack}, \bibinfo{author}{J.~Paix{\~a}o},
  \bibinfo{author}{S.~Da~Silva}, \bibinfo{author}{A.~Cunha~Jr},
  \bibinfo{author}{D.~Garcia~Cava}, \bibinfo{title}{Polynomial Chaos-Kriging
  metamodel for quantification of the debonding area in large wind turbine
  blades}, \bibinfo{journal}{Structural Health Monitoring}
  \bibinfo{volume}{21}~(\bibinfo{number}{2}) (\bibinfo{year}{2022})
  \bibinfo{pages}{666--682}, \bibinfo{note}{doi:
  \url{https://doi.org/10.1177/14759217211007956}}.

\bibitem[{Shang et~al.(2021)Shang, Ma, Yang, and Chao}]{shang2021efficient}
\bibinfo{author}{X.~Shang}, \bibinfo{author}{P.~Ma}, \bibinfo{author}{M.~Yang},
  \bibinfo{author}{T.~Chao}, \bibinfo{title}{An efficient polynomial
  chaos-enhanced radial basis function approach for reliability-based design
  optimization}, \bibinfo{journal}{Structural and Multidisciplinary
  Optimization} \bibinfo{volume}{63} (\bibinfo{year}{2021})
  \bibinfo{pages}{789--805}, \bibinfo{note}{doi:
  \url{https://doi.org/10.1007/s00158-020-02730-0}}.

\bibitem[{Torre et~al.(2019)Torre, Marelli, Embrechts, and
  Sudret}]{torre2019data}
\bibinfo{author}{E.~Torre}, \bibinfo{author}{S.~Marelli},
  \bibinfo{author}{P.~Embrechts}, \bibinfo{author}{B.~Sudret},
  \bibinfo{title}{Data-driven polynomial chaos expansion for machine learning
  regression}, \bibinfo{journal}{Journal of Computational Physics}
  \bibinfo{volume}{388} (\bibinfo{year}{2019}) \bibinfo{pages}{601--623},
  \bibinfo{note}{doi: \url{https://doi.org/10.1016/j.jcp.2019.03.039}}.

\bibitem[{Nado et~al.(2021)Nado, Band, Collier, Djolonga, Dusenberry, Farquhar,
  Feng, Filos, Havasi, Jenatton et~al.}]{nado2021uncertainty}
\bibinfo{author}{Z.~Nado}, \bibinfo{author}{N.~Band},
  \bibinfo{author}{M.~Collier}, \bibinfo{author}{J.~Djolonga},
  \bibinfo{author}{M.~W. Dusenberry}, \bibinfo{author}{S.~Farquhar},
  \bibinfo{author}{Q.~Feng}, \bibinfo{author}{A.~Filos},
  \bibinfo{author}{M.~Havasi}, \bibinfo{author}{R.~Jenatton}, et~al.,
  \bibinfo{title}{Uncertainty Baselines: Benchmarks for uncertainty \&
  robustness in deep learning}, \bibinfo{journal}{arXiv preprint
  arXiv:2106.04015} \bibinfo{note}{{URL}:
  \url{https://doi.org/10.48550/arXiv.2106.04015}}.

\bibitem[{Li et~al.(2022{\natexlab{a}})Li, Yin, and Du}]{li2022uncertainty}
\bibinfo{author}{H.~Li}, \bibinfo{author}{J.~Yin}, \bibinfo{author}{X.~Du},
  \bibinfo{title}{Uncertainty Quantification of Physics-Based Label-Free Deep
  Learning and Probabilistic Prediction of Extreme Events}, in:
  \bibinfo{booktitle}{International Design Engineering Technical Conferences
  and Computers and Information in Engineering Conference}, vol.
  \bibinfo{volume}{86236}, \bibinfo{organization}{American Society of
  Mechanical Engineers}, \bibinfo{pages}{V03BT03A001}, \bibinfo{note}{doi:
  \url{https://doi.org/10.1115/DETC2022-88277}},
  \bibinfo{year}{2022}{\natexlab{a}}.

\bibitem[{Neal(1996)}]{Neal1996}
\bibinfo{author}{R.~M. Neal}, \bibinfo{title}{{Bayesian Learning for Neural
  Networks}}, \bibinfo{publisher}{Springer-Verlag New York},
  \bibinfo{address}{New York, NY}, \bibinfo{note}{doi:
  \url{https://doi.org/10.1007/978-1-4612-0745-0}}, \bibinfo{year}{1996}.

\bibitem[{Williams(1996)}]{williams1996computing}
\bibinfo{author}{C.~Williams}, \bibinfo{title}{Computing with infinite
  networks}, \bibinfo{journal}{Advances in Neural Information Processing
  Systems} \bibinfo{volume}{9}.

\bibitem[{Lee et~al.(2018)Lee, Bahri, Novak, Schoenholz, Pennington, and
  Sohl-Dickstein}]{lee2017deep}
\bibinfo{author}{J.~Lee}, \bibinfo{author}{Y.~Bahri},
  \bibinfo{author}{R.~Novak}, \bibinfo{author}{S.~S. Schoenholz},
  \bibinfo{author}{J.~Pennington}, \bibinfo{author}{J.~Sohl-Dickstein},
  \bibinfo{title}{Deep Neural Networks as {G}aussian Processes}, in:
  \bibinfo{booktitle}{ICLR}, \bibinfo{year}{2018}.

\bibitem[{Novak et~al.(2018)Novak, Xiao, Lee, Bahri, Yang, Hron, Abolafia,
  Pennington, and Sohl-Dickstein}]{novak2018bayesian}
\bibinfo{author}{R.~Novak}, \bibinfo{author}{L.~Xiao},
  \bibinfo{author}{J.~Lee}, \bibinfo{author}{Y.~Bahri},
  \bibinfo{author}{G.~Yang}, \bibinfo{author}{J.~Hron}, \bibinfo{author}{D.~A.
  Abolafia}, \bibinfo{author}{J.~Pennington},
  \bibinfo{author}{J.~Sohl-Dickstein}, \bibinfo{title}{Bayesian deep
  convolutional networks with many channels are {G}aussian processes}, in:
  \bibinfo{booktitle}{NIPS Workshop on Bayesian Deep Learning},
  \bibinfo{year}{2018}.

\bibitem[{Garriga-Alonso et~al.(2019)Garriga-Alonso, Rasmussen, and
  Aitchison}]{garriga2018deep}
\bibinfo{author}{A.~Garriga-Alonso}, \bibinfo{author}{C.~E. Rasmussen},
  \bibinfo{author}{L.~Aitchison}, \bibinfo{title}{Deep convolutional networks
  as shallow {G}aussian processes}, in: \bibinfo{booktitle}{ICLR},
  \bibinfo{year}{2019}.

\bibitem[{Cho and Saul(2009)}]{cho2009kernel}
\bibinfo{author}{Y.~Cho}, \bibinfo{author}{L.~Saul}, \bibinfo{title}{Kernel
  methods for deep learning}, \bibinfo{journal}{Advances in Neural Information
  Processing Systems} \bibinfo{volume}{22}.

\bibitem[{Wilson et~al.(2016)Wilson, Hu, Salakhutdinov, and
  Xing}]{wilson2016deep}
\bibinfo{author}{A.~G. Wilson}, \bibinfo{author}{Z.~Hu},
  \bibinfo{author}{R.~Salakhutdinov}, \bibinfo{author}{E.~P. Xing},
  \bibinfo{title}{Deep kernel learning}, in: \bibinfo{booktitle}{Artificial
  intelligence and statistics}, \bibinfo{organization}{PMLR},
  \bibinfo{pages}{370--378}, \bibinfo{year}{2016}.

\bibitem[{Damianou and Lawrence(2013)}]{damianou2013deep}
\bibinfo{author}{A.~Damianou}, \bibinfo{author}{N.~D. Lawrence},
  \bibinfo{title}{Deep {G}aussian processes}, in:
  \bibinfo{booktitle}{Artificial intelligence and statistics},
  \bibinfo{organization}{PMLR}, \bibinfo{pages}{207--215}, \bibinfo{note}{doi:
  \url{https://doi.org/10.48550/arXiv.1211.0358}}, \bibinfo{year}{2013}.

\bibitem[{Bui et~al.(2016)Bui, Hern{\'a}ndez-Lobato, Hernandez-Lobato, Li, and
  Turner}]{bui2016deep}
\bibinfo{author}{T.~Bui}, \bibinfo{author}{D.~Hern{\'a}ndez-Lobato},
  \bibinfo{author}{J.~Hernandez-Lobato}, \bibinfo{author}{Y.~Li},
  \bibinfo{author}{R.~Turner}, \bibinfo{title}{Deep Gaussian processes for
  regression using approximate expectation propagation}, in:
  \bibinfo{booktitle}{International Conference on Machine Learning},
  \bibinfo{organization}{PMLR}, \bibinfo{pages}{1472--1481},
  \bibinfo{year}{2016}.

\bibitem[{Salimbeni and Deisenroth(2017)}]{salimbeni2017doubly}
\bibinfo{author}{H.~Salimbeni}, \bibinfo{author}{M.~Deisenroth},
  \bibinfo{title}{Doubly stochastic variational inference for deep Gaussian
  processes}, \bibinfo{journal}{Advances in Neural Information Processing
  Systems} \bibinfo{volume}{30}.

\bibitem[{Havasi et~al.(2018)Havasi, Hern{\'a}ndez-Lobato, and
  Murillo-Fuentes}]{havasi2018inference}
\bibinfo{author}{M.~Havasi}, \bibinfo{author}{J.~M. Hern{\'a}ndez-Lobato},
  \bibinfo{author}{J.~J. Murillo-Fuentes}, \bibinfo{title}{Inference in deep
  {G}aussian processes using stochastic gradient {H}amiltonian {M}onte
  {C}arlo}, \bibinfo{journal}{Advances in Neural Information Processing
  Systems} \bibinfo{volume}{31}.

\bibitem[{Fuge et~al.(2014)Fuge, Peters, and Agogino}]{fuge2014machine}
\bibinfo{author}{M.~Fuge}, \bibinfo{author}{B.~Peters},
  \bibinfo{author}{A.~Agogino}, \bibinfo{title}{Machine learning algorithms for
  recommending design methods}, \bibinfo{journal}{Journal of Mechanical Design}
  \bibinfo{volume}{136}~(\bibinfo{number}{10}) (\bibinfo{year}{2014})
  \bibinfo{pages}{101103}, \bibinfo{note}{doi:
  \url{https://doi.org/10.1115/1.4028102}}.

\bibitem[{Panchal et~al.(2019)Panchal, Fuge, Liu, Missoum, and
  Tucker}]{panchal2019machine}
\bibinfo{author}{J.~H. Panchal}, \bibinfo{author}{M.~Fuge},
  \bibinfo{author}{Y.~Liu}, \bibinfo{author}{S.~Missoum},
  \bibinfo{author}{C.~Tucker}, \bibinfo{title}{Machine learning for engineering
  design}, \bibinfo{journal}{Journal of Mechanical Design}
  \bibinfo{volume}{141}~(\bibinfo{number}{11}), \bibinfo{note}{doi:
  \url{https://doi.org/10.1115/1.4044690}}.

\bibitem[{Vale et~al.(2003)Vale, Shea et~al.}]{vale2003machine}
\bibinfo{author}{C.~A. Vale}, \bibinfo{author}{K.~Shea}, et~al.,
  \bibinfo{title}{A machine learning-based approach to accelerating
  computational design synthesis}, in: \bibinfo{booktitle}{DS 31: Proceedings
  of ICED 03, the 14th International Conference on Engineering Design,
  Stockholm}, \bibinfo{pages}{183--184}, \bibinfo{year}{2003}.

\bibitem[{Fan et~al.(2020)Fan, Zeng, Sun, and Liu}]{fan2020finding}
\bibinfo{author}{C.~Fan}, \bibinfo{author}{L.~Zeng}, \bibinfo{author}{Y.~Sun},
  \bibinfo{author}{Y.-Y. Liu}, \bibinfo{title}{Finding key players in complex
  networks through deep reinforcement learning}, \bibinfo{journal}{Nature
  Machine Intelligence} \bibinfo{volume}{2}~(\bibinfo{number}{6})
  (\bibinfo{year}{2020}) \bibinfo{pages}{317--324}, \bibinfo{note}{doi:
  \url{https://doi.org/10.1038/s42256-020-0177-2}}.

\bibitem[{Jiang et~al.(2020)Jiang, Xiong, Zhang, and Rosen}]{jiang2020machine}
\bibinfo{author}{J.~Jiang}, \bibinfo{author}{Y.~Xiong},
  \bibinfo{author}{Z.~Zhang}, \bibinfo{author}{D.~W. Rosen},
  \bibinfo{title}{Machine learning integrated design for additive
  manufacturing}, \bibinfo{journal}{Journal of Intelligent Manufacturing}
  (\bibinfo{year}{2020}) \bibinfo{pages}{1--14}\bibinfo{note}{Doi:
  \url{https://doi.org/10.1007/s10845-020-01715-6}}.

\bibitem[{Moosavi et~al.(2020)Moosavi, Jablonka, and Smit}]{moosavi2020role}
\bibinfo{author}{S.~M. Moosavi}, \bibinfo{author}{K.~M. Jablonka},
  \bibinfo{author}{B.~Smit}, \bibinfo{title}{The role of machine learning in
  the understanding and design of materials}, \bibinfo{journal}{Journal of the
  American Chemical Society} \bibinfo{volume}{142}~(\bibinfo{number}{48})
  (\bibinfo{year}{2020}) \bibinfo{pages}{20273--20287}, \bibinfo{note}{doi:
  \url{https://doi.org/10.1021/jacs.0c09105}}.

\bibitem[{Tao et~al.(2021)Tao, Xu, Li, and Lu}]{tao2021machine}
\bibinfo{author}{Q.~Tao}, \bibinfo{author}{P.~Xu}, \bibinfo{author}{M.~Li},
  \bibinfo{author}{W.~Lu}, \bibinfo{title}{Machine learning for perovskite
  materials design and discovery}, \bibinfo{journal}{NPJ Computational
  Materials} \bibinfo{volume}{7}~(\bibinfo{number}{1}) (\bibinfo{year}{2021})
  \bibinfo{pages}{1--18}, \bibinfo{note}{doi:
  \url{https://doi.org/10.1038/s41524-021-00495-8}}.

\bibitem[{Moustapha and Sudret(2019)}]{moustapha2019surrogate}
\bibinfo{author}{M.~Moustapha}, \bibinfo{author}{B.~Sudret},
  \bibinfo{title}{Surrogate-assisted reliability-based design optimization: a
  survey and a unified modular framework}, \bibinfo{journal}{Structural and
  Multidisciplinary Optimization} \bibinfo{volume}{60}~(\bibinfo{number}{5})
  (\bibinfo{year}{2019}) \bibinfo{pages}{2157--2176}, \bibinfo{note}{doi:
  \url{https://doi.org/10.1007/s00158-019-02290-y}}.

\bibitem[{Perera et~al.(2019)Perera, Wickramasinghe, Nik, and
  Scartezzini}]{perera2019machine}
\bibinfo{author}{A.~Perera}, \bibinfo{author}{P.~Wickramasinghe},
  \bibinfo{author}{V.~M. Nik}, \bibinfo{author}{J.-L. Scartezzini},
  \bibinfo{title}{Machine learning methods to assist energy system
  optimization}, \bibinfo{journal}{Applied Energy} \bibinfo{volume}{243}
  (\bibinfo{year}{2019}) \bibinfo{pages}{191--205}, \bibinfo{note}{doi:
  \url{https://doi.org/10.1016/j.apenergy.2019.03.202}}.

\bibitem[{Lei et~al.(2019)Lei, Liu, Du, Zhang, and Guo}]{lei2019machine}
\bibinfo{author}{X.~Lei}, \bibinfo{author}{C.~Liu}, \bibinfo{author}{Z.~Du},
  \bibinfo{author}{W.~Zhang}, \bibinfo{author}{X.~Guo}, \bibinfo{title}{Machine
  learning-driven real-time topology optimization under moving morphable
  component-based framework}, \bibinfo{journal}{Journal of Applied Mechanics}
  \bibinfo{volume}{86}~(\bibinfo{number}{1}) (\bibinfo{year}{2019})
  \bibinfo{pages}{011004}, \bibinfo{note}{doi:
  \url{https://doi.org/10.1115/1.4041319}}.

\bibitem[{Hinton and Salakhutdinov(2006)}]{hinton2006reducing}
\bibinfo{author}{G.~E. Hinton}, \bibinfo{author}{R.~R. Salakhutdinov},
  \bibinfo{title}{Reducing the dimensionality of data with neural networks},
  \bibinfo{journal}{Science} \bibinfo{volume}{313}~(\bibinfo{number}{5786})
  (\bibinfo{year}{2006}) \bibinfo{pages}{504--507}, \bibinfo{note}{doi:
  \url{https://doi.org/10.1126/science.1127647}}.

\bibitem[{Qian et~al.(2022)Qian, Tan, and Ye}]{qian2022adaptive}
\bibinfo{author}{C.~Qian}, \bibinfo{author}{R.~K. Tan},
  \bibinfo{author}{W.~Ye}, \bibinfo{title}{An adaptive artificial neural
  network-based generative design method for layout designs},
  \bibinfo{journal}{International Journal of Heat and Mass Transfer}
  \bibinfo{volume}{184} (\bibinfo{year}{2022}) \bibinfo{pages}{122313},
  \bibinfo{note}{doi:
  \url{https://doi.org/10.1016/j.ijheatmasstransfer.2021.122313}}.

\bibitem[{Regenwetter et~al.(2022)Regenwetter, Nobari, and
  Ahmed}]{regenwetter2022deep}
\bibinfo{author}{L.~Regenwetter}, \bibinfo{author}{A.~H. Nobari},
  \bibinfo{author}{F.~Ahmed}, \bibinfo{title}{Deep generative models in
  engineering design: A review}, \bibinfo{journal}{Journal of Mechanical
  Design} \bibinfo{volume}{144}~(\bibinfo{number}{7}) (\bibinfo{year}{2022})
  \bibinfo{pages}{071704}, \bibinfo{note}{doi:
  \url{https://doi.org/10.1115/1.4053859}}.

\bibitem[{Hamdia et~al.(2019)Hamdia, Ghasemi, Bazi, AlHichri, Alajlan, and
  Rabczuk}]{hamdia2019novel}
\bibinfo{author}{K.~M. Hamdia}, \bibinfo{author}{H.~Ghasemi},
  \bibinfo{author}{Y.~Bazi}, \bibinfo{author}{H.~AlHichri},
  \bibinfo{author}{N.~Alajlan}, \bibinfo{author}{T.~Rabczuk}, \bibinfo{title}{A
  novel deep learning based method for the computational material design of
  flexoelectric nanostructures with topology optimization},
  \bibinfo{journal}{Finite Elements in Analysis and Design}
  \bibinfo{volume}{165} (\bibinfo{year}{2019}) \bibinfo{pages}{21--30},
  \bibinfo{note}{doi: \url{https://doi.org/10.1016/j.finel.2019.07.001}}.

\bibitem[{Yang et~al.(2018)Yang, Li, Catherine~Brinson, Choudhary, Chen, and
  Agrawal}]{yang2018microstructural}
\bibinfo{author}{Z.~Yang}, \bibinfo{author}{X.~Li},
  \bibinfo{author}{L.~Catherine~Brinson}, \bibinfo{author}{A.~N. Choudhary},
  \bibinfo{author}{W.~Chen}, \bibinfo{author}{A.~Agrawal},
  \bibinfo{title}{Microstructural materials design via deep adversarial
  learning methodology}, \bibinfo{journal}{Journal of Mechanical Design}
  \bibinfo{volume}{140}~(\bibinfo{number}{11}), \bibinfo{note}{doi:
  \url{https://doi.org/10.1115/1.4041371}}.

\bibitem[{Alizadeh et~al.(2020)Alizadeh, Allen, and
  Mistree}]{alizadeh2020managing}
\bibinfo{author}{R.~Alizadeh}, \bibinfo{author}{J.~K. Allen},
  \bibinfo{author}{F.~Mistree}, \bibinfo{title}{Managing computational
  complexity using surrogate models: a critical review},
  \bibinfo{journal}{Research in Engineering Design}
  \bibinfo{volume}{31}~(\bibinfo{number}{3}) (\bibinfo{year}{2020})
  \bibinfo{pages}{275--298}, \bibinfo{note}{doi:
  \url{https://doi.org/10.1007/s00163-020-00336-7}}.

\bibitem[{Kennedy and O'Hagan(2001)}]{kennedy2001bayesian}
\bibinfo{author}{M.~C. Kennedy}, \bibinfo{author}{A.~O'Hagan},
  \bibinfo{title}{Bayesian calibration of computer models},
  \bibinfo{journal}{Journal of the Royal Statistical Society: Series B
  (Statistical Methodology)} \bibinfo{volume}{63}~(\bibinfo{number}{3})
  (\bibinfo{year}{2001}) \bibinfo{pages}{425--464}, \bibinfo{note}{doi:
  \url{https://doi.org/10.1111/1467-9868.00294}}.

\bibitem[{Cheng et~al.(2020{\natexlab{b}})Cheng, Lu, Ling, and
  Zhou}]{cheng2020surrogate}
\bibinfo{author}{K.~Cheng}, \bibinfo{author}{Z.~Lu}, \bibinfo{author}{C.~Ling},
  \bibinfo{author}{S.~Zhou}, \bibinfo{title}{Surrogate-assisted global
  sensitivity analysis: an overview}, \bibinfo{journal}{Structural and
  Multidisciplinary Optimization} \bibinfo{volume}{61}~(\bibinfo{number}{3})
  (\bibinfo{year}{2020}{\natexlab{b}}) \bibinfo{pages}{1187--1213},
  \bibinfo{note}{doi: \url{https://doi.org/10.1007/s00158-019-02413-5}}.

\bibitem[{Chatterjee et~al.(2019)Chatterjee, Chakraborty, and
  Chowdhury}]{chatterjee2019critical}
\bibinfo{author}{T.~Chatterjee}, \bibinfo{author}{S.~Chakraborty},
  \bibinfo{author}{R.~Chowdhury}, \bibinfo{title}{A critical review of
  surrogate assisted robust design optimization}, \bibinfo{journal}{Archives of
  Computational Methods in Engineering}
  \bibinfo{volume}{26}~(\bibinfo{number}{1}) (\bibinfo{year}{2019})
  \bibinfo{pages}{245--274}, \bibinfo{note}{doi:
  \url{https://doi.org/10.1007/s11831-017-9240-5}}.

\bibitem[{Viana et~al.(2009)Viana, Haftka, and Steffen}]{viana2009multiple}
\bibinfo{author}{F.~A. Viana}, \bibinfo{author}{R.~T. Haftka},
  \bibinfo{author}{V.~Steffen}, \bibinfo{title}{Multiple surrogates: how
  cross-validation errors can help us to obtain the best predictor},
  \bibinfo{journal}{Structural and Multidisciplinary Optimization}
  \bibinfo{volume}{39}~(\bibinfo{number}{4}) (\bibinfo{year}{2009})
  \bibinfo{pages}{439--457}, \bibinfo{note}{doi:
  \url{https://doi.org/10.1007/s00158-008-0338-0}}.

\bibitem[{Jin et~al.(2003)Jin, Du, and Chen}]{jin2003use}
\bibinfo{author}{R.~Jin}, \bibinfo{author}{X.~Du}, \bibinfo{author}{W.~Chen},
  \bibinfo{title}{The use of metamodeling techniques for optimization under
  uncertainty}, \bibinfo{journal}{Structural and Multidisciplinary
  Optimization} \bibinfo{volume}{25}~(\bibinfo{number}{2})
  (\bibinfo{year}{2003}) \bibinfo{pages}{99--116}, \bibinfo{note}{doi:
  \url{https://doi.org/10.1007/s00158-002-0277-0}}.

\bibitem[{Hu and Mahadevan(2016{\natexlab{a}})}]{hu2016single}
\bibinfo{author}{Z.~Hu}, \bibinfo{author}{S.~Mahadevan}, \bibinfo{title}{A
  single-loop kriging surrogate modeling for time-dependent reliability
  analysis}, \bibinfo{journal}{Journal of Mechanical Design}
  \bibinfo{volume}{138}~(\bibinfo{number}{6}), \bibinfo{note}{doi:
  \url{https://doi.org/10.1115/1.4033428}}.

\bibitem[{Gaspar et~al.(2014)Gaspar, Teixeira, and
  Soares}]{gaspar2014assessment}
\bibinfo{author}{B.~Gaspar}, \bibinfo{author}{A.~P. Teixeira},
  \bibinfo{author}{C.~G. Soares}, \bibinfo{title}{Assessment of the efficiency
  of Kriging surrogate models for structural reliability analysis},
  \bibinfo{journal}{Probabilistic Engineering Mechanics} \bibinfo{volume}{37}
  (\bibinfo{year}{2014}) \bibinfo{pages}{24--34}, \bibinfo{note}{doi:
  \url{https://doi.org/10.1016/j.probengmech.2014.03.011}}.

\bibitem[{Zhang et~al.(2019{\natexlab{c}})Zhang, Wang, and
  S{\o}rensen}]{zhang2019reif}
\bibinfo{author}{X.~Zhang}, \bibinfo{author}{L.~Wang}, \bibinfo{author}{J.~D.
  S{\o}rensen}, \bibinfo{title}{REIF: a novel active-learning function toward
  adaptive Kriging surrogate models for structural reliability analysis},
  \bibinfo{journal}{Reliability Engineering \& System Safety}
  \bibinfo{volume}{185} (\bibinfo{year}{2019}{\natexlab{c}})
  \bibinfo{pages}{440--454}, \bibinfo{note}{doi:
  \url{https://doi.org/10.1016/j.ress.2019.01.014}}.

\bibitem[{Yan and Zhou(2019)}]{yan2019adaptive}
\bibinfo{author}{L.~Yan}, \bibinfo{author}{T.~Zhou}, \bibinfo{title}{Adaptive
  multi-fidelity polynomial chaos approach to Bayesian inference in inverse
  problems}, \bibinfo{journal}{Journal of Computational Physics}
  \bibinfo{volume}{381} (\bibinfo{year}{2019}) \bibinfo{pages}{110--128},
  \bibinfo{note}{doi: \url{https://doi.org/10.1016/j.jcp.2018.12.025}}.

\bibitem[{Zhang et~al.(2020)Zhang, Apley, and Chen}]{zhang2020bayesian}
\bibinfo{author}{Y.~Zhang}, \bibinfo{author}{D.~W. Apley},
  \bibinfo{author}{W.~Chen}, \bibinfo{title}{Bayesian optimization for
  materials design with mixed quantitative and qualitative variables},
  \bibinfo{journal}{Scientific Reports}
  \bibinfo{volume}{10}~(\bibinfo{number}{1}) (\bibinfo{year}{2020})
  \bibinfo{pages}{1--13}, \bibinfo{note}{doi:
  \url{https://doi.org/10.1038/s41598-020-60652-9}}.

\bibitem[{{US NSTC}(2011)}]{national2011materials}
\bibinfo{author}{{US NSTC}}, \bibinfo{title}{{Materials Genome Initiative} for
  global competitiveness}, \bibinfo{publisher}{Executive Office of the
  President, National Science and Technology Council}, \bibinfo{year}{2011}.

\bibitem[{Lander et~al.(2021)Lander, Koizumi, Christodoulou, Sapochak,
  Friedersdorf, and Warren}]{lander2021materials}
\bibinfo{author}{E.~Lander}, \bibinfo{author}{K.~Koizumi},
  \bibinfo{author}{J.~Christodoulou}, \bibinfo{author}{L.~Sapochak},
  \bibinfo{author}{L.~E. Friedersdorf}, \bibinfo{author}{J.~Warren},
  \bibinfo{title}{Materials genome initiative strategic plan (2021)},
  \bibinfo{journal}{National Science And Technology Council} .

\bibitem[{McDowell et~al.(2019)McDowell, Scott et~al.}]{mcdowell2019creating}
\bibinfo{author}{D.~McDowell}, \bibinfo{author}{J.~Scott}, et~al.,
  \bibinfo{title}{{Creating the Next-Generation Materials Genome Initiative
  Workforce}}, \bibinfo{type}{Tech. Rep.}, \bibinfo{institution}{The Minerals
  Metals and Materials Society}, \bibinfo{year}{2019}.

\bibitem[{de~Pablo et~al.(2019)de~Pablo, Jackson, Webb, Chen, Moore, Morgan,
  Jacobs, Pollock, Schlom, Toberer et~al.}]{de2019new}
\bibinfo{author}{J.~J. de~Pablo}, \bibinfo{author}{N.~E. Jackson},
  \bibinfo{author}{M.~A. Webb}, \bibinfo{author}{L.-Q. Chen},
  \bibinfo{author}{J.~E. Moore}, \bibinfo{author}{D.~Morgan},
  \bibinfo{author}{R.~Jacobs}, \bibinfo{author}{T.~Pollock},
  \bibinfo{author}{D.~G. Schlom}, \bibinfo{author}{E.~S. Toberer}, et~al.,
  \bibinfo{title}{New frontiers for the materials genome initiative},
  \bibinfo{journal}{NPJ Computational Materials}
  \bibinfo{volume}{5}~(\bibinfo{number}{1}) (\bibinfo{year}{2019})
  \bibinfo{pages}{1--23}, \bibinfo{note}{doi:
  \url{https://doi.org/10.1038/s41524-019-0173-4}}.

\bibitem[{Christodoulou et~al.(2021)Christodoulou, Friedersdorf, Sapochak, and
  Warren}]{christodoulou2021second}
\bibinfo{author}{J.~Christodoulou}, \bibinfo{author}{L.~E. Friedersdorf},
  \bibinfo{author}{L.~Sapochak}, \bibinfo{author}{J.~A. Warren},
  \bibinfo{title}{The second decade of the {Materials Genome Initiative}},
  \bibinfo{note}{doi: \url{https://doi.org/10.1007/s11837-021-05008-y}},
  \bibinfo{year}{2021}.

\bibitem[{Sasaki and Igarashi(2019)}]{sasaki2019topology}
\bibinfo{author}{H.~Sasaki}, \bibinfo{author}{H.~Igarashi},
  \bibinfo{title}{Topology optimization accelerated by deep learning},
  \bibinfo{journal}{IEEE Transactions on Magnetics}
  \bibinfo{volume}{55}~(\bibinfo{number}{6}) (\bibinfo{year}{2019})
  \bibinfo{pages}{1--5}, \bibinfo{note}{doi:
  \url{https://doi.org/10.1109/TMAG.2019.2901906}}.

\bibitem[{Kallioras et~al.(2020)Kallioras, Kazakis, and
  Lagaros}]{kallioras2020accelerated}
\bibinfo{author}{N.~A. Kallioras}, \bibinfo{author}{G.~Kazakis},
  \bibinfo{author}{N.~D. Lagaros}, \bibinfo{title}{Accelerated topology
  optimization by means of deep learning}, \bibinfo{journal}{Structural and
  Multidisciplinary Optimization} \bibinfo{volume}{62}~(\bibinfo{number}{3})
  (\bibinfo{year}{2020}) \bibinfo{pages}{1185--1212}, \bibinfo{note}{doi:
  \url{https://doi.org/10.1007/s00158-020-02545-z}}.

\bibitem[{Xiao et~al.(2019)Xiao, Nazarian, and Bogdan}]{xiao2019self}
\bibinfo{author}{Y.~Xiao}, \bibinfo{author}{S.~Nazarian},
  \bibinfo{author}{P.~Bogdan}, \bibinfo{title}{Self-optimizing and
  self-programming computing systems: A combined compiler, complex networks,
  and machine learning approach}, \bibinfo{journal}{IEEE transactions on Very
  Large Scale Integration (VLSI) Systems}
  \bibinfo{volume}{27}~(\bibinfo{number}{6}) (\bibinfo{year}{2019})
  \bibinfo{pages}{1416--1427}, \bibinfo{note}{doi:
  \url{https://doi.org/10.1109/TVLSI.2019.2897650}}.

\bibitem[{Hu and Mahadevan(2016{\natexlab{b}})}]{hu2016global}
\bibinfo{author}{Z.~Hu}, \bibinfo{author}{S.~Mahadevan}, \bibinfo{title}{Global
  sensitivity analysis-enhanced surrogate (GSAS) modeling for reliability
  analysis}, \bibinfo{journal}{Structural and Multidisciplinary Optimization}
  \bibinfo{volume}{53}~(\bibinfo{number}{3})
  (\bibinfo{year}{2016}{\natexlab{b}}) \bibinfo{pages}{501--521},
  \bibinfo{note}{doi: \url{https://doi.org/10.1007/s00158-015-1347-4}}.

\bibitem[{Li et~al.(2021{\natexlab{b}})Li, Wang, Li, and Wang}]{li2021improved}
\bibinfo{author}{J.~Li}, \bibinfo{author}{B.~Wang}, \bibinfo{author}{Z.~Li},
  \bibinfo{author}{Y.~Wang}, \bibinfo{title}{An improved active learning method
  combing with the weight information entropy and Monte Carlo simulation of
  efficient structural reliability analysis}, \bibinfo{journal}{Proceedings of
  the Institution of Mechanical Engineers, Part C: Journal of Mechanical
  Engineering Science} \bibinfo{volume}{235}~(\bibinfo{number}{19})
  (\bibinfo{year}{2021}{\natexlab{b}}) \bibinfo{pages}{4296--4313},
  \bibinfo{note}{doi: \url{https://doi.org/10.1177/0954406220973233}}.

\bibitem[{Alibrandi et~al.(2022)Alibrandi, Andersen, and
  Zio}]{alibrandi2022informational}
\bibinfo{author}{U.~Alibrandi}, \bibinfo{author}{L.~V. Andersen},
  \bibinfo{author}{E.~Zio}, \bibinfo{title}{Informational probabilistic
  sensitivity analysis and active learning surrogate modelling},
  \bibinfo{journal}{Probabilistic Engineering Mechanics}
  (\bibinfo{year}{2022}) \bibinfo{pages}{103359}\bibinfo{note}{Doi:
  \url{https://doi.org/10.1016/j.probengmech.2022.103359}}.

\bibitem[{Sadoughi et~al.(2018)Sadoughi, Hu, MacKenzie, Eshghi, and
  Lee}]{sadoughi2018sequential}
\bibinfo{author}{M.~K. Sadoughi}, \bibinfo{author}{C.~Hu},
  \bibinfo{author}{C.~A. MacKenzie}, \bibinfo{author}{A.~T. Eshghi},
  \bibinfo{author}{S.~Lee}, \bibinfo{title}{Sequential exploration-exploitation
  with dynamic trade-off for efficient reliability analysis of complex
  engineered systems}, \bibinfo{journal}{Structural and Multidisciplinary
  Optimization} \bibinfo{volume}{57}~(\bibinfo{number}{1})
  (\bibinfo{year}{2018}) \bibinfo{pages}{235--250}, \bibinfo{note}{doi:
  \url{https://doi.org/10.1007/s00158-017-1748-7}}.

\bibitem[{Afshari et~al.(2022)Afshari, Enayatollahi, Xu, and
  Liang}]{afshari2022machine}
\bibinfo{author}{S.~S. Afshari}, \bibinfo{author}{F.~Enayatollahi},
  \bibinfo{author}{X.~Xu}, \bibinfo{author}{X.~Liang}, \bibinfo{title}{Machine
  learning-based methods in structural reliability analysis: A review},
  \bibinfo{journal}{Reliability Engineering \& System Safety}
  \bibinfo{volume}{219} (\bibinfo{year}{2022}) \bibinfo{pages}{108223},
  \bibinfo{note}{doi: \url{https://doi.org/10.1016/j.ress.2021.108223}}.

\bibitem[{Frazier(2018)}]{frazier2018bayesian}
\bibinfo{author}{P.~I. Frazier}, \bibinfo{title}{Bayesian optimization}, in:
  \bibinfo{booktitle}{Recent advances in optimization and modeling of
  contemporary problems}, \bibinfo{publisher}{INFORMS},
  \bibinfo{pages}{255--278}, \bibinfo{note}{doi:
  \url{https://doi.org/10.1287/educ.2018.0188}}, \bibinfo{year}{2018}.

\bibitem[{Shen and Huan(2021)}]{shen2021bayesian}
\bibinfo{author}{W.~Shen}, \bibinfo{author}{X.~Huan}, \bibinfo{title}{Bayesian
  sequential optimal experimental design for nonlinear models using policy
  gradient reinforcement learning}, \bibinfo{journal}{arXiv preprint
  arXiv:2110.15335} \bibinfo{note}{Doi:
  \url{https://doi.org/10.48550/arXiv.2110.15335}}.

\bibitem[{Goodfellow et~al.(2020)Goodfellow, Pouget-Abadie, Mirza, Xu,
  Warde-Farley, Ozair, Courville, and Bengio}]{goodfellow2020generative}
\bibinfo{author}{I.~Goodfellow}, \bibinfo{author}{J.~Pouget-Abadie},
  \bibinfo{author}{M.~Mirza}, \bibinfo{author}{B.~Xu},
  \bibinfo{author}{D.~Warde-Farley}, \bibinfo{author}{S.~Ozair},
  \bibinfo{author}{A.~Courville}, \bibinfo{author}{Y.~Bengio},
  \bibinfo{title}{Generative adversarial networks},
  \bibinfo{journal}{Communications of the ACM}
  \bibinfo{volume}{63}~(\bibinfo{number}{11}) (\bibinfo{year}{2020})
  \bibinfo{pages}{139--144}, \bibinfo{note}{doi:
  \url{https://doi.org/10.1145/3422622}}.

\bibitem[{Guo et~al.(2018)Guo, Lohan, Cang, Ren, and Allison}]{guo2018indirect}
\bibinfo{author}{T.~Guo}, \bibinfo{author}{D.~J. Lohan},
  \bibinfo{author}{R.~Cang}, \bibinfo{author}{M.~Y. Ren},
  \bibinfo{author}{J.~T. Allison}, \bibinfo{title}{An indirect design
  representation for topology optimization using variational autoencoder and
  style transfer}, in: \bibinfo{booktitle}{2018 AIAA/ASCE/AHS/ASC Structures,
  Structural Dynamics, and Materials Conference}, \bibinfo{pages}{0804},
  \bibinfo{note}{doi: \url{https://doi.org/10.2514/6.2018-0804}},
  \bibinfo{year}{2018}.

\bibitem[{Chen et~al.(2020)Chen, Chen, Xing, Xia, Zhu, Grundy, and
  Wang}]{chen2020wireframe}
\bibinfo{author}{J.~Chen}, \bibinfo{author}{C.~Chen},
  \bibinfo{author}{Z.~Xing}, \bibinfo{author}{X.~Xia},
  \bibinfo{author}{L.~Zhu}, \bibinfo{author}{J.~Grundy},
  \bibinfo{author}{J.~Wang}, \bibinfo{title}{Wireframe-based UI design search
  through image autoencoder}, \bibinfo{journal}{ACM Transactions on Software
  Engineering and Methodology (TOSEM)}
  \bibinfo{volume}{29}~(\bibinfo{number}{3}) (\bibinfo{year}{2020})
  \bibinfo{pages}{1--31}, \bibinfo{note}{doi:
  \url{https://doi.org/10.1145/3391613}}.

\bibitem[{Li et~al.(2022{\natexlab{b}})Li, Xie, and Sha}]{li2022predictive}
\bibinfo{author}{X.~Li}, \bibinfo{author}{C.~Xie}, \bibinfo{author}{Z.~Sha},
  \bibinfo{title}{A Predictive and Generative Design Approach for
  Three-Dimensional Mesh Shapes Using Target-Embedding Variational
  Autoencoder}, \bibinfo{journal}{Journal of Mechanical Design}
  \bibinfo{volume}{144}~(\bibinfo{number}{11})
  (\bibinfo{year}{2022}{\natexlab{b}}) \bibinfo{pages}{114501},
  \bibinfo{note}{doi: \url{https://doi.org/10.1115/1.4054906}}.

\bibitem[{Oh et~al.(2019)Oh, Jung, Kim, Lee, and Kang}]{oh2019deep}
\bibinfo{author}{S.~Oh}, \bibinfo{author}{Y.~Jung}, \bibinfo{author}{S.~Kim},
  \bibinfo{author}{I.~Lee}, \bibinfo{author}{N.~Kang}, \bibinfo{title}{Deep
  generative design: Integration of topology optimization and generative
  models}, \bibinfo{journal}{Journal of Mechanical Design}
  \bibinfo{volume}{141}~(\bibinfo{number}{11}), \bibinfo{note}{doi:
  \url{https://doi.org/10.1115/1.4044229}}.

\bibitem[{Regenwetter and Ahmed(2022)}]{regenwetter2022towards}
\bibinfo{author}{L.~Regenwetter}, \bibinfo{author}{F.~Ahmed},
  \bibinfo{title}{Towards Goal, Feasibility, and Diversity-Oriented Deep
  Generative Models in Design}, \bibinfo{journal}{arXiv preprint
  arXiv:2206.07170,} \bibinfo{note}{Doi:
  \url{https://doi.org/10.48550/arXiv.2206.07170}}.

\bibitem[{Song et~al.(2013)Song, Choi, Lee, Zhao, and Lamb}]{song2013adaptive}
\bibinfo{author}{H.~Song}, \bibinfo{author}{K.~K. Choi},
  \bibinfo{author}{I.~Lee}, \bibinfo{author}{L.~Zhao},
  \bibinfo{author}{D.~Lamb}, \bibinfo{title}{Adaptive virtual support vector
  machine for reliability analysis of high-dimensional problems},
  \bibinfo{journal}{Structural and Multidisciplinary Optimization}
  \bibinfo{volume}{47}~(\bibinfo{number}{4}) (\bibinfo{year}{2013})
  \bibinfo{pages}{479--491}, \bibinfo{note}{doi:
  \url{https://doi.org/10.1007/s00158-012-0857-6}}.

\bibitem[{Basudhar and Missoum(2008)}]{basudhar2008adaptive}
\bibinfo{author}{A.~Basudhar}, \bibinfo{author}{S.~Missoum},
  \bibinfo{title}{Adaptive explicit decision functions for probabilistic design
  and optimization using support vector machines}, \bibinfo{journal}{Computers
  \& Structures} \bibinfo{volume}{86}~(\bibinfo{number}{19-20})
  (\bibinfo{year}{2008}) \bibinfo{pages}{1904--1917}, \bibinfo{note}{doi:
  \url{https://doi.org/10.1016/j.compstruc.2008.02.008}}.

\bibitem[{Sener and Savarese(2017)}]{sener2017active}
\bibinfo{author}{O.~Sener}, \bibinfo{author}{S.~Savarese},
  \bibinfo{title}{Active learning for convolutional neural networks: A core-set
  approach}, \bibinfo{journal}{arXiv preprint arXiv:1708.00489}
  \bibinfo{note}{Doi: \url{https://doi.org/10.48550/arXiv.1708.00489}}.

\bibitem[{Haut et~al.(2018)Haut, Paoletti, Plaza, Li, and
  Plaza}]{haut2018active}
\bibinfo{author}{J.~M. Haut}, \bibinfo{author}{M.~E. Paoletti},
  \bibinfo{author}{J.~Plaza}, \bibinfo{author}{J.~Li},
  \bibinfo{author}{A.~Plaza}, \bibinfo{title}{Active learning with
  convolutional neural networks for hyperspectral image classification using a
  new Bayesian approach}, \bibinfo{journal}{IEEE Transactions on Geoscience and
  Remote Sensing} \bibinfo{volume}{56}~(\bibinfo{number}{11})
  (\bibinfo{year}{2018}) \bibinfo{pages}{6440--6461}, \bibinfo{note}{doi:
  \url{https://doi.org/10.1109/TGRS.2018.2838665}}.

\bibitem[{Xiang et~al.(2020)Xiang, Chen, Bao, and Li}]{xiang2020active}
\bibinfo{author}{Z.~Xiang}, \bibinfo{author}{J.~Chen},
  \bibinfo{author}{Y.~Bao}, \bibinfo{author}{H.~Li}, \bibinfo{title}{An active
  learning method combining deep neural network and weighted sampling for
  structural reliability analysis}, \bibinfo{journal}{Mechanical Systems and
  Signal Processing} \bibinfo{volume}{140} (\bibinfo{year}{2020})
  \bibinfo{pages}{106684}, \bibinfo{note}{doi:
  \url{https://doi.org/10.1016/j.ymssp.2020.106684}}.

\bibitem[{Bao et~al.(2021)Bao, Xiang, and Li}]{bao2021adaptive}
\bibinfo{author}{Y.~Bao}, \bibinfo{author}{Z.~Xiang}, \bibinfo{author}{H.~Li},
  \bibinfo{title}{Adaptive subset searching-based deep neural network method
  for structural reliability analysis}, \bibinfo{journal}{Reliability
  Engineering \& System Safety} \bibinfo{volume}{213} (\bibinfo{year}{2021})
  \bibinfo{pages}{107778}, \bibinfo{note}{doi:
  \url{https://doi.org/10.1016/j.ress.2021.107778}}.

\bibitem[{Nguyen and Nguyen-Xuan(2020)}]{nguyen2020deep}
\bibinfo{author}{L.~C. Nguyen}, \bibinfo{author}{H.~Nguyen-Xuan},
  \bibinfo{title}{Deep learning for computational structural optimization},
  \bibinfo{journal}{ISA Transactions} \bibinfo{volume}{103}
  (\bibinfo{year}{2020}) \bibinfo{pages}{177--191}, \bibinfo{note}{doi:
  \url{https://doi.org/10.1016/j.isatra.2020.03.033}}.

\bibitem[{Asano and Noda(2018)}]{asano2018optimization}
\bibinfo{author}{T.~Asano}, \bibinfo{author}{S.~Noda},
  \bibinfo{title}{Optimization of photonic crystal nanocavities based on deep
  learning}, \bibinfo{journal}{Optics Express}
  \bibinfo{volume}{26}~(\bibinfo{number}{25}) (\bibinfo{year}{2018})
  \bibinfo{pages}{32704--32717}, \bibinfo{note}{doi:
  \url{https://doi.org/10.1364/OE.26.032704}}.

\bibitem[{Beland and Nair(2017)}]{beland2017bayesian}
\bibinfo{author}{J.~J. Beland}, \bibinfo{author}{P.~B. Nair},
  \bibinfo{title}{Bayesian optimization under uncertainty}, in:
  \bibinfo{booktitle}{NIPS BayesOpt 2017 workshop}, \bibinfo{year}{2017}.

\bibitem[{Mathern et~al.(2021)Mathern, Steinholtz, Sj{\"o}berg, {\"O}nnheim,
  Ek, Rempling, Gustavsson, and Jirstrand}]{mathern2021multi}
\bibinfo{author}{A.~Mathern}, \bibinfo{author}{O.~S. Steinholtz},
  \bibinfo{author}{A.~Sj{\"o}berg}, \bibinfo{author}{M.~{\"O}nnheim},
  \bibinfo{author}{K.~Ek}, \bibinfo{author}{R.~Rempling},
  \bibinfo{author}{E.~Gustavsson}, \bibinfo{author}{M.~Jirstrand},
  \bibinfo{title}{Multi-objective constrained Bayesian optimization for
  structural design}, \bibinfo{journal}{Structural and Multidisciplinary
  Optimization} \bibinfo{volume}{63}~(\bibinfo{number}{2})
  (\bibinfo{year}{2021}) \bibinfo{pages}{689--701}, \bibinfo{note}{doi:
  \url{https://doi.org/10.1007/s00158-020-02720-2}}.

\bibitem[{Frazier and Wang(2016)}]{frazier2016bayesian}
\bibinfo{author}{P.~I. Frazier}, \bibinfo{author}{J.~Wang},
  \bibinfo{title}{Bayesian optimization for materials design}, in:
  \bibinfo{booktitle}{Information Science for Materials Discovery and Design},
  \bibinfo{publisher}{Springer}, \bibinfo{pages}{45--75}, \bibinfo{note}{doi:
  \url{https://doi.org/10.1007/978-3-319-23871-5_3}}, \bibinfo{year}{2016}.

\bibitem[{Sharpe et~al.(2018)Sharpe, Seepersad, Watts, and
  Tortorelli}]{sharpe2018design}
\bibinfo{author}{C.~Sharpe}, \bibinfo{author}{C.~C. Seepersad},
  \bibinfo{author}{S.~Watts}, \bibinfo{author}{D.~Tortorelli},
  \bibinfo{title}{Design of mechanical metamaterials via constrained {B}ayesian
  optimization}, in: \bibinfo{booktitle}{International Design Engineering
  Technical Conferences and Computers and Information in Engineering
  Conference}, vol. \bibinfo{volume}{51753}, \bibinfo{organization}{American
  Society of Mechanical Engineers}, \bibinfo{pages}{V02AT03A029},
  \bibinfo{note}{doi: \url{https://doi.org/10.1115/DETC2018-85270}},
  \bibinfo{year}{2018}.

\bibitem[{Miguel et~al.(2022)Miguel, Lopez, Torii, and
  Beck}]{miguel2022reliability}
\bibinfo{author}{L.~F.~F. Miguel}, \bibinfo{author}{R.~H. Lopez},
  \bibinfo{author}{A.~J. Torii}, \bibinfo{author}{A.~T. Beck},
  \bibinfo{title}{Reliability-based optimization of multiple Folded Pendulum
  TMDs through Efficient Global Optimization}, \bibinfo{journal}{Engineering
  Structures} \bibinfo{volume}{266} (\bibinfo{year}{2022})
  \bibinfo{pages}{114524}, \bibinfo{note}{doi:
  \url{https://doi.org/10.1016/j.engstruct.2022.114524}}.

\bibitem[{Liu and Wang(2022)}]{liu2022metal}
\bibinfo{author}{D.~Liu}, \bibinfo{author}{Y.~Wang}, \bibinfo{title}{Metal
  Additive Manufacturing Process Design based on Physics Constrained Neural
  Networks and Multi-Objective Bayesian Optimization},
  \bibinfo{journal}{Manufacturing Letters} \bibinfo{volume}{33}
  (\bibinfo{year}{2022}) \bibinfo{pages}{817--827}, \bibinfo{note}{doi:
  \url{https://doi.org/10.1016/j.mfglet.2022.07.101}}.

\bibitem[{Le~Gratiet and Garnier(2014)}]{le2014recursive}
\bibinfo{author}{L.~Le~Gratiet}, \bibinfo{author}{J.~Garnier},
  \bibinfo{title}{Recursive co-kriging model for design of computer experiments
  with multiple levels of fidelity}, \bibinfo{journal}{International Journal
  for Uncertainty Quantification} \bibinfo{volume}{4}~(\bibinfo{number}{5}),
  \bibinfo{note}{doi:
  \url{https://doi.org/10.1615/Int.J.UncertaintyQuantification.2014006914}}.

\bibitem[{\'{A}lvarez et~al.(2012)\'{A}lvarez, Rosasco, and
  Lawrence}]{alvarez2011kernels}
\bibinfo{author}{M.~A. \'{A}lvarez}, \bibinfo{author}{L.~Rosasco},
  \bibinfo{author}{N.~D. Lawrence}, \bibinfo{title}{Kernels for Vector-Valued
  Functions: A Review}, \bibinfo{journal}{Foundations and Trends in Machine
  Learning} \bibinfo{volume}{4}~(\bibinfo{number}{3}) (\bibinfo{year}{2012})
  \bibinfo{pages}{195--266}, \bibinfo{note}{doi:
  \url{https://doi.org/10.1561/2200000036}}.

\bibitem[{Dwight and Han(2009)}]{dwight2009efficient}
\bibinfo{author}{R.~P. Dwight}, \bibinfo{author}{Z.-H. Han},
  \bibinfo{title}{Efficient uncertainty quantification using gradient-enhanced
  kriging}, \bibinfo{journal}{AIAA paper} \bibinfo{volume}{2276}
  (\bibinfo{year}{2009}) \bibinfo{pages}{2009}, \bibinfo{note}{doi:
  \url{https://doi.org/10.2514/6.2009-2276}}.

\bibitem[{Tran et~al.(2019)Tran, Tran, and Wang}]{tran2019constrained}
\bibinfo{author}{A.~Tran}, \bibinfo{author}{M.~Tran},
  \bibinfo{author}{Y.~Wang}, \bibinfo{title}{Constrained mixed-integer
  {Gaussian mixture Bayesian optimization} and its applications in designing
  fractal and auxetic metamaterials}, \bibinfo{journal}{Structural and
  Multidisciplinary Optimization} \bibinfo{volume}{59} (\bibinfo{year}{2019})
  \bibinfo{pages}{2131--2154}, \bibinfo{note}{doi:
  \url{https://doi.org/10.1007/s00158-018-2182-1}}.

\bibitem[{Paciorek and Schervish(2003)}]{paciorek2003nonstationary}
\bibinfo{author}{C.~Paciorek}, \bibinfo{author}{M.~Schervish},
  \bibinfo{title}{Nonstationary covariance functions for {G}aussian process
  regression}, \bibinfo{journal}{Advances in Neural Information Processing
  Systems} \bibinfo{volume}{16}.

\bibitem[{Heinonen et~al.(2016)Heinonen, Mannerstr{\"o}m, Rousu, Kaski, and
  L{\"a}hdesm{\"a}ki}]{heinonen2016non}
\bibinfo{author}{M.~Heinonen}, \bibinfo{author}{H.~Mannerstr{\"o}m},
  \bibinfo{author}{J.~Rousu}, \bibinfo{author}{S.~Kaski},
  \bibinfo{author}{H.~L{\"a}hdesm{\"a}ki}, \bibinfo{title}{Non-stationary
  {G}aussian process regression with {H}amiltonian {M}onte {C}arlo}, in:
  \bibinfo{booktitle}{Artificial Intelligence and Statistics},
  \bibinfo{organization}{PMLR}, \bibinfo{pages}{732--740},
  \bibinfo{year}{2016}.

\bibitem[{Remes et~al.(2017)Remes, Heinonen, and Kaski}]{remes2017non}
\bibinfo{author}{S.~Remes}, \bibinfo{author}{M.~Heinonen},
  \bibinfo{author}{S.~Kaski}, \bibinfo{title}{Non-stationary spectral kernels},
  \bibinfo{journal}{Advances in Neural Information Processing Systems}
  \bibinfo{volume}{30}.

\bibitem[{Schwabacher and Goebel(2007)}]{schwabacher2007survey}
\bibinfo{author}{M.~Schwabacher}, \bibinfo{author}{K.~Goebel},
  \bibinfo{title}{A Survey of Artificial Intelligence for Prognostics.}, in:
  \bibinfo{booktitle}{AAAI fall symposium: artificial intelligence for
  prognostics}, \bibinfo{organization}{Arlington, VA},
  \bibinfo{pages}{108--115}, \bibinfo{year}{2007}.

\bibitem[{Kefalas et~al.(2022)Kefalas, van Stein, Baratchi, Apostolidis, and
  Bäck}]{uqp3}
\bibinfo{author}{M.~Kefalas}, \bibinfo{author}{B.~van Stein},
  \bibinfo{author}{M.~Baratchi}, \bibinfo{author}{A.~Apostolidis},
  \bibinfo{author}{T.~Bäck}, \bibinfo{title}{An End-to-End Pipeline for
  Uncertainty Quantification and Remaining Useful Life Estimation: An
  Application on Aircraft Engines} \bibinfo{volume}{7} (\bibinfo{year}{2022})
  \bibinfo{pages}{245--260}, \bibinfo{note}{doi:
  \url{https://doi.org/10.36001/phme.2022.v7i1.3317}}.

\bibitem[{Lee and Mitici(2022)}]{uqp11}
\bibinfo{author}{J.~Lee}, \bibinfo{author}{M.~Mitici}, \bibinfo{title}{Deep
  reinforcement learning for predictive aircraft maintenance using
  Probabilistic Remaining-Useful-Life prognostics},
  \bibinfo{journal}{Reliability Engineering \& System Safety}
  (\bibinfo{year}{2022}) \bibinfo{pages}{108908}\bibinfo{note}{Doi:
  \url{https://doi.org/10.1016/j.ress.2022.108908}}.

\bibitem[{Mazaev et~al.(2021)Mazaev, Crevecoeur, and Van~Hoecke}]{uqp15}
\bibinfo{author}{G.~Mazaev}, \bibinfo{author}{G.~Crevecoeur},
  \bibinfo{author}{S.~Van~Hoecke}, \bibinfo{title}{Bayesian convolutional
  neural networks for remaining useful life prognostics of solenoid valves with
  uncertainty estimations}, \bibinfo{journal}{IEEE Transactions on Industrial
  Informatics} \bibinfo{volume}{17}~(\bibinfo{number}{12})
  (\bibinfo{year}{2021}) \bibinfo{pages}{8418--8428}, \bibinfo{note}{{doi}:
  \url{https://doi.org/10.1109/TII.2021.3078193}}.

\bibitem[{Zhu et~al.(2022)Zhu, Chen, Peng, and Ye}]{uqp17}
\bibinfo{author}{R.~Zhu}, \bibinfo{author}{Y.~Chen}, \bibinfo{author}{W.~Peng},
  \bibinfo{author}{Z.-S. Ye}, \bibinfo{title}{Bayesian deep-learning for RUL
  prediction: An active learning perspective}, \bibinfo{journal}{Reliability
  Engineering \& System Safety} \bibinfo{volume}{228} (\bibinfo{year}{2022})
  \bibinfo{pages}{108758}, \bibinfo{note}{doi:
  \url{https://doi.org/10.1016/j.ress.2022.108758}}.

\bibitem[{Yang et~al.(2022)Yang, Peng, Xie, and Wang}]{uqp1}
\bibinfo{author}{J.~Yang}, \bibinfo{author}{Y.~Peng}, \bibinfo{author}{J.~Xie},
  \bibinfo{author}{P.~Wang}, \bibinfo{title}{Remaining Useful Life Prediction
  Method for Bearings Based on LSTM with Uncertainty Quantification},
  \bibinfo{journal}{Sensors} \bibinfo{volume}{22}~(\bibinfo{number}{12})
  (\bibinfo{year}{2022}) \bibinfo{pages}{4549}, \bibinfo{note}{doi:
  \url{https://doi.org/10.3390/s22124549}}.

\bibitem[{Li et~al.(2020{\natexlab{b}})Li, Yang, Lee, Wang, and Rong}]{uqp10}
\bibinfo{author}{G.~Li}, \bibinfo{author}{L.~Yang}, \bibinfo{author}{C.-G.
  Lee}, \bibinfo{author}{X.~Wang}, \bibinfo{author}{M.~Rong}, \bibinfo{title}{A
  Bayesian deep learning RUL framework integrating epistemic and aleatoric
  uncertainties}, \bibinfo{journal}{IEEE Transactions on Industrial
  Electronics} \bibinfo{volume}{68}~(\bibinfo{number}{9})
  (\bibinfo{year}{2020}{\natexlab{b}}) \bibinfo{pages}{8829--8841},
  \bibinfo{note}{doi: \url{https://doi.org/10.1109/TIE.2020.3009593}}.

\bibitem[{Lin and Li(2022)}]{uqp14}
\bibinfo{author}{Y.-H. Lin}, \bibinfo{author}{G.-H. Li}, \bibinfo{title}{A
  Bayesian Deep Learning Framework for RUL Prediction Incorporating Uncertainty
  Quantification and Calibration}, \bibinfo{journal}{IEEE Transactions on
  Industrial Informatics} \bibinfo{note}{Doi:
  \url{https://doi.org/10.1109/TII.2022.3156965}}.

\bibitem[{Wei et~al.(2021)Wei, Gu, Ye, Wang, Xu, and Wu}]{uqp16}
\bibinfo{author}{M.~Wei}, \bibinfo{author}{H.~Gu}, \bibinfo{author}{M.~Ye},
  \bibinfo{author}{Q.~Wang}, \bibinfo{author}{X.~Xu}, \bibinfo{author}{C.~Wu},
  \bibinfo{title}{Remaining useful life prediction of lithium-ion batteries
  based on Monte Carlo Dropout and gated recurrent unit},
  \bibinfo{journal}{Energy Reports} \bibinfo{volume}{7} (\bibinfo{year}{2021})
  \bibinfo{pages}{2862--2871}, \bibinfo{note}{doi:
  \url{https://doi.org/10.1016/j.egyr.2021.05.019}}.

\bibitem[{Kong et~al.(2022)Kong, Zhang, and Mahadevan}]{kong2022bayesian}
\bibinfo{author}{Y.~Kong}, \bibinfo{author}{X.~Zhang},
  \bibinfo{author}{S.~Mahadevan}, \bibinfo{title}{Bayesian Deep Learning for
  Aircraft Hard Landing Safety Assessment}, \bibinfo{journal}{IEEE Transactions
  on Intelligent Transportation Systems}
  \bibinfo{volume}{23}~(\bibinfo{number}{10}) (\bibinfo{year}{2022})
  \bibinfo{pages}{17062--17076}, \bibinfo{note}{doi:
  \url{https://doi.org/10.1109/TITS.2022.3162566}}.

\bibitem[{Peng et~al.(2019)Peng, Ye, and Chen}]{peng2019bayesian}
\bibinfo{author}{W.~Peng}, \bibinfo{author}{Z.-S. Ye},
  \bibinfo{author}{N.~Chen}, \bibinfo{title}{Bayesian deep-learning-based
  health prognostics toward prognostics uncertainty}, \bibinfo{journal}{IEEE
  Transactions on Industrial Electronics}
  \bibinfo{volume}{67}~(\bibinfo{number}{3}) (\bibinfo{year}{2019})
  \bibinfo{pages}{2283--2293}, \bibinfo{note}{doi:
  \url{https://doi.org/10.1109/TIE.2019.2907440}}.

\bibitem[{Xiang et~al.(2023)Xiang, Qin, Luo, Wu, and
  Gryllias}]{XIANG2023110187}
\bibinfo{author}{S.~Xiang}, \bibinfo{author}{Y.~Qin}, \bibinfo{author}{J.~Luo},
  \bibinfo{author}{F.~Wu}, \bibinfo{author}{K.~Gryllias}, \bibinfo{title}{A
  concise self-adapting deep learning network for machine remaining useful life
  prediction}, \bibinfo{journal}{Mechanical Systems and Signal Processing}
  \bibinfo{volume}{191} (\bibinfo{year}{2023}) \bibinfo{pages}{110187}, ISSN
  \bibinfo{issn}{0888-3270}, \bibinfo{note}{doi:
  \url{https://doi.org/10.1016/j.ymssp.2023.110187}}.

\bibitem[{Xu et~al.(2019)Xu, Baraldi, Al-Dahidi, and Zio}]{uqp5}
\bibinfo{author}{M.~Xu}, \bibinfo{author}{P.~Baraldi},
  \bibinfo{author}{S.~Al-Dahidi}, \bibinfo{author}{E.~Zio},
  \bibinfo{title}{Fault Prognostics by an Ensemble of Echo State Networks in
  Presence of Event Based Measurements}, \bibinfo{journal}{Engineering
  Applications of Artificial Intelligence} \bibinfo{volume}{87}
  (\bibinfo{year}{2019}) \bibinfo{pages}{103346}, \bibinfo{note}{doi:
  \url{https://doi.org/10.1016/j.engappai.2019.103346}}.

\bibitem[{Zgraggen et~al.(2022)Zgraggen, Pizza, and Huber}]{uqp6}
\bibinfo{author}{J.~Zgraggen}, \bibinfo{author}{G.~Pizza},
  \bibinfo{author}{L.~G. Huber}, \bibinfo{title}{Uncertainty Informed Anomaly
  Scores with Deep Learning: Robust Fault Detection with Limited Data}, in:
  \bibinfo{booktitle}{PHM Society European Conference},
  vol.~\bibinfo{volume}{7}, \bibinfo{pages}{530--540}, \bibinfo{note}{doi:
  \url{https://doi.org/10.36001/phme.2022.v7i1.3342}}, \bibinfo{year}{2022}.

\bibitem[{Liao et~al.(2018)Liao, Zhang, and Liu}]{uqp8}
\bibinfo{author}{Y.~Liao}, \bibinfo{author}{L.~Zhang},
  \bibinfo{author}{C.~Liu}, \bibinfo{title}{Uncertainty prediction of remaining
  useful life using long short-term memory network based on bootstrap method},
  in: \bibinfo{booktitle}{2018 IEEE International Conference on Prognostics and
  Health Management (ICPHM)}, \bibinfo{organization}{IEEE},
  \bibinfo{pages}{1--8}, \bibinfo{note}{doi:
  \url{https://doi.org/10.1109/ICPHM.2018.8448804}}, \bibinfo{year}{2018}.

\bibitem[{Rigamonti et~al.(2017)Rigamonti, Baraldi, Zio, Roychoudhury, Goebel,
  and Poll}]{Rigamonti2017EnsembleOO}
\bibinfo{author}{M.~G. Rigamonti}, \bibinfo{author}{P.~Baraldi},
  \bibinfo{author}{E.~Zio}, \bibinfo{author}{I.~Roychoudhury},
  \bibinfo{author}{K.~Goebel}, \bibinfo{author}{S.~Poll},
  \bibinfo{title}{Ensemble of optimized echo state networks for remaining
  useful life prediction}, \bibinfo{journal}{Neurocomputing}
  \bibinfo{volume}{281} (\bibinfo{year}{2017}) \bibinfo{pages}{121--138},
  \bibinfo{note}{doi: \url{https://doi.org/10.1016/j.neucom.2017.11.062}}.

\bibitem[{Biggio et~al.(2021)Biggio, Wieland, Chao, Kastanis, and Fink}]{uqp2}
\bibinfo{author}{L.~Biggio}, \bibinfo{author}{A.~Wieland},
  \bibinfo{author}{M.~A. Chao}, \bibinfo{author}{I.~Kastanis},
  \bibinfo{author}{O.~Fink}, \bibinfo{title}{Uncertainty-Aware Prognosis via
  Deep Gaussian Process}, \bibinfo{journal}{IEEE Access} \bibinfo{volume}{9}
  (\bibinfo{year}{2021}) \bibinfo{pages}{123517--123527}, \bibinfo{note}{doi:
  \url{https://doi.org/10.1109/ACCESS.2021.3110049}}.

\bibitem[{Ellis et~al.(2022)Ellis, Heyns, and Schmidt}]{ELLIS2022108805}
\bibinfo{author}{B.~Ellis}, \bibinfo{author}{P.~S. Heyns},
  \bibinfo{author}{S.~Schmidt}, \bibinfo{title}{A hybrid framework for
  remaining useful life estimation of turbomachine rotor blades},
  \bibinfo{journal}{Mechanical Systems and Signal Processing}
  \bibinfo{volume}{170} (\bibinfo{year}{2022}) \bibinfo{pages}{108805},
  \bibinfo{note}{doi: \url{https://doi.org/10.1016/j.ymssp.2022.108805}}.

\bibitem[{Jankowiak et~al.(2020)Jankowiak, Pleiss, and Gardner}]{dspp}
\bibinfo{author}{M.~Jankowiak}, \bibinfo{author}{G.~Pleiss},
  \bibinfo{author}{J.~R. Gardner}, \bibinfo{title}{Deep Sigma Point Processes}
  \urlprefix\url{https://arxiv.org/abs/2002.09112}.

\end{thebibliography}

\appendix
\section{Some further discussions on Gaussian process regression}
\subsection{An extended discussion on kernels}
\label{sec:gpr_kernels_extended}

The class of Mat\'ern kernels represents a very general class of covariance functions, of which the squared exponential kernel is a special case. It offers a broad class of kernels with varying values of a smoothness parameter $\nu>0$ that controls the smoothness of the resulting approximation of the underlying function~\cite{rasmussen2006gaussian}. The Mat{\'e}rn covariance between the function outputs at two points are described as \cite{rasmussen2006gaussian}
\begin{equation}
k(\mathbf{x}, \mathbf{x}') = \sigma_f^2 \frac{1}{\Gamma(\nu) 2^{\nu-1}} \left(\frac{\sqrt{2\nu}}{\ell} dist\left(\mathbf{x}, \mathbf{x}'\right)\right)^{\nu} K_{\nu}\left(\frac{\sqrt{2\nu}}{\ell} dist\left(\mathbf{x}, \mathbf{x}'\right)\right)
\label{eq:MaternKernel}
\end{equation}
where $\Gamma(\cdot)$ is the Gamma function, $dist(\mathbf{x}, \mathbf{x}')$ is the Euclidean distance between points $\mathbf{x}$ and $\mathbf{x}'$, i.e., $dist(\mathbf{x}, \mathbf{x}') = |\mathbf{x} - \mathbf{x}'| = \sqrt{\sum_{d=1}^{D}(x_d - x_d’)^2}$, and $K_\nu$ is the modified Bessel function of the second kind and order $\nu$. A larger value of $\nu$ results in a smoother appropriated function. When $\nu \to \infty$, the Mat\'ern kernel becomes the squared exponential kernel. Another special case worth mentioning is when $\nu = 1/2$, the Mat\'ern kernel is equivalent to the absolute exponential kernel (sometimes also called the Ornstein-Uhlenbeck process kernel), which can be expressed as 
\begin{equation}
k(\mathbf{x}, \mathbf{x}') = \sigma_f^2 \exp{\left(-\frac{dist\left(\mathbf{x}, \mathbf{x}'\right)}{l} \right)}.
\label{eq:aekernel}
\end{equation}
GPR using this Mat\'ern 1/2 kernel yields rather unsmooth (rough) functions sampled from the Gaussian process prior and posterior. Additionally, observations do not inform predictions on input points far away from the points of observations, leading to poor generalization performance of the resulting GPR model. Two other special cases of the Mat\'ern kernels are $\nu = 3/2$ and $\nu = 5/2$. The resulting Mat\'ern 3/2 kernel and Mat\'ern 5/2 kernel are not infinitely differentiable, unlike the squared exponential kernel, but at least once (Mat\'ern 3/2) or twice differentiable ($\nu = 5/2$). These two kernels may be useful in cases where intermediate solutions between the unsmooth Mat\'ern 1/2 kernel and the perfectly smooth squared exponential kernel are needed to approximate functions that are expected to be somewhat smooth yet not perfectly smooth. 

The Mat\'ern kernel in Eq. (\ref{eq:MaternKernel}) has a single length scale $l$ and is of an isotropic form. Like the ARD squared exponential kernel shown in Eq. (\ref{eq:ardsekernel}), an anisotropic variant of the Mat\'ern kernel can be defined by introducing $D$ length scales, each depicting the relevance of an input dimension. The resulting ARD Mat\'ern kernel has a slightly modified term, $\sqrt{\sum_{d=1}^{D}\frac{(x_d - x_d')^2}{l_d^2}}$, in place of the original term, $\frac{\sqrt{\sum_{d=1}^{D}(x_d - x_d')^2}}{l}$ (i.e., $\frac{dist(\mathbf{x}, \mathbf{x}')}{l}$ in Eq. (\ref{eq:MaternKernel})). For $D$-dimensional input $\mathbf{x} \in \mathcal{X} \subseteq \mathbb{R}^d$, an anisotropic kernel is composed of $(D+1)$ hyperparameters, $\sigma_f,l_1,\dots,l_D$. 

\begin{figure}[!ht]
\includegraphics[width=\textwidth]{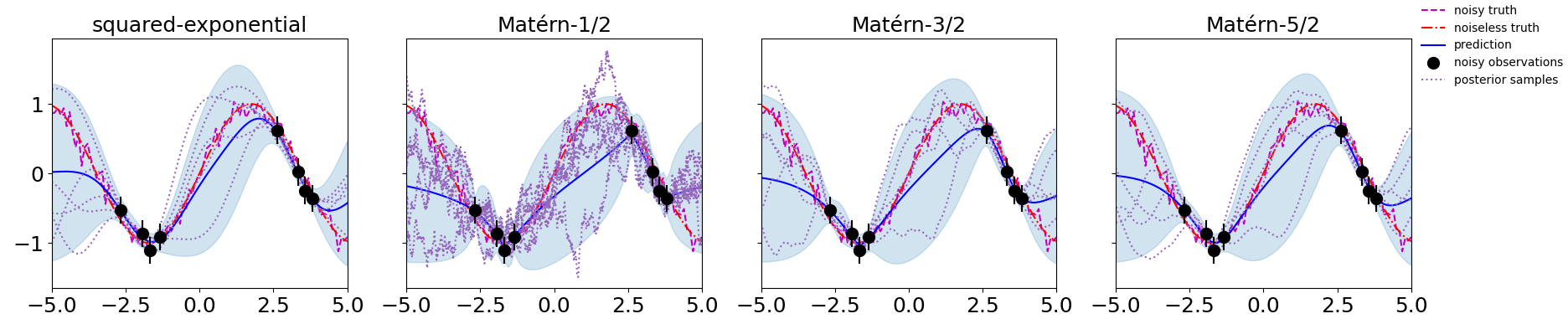}
\caption{Comparison of GPR models built using multiple kernels: squared-exponential $\left(\nu\to\infty\right)$, Mat{\'e}rn1/2 $\left(\nu = \frac{1}{2}\right)$, Mat{\'e}rn3/2 $\left(\nu = \frac{3}{2})\right)$, Mat{\'e}rn5/2 $\left(\nu = \frac{5}{2}\right)$, with the same eight training data points, along with five samples randomly drawn from the posterior.}
\label{fig:GPdiffKernels}
\end{figure}

\begin{figure}[!ht]
    \centering
    \includegraphics[scale=0.89]{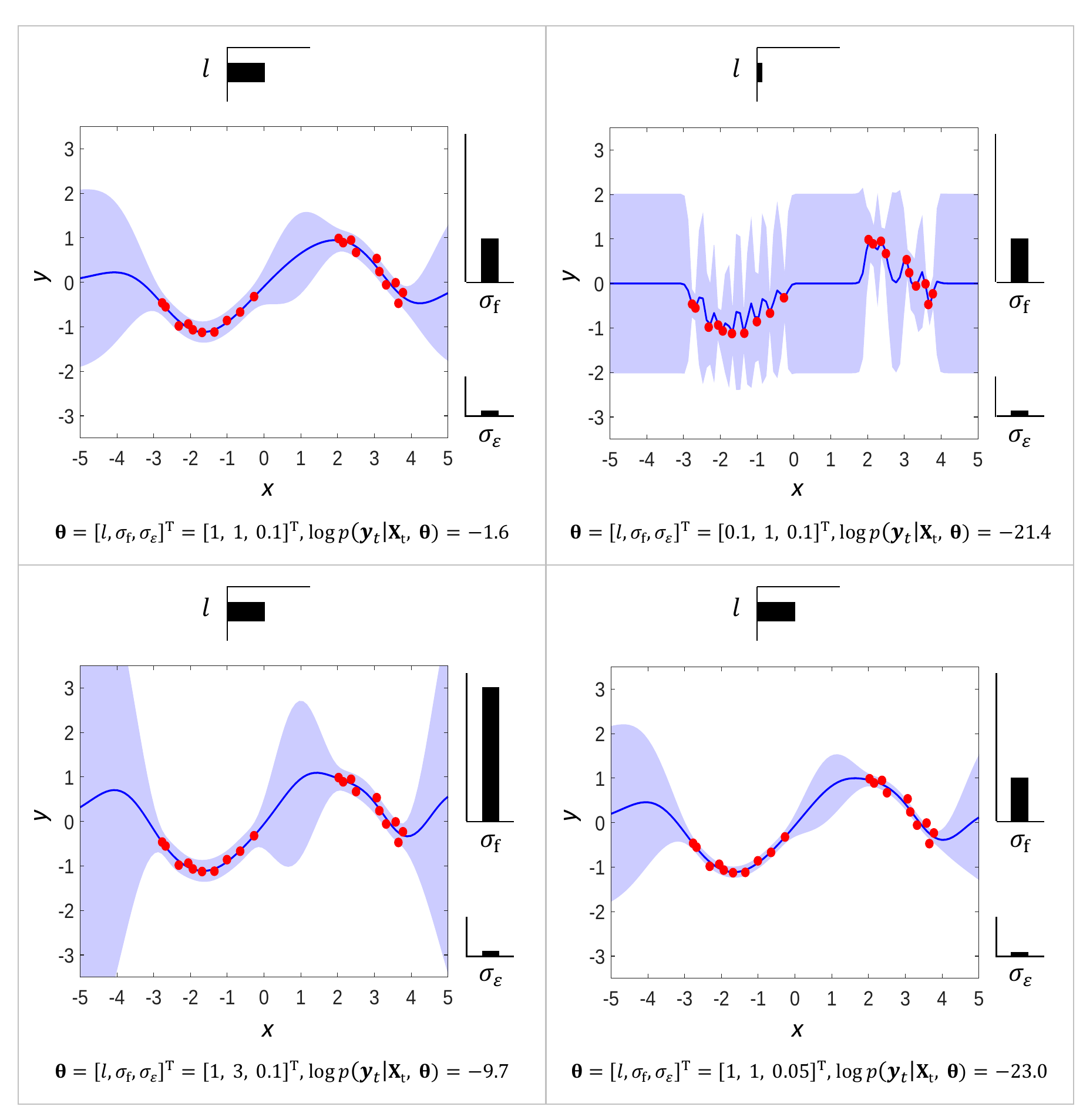}
    \caption{Effect of hyperparameters on the Gaussian process posterior for the 1D toy example used in Fig. \ref{fig:samples_prior_posterior}. Note that the confidence intervals shown collectively as light blue shade are derived from the posterior of (noisy) observations (function output plus noise); they are slightly wider than the confidence intervals for the underlying function shown in Fig. \ref{fig:samples_prior_posterior} due to the added Gaussian noise (see the discussion below Eqs. (\ref{eq:posteriorGPMeanWithZeroMean}) and ({\ref{eq:posteriorGPKernelWithZeroMean}}) in Sec. \ref{sec:basics_gpr}.d).}
    \label{fig:hyperparameters_effect}
\end{figure}

To illustrate the concept of kernels, Fig.~\ref{fig:GPdiffKernels} compares GPR models built using multiple commonly used kernels in a 1D example. As demonstrated in this figure, the squared-exponential kernel produces the smoothest GPR, whereas Mat{\'e}rn1/2 produces the roughest GPR (where the samples drawn from the posterior are equivalent to a Brownian motion). 
The intuition is that the larger the $\nu$ value, the smoother the underlying function. Specifically, when $\nu = 1/2$, the Gaussian process sampled from posterior with this kernel (Mat{\'e}rn1/2) corresponds to a Brownian motion (or equivalently, a Wiener process), whereas $\nu \to \infty$ smoothens the sampled Gaussian process because the posterior mean is infinitely differentiable (i.e., $\mathcal{C}^{\infty}$)~\cite{rasmussen2006gaussian}. 
The noiseless ground truth, $f(x) = \sin(0.9x)$, is plotted as dot-dashed magenta lines. Each noisy observation used for training is obtained based on the following observation model: $y = f(x) + \varepsilon$, where the Gaussian noise $\varepsilon \sim \mathcal{N}(0,0.1^2)$. Eight training observations are plotted as black dots, and five samples randomly drawn from the GPR posterior are plotted as dotted purple lines.

\subsection{Parametric study on effect of hyperparameter optimization}
\label{sec:gpr_hyperparameters_effect}
Figure \ref{fig:hyperparameters_effect} illustrates the effect of $l$, $\sigma_\mathrm{f}$, and $\sigma_\varepsilon$ on the Gaussian process posterior of observations $\mathbf{y}_*$ (each being function output $f$ plus noise $\varepsilon$) for the 1D toy example used in Fig. \ref{fig:samples_prior_posterior}. In each of the four cases considered, the values of the three hyperparameters and log marginal likelihood (see Eq. \ref{eq:LogMarginalLikelihood}) are shown right below the regression plot. In all four cases, the observation ($\mathbf{y}_*$) posterior has the same mean curve as the function ($\mathbf{f}_*$) posterior but a slightly larger variance at any input point due to the non-zero noise variance $\sigma_{\varepsilon}^2$, as discussed in Sec. \ref{sec:basics_gpr}.d. The length scale determines how quickly the correlation between the function values at two input points decays as they become farther away. Too small of an $l$ value (e.g., $l = 0.1$ in Fig. \ref{fig:hyperparameters_effect}) leads to an approximation that varies too quickly horizontally and yields too wide of uncertainty regions between training points. The signal amplitude $\sigma_\mathrm{f}$ depicts the maximum vertical variation of functions/observations drawn from the Gaussian process. A larger $\sigma_\mathrm{f}$ value (e.g., $\sigma_\mathrm{f} = 3$ in Fig. \ref{fig:hyperparameters_effect}) results in a larger maximum width of the confidence interval for a test point between or away from training points. It is an important hyperparameter for quantifying epistemic uncertainty, although it is difficult to derive an optimum value solely based on training data. The signal standard deviation $\sigma_\varepsilon$ controls the amount of (input-independent) noise in the observations. Too small of a $\sigma_\varepsilon$ value (e.g., $\sigma_\varepsilon = 0.05$ in Fig. \ref{fig:hyperparameters_effect}) results in an approximation that fails to capture the observational noise (aleatory uncertainty).

\subsection{Connections with neural networks and recent development}
\label{sec:gpr_nn}
Efforts to draw connections between GPR and neural networks dated back more than two decades, with the first study showing the equivalence between a Gaussian process and a fully-connected neural network with a single, \emph{infinite-width} hidden layer and an i.i.d. prior over the network parameters (weights and biases) \citep{Neal1996}. This equivalence is significant because using a Gaussian process prior over functions allows one to perform Bayesian inference in its exact form on neural networks using simple matrix operations (see the familiar formulae for Gaussian process posterior in Eqs. (\ref{eq:posteriorMeanWithZeroMean}) and (\ref{eq:posteriorVarianceWithZeroMean})) \citep{williams1996computing}. One obvious benefit is that one does not need to resort to iterative, more computationally expensive training algorithms, such as gradient descent and stochastic gradient descent, or approximate Bayesian inference methods for Bayesian neural networks (see Sec. \ref{sec:bnn}). As deep learning has been gaining popularity in recent years, significant extensions were made to draw such connections for standard DNNs \citep{lee2017deep} and DNNs with convolutional filters, or so-called deep convolutional neural networks \citep{novak2018bayesian, garriga2018deep}. 

\begin{figure}[!ht]
    \centering
    \includegraphics[scale=0.89]{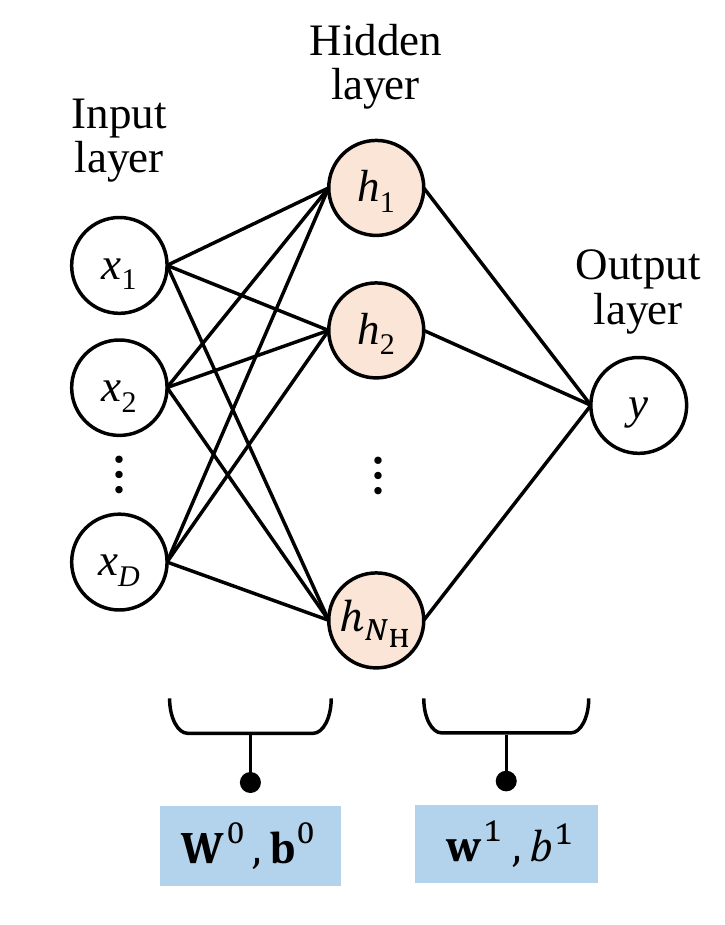}
    \caption{A single-hidden-layer neural network where the number of hidden units $N_\mathrm{H}$ could approach infinity, i.e., $N_\mathrm{H} \to \infty$. $\mathbf{W}^0$ and $\mathbf{b}^0$ conveniently denote the $N_\mathrm{H} \times D$ matrix of input-to-hidden weights and the vector of $ N_\mathrm{H}$ input-to-hidden biases. Similarly, $\mathbf{w}^1$ denote the vector of $ N_\mathrm{H}$ hidden-to-output weights, again, for notational convenience purposes.}
\label{fig:single_hidden_layerNN}
\end{figure}

Let us now briefly review the early work in~\cite{Neal1996}. We consider a fully-connected neural network with one hidden layer, illustrated in Fig. ~\ref{fig:single_hidden_layerNN}. To get to each hidden node $h_j$, $1 \leq j \leq N_\mathrm{H}$, where $N_\mathrm{H}$ is the number of hidden units, we first apply a linear transformation of input point $\mathbf{x}$ and then a nonlinear operation using an activation function $\psi(\cdot): \mathbb{R}^D \mapsto \mathbb{R}$. The resulting $j$-th hidden unit takes the following form:
\begin{equation}
h_j(\mathbf{x}) = \psi\left( b_j^{0} + \sum_{d=1}^D w_{dj}^{0} x_d \right),
\end{equation}
where $w^{0}_{dj}$ denotes the input-to-hidden weight from $x_d$ to $h_j$ and $b_j^{0}$ is the input-to-hidden bias for $h_j$. 
To get to the output node $y$ (assuming zero observation noise for simplicity, i.e., $y(\mathbf{x}) = f(\mathbf{x})$), we apply another linear transformation of the hidden units with hidden-to-output weights and a bias
\begin{equation}
y(\mathbf{x}) = b^{1} + \sum_{j=1}^{N_\mathrm{H}} w_{j}^{1} h_j(\mathbf{x}),
\label{eq:nn_outputnode}
\end{equation}
where $w^{1}_{j}$ denotes the hidden-to-output weight from $h_j$ to $y$, and $b^{1}$ is the hidden-to-output bias.

We assume (1) the prior of the hidden-to-output weights $w^{1}_{j}$ and bias $b$ follows independent zero-mean (often Gaussian) distributions with variances being $\sigma_{w^{1}}^2$ and $\sigma_b^2$, respectively, and (2) the input-to-hidden weights $w^{0}_{dj}$ and biases $b_j^{0}$ are i.i.d. It follows that the network output $y(\mathbf{x})$ in Eq. (\ref{eq:nn_outputnode}) is a summation over $(N_\mathrm{H} + 1)$ i.i.d. random variables \citep{Neal1996}. Based on the Central Limit Theorem, when $N_\mathrm{H} \to \infty$, i.e., when the width of the hidden layer approaches infinity, $\widehat y(\mathbf{x})$ will follow a Gaussian distribution. This Gaussian prior holds regardless of the distribution types of the $(N_\mathrm{H} + 1)$ random variables in the sum. Let us move on to look at any finite set of input points, $\mathbf{x}_1, \dots, \mathbf{x}_{N_*}$. As $N_\mathrm{H} \to \infty$, their network outputs, $\widehat y_1, \dots, \widehat y_{N_*}$, will be jointly Gaussian, according to the multidimensional Central Limit Theorem. It means that the joint distribution of the network outputs at any finite collection of input points is multivariate Gaussian, which exactly matches the definition of a Gaussian process discussed in Sec. \ref{sec:basics_gpr}.a. Thus, $\widehat y(\mathbf{x}) \sim \mathcal{GP}(m_\mathrm{nn}(\mathbf{x}),k_\mathrm{nn}(\mathbf{x}, \mathbf{x}'))$, a Gaussian process with the mean function $m_\mathrm{nn}(\cdot)$ and covariance function $k_\mathrm{nn}(\cdot)$. Since the hidden-to-output weights $w_{j}^{1}$ and bias $b^{1}$ have zero means, $m_\mathrm{nn} \equiv \mathbb{E}[\widehat y(\mathbf{x})] = 0$. The covariance function can be derived based on i.i.d. conditions and takes the following form:
\begin{equation}
k_\mathrm{nn}(\mathbf{x},\mathbf{x}') \equiv \mathbb{E}\left[\widehat y(\mathbf{x}) \widehat y(\mathbf{x}') \right] = \sigma_{b^1}^2 + \sum_{j=1}^{N_\mathrm{H}} \sigma^2_{w^{1}} \mathbb{E}\left[ h_j(\mathbf{x}) h_j(\mathbf{x}')  \right]
= \sigma_{b^1}^2 + \underbrace{N_\mathrm{H} \sigma_{w^{1}}^2}_{\omega^2} \underbrace{ \mathbb{E}\left[ h_j(\mathbf{x}) h_j(\mathbf{x}')  \right]}_{\mathcal{C}(\mathbf{x}, \mathbf{x}')},
\label{eq:nn_gpr_kernel}
\end{equation}
where the prior variance $\sigma_{w^{1}}^2$ of each hidden-to-output weight is set to scale carefully as $\omega^2 / N_\mathrm{H}$ for some fixed ``unscaled” variance $\omega^2$ and $\mathcal{C}(\mathbf{x}, \mathbf{x}')$ need to be evaluated for all $\mathbf{x}$ in the training set and all $\mathbf{x}’$ in the training and test sets. $\mathcal{C}(\mathbf{x}, \mathbf{x}')$ has an analytic form for certain types of activation functions such as the error function (or Gaussian nonlinearities) \citep{williams1996computing,rasmussen2006gaussian}, one-sided polynomial functions \citep{cho2009kernel}, and ReLU (rectified linear unit) \citep{lee2017deep}. As a result, infinitely wide Bayesian neural networks give rise to a new family of GPR kernels. An interesting and attractive property of these neural networks is that all network parameters are often initialized as independent zero-mean Gaussians, some with properly scaled variances, and the kernel parameters (e.g., ``unscaled” prior variances of weights and prior variances of biases) may be the only parameters that need to be optimized.

What has been discussed in this subsection represents a category of approaches for combining the strengths of GPR (exact Bayesian inference, distance awareness, etc.) with those of neural networks (feature extraction from high-dimensional inputs  (large $D$), ability to model nonlinearities, etc.). These approaches explore the direct theoretical relationship between infinitely wide neural networks and GPR. Another category of approaches uses GPR with standard kernels (such as the squared exponential kernel in Eq. (\ref{eq:sekernel})) whose inputs are feature representations in the hidden space learned by a neural network \citep{wilson2016deep, liu2020simple, fortuin2021deep, van2021feature}.  These approaches are often called \emph{deep kernel learning}. The network weights, biases, and GPR kernel parameters can be jointly optimized end-to-end, which is straightforward to implement using gradient descent or stochastic gradient descent. These approaches excel in OOD detection thanks to the distance awareness property of GPR and offer a solution to improving the scalability of GPR to high-dimensional inputs. A drawback is that overparameterization associated with a DNN (e.g., a deep convolutional neural network) may make the network prone to overfitting. Another issue is \emph{feature collapse} \citep{van2020uncertainty}, which needs to be carefully addressed to preserve input distances in the hidden space. This issue will be discussed along with a representative approach in this category called spectral-normalized neural Gaussian process (SNGP) in Sec. \ref{sec:deterministic_uq}. A third category of approaches aims to mimic the many-layer architecture of a DNN by stacking Gaussian processes on top of one another in a hierarchical form \citep{damianou2013deep,  bui2016deep, salimbeni2017doubly, havasi2018inference}. The resulting deep Gaussian processes are probabilistic ML models with the UQ capability brought in by GPR and the added flexibility to learn complex mappings from datasets that can be small or large. However, the performance gains over standard GPR comes at a cost: exact Bayesian inference by deep Gaussian processes can be prohibitively expensive due to the computationally demanding need to compute the inverse and determinant of the covariance matrix. Therefore, almost all deep Gaussian process approaches adopt appropriate inference techniques for efficient model training that use only a small set of the so-called inducing points to build covariance matrixes \citep{bui2016deep, salimbeni2017doubly, havasi2018inference}.

\section{UQ of ML models in engineering design}
\label{sec:UQ_design}

\subsection{Needs of ML models in engineering design}
\label{sec:needs_ML}

In recent years, the rapid advancement of high-performance computing and data analytics techniques has made ML a game changer for engineering design. In particular, ML enables engineers and designers to relax simplifications and assumptions that are usually needed in conventional design paradigms~\cite{fuge2014machine,panchal2019machine}, accelerate the design process by shortening the required design cycles~\cite{vale2003machine}, and handle the design of highly complex systems with large numbers of design variables \cite{fan2020finding,jiang2020machine}. These benefits provided by data-driven ML models are particularly appealing for simulation-based engineering system design, which usually entails costly simulations.

\begin{figure}[!ht]
    \centering
    \includegraphics[scale=0.86]{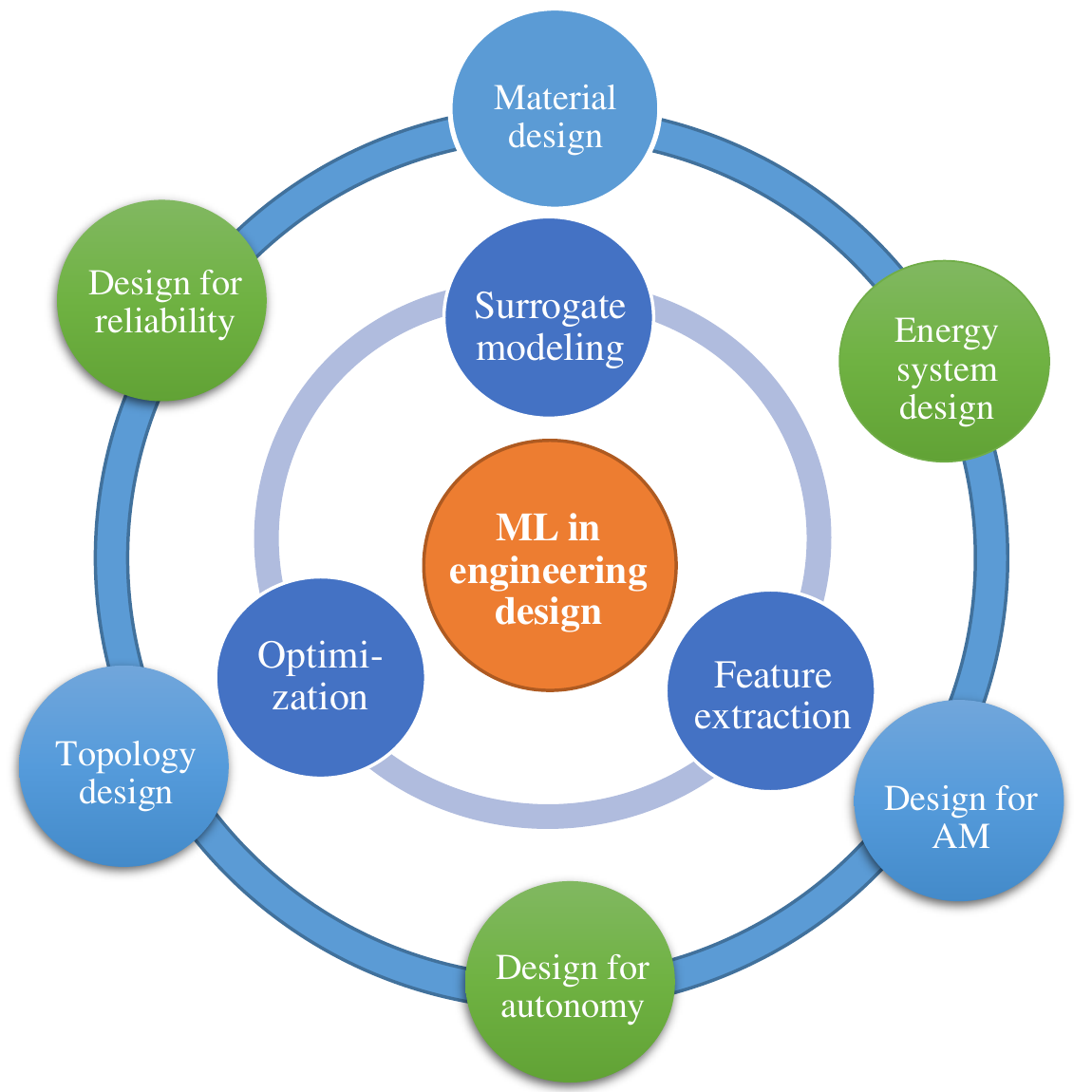}
    \caption{ML-enabled techniques in engineering design and applications.}
    \label{fig:engineering_design}
\end{figure}

As shown in Fig. \ref{fig:engineering_design}, ML revolutionizes engineering design mainly through three categories of ML-enabled capabilities: feature extraction, surrogate modeling, and optimization. Approaches in each of these three categories have been applied to solve challenging engineering design problems in various applications, such as discovery and design of engineering materials~\cite{moosavi2020role,tao2021machine}, design for reliability \cite{moustapha2019surrogate}, energy system design \cite{perera2019machine}, and topology optimization \cite{lei2019machine}, to name a few.

\begin{enumerate}[label=\roman*.]
        \item {\bf{Feature extraction}}: Extracting informative features from massive volumes of raw data is a representative use case of ML in engineering design. In this regard, ML, particularly deep learning, has become more and more prevalent in engineering design due to its salient characteristic of automatically extracting feature representations from high-dimensional data in its raw form. Specifically, in the context of engineering design, the powerful representation learning ability has been frequently utilized in two types of design activities, namely (1) dimension reduction, which is to reduce the dimensionality of design problems, and (2) generative design, which is to generate candidate designs subject to certain design constraints ~\cite{hinton2006reducing,qian2022adaptive,regenwetter2022deep}. 

        \begin{enumerate}
            \item For {\it {dimension reduction}}, autoencoder, as an unsupervised learning technique, has been commonly adopted to learn efficient codings and compressed knowledge representations from unlabeled data~\cite{hinton2006reducing}. More specifically, an autoencoder consists of an encoder and a decoder: the encoder transforms high-dimensional data into a low-dimensional representation through a \lq\lq bottleneck" layer of neurons, while the decoder recovers the high-dimensional data from the low-dimensional code. The encoder and decoder are trained together to minimize the discrepancy between the original data and its reconstruction. Due to their powerful representation capacity, autoencoders and their variants (e.g., sparse autoencoders and variational autoencoders (VAEs)) have been actively employed to extract important features, supporting diverse engineering design tasks~\cite{hamdia2019novel}. 

            \item For {\it generative design}, researchers have investigated ML approaches to aid the design process through automatic design synthesis. In short, generative design is an iterative process of using algorithms to facilitate the exploration of thousands of design variants as guided by the parameters outlined in the study setup to approach an optimal design that meets the performance target. Towards this end, ML has contributed substantially to automating the process of generative design, which is often referred to as automatic design synthesis in the design community. In essence, automatic design synthesis is to learn a generative model from existing designs and then generate new designs meeting design requirements (e.g., performance targets and cost constraints) based on the compact representations of training data in the hidden space. In particular, VAEs and GANs are two popular classes of ML algorithms for generative design~\cite{yang2018microstructural}.
        \end{enumerate}
        
        \item {\bf{Surrogate modeling}}: It is a process of using ML models as emulators of computationally expensive computer simulation models in engineering design \cite{jin2001comparative}. With the development of computational mechanics and advanced numerical solvers, computer simulations are getting increasingly sophisticated. The high-fidelity computer simulations allow us to accurately predict complicated physical phenomena without performing large numbers of expensive physical experiments, thereby accelerating the design of engineering systems to meet mission-specific requirements. Although high-fidelity simulations significantly enhance our predictive capability, they present notable challenges to engineering design due to the high computational demand and burden often associated with them. ML models play a vital role in addressing this challenge by maintaining the same predictive capability level as high-fidelity simulations while significantly reducing the computational effort required to make high-fidelity predictions \cite{alizadeh2020managing}. The basic idea of ML-enabled surrogate modeling is to replace an expensive-to-evaluate high-fidelity simulation model with a much ``cheaper" mathematical surrogate, essentially an ML model. Over the past few decades, various surrogate modeling methods have been proposed for different purposes within engineering design, including model calibration \cite{kennedy2001bayesian}, reliability analysis \cite{bichon2008efficient}, sensitivity analysis \cite{cheng2020surrogate}, and optimization \cite{chatterjee2019critical}. These existing surrogate modeling methods can be broadly classified into two groups:
        \begin{enumerate}
            \item {\it {Global surrogate modeling for general purposes:}} This class of surrogate models is constructed for the general purpose of design optimization and tries to achieve a good prediction accuracy in the whole design region of interest \cite{viana2009multiple,jin2003use,jin2002sequential}. More specifically, let us use $\hat y = \hat G({\bf{x}})$ to represent the surrogate model of a computer simulation model $y = G({\bf{x}}), {\bf{x}} \in {\Omega _{\bf{x}}}$, where ${\Omega _{\bf{x}}}$ is the prediction domain of the inputs. In global surrogate modeling, we are concerned about the prediction accuracy of $\hat y = \hat G({\bf{x}})$ for all ${\bf{x}} \in {\Omega _{\bf{x}}}$. Because of this, the training data for ML model construction needs to spread throughout the whole prediction domain ${\Omega _{\bf{x}}}$, with those in nonlinear regions being denser and the others in relatively smoother regions being more sparse. Various sampling techniques have been developed to efficiently construct globally accurate surrogate models using ML. Some examples of the techniques include MSE-based methods, the A-optimality criterion, and maximin scaled distance approaches \cite{jin2001comparative}. The goal of global surrogate modeling is to construct a surrogate that is fully representative of the original computer simulation model. Since the surrogate model is not constructed for any specific purposes and the prediction accuracy has been verified for all ${\bf{x}} \in {\Omega _{\bf{x}}}$, it can be used for any purposes, such as design optimization, uncertainty analysis, and sensitivity analysis, after its construction. In addition, the UQ calibration metrics presented in Sec.~\ref{sec:calibration_metrics} and Sec.~\ref{sec:nll} can be used to quantify the prediction accuracy of a global surrogate model, if the test data is representative of the design domain ${\Omega _{\bf{x}}}$. 
            \item {\it {Local surrogate modeling for specific purposes:}} Instead of achieving good prediction accuracy in the whole design region, this group of surrogate models only focuses on prediction in very localized design regions, such as the limit state regions in design for reliability problems \cite{bichon2008efficient,hu2016single,gaspar2014assessment,zhang2019reif} and important regions for model calibration purposes \cite{yan2019adaptive}. In local surrogate modeling, we are concerned about the prediction accuracy of $\hat y = \hat G({\bf{x}})$ for ${\bf{x}} \in {\tilde \Omega _{\bf{x}}}$, where ${\tilde \Omega _{\bf{x}}} \subset {\Omega _{\bf{x}}}$ is a subset of the prediction domain of the inputs. This sub-domain ${\tilde \Omega _{\bf{x}}}$ varies with the specific purpose of the surrogate modeling. For example, when the surrogate model is constructed for the purpose of reliability analysis, which is a classification problem, ${\tilde \Omega _{\bf{x}}}$ will be the regions along the limit state or classification boundary. When the surrogate model is constructed for optimization, ${\tilde \Omega _{\bf{x}}}$ will be the regions where the optima locate. As a result, the training data for surrogate modeling will be concentrated in those localized regions instead of spreading evenly throughout the whole prediction domain of the inputs. Because we only concentrate on a sub-domain ${\tilde \Omega _{\bf{x}}}$ of the input space, ${\Omega _{\bf{x}}}$, the local surrogate model $\hat y = \hat G({\bf{x}}), {\bf{x}} \in {\tilde \Omega _{\bf{x}}}$ only partially represents the original simulation model (i.e., the surrogate is an accurate representation of the simulation model only in the sub-domain of the design space). Moreover, since the sub-domain ${\tilde \Omega _{\bf{x}}}$ is usually unknown during the construction of the surrogate model, learning functions (also called acquisition functions in some methods) are needed to identify these localized sub-domains adaptively based on the currently available information about the underlying simulation model (ground truth). Because the surrogate model is constructed for a specific purpose (e.g.,  model calibration, reliability analysis, or optimization), its accuracy also needs to be quantified using metrics tailored for that specific purpose. For example, a metric used to check the prediction accuracy of the surrogate model for reliability analysis may not be appropriate for constructing a surrogate model for design optimization.
        \end{enumerate} 
        
        \item {\bf{Optimization}}: Engineering design problems are essentially optimization problems. Conventional gradient-based optimizers often have difficulties in finding global optima. Even though evolutionary optimization methods can overcome some of the limitations of gradient-based optimizers, the former methods are likely to require much larger numbers of function evaluations, which could become prohibitively costly for high-fidelity simulation models in many engineering design problems. ML-based or ML-assisted optimization methods have been proposed to tackle this challenge, resulting in a new family of optimization methods collectively named gradient-free ML-based optimization. One representative example of this family is Bayesian optimization \cite{jones1998efficient}. ML-based optimization transforms the way that engineering systems are designed in many fields, such as new materials \cite{zhang2020bayesian}. 
        It is worth noting that the Materials Genome Initiative~\cite{national2011materials,lander2021materials,lander2021materials,mcdowell2019creating,de2019new,christodoulou2021second}, firstly debuted in 2011, was embedded in the context of designing new materials using ML and optimization to significantly reduce the research and development time. 
        Moreover, the development of deep learning methods in recent years even allows designers to bypass complicated design optimization by directly generating candidate designs for a particular application. Some examples include the ML-based topology optimization \cite{sasaki2019topology,kallioras2020accelerated} and deep learning-enabled design of large-scale complex networks \cite{xiao2019self}.

\end{enumerate}

\subsection{Role of UQ of ML models in engineering design}
An indispensable step for the above-reviewed three categories of ML-enabled techniques (i.e., feature extraction, surrogate modeling, and optimization) is UQ of ML models. For example, for ML-enabled feature extraction in engineering design, quantifying the predictive uncertainty of ML models play an important role in (1) ensuring the extracted features are representative of the original data sources, (2) eliminating the ill-posedness of inverse problems in generative design, and (3) accounting for variability across input features. 

For surrogate modeling in engineering design, an essential step in building an accurate surrogate model (global or local surrogate) is the collection of training data. However, an initial set of training data is usually insufficient to build a surrogate model with satisfactory prediction accuracy. A subsequent refinement step sometimes is needed to improve the prediction accuracy of the surrogate model. Due to the high computational effort required to collect training data from high-fidelity simulations in engineering design, it is desirable to reduce the number of training data points or refinement iterations for surrogate modeling as much as possible. Over the past few decades, numerous refinement strategies have been developed in engineering design to minimize the number of iterations in collecting training data for the purpose of improving the performance of surrogate models. Even though these refinement strategies may differ from each other, they share one notable starting point: {\it quantifying the predictive uncertainty of the surrogate model for any given input}. 



For instance, the most commonly used refinement method for global surrogate modeling is to identify new training data by maximizing the variance of the prediction of the surrogate model ~\cite{jin2002sequential}. That is a mean squared error-method as mentioned above in \ref{sec:needs_ML}. In a GPR model, the variance of the prediction can be directly obtained from the surrogate model. For other types of surrogate models, however, the predictive uncertainty needs to be quantified using a separate UQ method. Moreover, UQ of ML models becomes particularly important, if local surrogate models need to be constructed for engineering design. In the context of local surrogate modeling, learning functions (also called acquisition functions), such as the expected improvement (EI) function in GPR-based surrogate modeling, are required to identify new training data in critical local regions (i.e., ${\tilde \Omega _{\bf{x}}}$ mentioned in \ref{sec:needs_ML}) of the input space. The new training data will then be used to refine the surrogate. Many (20+) learning functions have been proposed in recent years for local surrogate modeling of various purposes (e.g., surrogate construction, reliability analysis, and optimization). These learning functions look into multiple quantitative metrics to examine different aspects crucial to the iterative improvement of surrogate models, such as classification error \cite{hu2016global}, information entropy \cite{li2021improved,alibrandi2022informational}, and exploitation and exploration \cite{sadoughi2018sequential}, among others. A detailed review of various learning functions for local surrogate modeling for reliability analysis is available in Ref. \cite{afshari2022machine}. To the best of our knowledge, nearly all the learning functions for local surrogate modeling heavily rely on UQ of ML models. Let us take a look at two well-known learning functions for local surrogate modeling in reliability-based design optimization: the expected feasibility function (EFF) \cite{bichon2008efficient} and the U function \cite{echard2011ak}. They are mathematically described as follows:
\begin{subequations} \label{Eq:9}
\begin{gather}
EFF({\bf{x}}) = \int_{e - \tau }^{e + \tau } {\left[ {\tau  - \left| {e - y} \right|} \right]{p_{\hat y({\bf{x}})}}(y)} dy, \label{eq:EFF} \\
U({\bf{x}}) = {{\left| {{\mu _{\hat y}}({\bf{x}}) - e} \right|} \over {{\sigma _{\hat y}}({\bf{x}})}}, \label{eq:U}
\end{gather} 
\end{subequations}
where $e$ is the failure threshold used to define the limit state, $y = e$, that separates the failure region ($y > e$) from the safe region ($y \leq e$), $\tau$ is half the width of a two-sided critical interval in the vincinity of the limit state ($y = e$), often set as two times the standard deviation of the ML model prediction, i.e., $\tau = 2{\sigma _{\hat y}}({\bf{x}})$, ${\mu _{\hat y}}({\bf{x}})$ and ${\sigma _{\hat y}}({\bf{x}})$ are, respectively, the mean and standard deviation of the ML prediction with respect to the input $\bf{x}$, and ${p_{\hat y({\bf{x}})}}(y)$ is the probability density function of $y$ for given input ${\bf{x}}$ predicted by the ML model. 

As shown in the above two equations, UQ of ML models plays an essential role in the construction of such learning functions. This observation also applies to the other learning functions in local surrogate modeling. It is commonly referred as adaptive surrogate modeling in the literature. In general, the identification of the sub-domain ${\tilde \Omega _{\bf{x}}}$ (see \ref{sec:needs_ML}) relies on the learning functions in local surrogate modeling, where UQ of ML models plays a foundational role towards the establishment of these learning functions. 

Similar to local surrogate modeling, ML-enabled optimization in engineering design also depends heavily on the ability to quantify the predictive uncertainty of ML models, which is essential for ML models to exploit and explore the design domain to efficiently identify optimal designs. Examples of such ML-based optimizers include Bayesian optimization \cite{frazier2018bayesian} and deep reinforcement learning-based optimization \cite{shen2021bayesian}. Specifically for Bayesian optimization, a trade-off between exploitation and exploration is balanced through a learning/acquisition function, which is very similar to that in local surrogate modeling discussed above. Some popular learning functions include the probability of improvement, EI, upper confidence bound, and knowledge gradient (a generalization of EI). Taking the EI function for a minimization problem as an example, this function is mathematically defined as \cite{jones1998efficient}.

\begin{equation}\label{eq:ei}
EI({\bf{x}}) = ({f_{\min }} - {\mu _{\hat y}}({\bf{x}}))\Phi \left( {{{{f_{\min }} - {\mu _{\hat y}}({\bf{x}})} \over {{\sigma _{\hat y}}({\bf{x}})}}} \right) + {\sigma _{\hat y}}({\bf{x}})\phi \left( {{{{f_{\min }} - {\mu _{\hat y}}({\bf{x}})} \over {{\sigma _{\hat y}}({\bf{x}})}}} \right),
\end{equation}
where ${f_{\min }}$ is the current best function value obtained from the existing training data \cite{jones1998efficient}. As indicated in this equation, ${{\mu _{\hat y}}({\bf{x}})}$ and ${{\sigma _{\hat y}}({\bf{x}})}$ are two essential elements of the EI function. UQ of ML models is needed to obtain these two terms, and more fundamentally, the probability distribution of $\hat y$ is required to derive a learning/acquisition function such as the EI function in Eq. (\ref{eq:ei}). Defining such a function makes it possible to accelerate design optimization through ML. This characteristic is very similar to that of learning functions in local surrogate modeling. 

In a broad sense, adaptive surrogate modeling-based design optimization can also be classified as a type of local surrogate model since a learning function is used to adaptively identify critical local regions that are important for the specific purpose of identifying a maximum or minimum. Moreover, a global surrogate model and a local surrogate model are interchangeable during the process of ML model construction. For example, we usually start with a global surrogate model in order to construct a local surrogate model because the critical local regions are unknown and need to be identified using a learning function based on the UQ of an ML model. After constructing a local surrogate model for a specific purpose (e.g., reliability analysis, optimization), we can always convert this local surrogate into a global one if we want to expand the prediction domain to the whole design domain. Regardless of whether design optimization leverages local or global surrogate modeling, UQ of ML models is almost always the foundation of the three categories of ML-enabled capabilities in engineering design described in \ref{sec:needs_ML}.

\subsection{State of knowledge and gaps}
Driven by the increasing needs of various engineering design problems (e.g., design for reliability, design for additive manufacturing, new material design, energy system design, etc) as illustrated in Fig. \ref{fig:engineering_design}, the three categories of ML-enabled techniques established upon UQ of ML models (see \ref{sec:needs_ML}) have been extensively studied in the literature. Next, we elaborate the current state-of-the-art literature and highlight research gaps that need further investigation and efforts from three aspects: feature extraction, surrogate modeling, and optimization. 

According to our literature survey, studies on feature extraction in engineering design mostly implement neural network-based approaches, such as those based on variants of autoencoders and GANs as mentioned in \ref{sec:needs_ML} ~\cite{goodfellow2020generative}. For example,~\citet{guo2018indirect} tackled the topology design of a heat conduction system using the latent representation produced by a VAE.~\citet{chen2020wireframe} trained a wireframe image autoencoder with a large database of unlabeled real-application user interface (UI) designs to serve as a UI search engine for the purpose of supporting UI design in software development.~\citet{li2022predictive} developed a target-embedding VAE neural network and explored its usage in the design of 3D car body and mugs. In recent years, the idea of using ML for automatic design synthesis has also gained increasing popularity~\cite{regenwetter2022deep,oh2019deep,regenwetter2022towards}, especially in the mechanical design community. For instance,~\citet{zhang20193d} used an unsupervised VAE to learn a generative model from a corpus of existing 3D glider designs and demonstrated the utility of the VAE in the 3D outer shape design of gliders.~\citet{chen2018b} developed a generative model established upon a GAN for synthesizing smooth curves, in which the generator first synthesized parameters for rational B\'ezier curves, and then transformed those parameters into discrete point representations. In another study, ~\citet{chen2019synthesizing} considered the interpart dependencies and proposed a GAN-based generative model for synthesizing designs by decomposing the synthesis into synthesizing each part conditioned on its corresponding parent part. The UQ methods for ML models presented in Sec.~\ref{sec:UQ_methods} can be directly applied to the aforementioned neural network models to improve the effectiveness of feature extraction in engineering design by enabling dimension reduction or generative design under uncertainty. However, as of now, only a limited number of studies have touched on topics to investigate the UQ of neural networks used in feature extraction. 

For global surrogate modeling, approaches have been investigated using various ML methods, including GPR models, neural networks (both regular artificial neural networks and DNNs), support vector regression, random forest, etc. For local surrogate modeling, however, most current approaches are developed based on GPR models. This is largely attributed to the capability of GPR to analytically quantify the predictive uncertainty in the form of a Gaussian distribution that is convenient to use. In fact, most of the learning functions for local surrogate model-based reliability analysis are derived or developed based on GPR models. For example, learning functions in closed forms as given in Eqs. (\ref{eq:EFF}) and (\ref{eq:U}) have been derived for GPR models. Quantifying the predictive uncertainty of GPR models in the Gaussian form facilitates an efficient evaluation of various learning functions for the refinement of local surrogates. In addition to GPR-based local surrogate modeling methods, a few approaches have also been proposed for local surrogate modeling based on UQ of support vector regression models \cite{song2013adaptive,basudhar2008adaptive}.  In recent years, with the rapid development of deep learning techniques and the capability of quantifying the prediction uncertainty of deep learning models, local surrogate modeling methods have been studied for deep neural networks to achieve “active learning” \cite{sener2017active,haut2018active,xiang2020active}. For instance, \citet{xiang2020active} proposed an active learning method for DNN-based structural reliability analysis by extending a weighted sampling method from GPR models to DNNs. This extension allows for selecting new training data for refining DNN models for reliability analysis. Similarly, \citet{bao2021adaptive} extended the subset sampling method to DNNs, resulting in an adaptive DNN method for structural reliability analysis. Even though active learning for local surrogate modeling has great potential in reducing the size of training data required to build accurate surrogate models, it is still in the early development stage for other ML models beyond GPR models. In particular, many existing UQ methods for deep learning models are still far from GPR's scientific rigor and theoretical soundness because few can stand strict UQ tests pertaining to uncertainty calibration, decomposition, and attribution. Additionally, even fewer methods offer principled ways to reduce the predictive uncertainty of deep neural networks. With UQ methods for ML models (as reviewed in Sec. \ref{sec:UQ_methods}) getting more and more mature, we foresee that active learning for local surrogate modeling will also become a very active research topic for ML models other than GPR models. 
 

Similar to local surrogate modeling, even though some deep learning-based optimization methods have been developed recently \cite{nguyen2020deep,asano2018optimization}, ML-enabled optimization has mostly been studied using GPR models, resulting in a group of Bayesian optimization-based engineering design methods \cite{beland2017bayesian,mathern2021multi,zhang2020bayesian}, whose applications include material design \cite{frazier2016bayesian,sharpe2018design}, design for reliability \cite{miguel2022reliability}, and design for additive manufacturing \cite{liu2022metal}. Because GPR is a flexible and versatile framework, which means it can be fairly easy to extend to other problems and applications, numerous extensions have been considered to adopt GPR models in different settings under the big umbrella of ``Bayesian optimization". These extensions include, but are not limited to, using multi-fidelity strategy to reduce the required number of high-fidelity samples in GPR-based Bayesian optimization~\cite{le2014recursive}, Bayesian optimization for multi-output response~\cite{alvarez2011kernels}, enhancing Bayesian optimization through gradient information during the construction of a GPR model~\cite{dwight2009efficient}, Bayesian optimization for problems with mixed-integer design variables (also known as mixed-variables)~\cite{tran2019constrained}, and Bayesian optimization based on heteroscedastic or non-stationary GPR models~\cite{le2005heteroscedastic,paciorek2003nonstationary,heinonen2016non,remes2017non}. 

Based on the above reviews, we can conclude that the UQ methods for ML models reviewed in Sec.~\ref{sec:UQ_methods} provide valuable tools to fill the gaps in the following three major activities of ML-based engineering design: ML-enabled feature extraction, surrogate modeling, and optimization.

\begin{enumerate}[label=\alph*.]
        \item {\bf{Enabling uncertainty-informed surrogate modeling and optimization}}: The UQ methods for neural networks presented in Secs. \ref{sec:bnn} and \ref{sec:neuralnetworkensemble} enable us to extend various local surrogate modeling and optimization methods, which are originally developed for GPR models, to various neural network-based ML models. This opportunity is especially important for deep neural networks that are gaining popularity in the engineering design community.
        \item {\bf{Accounting for aleatory uncertainty in ML-based engineering design}}: Most current methods for global surrogate modeling, local surrogate modeling, and ML-based optimization lack the capability of considering input-dependent aleatory uncertainty during the local surrogate modeling or optimization. UQ methods newly developed in the ML community such as the neural network ensemble method reviewed in Sec. \ref{sec:neuralnetworkensemble} offer opportunities to address this important issue.
        \item {\bf{Reducing computational cost}}: Computationally efficient UQ methods are needed to quantify the predictive uncertainty of ML models, since local/global surrogate modeling and its applications to design optimization more than often require multiple UQ runs, with each run at a different input sample (e.g., for the iterative refinement of a surrogate or search for a global optimum). A computationally expensive UQ procedure could significantly increase the overhead time for surrogate modeling or design optimization, which may diminish the benefits of using an ML model in engineering design. To enable the wide adoption of UQ for ML in engineering design, the UQ method should be able to not only accurately quantify the predictive uncertainty, but also be very efficient in doing that. The methods presented in Sec. \ref{sec:bnn} and \ref{sec:neuralnetworkensemble} have great potential to address this issue. 
\end{enumerate}

In summary, UQ of ML models is essential for ML-based engineering design to enable accelerated design optimization and analysis and scale design optimization to large-scale problems. The approaches presented in Sec. \ref{sec:UQ_methods} could lead to a paradigm shift in various engineering design applications (e.g., materials, energy systems, additive manufacturing, to name a few) in the long term.


\section{UQ of ML models in prognostics}
\subsection{Introduction to prognostics}\label{sec:Intro_prognostics}
Prognostics aims to predict the future evolution of the health condition of systems, components, or processes based on their current state, the past evolution of the health condition, and the future predicted or planned usage or operating profile \cite{olgareview}. If no additional information on the future usage or operating profile is available, it is often assumed that the system will be operated in the same way as it was operated in the past.

Generally, two different types of data-driven approaches for predicting the RUL can be distinguished \cite{schwabacher2007survey}: 
\begin{enumerate}
    \item Identifying a health indicator and predicting its trend until a defined threshold is reached.
    \item Directly mapping the extracted features or raw measurements as in the case of DL to the RUL.
\end{enumerate}

For the first approach, the focus is on identifying a specific parameter or health indicator that is indicative of the health state of the system or component being monitored. This degradation indicator could be a physical measurement, a derived relevant feature or a combination of several degradation indicators that change over time as the system undergoes degradation. Once the health indicator is identified, the next step is to predict its trend over time. This involves using various predictive modeling techniques, such as regression or time-series analysis, to estimate how the health indicator evolves as the system degrades over time. The goal is to predict when the health indicator will reach a defined threshold, indicating that the system or component is reaching the end of its useful life.

For the second approach, instead of focusing on predicting the trend of a specific health indicator, the predictive model directly maps either the extracted features or, in the case of deep learning, directly from the raw measurements of the system or component to the RUL.

\subsection{Sources of uncertainty in prognostics}\label{sec:uncertainty_prognostics}

In prognostics, there are several sources of uncertainty that can affect the quality of RUL predictions. These uncertainties can originate from diverse factors, and depending on the system, they can impact the RUL prediction to various degrees \cite{sankararaman2015uncertainty}.

While measurement and model uncertainty are common sources of uncertainty in all disciplines and are also encountered in prognostics, some additional challenges for prognostics in terms of uncertainty include the uncertainty of the future usage and operating profiles, the quality and the limited availability of representative time-to-failure trajectories, high variability of operating conditions, and the dependence on external factors and environmental conditions and their impact on system degradation. Moreover, since failure modes and their mechanisms play a crucial role in the evolution of component and system degradation, the precise degradation mechanisms leading to failures may not be fully understood or may involve complex interactions. Such uncertainty in failure modes adds an additional source of uncertainties to the predictions.

\subsection{DL for prognostics}\label{sec:ML_prognostics}
The great advantage brought by DL approaches in the context of prognostics stems from their ability to automatically process high-dimensional, heterogeneous - and often noisy - sensor data in an end-to-end fashion, learn the features automatically and reduce the necessity for hand-crafted feature extraction to the minimum ~\cite{biggio2020prognostics}. This concept has given rise to extensive research showcasing the prediction capabilities of modern DL algorithms in the context of prognostics. Nevertheless, most of these approaches are designed to output a single-point estimate of the RUL of the considered industrial or infrastructure assets (\cite{olgareview,biggio2020prognostics, WANG202081} and the references therein). This is the case since standard neural networks' outputs are deterministic and are not typically accompanied by a meaningful probabilistic interpretation. This is undesirable in the context of prognostics. Sensor data are frequently distorted by multiple sources of noise and, training data is often limited in scope and fails to represent the full range of conditions that may arise in real-world scenarios.  Consequently, there is a significant risk of encountering high levels of epistemic uncertainties, which must be quantified and communicated to the decision makers.

\subsection{State-of-the-art uncertainty-aware DL approaches for prognostics}\label{sec:uncertainty_aware_DL_prognostics}

The emergence of DNNs has contributed to mitigating the two aforementioned issues, providing a highly expressive class of methods capable of efficiently processing large-scale datasets (see \cite{biggio2020prognostics, olgareview, WANG202081} and the references therein). Since standard DL approaches do not naturally incorporate UQ routines, using neural networks in prognostics has come at the price of neglecting UQ, hence providing simple point-estimate predictions as outputs. Only recently, thanks to recent advances in BNNs, more efforts have been spent in designing uncertainty-aware DL techniques for prognostics.

One of the simplest strategies to enable UQ of DNNs is MC dropout. As explained in Section \ref{sec:mcdropout}, this method is based on activating dropout layers at inference time, thereby, making the neural network's forward pass stochastic. Thanks to its intuitive rationale and relatively straightforward implementation, it is not surprising that the majority of uncertainty-aware DL methods for prognostics have been established on this strategy in combination with standard neural network architectures, such as fully-connected neural networks \cite{uqp3,uqp13}, CNNs \cite{uqp11,uqp15,uqp17}, and RNNs \cite{uqp1,uqp10,uqp14,uqp16,kong2022bayesian,peng2019bayesian}. Engineered systems to which MC dropout has been applied in prognostics include lithium-ion batteries \cite{uqp13,uqp16,uqp17}, turbofan engines \cite{uqp3,uqp11,uqp14, XIANG2023110187}, bearings \cite{uqp1,uqp17}, solenoid valves \cite{uqp15}, hydraulic mechanisms \cite{uqp10}, and circuit breakers \cite{uqp10}. While most of the studies have applied existing MC dropout implementations to prognostics, in \cite{uqp10}, the authors propose an adapted framework to model epistemic and aleatory uncertainty by means of MC dropout and a final aleatory layer with two nodes representing the parameters of either a Gaussian or two-parameter Weibull distribution. By appropriately sampling from the weight distribution entailed by the MC dropout and from the output distribution of the final aleatory layer, the authors are able to extract and disentangle epistemic and aleatory uncertainty.

Besides MC dropout, ensemble methods \cite{uqp5,uqp6, uqp8,Rigamonti2017EnsembleOO} and deep Gaussian processes \cite{uqp2, ELLIS2022108805} have also been used in prognostics. In particular, in \cite{Rigamonti2017EnsembleOO}, an ensemble of Echo State Networks (ESNs), a type of reservoir computing method, aggregated with an additional ESN on top of the ensemble to estimate the residual variance, is used to predict the RUL and the associated prediction intervals. The model is tested both on toy cases and on real industrial datasets and is shown to yield good performance. In another research study, Deep Gaussian Processes \cite{damianou2013deep,dspp}, have been employed for the prediction of the RUL on a dataset of turbofan engines \cite{uqp2}. The advantage of these techniques lies in the fact that they combine the probabilistic nature of standard GPR and the expressive power of DNNs. In addition, contrarily to vanilla GPR, they can be applied to the \enquote{big-data} regime, which is very common in prognostics. The results show that deep Gaussian processes perform well in the task of RUL prediction, outperforming a number of deep learning baseline methods.

\section{Demonstration of Instability of MC Dropout}\label{appendix_MCDropout}

\begin{figure}[!ht]
\includegraphics[width=\textwidth]{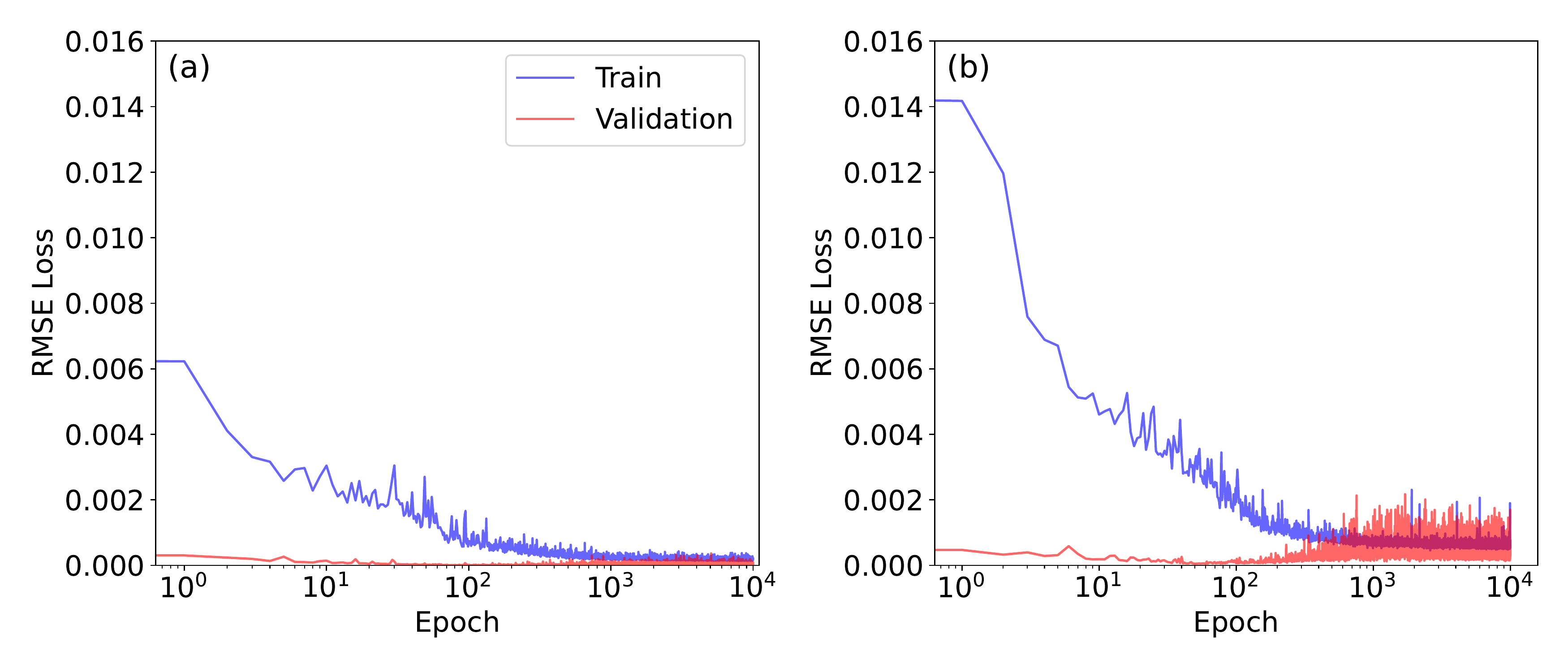}
\caption{Training and validation losses for MC dropout models with dropout rate of (a) 0.05 and (b) 0.2 respectively.}
\label{fig:mcd_loss}
\end{figure}

In Section \ref{sec:mcdropout}, we mention the instability of the MC dropout model arising from even slight variations in hyperparameters, such as model size, training epochs and dropout rate. In this appendix, we first show the training and validation losses for two MC dropout models trained with the same data of the toy example from Section \ref{sec:toy_example} in Fig.~\ref{fig:mcd_loss}. The two MC dropout models have the same architecture (3 ResNet blocks as shown in Fig.~\ref{fig:cs2_architecture}), but have different dropout rates. In this case, the MC dropout model converged at around 500 epochs, but no over-fitting is observed until 10000 epochs. Next, we plot the uncertainty maps for various configurations of the MC dropout model in Table~\ref{tab:MC_dropout_instable}. The uncertainty maps are highly inconsistent, thus leading to our conclusion about the instability of MC dropout.

\begin{table}[!ht]
\centering
\begin{tabular}{cc|ccc|}
\cline{3-5}
    &   & \multicolumn{3}{c|}{\textbf{Number of training epochs}}                                \\ \cline{3-5} 
    &   & \multicolumn{1}{c|}{200} & \multicolumn{1}{c|}{500} & 1000 \\ \cline{3-5} 
    &   & \multicolumn{3}{c|}{\cellcolor[HTML]{C0C0C0}Dropout rate: 0.05} \\ \hline
\multicolumn{1}{|c|}{\multirow{2}{*}[10.5ex]{\rotatebox[origin=c]{90}{\textbf{Number of ResNet blocks}}}}  & \multirow{1}{*}[8.5ex]{1} & \multicolumn{1}{c|}{\includegraphics[width=\afigwidth]{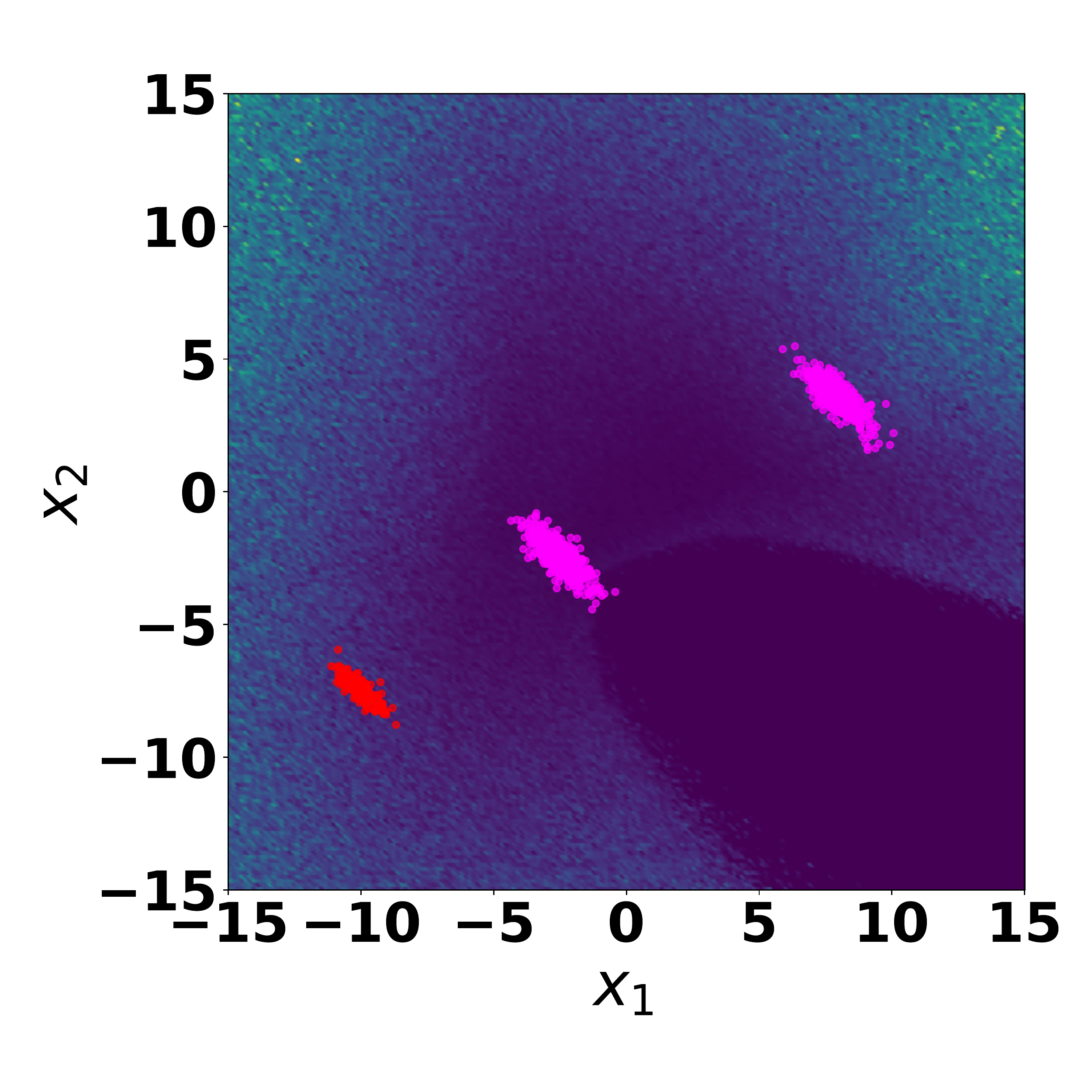}}   & \multicolumn{1}{c|}{\includegraphics[width=\afigwidth]{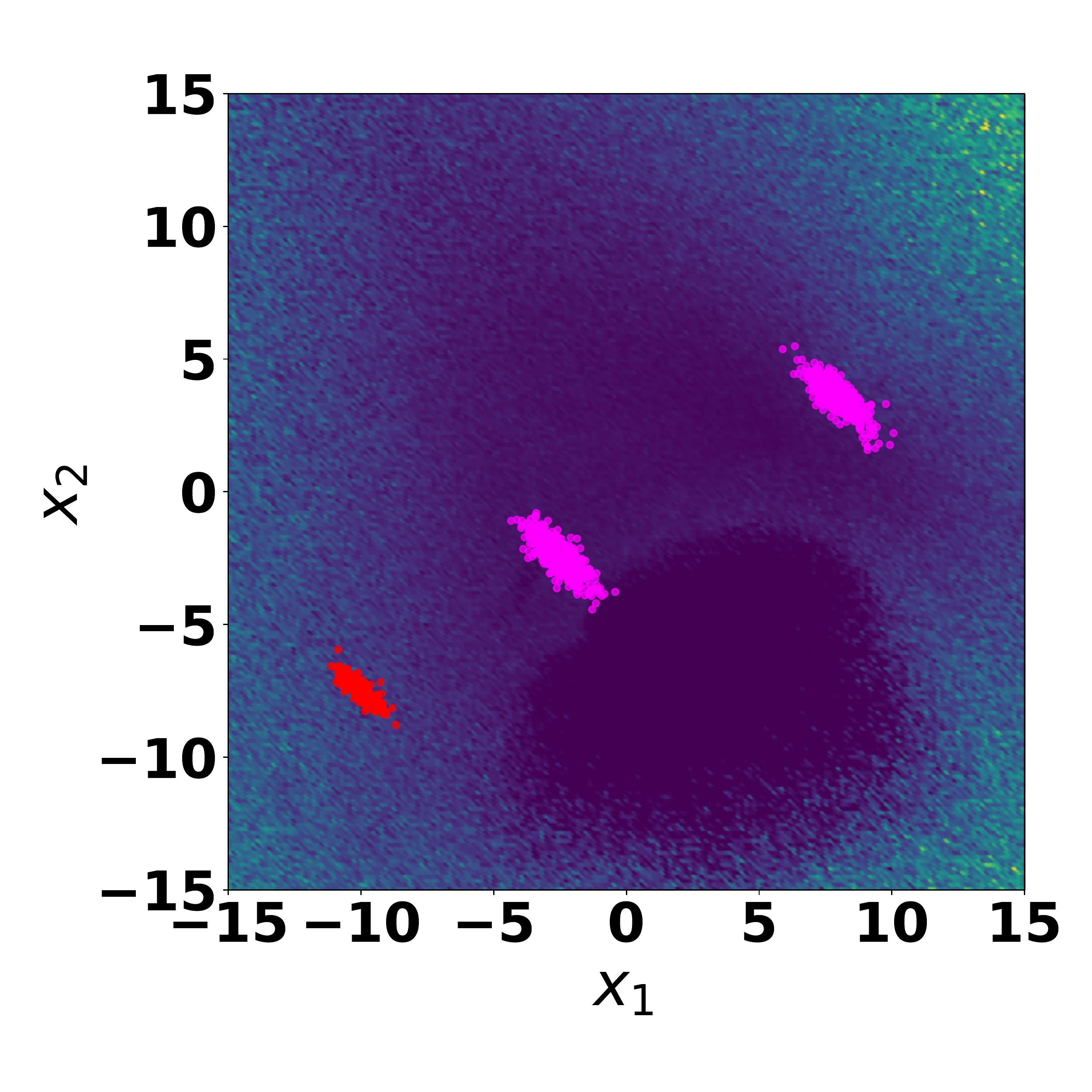}}   & \includegraphics[width=\afigwidth]{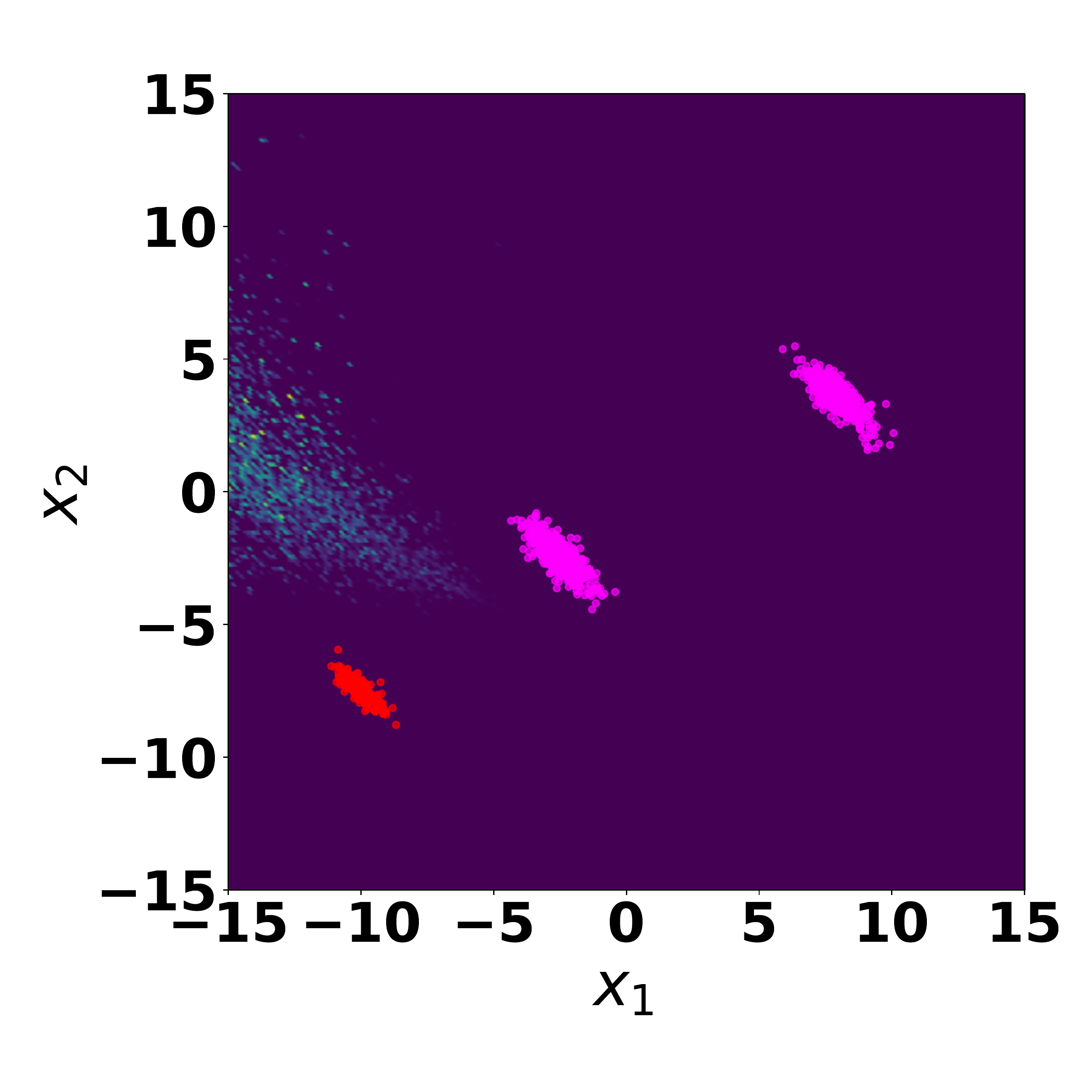}    \\ \cline{2-5} 
\multicolumn{1}{|c|}{}  & \multirow{1}{*}[8.5ex]{3} & \multicolumn{1}{c|}{\includegraphics[width=\afigwidth]{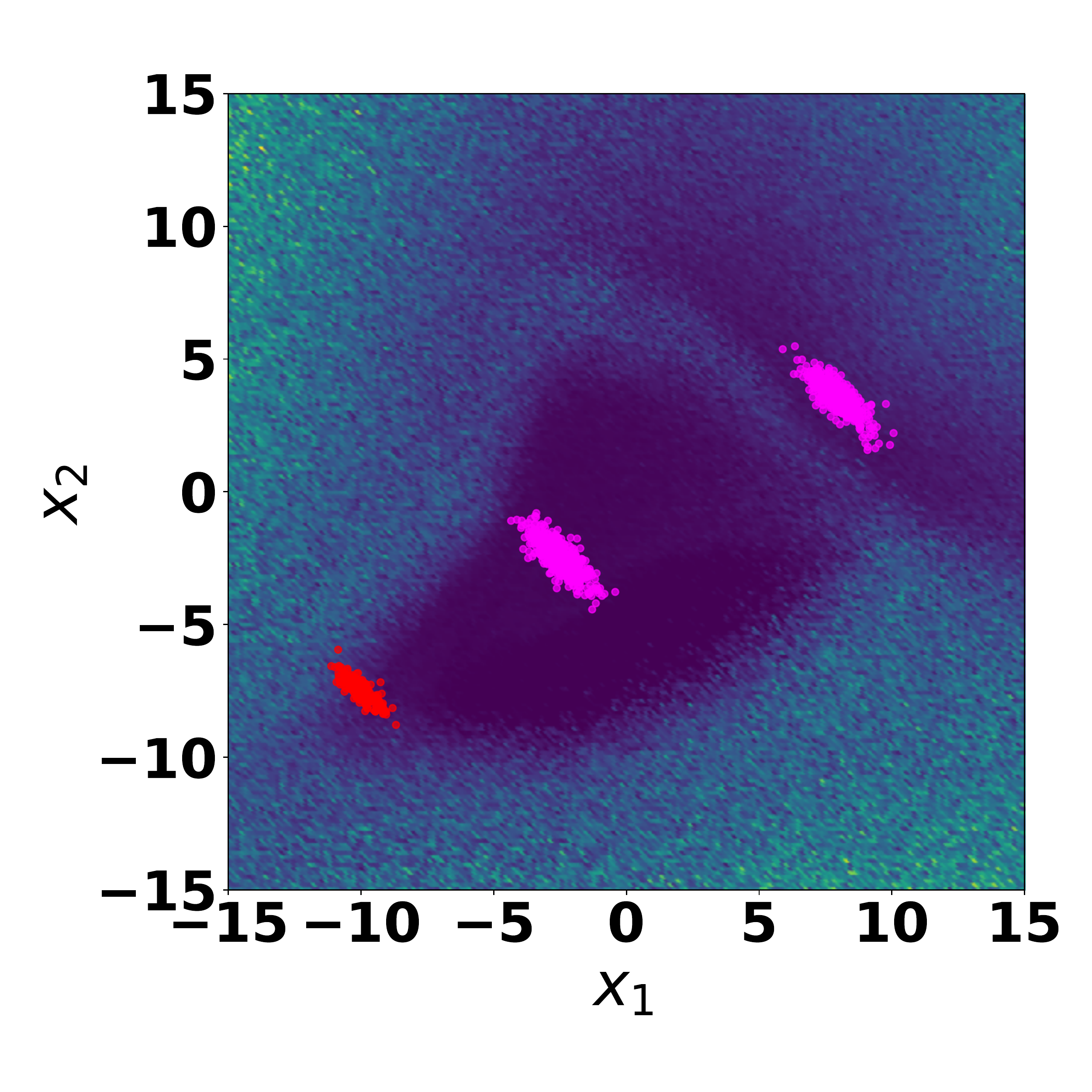}}   & \multicolumn{1}{c|}{\includegraphics[width=\afigwidth]{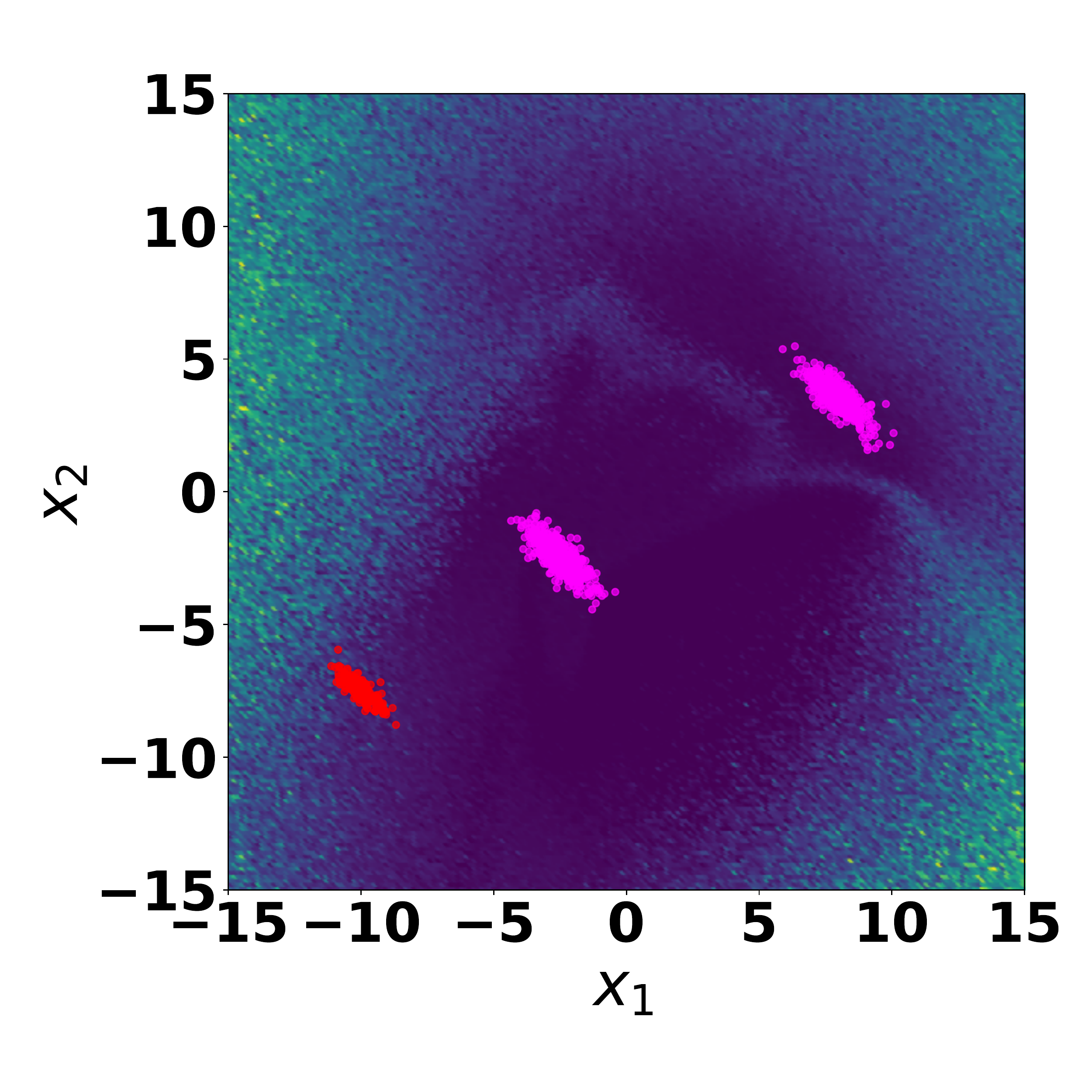}}   & \includegraphics[width=\afigwidth]{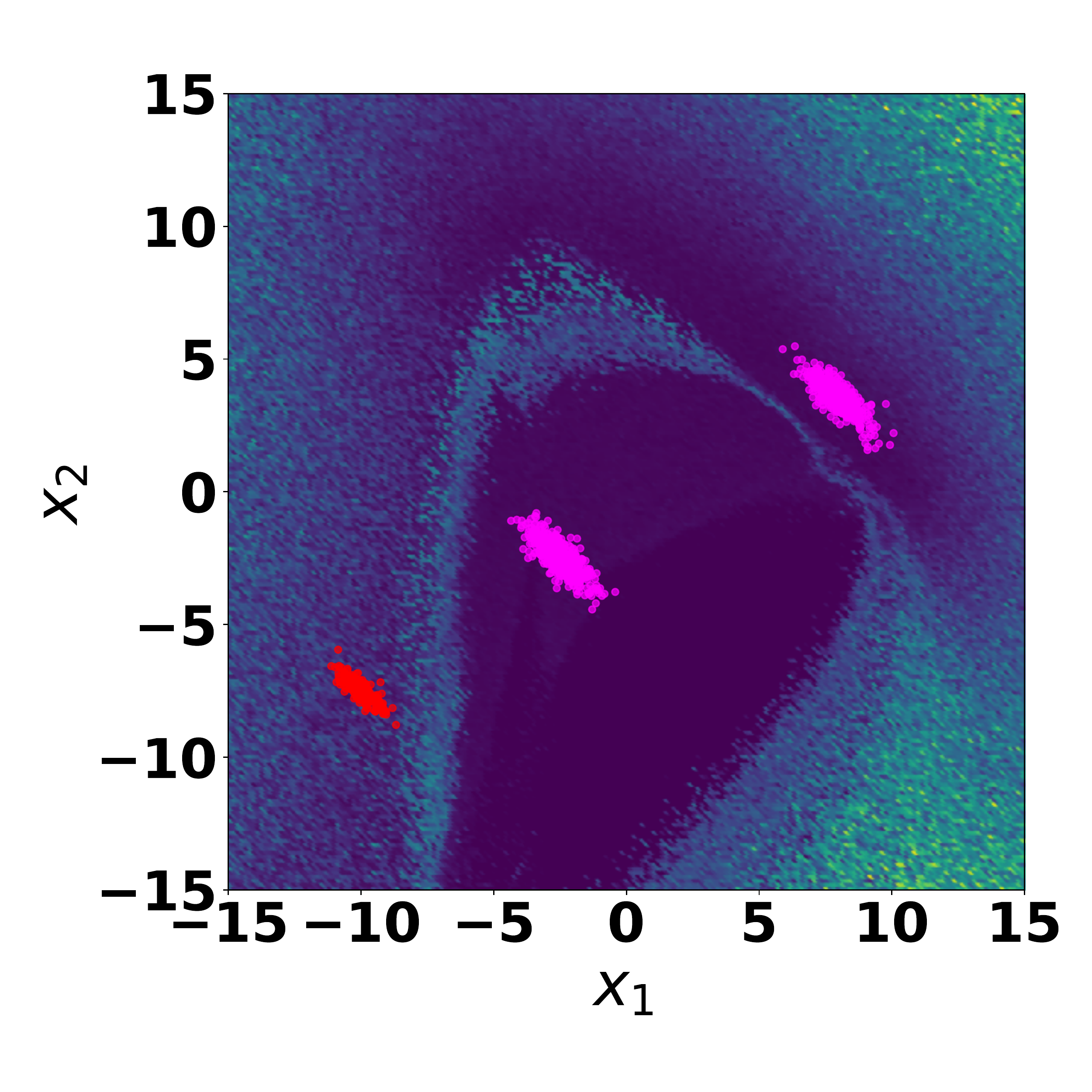}    \\ \hline
                      &   & \multicolumn{3}{c|}{\cellcolor[HTML]{C0C0C0}Dropout rate: 0.1}  \\ \hline
\multicolumn{1}{|c|}{\multirow{2}{*}[10.5ex]{\rotatebox[origin=c]{90}{\textbf{Number of ResNet blocks}}}}   & \multirow{1}{*}[8.5ex]{1} & \multicolumn{1}{c|}{\includegraphics[width=\afigwidth]{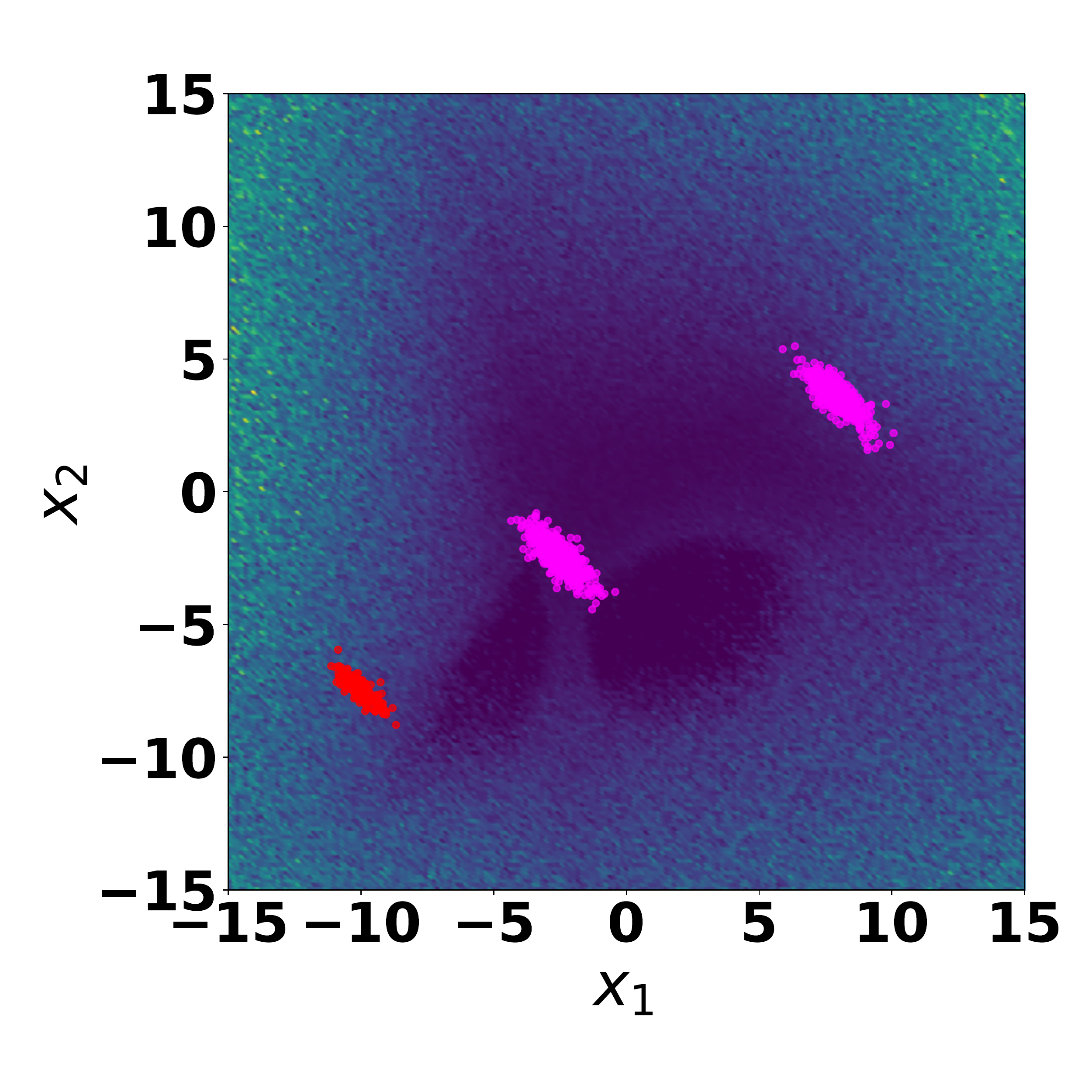}}   & \multicolumn{1}{c|}{\includegraphics[width=\afigwidth]{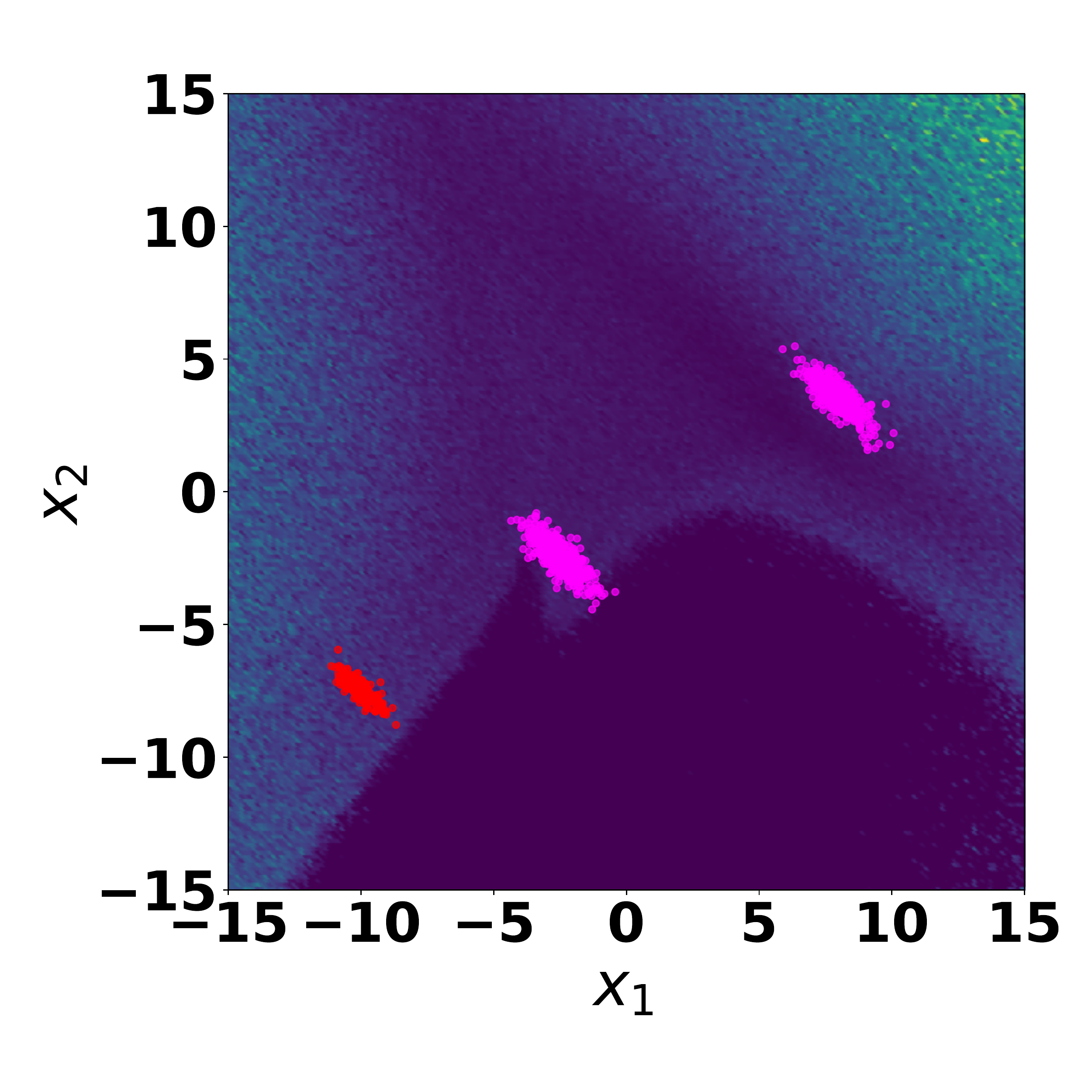}}   & \includegraphics[width=\afigwidth]{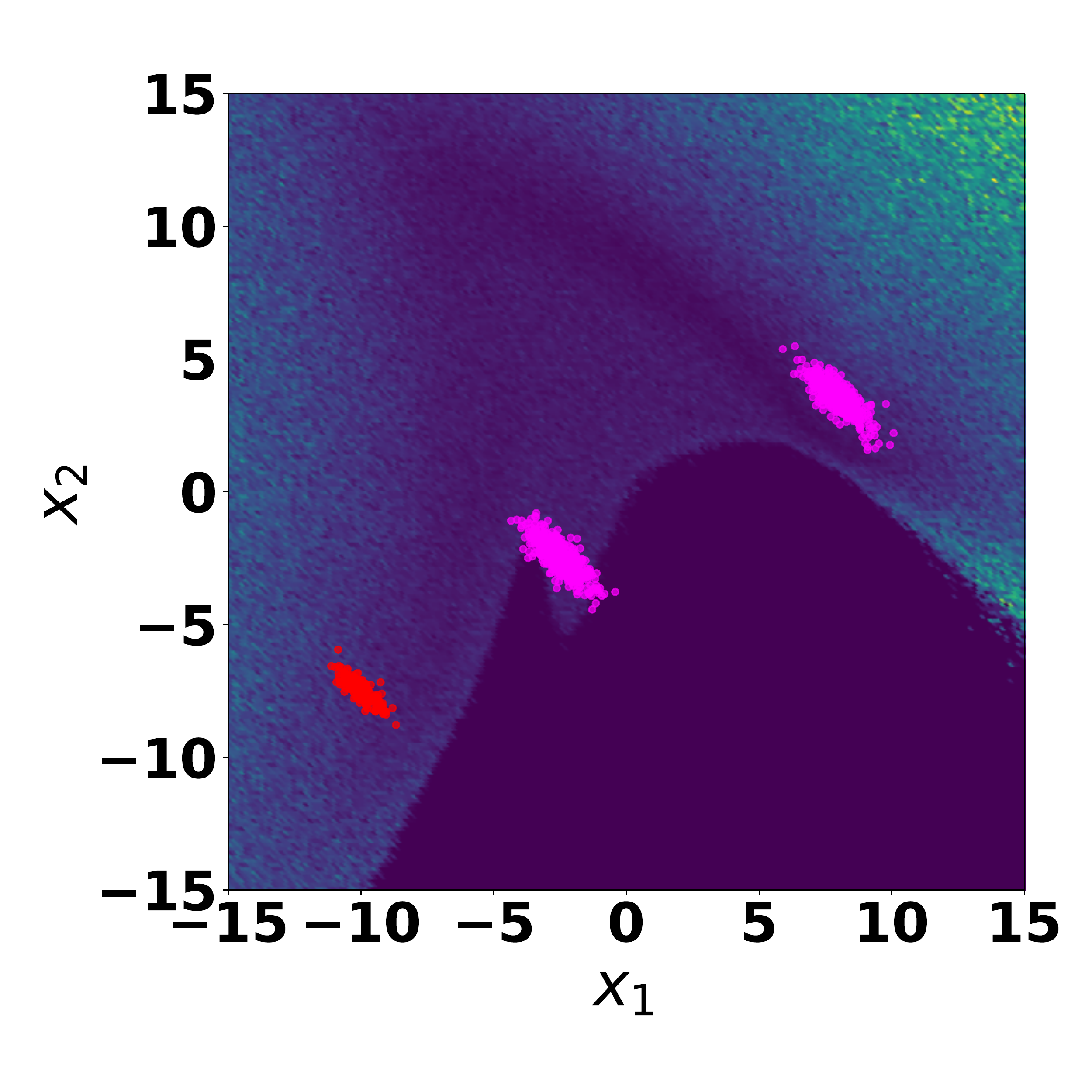}    \\ \cline{2-5} 
\multicolumn{1}{|c|}{} & \multirow{1}{*}[8.5ex]{3} & \multicolumn{1}{c|}{\includegraphics[width=\afigwidth]{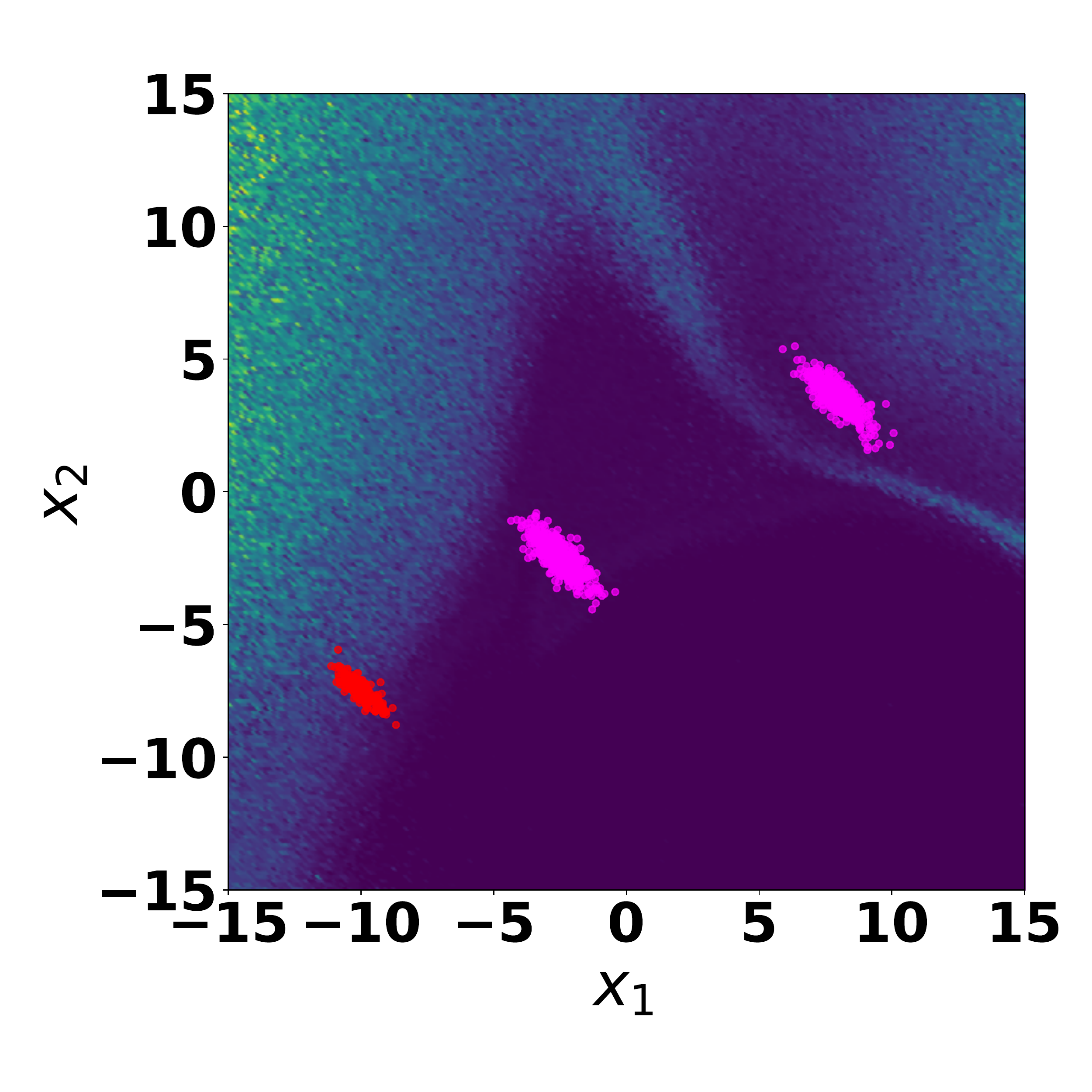}}  & \multicolumn{1}{c|}{\includegraphics[width=\afigwidth]{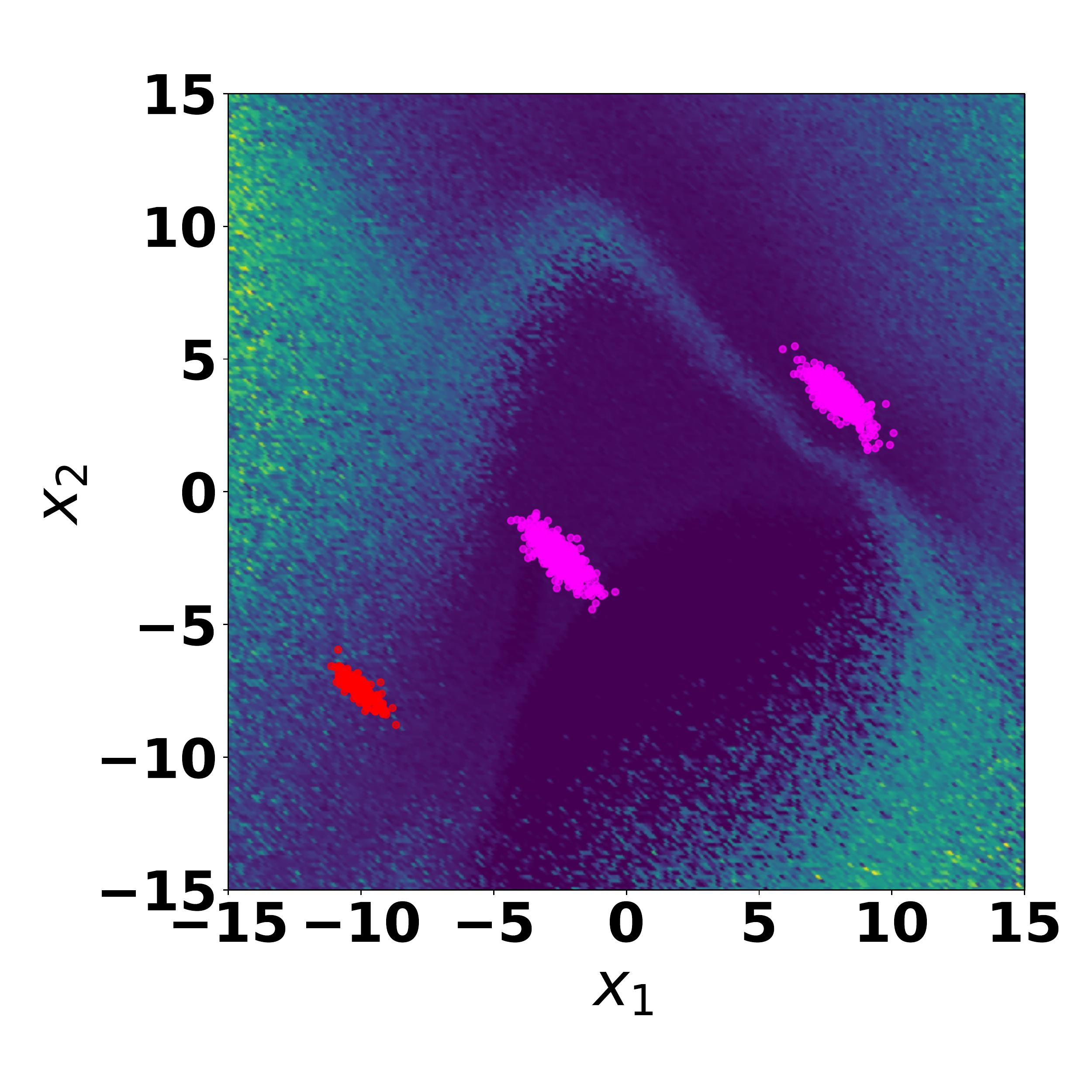}}  & \includegraphics[width=\afigwidth]{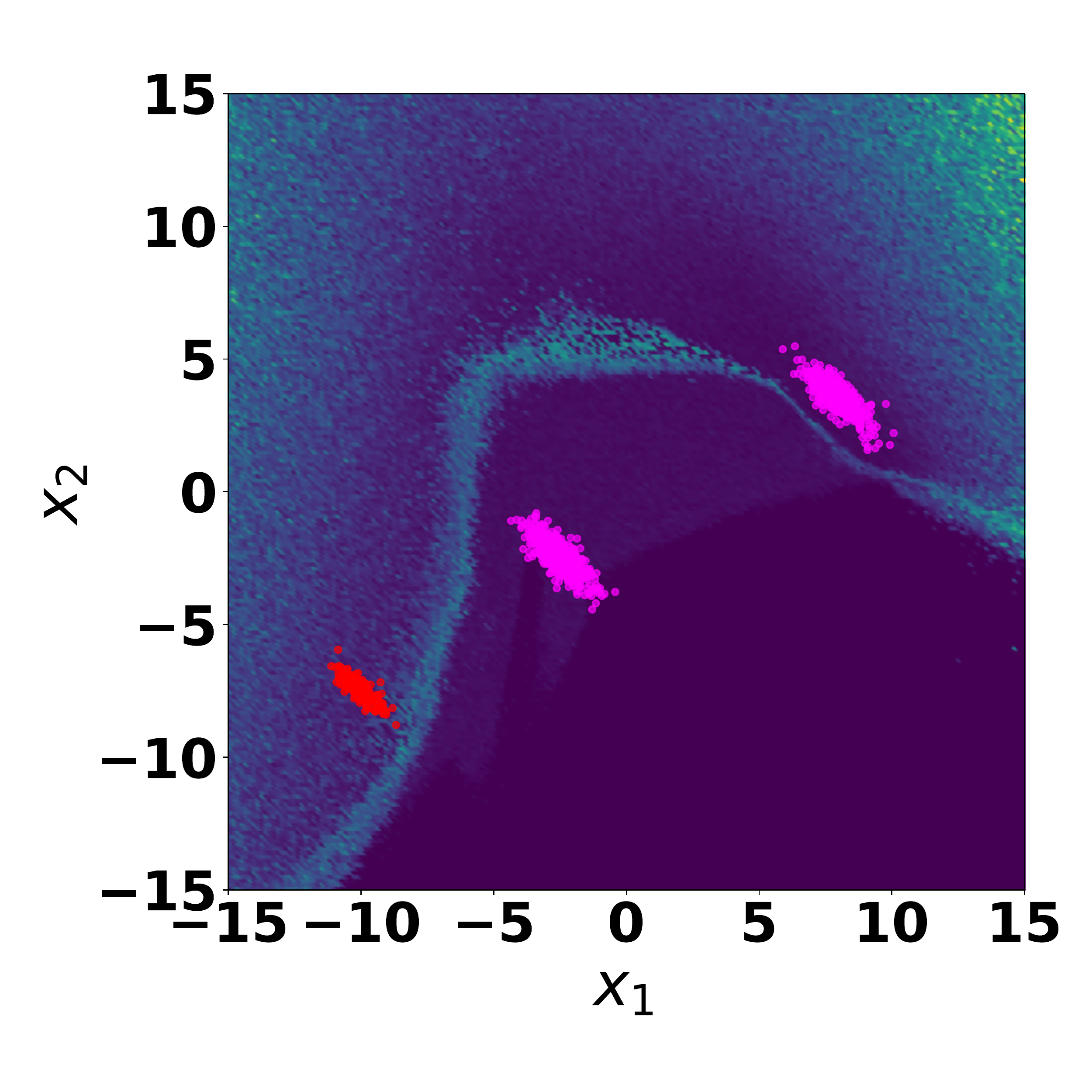}   \\ \hline
                                                                &   & \multicolumn{3}{c|}{\cellcolor[HTML]{C0C0C0}Dropout rate: 0.2}  \\ \hline
\multicolumn{1}{|c|}{\multirow{2}{*}[10.5ex]{\rotatebox[origin=c]{90}{\textbf{Number of ResNet blocks}}}}                                         & \multirow{1}{*}[8.5ex]{1} & \multicolumn{1}{c|}{\includegraphics[width=\afigwidth]{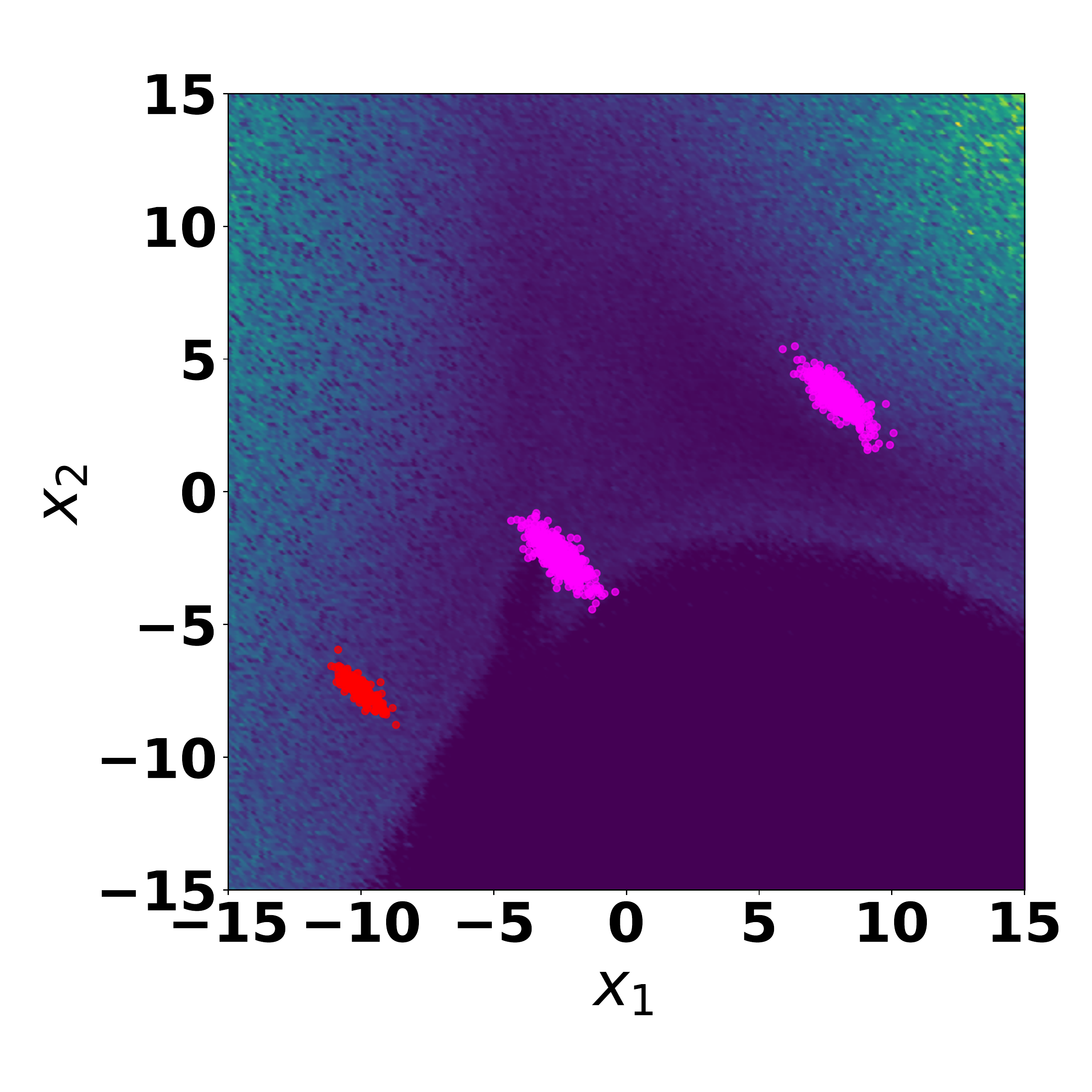}}  & \multicolumn{1}{c|}{\includegraphics[width=\afigwidth]{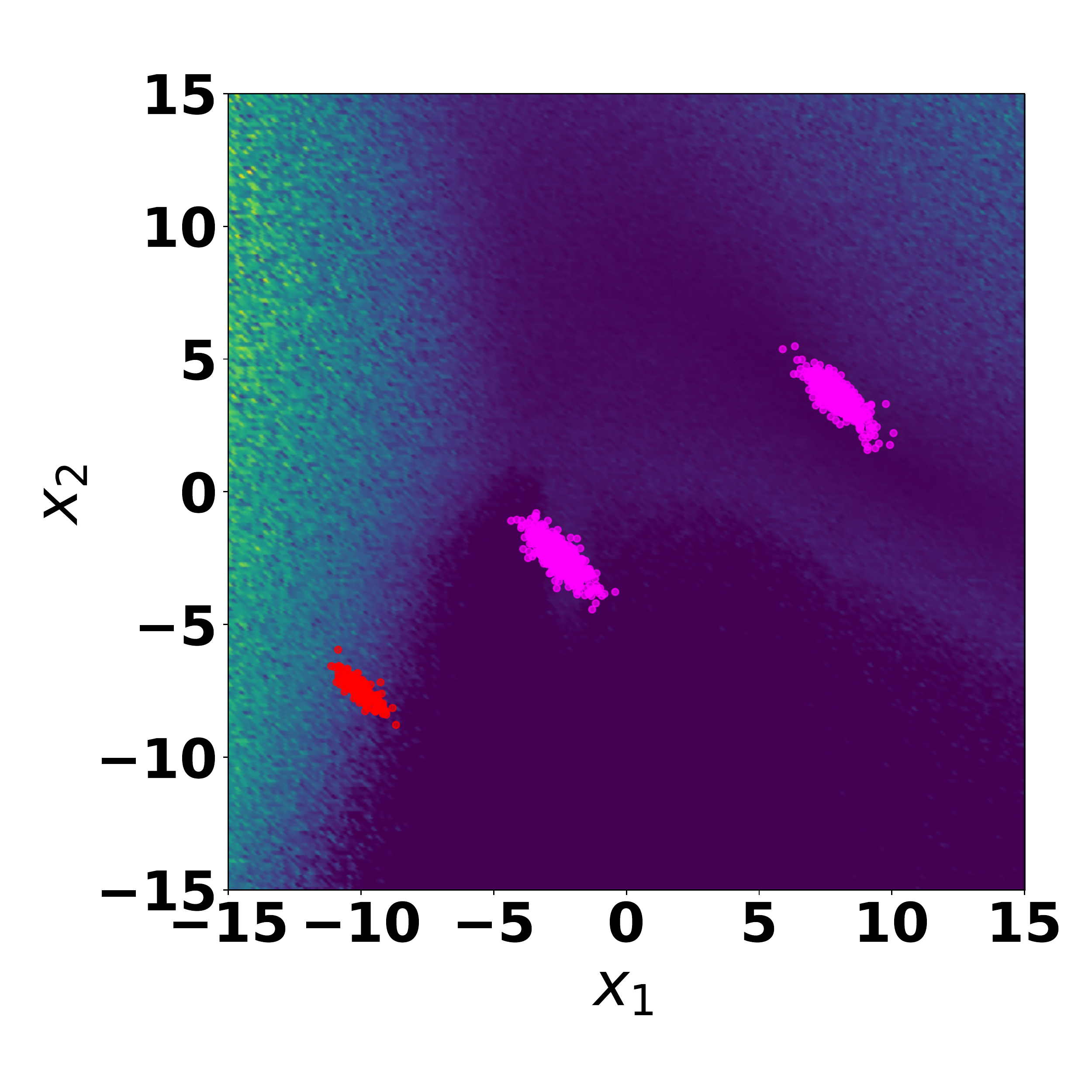}}  & \includegraphics[width=\afigwidth]{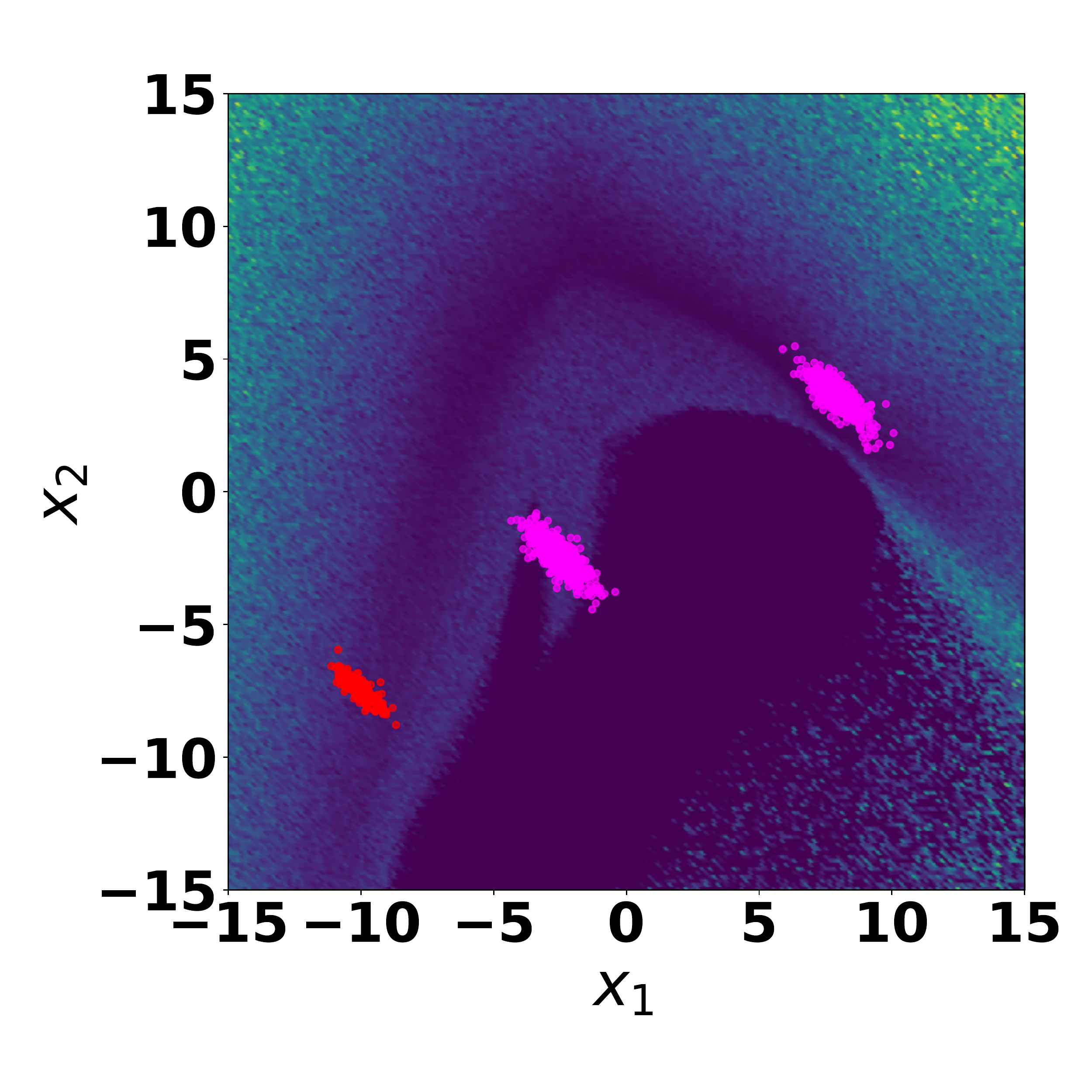}   \\ \cline{2-5} 
\multicolumn{1}{|c|}{}  & \multirow{1}{*}[8.5ex]{3} & \multicolumn{1}{c|}{\includegraphics[width=\afigwidth]{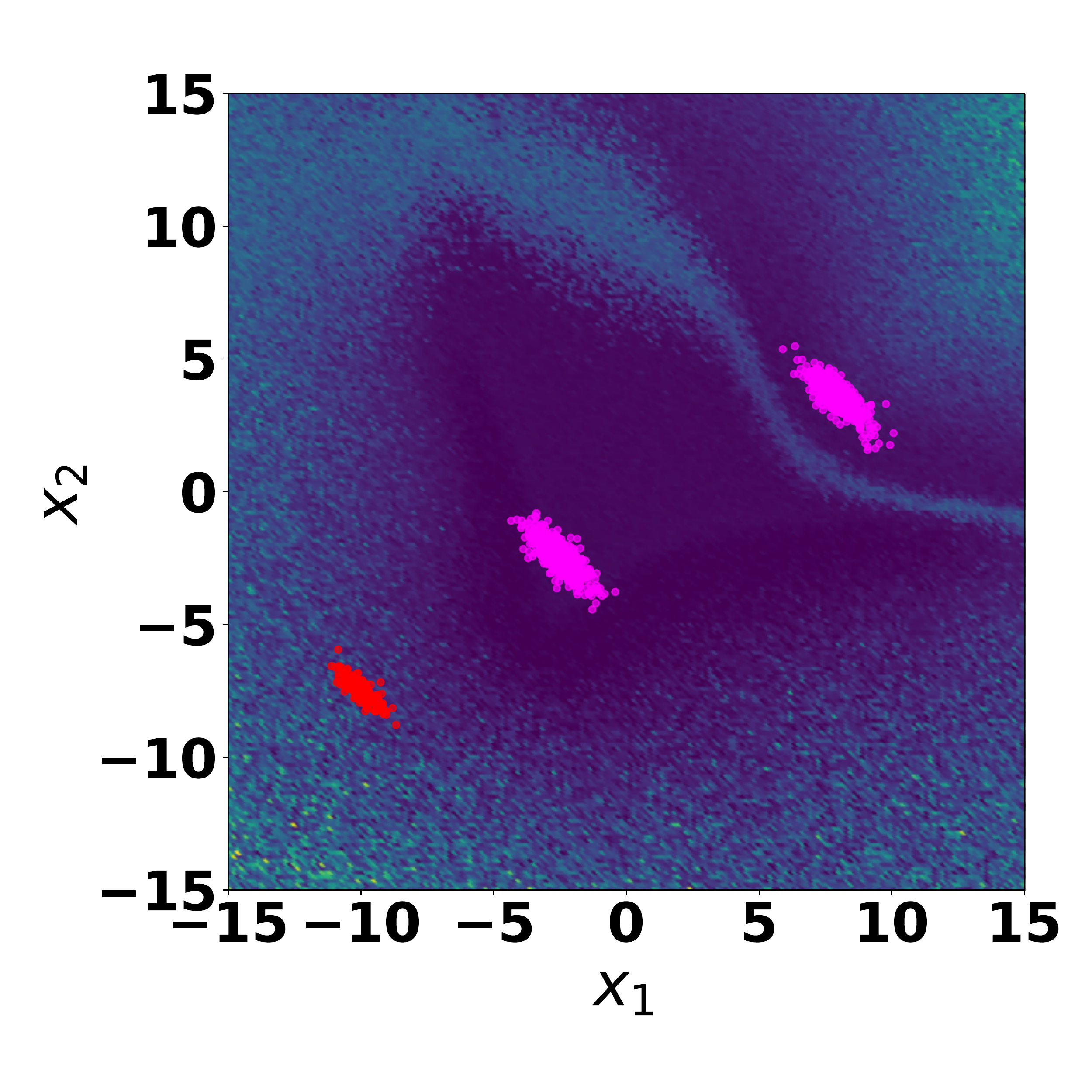}}  & \multicolumn{1}{c|}{\includegraphics[width=\afigwidth]{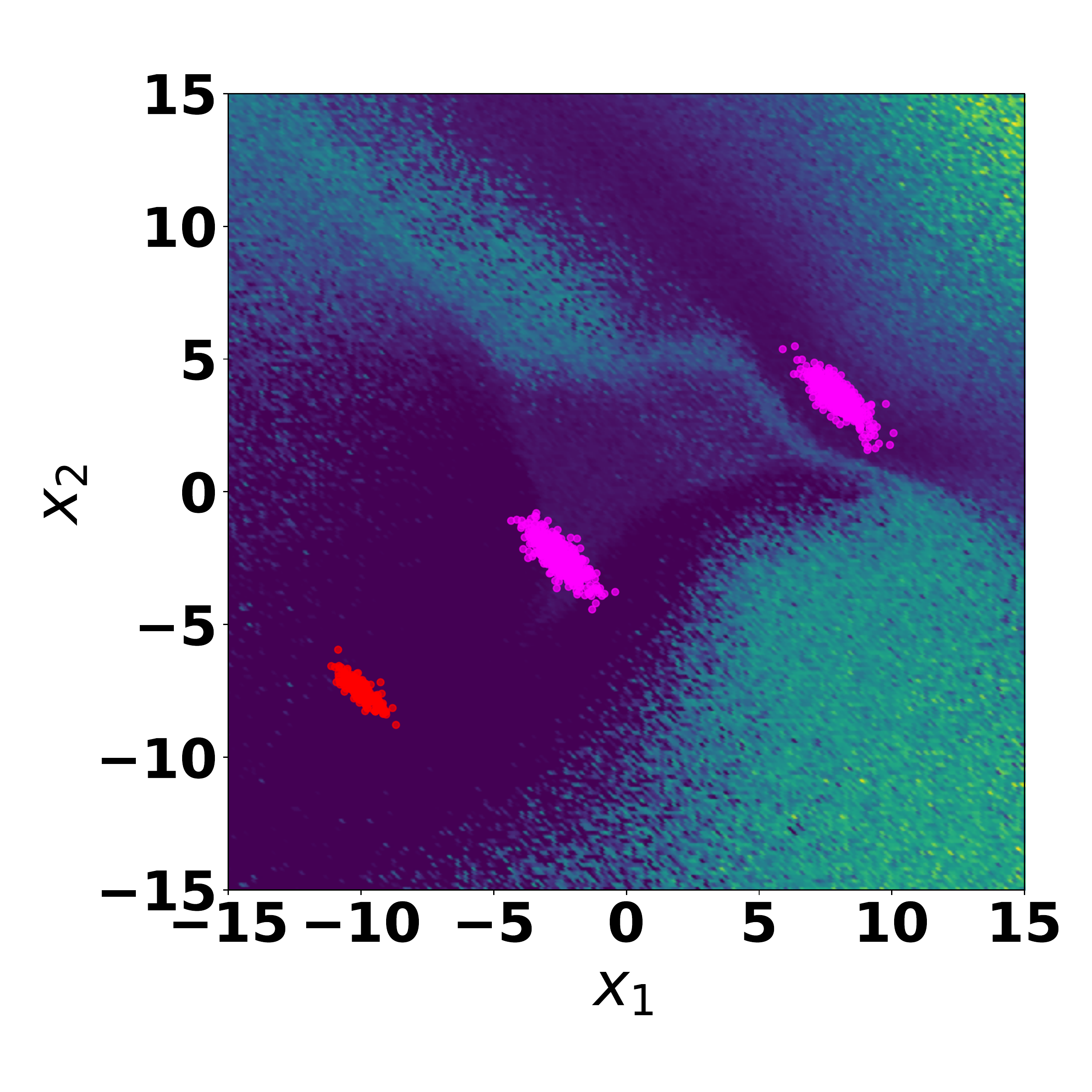}}  & \includegraphics[width=\afigwidth]{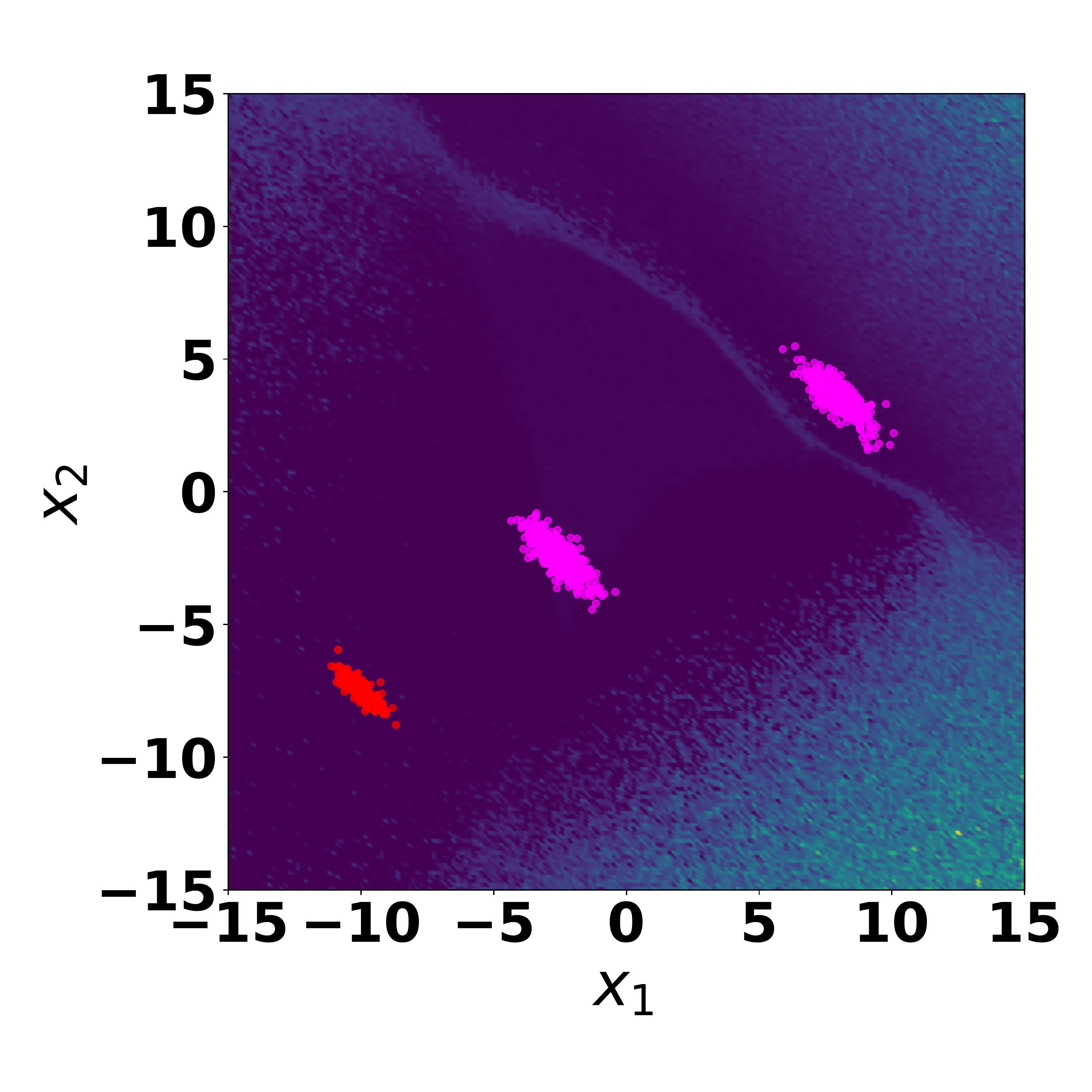}   \\ \hline
\end{tabular}
\\
\includegraphics[scale=0.5]{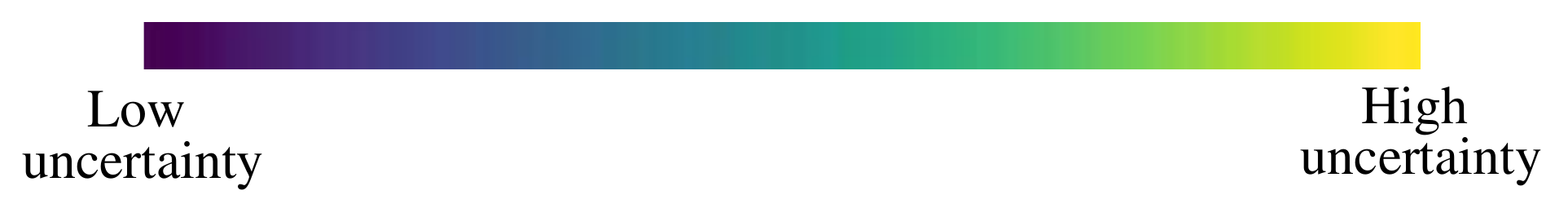}

\caption{Demonstration of the instability associated with uncertainty maps of MC dropout with respect to dropout rate, number of training epochs, and ResNet architecture.}
\label{tab:MC_dropout_instable}
\end{table}

\end{document}